\documentclass[pdflatex,sn-basic]{sn-jnl}
\usepackage{enumitem}
\usepackage{color}
\definecolor{SilverChalice}{rgb}{0.643,0.643,0.643}
\usepackage{pifont}
\usepackage{adjustbox}
\usepackage{graphicx}%
\usepackage{multirow}%
\usepackage{amsmath,amssymb,amsfonts}%
\usepackage{amsthm}%
\usepackage{mathrsfs}%
\usepackage[title]{appendix}%
\usepackage{xcolor}%
\usepackage{textcomp}%
\usepackage{manyfoot}%
\usepackage{booktabs}%
\usepackage{algorithm}%
\usepackage{algorithmicx}%
\usepackage{algpseudocode}%
\usepackage{listings}%
\usepackage{stfloats}
\usepackage{caption}
\usepackage{subcaption}
\usepackage{float}

\theoremstyle{thmstyleone}%
\theoremstyle{thmstyletwo}%

\theoremstyle{thmstylethree}%

\begin{document}

\title[Article Title]{Uncertainty-Aware Post-Hoc Calibration:\\ Mitigating Confidently Incorrect Predictions Beyond Calibration Metrics}

\author[1]{\fnm{Hassan} \sur{Gharoun}}\email{Hassan.Gharoun@student.uts.edu.au}

\author[1]{\fnm{Mohammad Sadegh} \sur{Khorshidi}}\email{Mohammadsadegh.khorshidialikordi@student.uts.edu.au}

\author[1]{\fnm{Kasra} \sur{Ranjbarigderi}}\email{Kasra.Ranjbarigderi@student.uts.edu.au}

\author[1]{\fnm{Fang} \sur{Chen}}\email{Fang.Chen@uts.edu.au}

\author*[1,2]{\fnm{Amir Hossein} \sur{Gandomi}}\email{Gandomi@uts.edu.au}

\affil*[1]{\orgdiv{Faculty of Engineering \& IT}, \orgname{University of Technology Sydney}, \orgaddress{\street{Broadway}, \city{Sydney}, \postcode{2007}, \state{NSW}, \country{Australia}}}

\affil[2]{\orgdiv{University Research and Innovation Center (EKIK)}, \orgname{Óbuda University}, \orgaddress{\city{Budapest}, \postcode{1034}, \state{Budapest}, \country{Hungary}}}

\abstract{
Despite extensive research on neural network calibration, existing methods typically apply global transformations that treat all predictions uniformly, overlooking the heterogeneous reliability of individual predictions. Furthermore, the relationship between improved calibration and effective uncertainty-aware decision-making remains largely unexplored. This paper presents a post-hoc calibration framework that leverages prediction reliability assessment to jointly enhance calibration quality and uncertainty-aware decision-making. The framework employs proximity-based conformal prediction to stratify calibration samples into putatively correct and putatively incorrect groups based on semantic similarity in feature space. A dual calibration strategy is then applied: standard isotonic regression calibrated confidence in putatively correct predictions, while underconfidence-regularized isotonic regression reduces confidence toward uniform distributions for putatively incorrect predictions, facilitating their identification for further investigations. A comprehensive evaluation is conducted using calibration metrics, uncertainty-aware performance measures, and empirical conformal coverage. Experiments on CIFAR-10 and CIFAR-100 with BiT and CoAtNet backbones show that the proposed method achieves lower confidently incorrect predictions, and competitive Expected Calibration Error compared with isotonic and focal-loss baselines. This work bridges calibration and uncertainty quantification through instance-level adaptivity, offering a practical post-hoc solution that requires no model retraining while improving both probability alignment and uncertainty-aware decision-making.
}


\maketitle

\section{Introduction}\label{sec:Intro}
Artificial intelligence (AI) and machine learning (ML) have become an essential tool for solving complex problems across various domains, demonstrating impressive predictive performance. The successful deployment and adoption of AI and ML-powered decision support tools in real-world settings and applied domains requires models not only produce accurate classification results but also display reliable confidence \citep{carneiro2020deep}. 

Confidence refers to the model's belief in its prediction, typically represented as a probability score assigned to each class label \citep{carse2022calibration,guo2017calibration}. This score indicates how likely the model estimates the input belongs to a particular class. Ideally, if a model classifies an instance with 80\% confidence, this means that—over a large number of similar cases—approximately 80\% of such predictions should be correct.

In this context, calibration refers to the alignment between a model’s predicted probabilities and the true likelihood of correctness, often achieved by adjusting these probabilities to improve their reliability \citep{kull2019beyond}. Similarly, miscalibration occurs when the predicted confidence scores do not align with the actual probability of correctness. A model may assign a high probability to a prediction, but if it is miscalibrated, this confidence value may not accurately reflect the true likelihood of the classification being correct.

The scientific community has proposed various metrics to quantify the agreement between predicted probabilities and observed event frequencies, with the Expected Calibration Error (ECE) \citep{naeini2015obtaining} being among the most established and widely adopted. ECE, mathematically presented by Eq.~\ref{Eq:ECE}, partitions the prediction space into bins according to predicted probabilities and measures the discrepancy between the mean confidence and empirical accuracy within each bin, with the final value computed as the weighted average of these differences across all bins.

\begin{equation}
    \mathrm{ECE} = \frac{1}{N} \sum_{m=1}^{M} |b_m| \, \big| \mathrm{acc}(b_m) - \mathrm{conf}(b_m) \big|
\label{Eq:ECE}
\end{equation}

\noindent where $N$ is the total number of samples, $|b_m|$ is the number of samples in bin $b_m$, $\mathrm{acc}(b_m)$ and $\mathrm{conf}(b_m)$ are the the accuracy and confidence of bucket $m$ calculated by Eq.~\ref{Eq:ECE_acc_conf}.

\begin{equation}
    \mathrm{acc}(b_m) = \frac{1}{|b_m|} \sum_{i \in B_m} \mathbf{1}(\hat{y}_i = y_i), 
    \quad
    \mathrm{conf}(b_m) = \frac{1}{|b_m|} \sum_{i \in B_m} p_i.
    \label{Eq:ECE_acc_conf}
\end{equation}

While ECE provides an aggregate measure of calibration error across all bins, the Maximum Calibration Error (MCE) \citep{naeini2015obtaining} is another widely adopted metric that captures the most pronounced deviation within a single bin, formulated by Eq. \ref{Eq:MCE}.

\begin{equation}
    \mathrm{MCE} = \max_{m \in \{1, \dots, M\}} \big| \mathrm{acc}(b_m) - \mathrm{conf}(b_m) \big|.
    \label{Eq:MCE}
\end{equation}

When models exhibit miscalibration, calibration methods adjust confidence estimates to improve probability-frequency alignment \citep{wang2023calibration}. As immediately apparent from their mathematical definitions, ECE and conventional calibration metrics lack the capacity to quantify the variability or inconsistency in prediction confidence estimates. Variability in confidence estimates may arise either from inherent noise in the data (aleatoric uncertainty) or from the model’s limited knowledge of its parameters (epistemic uncertainty). This variability in prediction confidence estimates is defined as uncertainty in ML and AI predictions \citep{gharoun2025proximitybased}. The existence of various uncertainty sources requires models to explicitly evaluate the reliability of the confidence estimates to foster trust. The literature has established various methods known as uncertainty quantification techniques that estimate prediction uncertainty by generating multiple probability estimates for individual instances through methods such as ensemble modeling or Bayesian methods \citep{abdar2021review}, thereby quantifying the variability in model predictions rather than relying on single probability estimates. 

The breadth of the predictive distribution obtained through uncertainty quantification techniques serves as a measure of the model's uncertainty in its prediction for a given instance. In the literature, Prediction entropy (PE) is extensively used to quantify the uncertainty associated with individual predictions by measuring the spread of the prediction probability distribution. Mathematically, PE can be expressed as Eq.~\ref{Eq:PE}:

\begin{equation}
    PE(\mathbf{x}) = - \sum_{c=1}^{C} \mu_{\text{pred}} (\mathbf{x}, c) \log[\mu_{\text{pred}} (\mathbf{x}, c)]
    \label{Eq:PE}
\end{equation}

\noindent where $\mu_{\text{pred}}(\mathbf{x}, c)$ represents the mean predicted probability for class c given input $\mathbf{x}$.

Extensive research has typically employed threshold-based decision rules, under which predictions with uncertainty estimates (such as PE), exceeding a predetermined threshold have been classified as uncertain, whereas those below the threshold have been regarded as certain. The underlying rationale is that uncertain predictions are flagged for human review, thereby enabling an additional expert opinion or triggering a safety protocol to validate or override the model’s decision.

In the ideal scenario, correct predictions would consistently be associated with certainty, whereas incorrect predictions would be flagged as uncertain. Such alignment enables users to trust predictions identified as certain, while those labeled uncertain provide a clear signal for further evaluation.

To establish the relationship between entropy values and mean confidence requirements, consider the binary classification. The entropy curve reaches its peak when both classes are equally likely, i.e., each probability is about 0.5, which yields a maximum of 1.0 under base-2 logarithms. To achieve low entropy values significantly below this maximum, the model must assign probabilities that strongly favor one class. For instance, when using log base~2 and setting an entropy threshold of $\tau = 0.1$, the required probability for the dominant class approaches $0.93$. This mathematical constraint extends to multiclass problems where the maximum entropy scales as $\log(C)$ for $C$ classes, making low entropy thresholds increasingly demanding of model confidence. The fundamental structure of entropy thus creates a mathematical requirement for high-confidence predictions when implementing low-uncertainty filtering, independent of whether such confidence levels are empirically warranted.

After establishing the mathematical relationship between prediction entropy and confidence requirements, along with the definition and purpose of calibration metrics, it becomes possible to distinguish between calibration methods and uncertainty-aware decision-making. Neural networks have been shown to be prone to producing overconfident predictions \citep{guo2017calibration}, meaning they tend to be miscalibrated. Calibration methods aim to adjust these overconfident estimates to improve probability-frequency alignment, with calibration metrics such as ECE measuring the degree of miscalibration. However, while ECE and similar metrics provide valuable information about the statistical alignment between confidence estimates and outcome frequencies, these metrics do not necessarily indicate whether a model exhibits uncertainty-aware behavior, for example, distinguishing certain correct predictions from uncertain incorrect ones in a manner that is useful for decision making.

Consider the ECE measuring the alignment between average accuracy and average confidence across prediction bins. This aggregation can mask problematic scenarios where incorrect predictions exhibit confidence levels similar to correct predictions within the same bin. Specifically, high-confidence bins may contain incorrect predictions that express unwarranted certainty, yet these miscalibrated instances become obscured when averaged with well-calibrated correct predictions.

\noindent This reveals fundamental limitations:  
\begin{itemize}
    \item ECE's bin-wise averaging can yield acceptable calibration scores even when the model fails to distinguish between justifiably certain correct predictions and inappropriately certain incorrect predictions—the hallmark of uncertainty-aware behavior. It is important to recognize that calibration techniques and metrics are not designed to capture instance-level uncertainty awareness. Therefore, calibration metrics should be complemented by analyses examining whether models appropriately adjust confidence based on prediction correctness—assigning high confidence to correct predictions and low confidence to incorrect ones.
    \item Existing calibration methods apply uniform transformations across all predictions regardless of their correctness likelihood. However, calibrating incorrect predictions is generally uninformative, as probability adjustments cannot convert an incorrect prediction into a correct one. This limitation, consistent with that of calibration metrics noted above, further challenges the pursuit of uncertainty-aware decision-making.
\end{itemize}

\noindent \textbf{Contribution.} Accordingly, this study addresses these limitations by introducing an uncertainty-aware calibration method. The proposed approach seeks to enhance uncertainty-aware decision-making through selective post-hoc calibration, a direction that, to the best of the authors’ knowledge, has been largely overlooked in the literature. To move toward the ideal of uncertainty-aware behavior, the method integrates conformal prediction to separate predictions into putatively correct and putatively incorrect groups. The putatively correct predictions are calibrated to align confidence estimates with observed outcome frequencies, while the putatively incorrect predictions are intentionally flattened toward the lowest possible confidence values. In this way, the method preserves prediction accuracy while promoting uncertainty-awareness: predictions that are likely correct retain meaningful calibration, whereas predictions that are likely incorrect are explicitly down-weighted to reflect the model’s limitations. Importantly, the approach adheres to the principle of post-hoc calibration by adjusting confidence estimates without altering the predicted class labels.

The remainder of this paper is organized as follows. Section \ref{sec:Background} reviews the relevant literature on calibration, uncertainty quantification, and their applications in decision-making systems. Section \ref{sec:Method} details the proposed methodology. Section \ref{sec:Experiments} describes the experimental setup, including datasets, model architectures, evaluation metrics, and implementation details. Section \ref{sec:Results} reports the results and provides a discussion of the proposed approach’s effects on calibration quality, uncertainty estimation, and operational efficiency. Finally, Section \ref{sec:Conclusion} summarizes the study and outlines directions for future research.

\section{Background} \label{sec:Background}
This section is organized into three parts: Section \ref{subsec:Backgroun_Calibration} reviews calibration methodologies in ML; Section \ref{subsec:Backgroun_UQ} examines uncertainty quantification techniques and their application to decision-making; and Section \ref{subsec:Backgroun_Gap} synthesizes the identified research gaps and introduces the proposed framework.

\subsection{Calibration} \label{subsec:Backgroun_Calibration}
Several Studies have demonstrated that ML models, especially neural networks, are prone to generate overly confident predictions, resulting in a discrepancy between predicted probabilities and actual outcomes \citep{gal2016dropout,guo2017calibration}. Thus, calibration methods have been introduced to refine confidence scores (predicted probabilities), aligning them more closely with ground truth. 

Calibration methods in ML can be broadly categorized into two groups. Recently emerged research direction which have gained prominence in recent years, comprises training-time techniques, incorporate calibration objectives during model optimization by modifying the loss function or network architecture, thereby enabling models to learn calibrated probability estimates as part of the training procedure. A straightforward method is label smoothing \citep{muller2019does}, which replaces hard labels (e.g., [0, 1, 0]) with softened distributions (e.g., [0.05, 0.9, 0.05]) to prevent excessive model confidence and discourage extreme logit values that lead to miscalibration. Focal loss, although originally designed to address class imbalance in object detection, has become one of the most widely recognized loss functions with secondary benefits for calibration. Its mechanism of down-weighting easily classified examples suppresses overconfident predictions, yielding improved probability calibration as an indirect effect \citep{mukhoti2020calibrating}. Inspired by the adoption of focal loss, numerous adaptations have been introduced to enhance its effectiveness. \cite{wang2021rethinking} developed Inverse Focal Loss that reverses the mechanism of standard focal loss by emphasizing underconfident but correct predictions instead of hard misclassified ones. \cite{tao2023dual} proposed Dual Focal Loss that integrates standard focal loss, which emphasizes hard-to-classify examples, with a reverse-focused term that promotes confident predictions for easy, correctly classified examples. \cite{ghosh2022adafocal} introduced Adaptive Focal Loss, which employs adaptive weighting based on the degree of miscalibration of each prediction—assigning higher penalties to overconfident predictions and lower penalties to well-calibrated ones.

In parallel to focal loss and its variants, an extensive line of research has revisited the cross-entropy formulation by embedding regularization mechanisms to improve calibration. \cite{kumar2018trainable} proposed the Maximum Mean Calibration Error (MMCE), a trainable calibration metric that directly optimizes calibration during neural network training by leveraging kernel mean embeddings in a reproducing kernel Hilbert space. \cite{karandikar2021soft} proposed the soft calibration objective as a differentiable loss function that regularizes neural networks to produce confidence scores aligned with true correctness likelihoods. It replaces discrete binning (as used in ECE) with soft binning via Gaussian kernels, enabling gradient-based optimization of calibration during training. \cite{bohdal2021meta} proposed meta-calibration that trains a secondary model (meta-model) to improve the calibration of predicted probabilities by learning from both the model’s confidence scores and other auxiliary features. While meta-calibration offers a more flexible and instance-aware approach, it does not directly optimize calibration as part of the primary model's training objective, which means the base model itself remains agnostic to calibration during its learning process. \cite{hui2020evaluation} used the Brier Score as a training loss function to encourage the model to produce well-calibrated probability estimates by minimizing the squared difference between predicted probabilities and true class labels. \cite{li2022ultra} employed Kullback-Leibler (KL) divergence to reduce the discrepancy between predicted and ground-truth label distributions. 

The second group comprises post-hoc calibration methods, which represent the earliest family of calibration approaches. Post-hoc calibration methods are applied after model training to refine predicted probabilities without modifying the underlying model parameters. One of the earliest post-hoc calibration techniques is Platt Scaling \citep{platt1999probabilistic}, which fits a logistic regression model to the output confidence scores of a classifier, converting them into calibrated probabilities. A more recent and commonly adopted technique is Temperature Scaling, which adjusts the logits produced by a neural network using a single temperature parameter optimized on a validation set \citep{guo2017calibration}. Variants of temperature scaling, such as parameterized temperature scaling, have also been developed to enhance its flexibility and performance \citep{tomani2022parameterized}. 

Another widely used post-hoc method is Isotonic Regression, a non-parametric approach that learns a piecewise constant function to map predicted scores to calibrated probabilities \cite{fawcett2007pav}. Isotonic regression has served as a cornerstone in post-hoc calibration due to its non-parametric nature and ability to enforce monotonicity between predicted scores and empirical correctness. Its effectiveness has inspired a series of methodological extensions and refinements in recent years. Classical isotonic regression produces stepwise, piecewise-constant calibration functions that often exhibit discontinuities. To address this limitation, several smoothing approaches have been proposed. \cite{jiang2011smooth} and \cite{huang2021calibrating} developed Smooth Isotonic Regression (SIR), which applies Piecewise Cubic Hermite Interpolating Polynomial (PCHIP) interpolation to isotonic regression outputs, yielding continuous monotone calibration curves with improved generalization. Complementarily, \cite{allikivi2019non} introduced Non-parametric Bayesian Isotonic Calibration, which embeds isotonic regression in a Bayesian framework by placing priors over piecewise linear isotonic maps and computing posterior mean calibration functions via Monte Carlo sampling, thereby enforcing smoothness while mitigating overconfidence at score extremes. Alternative extensions modify isotonic regression through regularization mechanisms. \cite{naeini2016binary} introduced Near-Isotonic Regression (ENIR), which relaxes the strict monotonicity constraint by allowing limited violations penalized via a regularization term. \cite{machado2024post} proposed a regularized isotonic calibration that adds a smoothness-inducing penalty to avoid overfitting and simultaneously constrains the isotonic mapping to preserve discrimination (AUC), striking a balance between calibration fidelity and ranking consistency. \cite{nyberg2021reliably} developed Reliably Calibrated Isotonic Regression (RCIR), which augments standard isotonic binning with Bayesian credible interval constraints to ensure statistical reliability in imbalanced settings through greedy bin-merging algorithms. Recent works have extended isotonic regression beyond binary settings. \cite{berta2024classifier}  generalized isotonic regression to multi-class calibration by introducing multidimensional ROC surfaces with novel ROC monotony constraints, implemented through recursive probability simplex splitting to guarantee calibration while controlling overfitting. \cite{aradimproving} proposed normalization-aware methods, including Normalization-Aware Flattened Isotonic Regression (NA-FIR) and Sorted Cumulative Isotonic Regression (SCIR), which incorporate probability simplex constraints directly into optimization, overcoming Category Independence assumptions inherent in one-vs-rest approaches. Beyond methodological advances, isotonic regression has been successfully adapted to specialized domains. \cite{fonseca2017calibration} embedded isotonic calibration within a temporal recalibration framework for probability-of-default modeling, demonstrating improved long-term stability under temporal drift compared with logistic and sigmoid calibration across multiple credit datasets. In notable research, \cite{pernot2023stratification} extended the use of isotonic regression to calibrate prediction variance in regression tasks. By mapping predicted variances to empirical squared errors through monotone transformations, the method improved alignment between predicted and observed variances.

These calibration methods enhance the alignment between predicted probabilities and observed outcomes; however, they raise an important question: \textit{Do improvements in calibration necessarily translate into more effective uncertainty-aware decision making?} While calibration ensures that predicted probabilities statistically correspond to empirical outcome frequencies at the population level, uncertainty quantification characterizes the reliability and variability of individual predictions. A key research gap, therefore, is that despite extensive efforts to improve calibration, the influence of calibration on the effectiveness of uncertainty-aware decision processes remains largely overlooked in existing studies.

\subsection{Uncertainty Quantification} \label{subsec:Backgroun_UQ}
In neural networks, uncertainty quantification requires addressing the fact that these models typically produce a deterministic output for a given input by optimizing network parameters $\omega$ through point estimation \citep{aseeri2021uncertainty}. However, this approach does not account for the variability in predictions, thereby limiting the model’s ability to represent uncertainty. Placing a probability distribution over the model parameters enables predictions to be represented probabilistically, facilitating uncertainty quantification. Methods like Bayesian inference—such as Markov chain Monte Carlo (MCMC) \citep{gamerman2006markov}, Variational Inference (VI) \citep{graves2011practical}, Monte Carlo Dropout (MCD) \citep{gal2016dropout}, Variational Autoencoders (VAE) \citep{kingma2013auto}, and Bayes By Backprop (BBB)\citep{fortunato2017bayesian}—are often used to estimate the posterior distribution of parameters. Additionally, ensemble learning is a popular approach for uncertainty quantification, where diverse models with varying architectures, parameters, or training data produce probabilistic outputs rather than single-point predictions \citep{gharoun2024trustinformed}. Numerous studies \citep{aseeri2021uncertainty,martin2024uncertainty,senousy2021mcua,carneiro2020deep,habibpour2021uncertainty,habibpour2023uncertainty,westermann2021using,yao2024uncertainty,aguilar2022uncertainty} have employed the aforementioned methods to estimate the uncertainty of ML models, particularly neural networks, to improve decision-making by effectively communicating prediction uncertainty. These approaches typically generate predictions accompanied by uncertainty estimates, often in the form of prediction entropy, allowing highly uncertain outputs to be flagged for human review. 

Building upon established uncertainty quantification methods, recent studies have shifted focus toward enhancing uncertainty-aware decision-making through direct improvements in model confidences. \cite{tabarisaadi2022optimized} introduced a loss function that simultaneously optimizes prediction accuracy and uncertainty estimates. This loss function combines cross-entropy with KL divergence to measure the separation between uncertainty densities for correct and incorrect predictions. However, despite comparing against uncertainty quantification methods like MC-Dropout and Bayesian approaches, their method only optimizes standard softmax confidence without implementing actual uncertainty estimation. Additionally, while the authors report improvements in Uncertainty Accuracy (UAcc), the absence of calibration metrics makes it impossible to determine whether this represents a genuine improvement in uncertainty quantification or merely increased overconfidence. The KL divergence loss could be pushing the model toward extreme confidence values, which would improve UAcc, while the impact on calibration is not studied. In notable work, \cite{shamsi2021uncertainty} proposes two alternative loss functions for training MC-Dropout networks: one combining cross-entropy with mean predictive entropy to increase uncertainty separation between correct and incorrect predictions, and another combining cross-entropy with ECE to directly optimize calibration during training. The evaluation is limited to binary classification on 2D synthetic datasets (two-moon and blobs) and UAcc, leaving unclear whether the approach scales to multiclass problems where calibration is significantly more challenging. A further concern lies in the interpretation of the UAcc metric. The UAcc metric is inherently sensitive to class imbalance between correctly and incorrectly predicted samples. Its improvement may arise either from a genuine reduction in confidently incorrect predictions—typically a minority subset—or from an inflated proportion of confidently correct predictions, which generally dominate the dataset. This distinction is critical, as confidently incorrect predictions pose a greater safety risk in uncertainty-aware decision systems. Without disaggregating these components, reported gains in UAcc cannot be conclusively attributed to improved uncertainty discrimination.

\subsection{Limitations of Prior Works and Proposed Advancements} \label{subsec:Backgroun_Gap}
The reviewed literature reveals two critical limitations in current calibration research. First, existing calibration methods—whether training-time or post-hoc—predominantly focus on improving population-level alignment between predicted probabilities and empirical frequencies, while largely overlooking their impact on uncertainty-aware decision-making. The sole exception is \cite{pernot2023stratification}, which adapted isotonic regression to align predicted variance with actual variance in regression tasks, demonstrating the potential for calibration techniques to enhance uncertainty quantification. Second, calibration methods universally apply global transformations that treat all predictions uniformly, failing to account for heterogeneous prediction reliability across instances. 

Concurrently, the uncertainty quantification literature exhibits two distinct streams: The first focuses on developing uncertainty quantification techniques to measure prediction uncertainty without explicitly improving downstream decision-making. The second, more recent trend, attempts to enhance uncertainty-aware decision processes but often remains constrained to binary tasks or lacks proper evaluation protocols.

To address these gaps, this study proposes an uncertainty-aware post-hoc calibration framework that introduces instance-level adaptivity through conformal prediction-based sample stratification. A held-out conformal set enables proximity-based conformal prediction to flag individual samples as putatively correct or putatively incorrect. Putatively correct samples undergo standard isotonic calibration to preserve confidence in reliable predictions, while putatively incorrect samples are calibrated toward uniform probability distributions, enabling their identification as uncertain during decision processes and subsequent deferral to human review. This dual-pathway approach bridges calibration and uncertainty quantification, offering the first post-hoc method that leverages prediction reliability assessment to simultaneously improve calibration quality and uncertainty-aware decision-making effectiveness.

\section{Methodology} \label{sec:Method}
This section presents the details of the proposed novel post-hoc calibration framework, in which the limitations of uniform calibration approaches are mitigated by leveraging conformal prediction to discriminate between predictions of varying reliability. Semantic proximity-based conformal prediction is employed to partition calibration samples into putatively correct and putatively incorrect groups. Separate isotonic regressors are then fitted on these groups, enabling differentiated calibration strategies that adjust predicted probabilities according to the underlying reliability of the samples. The following sections present the proposed method in detail.

\subsection{Preliminary}
Isotonic regression, a classical nonparametric estimator of monotonic functions \citep{robertson1988order}, was first leveraged for probability calibration in binary classification by \cite{zadrozny2002transforming}. For a set of predictions and corresponding ground-truth labels, isotonic regression finds a monotonically non-decreasing function that minimizes the mean squared error.

Given uncalibrated probabilities $p_1, p_2, \ldots, p_n$ and binary ground-truth $y_1, y_2, \ldots, y_n$, isotonic regression solves Eq.~\ref{Eq:isototnic_Regression}:

\begin{equation}
    \min_{f} \sum_{i=1}^{n} \bigl(y_i - f(p_i)\bigr)^2
    \label{Eq:isototnic_Regression}
\end{equation}

\noindent subject to the monotonicity constraint Eq.~\ref{Eq:monotonicity_cons}:

\begin{equation}
    p_i \leq p_j \;\;\Rightarrow\;\; f(p_i) \leq f(p_j)
    \label{Eq:monotonicity_cons}
\end{equation}

\noindent In the context of probability calibration, $f(p_i)$ correspond to calibrated predicted probabilities and $y_i$ represents the true outcome. The optimization problem can be formulated as finding values $r_1, r_2, \ldots, r_n$ such that:

\begin{equation}
    \min_{r_1, r_2, \ldots, r_n} \sum_{i=1}^{n} \bigl(y_i - r_i\bigr)^2
    \label{Eq:finding_calibrated_p}
\end{equation}

\noindent subject to the constraint $r_1\le r_2\le \ldots\le r_n$. This constrained optimization ensures that higher predicted probabilities correspond to higher calibrated probabilities, maintaining the discriminative ability of the original model.

The solution is obtained using the Pool Adjacent Violators Algorithm (PAVA) \citep{robertson1988order}. The algorithm operates by iteratively identifying adjacent pairs that violate the monotonicity constraint and merging them into blocks. When violations are detected, adjacent points are combined into a contiguous block B, and all points within the block are assigned the same calibrated value computed as the block average (Eq.~\ref{Eq:calibrated_values}):

\begin{equation}
    f(p_i) = \frac{1}{|B|}\sum_{i \in B} y_i
    \label{Eq:calibrated_values}
\end{equation}

\noindent This procedure continues iteratively until no monotonicity violations remain, guaranteeing convergence to the global optimum. The resulting calibration function is piecewise-constant, non-decreasing, and minimizes the squared deviation between predicted probabilities and observed outcomes.

For multi-class calibration scenarios, isotonic regression was applied using a one-versus-all approach. Each class c was treated as a binary classification problem where the positive class consists of samples with true label c, and the negative class comprises all remaining samples. This decomposition allows separate calibration functions to be learned for each class, enabling the method to handle complex multi-class probability distributions.

\subsection{Dual Isotonic Calibration via Proximity-based Conformal Prediction}
\subsubsection{Problem Formulation}
Let $f_{\theta} : \mathcal{X} \rightarrow \Delta^{C}$ be a neural network with parameters $\theta$ that maps input space $\mathcal{X}$ to probability simplex $\Delta^{C}$ for $C$ classes. This study aims to learn a calibration function 
$g : \Delta^{C} \rightarrow \Delta^{C}$ such that the calibrated predictions 
$\bar{p}_i = g(f_{\theta}(x_i))$ exhibit improved uncertainty-aware decision making while maintaining predictive performance.

\subsubsection{Preprocessing} \label{sec:Preprocessing}
Let a dataset be denoted by $\mathcal{D} = \{(x_i, y_i)\}_{i=1}^N$, where $x_i \in \mathbb{R}^d$ is the input feature vector, $y_i \in \{1,2,\ldots,C\}$ is the classe labels, and $d$ is the input dimensionality and and $C$ is the number of classes.  The dataset is partitioned into four disjoint subsets: $\mathcal{D} = \mathcal{D}_{\text{train}} \;\cup\; \mathcal{D}_{\text{conf}} \;\cup\; \mathcal{D}_{\text{cal}} \;\cup\; \mathcal{D}_{\text{test}}$, corresponding to training, conformal, calibration, and testing sets.

\subsubsection{Model Training with Monte-Carlo Dropout}
A neural network $f_{\theta}$, parameterized by weights $\theta$, is trained on the training set $\mathcal{D}_{\text{train}}$. 
For a given input $x$, the ideal Bayesian posterior predictive distribution is defined as Eq.~\ref{Eq:Posterior_dist}:

\begin{equation}
    p(y=c \mid x, \mathcal{D}_{\text{train}}) = \int p(y \mid x, \theta) \, p(\theta \mid \mathcal{D}_{\text{train}}) \, d\theta
    \label{Eq:Posterior_dist}
\end{equation}

\noindent where $p(\theta \mid \mathcal{D}_{\text{train}})$ represents the posterior over the model parameters. This study employs the Monte-Carlo dropout (MCD) as an approximation. 
At inference, dropout is retained and $T$ stochastic forward passes are performed, producing a collection of predictive probability vectors, presented by Eq.~\ref{Eq:MCD}:

\begin{equation}
    \mathcal{P}(x) = \{\mathbf{p}_t(x) = f_{\theta_t}(x)\}_{t=1}^T, 
    \quad \mathbf{p}_t(x) \in \Delta^{C}
    \label{Eq:MCD}
\end{equation}

\noindent where $\Delta^{C}$ denotes the probability simplex over $C$ classes and $\theta_t$ corresponds to a subnetwork realization induced by dropout. 

The \textit{empirical mean} of these stochastic predictions provides an estimate of the posterior predictive distribution, calculated by Eq.~\ref{Eq:mean_pred_MCD}:

\begin{equation}
    \hat{\mathbf{p}}(x) = \frac{1}{T} \sum_{t=1}^T \mathbf{p}_t(x)
    \label{Eq:mean_pred_MCD}
\end{equation}

The corresponding \textit{prediction entropy} is computed as Eq.\ref{Eq:normalized_PE}, which serves as a uncertainty quantification measure for decision-making. The normalization factor $(1/\log C)$ scales the entropy to the range $[0, 1]$, with $0$ indicating complete certainty and $1$ representing maximum uncertainty.
\begin{equation}
    PE(x) = -\frac{1}{\log C} \sum_{c=1}^C \hat{p}_c(x) \log \hat{p}_c(x)
    \label{Eq:normalized_PE}
\end{equation}

\subsubsection{Pseudo-Correctness Stratification via Conformal Prediction}
Feature-space proximity-based conformal prediction was employed to identify putatively correct and putatively incorrect predictions. The term \textit{“putatively”} is used to indicate that predictions are assumed to be correct or incorrect based on conformity evidence, reflecting expected rather than guaranteed reliability. 

For the calibration set $D_{cal}$, samples with similar feature representations were identified by computing k-nearest neighbors using the conformal set $D_{conf}$ as the reference pool. This similarity search utilized the Facebook AI Similarity Search (FAISS) library to efficiently identify samples with minimal Euclidean distance in the feature representation space. Accordingly, the conformal set $\mathcal{D}_{\text{conf}}$ is indexed in a FAISS structure for efficient retrieval. For each calibration sample $x \in \mathcal{D}_{\text{cal}}$, the $k$ nearest neighbors $\mathcal{N}_k(x) = \{ (x_j, y_j) \}_{j=1}^k$ are retrieved from $\mathcal{D}_{\text{conf}}$. For each neighbor $(x_j, y_j) \in \mathcal{N}_k(x)$, a \textit{non-conformity score} is calculated using Eq.~\ref{Eq:probability-based-non-conformity}:

\begin{equation}
    \alpha_j = 1 - \hat{p}_{y_j}(x_j)
    \label{Eq:probability-based-non-conformity}
\end{equation}

\noindent $\hat{p}_{y_j}(x_j)$ is the mean predicted probability (via MC dropout) assigned by the model to the true class $y_j$ for neighbor $x_j$.

The conformal quantile is defined as Eq.~\ref{Eq:conformal_quantile}: 
\begin{equation}
    q_{1-\alpha}(x) = \text{Quantile}_{1-\alpha}\{\alpha_j : (x_j, y_j) \in \mathcal{N}_k(x)\}
    \label{Eq:conformal_quantile}
\end{equation}

\noindent where $q_{1-\alpha}(x)$ represents the $(1-\alpha)$-quantile of the non-conformity scores across the neighborhood of $x$, with $\alpha$ denoting the miscoverage rate.

The prediction set for $x$ is given by Eq.~\ref{Eq:prediction_set}:
\begin{equation}
    \Gamma(x) = \{c \in \{1, \ldots, C\} \;|\; 1 - \hat{p}_c(x) \leq q_{1-\alpha}(x)\}
    \label{Eq:prediction_set}
\end{equation}

\noindent where $\Gamma(x)$ represent the conformal prediction set containing the labels deemed plausible for $x$.

Finally, a singleton matching strategy was employed to stratify samples. A calibration sample was labeled putatively correct only when its conformal prediction set was a singleton ($|\Gamma(x)| = 1$) and this single class matched the neural network’s predicted label. All remaining samples, including those with multi-class conformal sets or mismatched singletons, were labeled putatively incorrect.

This process stratified the calibration set $D_{cal}$ into two distinct groups: putatively correct predictions and putatively incorrect predictions. The same conformal prediction procedure was subsequently applied to the test set $D_{test}$, assigning each test instance a putative correctness label for downstream calibration treatment.

\subsubsection{Dual Isotonic Calibration}
Instead of applying a single post-hoc calibrator to all samples, two calibrator types are trained and used conditionally:
\begin{itemize}
    \item a standard isotonic calibrator for putatively correct samples,
    \item an underconfidence-regularized isotonic calibrator for putatively incorrect samples.
\end{itemize}

\noindent Formally, for each class $c \in \{1, \ldots, C\}$:  

\begin{itemize}[label=--] 
    \item For putatively correct samples:
    \begin{equation}
        g_c^{\text{std}}(p) = \text{IsoReg}\left(\{(p_c(x), \mathbb{I}[y(x) = c]) : x \in \mathcal{D}_{\text{cal}}^{\text{correct}}\}\right)
        \label{Eq:correct_iso_reg}
    \end{equation}
    \item For putatively incorrect samples:
    \begin{equation}
        g_c^{\text{und}}(p) = \text{IsoReg}\left(\left\{(p_c(x), \, \beta \cdot p_c(x) + (1-\beta)\cdot \tfrac{1}{C}) : x \in \mathcal{D}_{\text{cal}}^{\text{incorrect}}\right\}\right)
        \label{Eq:incorrect_iso_reg}
    \end{equation}
\end{itemize}

\noindent where $IsoReg$ is class-wise isotonic regression, $\beta$ is the underconfidence factor, and $\tfrac{1}{C}$ represents the uniform probability for class $c$.

The underconfidence-regularized isotonic regression modifies training targets prior to fitting in order to systematically reduce overconfidence while preserving monotonicity. The transformation is defined as Eq:~\ref{Eq:underconfidence_logic}:

\begin{equation}
    \tilde{y}_{i,c} = \beta \cdot p_{i,c} + (1 - \beta) \cdot \tfrac{1}{C}
    \label{Eq:underconfidence_logic}
\end{equation}

\noindent where $p_{i,c}$ denotes the original predicted probability for class $c$ on sample $i$, $C$ is the total number of classes, and $\beta \in [0,1]$ is the underconfidence factor. This operation regularizes the targets by pulling them toward the uniform distribution $1/C$.

For potentially incorrect samples, the use of binary indicator targets $\mathbb{I}[y(x) = c]$ is problematic, since the model’s predicted label is likely to be erroneous. 
Fitting isotonic regression directly to such labels would drive the calibrator to assign high confidence to systematically incorrect predictions. To avoid this, the targets are regularized by incorporating a uniform mixture component $\tfrac{1}{C}$, which corresponds to maximum entropy (i.e., minimum information). This uniform prior serves as the most appropriate baseline when prediction quality is questionable, ensuring that calibration does not reinforce incorrect predictions.  

Moreover, the inclusion of this maximum-entropy reference makes such samples more likely 
to be detected during subsequent uncertainty-aware decision making.

Thus, For any probability $p_{i,c}$,when $\beta < 1$, the transformed target $\tilde{y}_{i,c}$ satisfies:
\begin{itemize}
    \item $\tilde{y}_{i,c} < p_{i,c}$ when $p_{i,c} > \tfrac{1}{C}$ (high-confidence cases),
    \item $\tilde{y}_{i,c} > p_{i,c}$ when $p_{i,c} < \tfrac{1}{C}$ (low-confidence cases).
\end{itemize}

Thus, confident predictions are downweighted, while low-confidence predictions are slightly elevated. Despite this modification, isotonic regression remains monotone. 

Since isotonic regression is applied independently to each class using the one-versus-all approach, the resulting calibrated probabilities across all classes may not sum to unity. This occurs because each class probability is transformed according to its own monotonic calibration function, which can distort the original normalization constraint that ensures probabilities sum to one. Therefore, after applying either standard or underconfidence-regularized isotonic regression, a renormalization step is done to restore the probability distribution property.

\subsubsection{Calibration at Inference Time}
At inference, each test sample is processed through a multi-stage pipeline. First, $T$ forward
passes with MCD generate stochastic predictions $\mathbf{p}_t(x)\,\}_{t=1}^{T}$,
which are averaged to obtain the mean predicted probabilities $\hat{\mathbf{p}}(x)$. Second, the sample’s features is used to retrieve $k$-nearest neighbors from the indexed conformal
set $\mathcal{D}_{\mathrm{conf}}$ via FAISS. These neighbors’ conformity scores determine the
sample-specific nonconformity threshold, from which a conformal prediction set
$\Gamma(x)$ is constructed. Third, the singleton-matching strategy classifies the sample as
putatively correct or putatively incorrect.
Finally, depending on this stratification, either standard isotonic regression (for putatively
correct samples) or underconfidence--regularized isotonic regression (for putatively incorrect samples) is applied to each class probability independently. The calibrated probabilities are then
renormalized to sum to one, yielding the final calibrated distribution $\bar{\mathbf{p}}(x)$. Predictive entropy is computed from $\bar{\mathbf{p}}(x)$ to enable uncertainty-aware decision making.


\section{Experiments} \label{sec:Experiments}

\subsection{Configuration}
This study employs transfer learning (TL) by using pre-trained models as feature extractors. These models, originally trained on large datasets like ImageNet, had their final fully connected layers removed while keeping the remaining layers frozen. This allows them to retain their learned representations and extract meaningful features from other datasets. This study utilized two pre-trained model: BigTransfer (BiT) \citep{kolesnikov2020big}, and CoAtNet \citep{dai2021coatnet}. The output feature vectors were standardized to 256 dimensions per image. Using different pre-trained models with diverse architectures allows to examine whether backbone variations influence the obtained results and impact confidence calibration.

\subsection{Datasets}
In this study, evaluation was conducted on three benchmark datasets: CIFAR-10 (10 classes), CIFAR-100 with coarse superclass labels (20 classes, denoted CIFAR-100-S), and CIFAR-100 with fine-grained class labels (100 classes, denoted CIFAR-100-F). Each dataset consists of 60,000 color images of size $32 \times 32$, originally partitioned into 50,000 training images and 10,000 test images. In this work, however, the original training and test splits were merged into a single dataset of 60,000 images, which was then randomly divided following the data partitioning protocol described in Section \ref{sec:Preprocessing}. Both CIFAR-10 and CIFAR-100 are publicly accessible through the TensorFlow Datasets library.

This dataset selection enables evaluation across varying classification complexities: CIFAR-10 provides a baseline for general object recognition where model confidence is typically high, CIFAR-100-S examines intermediate-difficulty coarse-grained categorization, and CIFAR-100-F tests the framework under challenging fine-grained classification, where conformal stratification accuracy becomes critical.

\subsection{Evaluation}
The evaluation framework employed in this study comprises two complementary components to provide a comprehensive assessment of the proposed dual calibration method.

\begin{itemize}
    \item Standard Performance and Calibration Metrics: Traditional model evaluation is conducted using established metrics, including F1-score, accuracy, and Area Under the Curve (AUC), to assess predictive performance. Additionally, calibration quality is evaluated through ECE, MCE, as introduced in the section \ref{sec:Intro}, and Brier Score to quantify the alignment between predicted probabilities and actual outcomes. The Brier Score measures the mean squared difference between predicted probabilities and the actual outcomes, providing a comprehensive assessment of both calibration and sharpness. For a multi-class classification problem, the Brier Score is calculated as Eq.~\ref{Eq:BS}:
    \begin{equation}
        \text{BS} = \frac{1}{N} \sum_{i=1}^{N} \sum_{c=1}^{C} (p_{i,c} - y_{i,c})^2\\
        \label{Eq:BS}
    \end{equation}
    where $N$ is the number of predictions, $p_{i,c}$ is the predicted probability for the class $c$ for sample $i$, and $y_{i,c}$ is the one-hot encoded true label (1 if sample $i$ belongs to class $c$, 0 otherwise). Lower Brier Scores indicate better overall prediction quality, as the metric simultaneously penalizes both poor calibration and poor discrimination \citep{gharoun2025leveraging}.

    \item Uncertainty-Aware Evaluation Metrics: Since this study aims to analyze the impact of the proposed method on predictive uncertainty estimates, a specialized evaluation approach is adopted utilizing a modified confusion matrix designed for uncertainty measurement, as introduced by \cite{asgharnezhad2022objective}. This framework applies an entropy threshold to categorize predictions as either certain or uncertain, resulting in four distinct outcome categories: True Certainty (TC), where the prediction is both correct and certain; True Uncertainty (TU), where the prediction is incorrect and appropriately flagged as uncertain; False Uncertainty (FU), where the prediction is correct but incorrectly classified as uncertain; and False Certainty (FC), where the prediction is incorrect yet classified as certain.

    \begin{figure}[H]
          \centering
          \captionsetup[subfloat]{font=tiny}
          {\includegraphics[width=0.65\columnwidth]{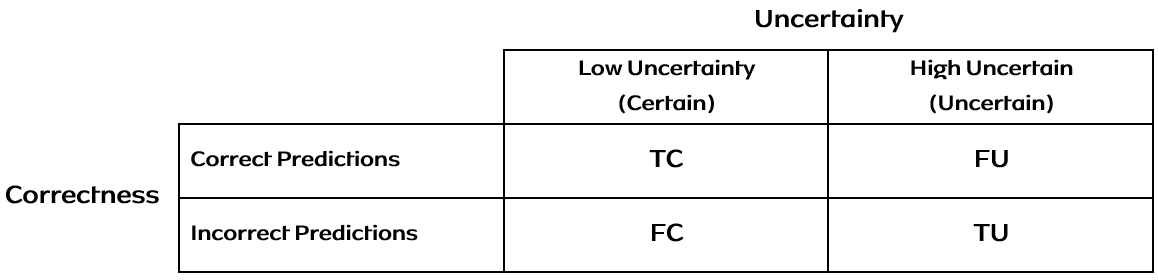}}\par
          \caption{Uncertainty-Aware Outcome Categorization}
          \label{fig:U_conf_mat}
    \end{figure}
    
    Slicing predictions into the four-category space of TC, TU, FU, and FC, as illustrated by Figure~\ref{fig:U_conf_mat}, enables the calculation of performance metrics analogous to traditional confusion matrix metrics, but specifically designed for uncertainty evaluation. In this study, the following uncertainty-aware metrics are used:
    \begin{itemize}
        \item Uncertainty Accuracy (UAcc): measures the overall model accuracy in uncertainty-aware decision-making, representing the proportion of predictions that are correctly classified in terms of both their correctness and uncertainty status, calculated by Eq.~\ref{Eq:UACC}.
        \begin{equation}
            UAcc= \frac{TU+TC}{TU+TC+FU+FC}
        \label{Eq:UACC}
        \end{equation}
        \item Uncertainty True Positive Rate (UTPR): shows the performance of the model in correctly identifying certain predictions among all predictions that should be certain (i.e., correct predictions), computed by Eq.~\ref{Eq:UTPR}. A higher UTPR indicates better ability to confidently classify correct predictions, thereby reducing unnecessary review burden.
        \begin{equation}
            UTPR= \frac{TC}{TC+FU}
        \label{Eq:UTPR}
        \end{equation}
        \item Uncertainty False Positive Rate (UFPR): quantifies the rate at which incorrect predictions are misclassified as certain among all predictions that should be uncertain (i.e., incorrect predictions), formulated by Eq.~\ref{Eq:UFPR}. A lower UFPR indicates better safety, as fewer incorrect predictions are inappropriately flagged as reliable.
        \begin{equation}
            UFPR= \frac{FC}{FC+TU}
        \label{Eq:UFPR}
        \end{equation}
        \item Uncertainty G-Mean: provides a balanced measure that simultaneously considers both the model's ability to confidently handle correct predictions (UTPR) and its capacity to appropriately flag incorrect predictions as uncertain (Uncertainty Specificity), calculated by Eq.~\ref{Eq:UGMEAN}. Uncertainty Specificity can be obtained by $1-\text{UFPR}$. This metric is particularly valuable for interpreting results as it captures the trade-off between operational efficiency and safety in uncertainty-aware systems, with higher values indicating superior overall uncertainty quantification performance.
        \begin{equation}
            UG-Mean= \sqrt{\text{UTPR} \times (1 - \text{UFPR})}
        \label{Eq:UGMEAN}
        \end{equation}
    \end{itemize}

\end{itemize}

\section{Results and Discussion} \label{sec:Results}
This section presents the empirical evaluation of the proposed calibration method compared against standard isotonic regression calibration.

In this study, Keras Tuner with Bayesian optimization was employed to optimize neural network architectures. The Bayesian search algorithm explored a predefined hyperparameter space to identify configurations that maximize validation accuracy through efficient sampling of the parameter space. The evaluation was conducted across 30 independent runs to ensure statistical robustness. For each run involving post-hoc calibration methods, the dataset was randomly partitioned into training (55\%), conformal (15\%), calibration (15\%), and test (15\%) subsets. For the proposed dual calibration method, two key hyperparameters were fixed: the neighborhood size for proximity-based conformal prediction was set to $K=20$, and the significance level for conformal prediction was set to $\alpha=0.01$. The underconfidence ratio $\beta$ was optimized through random search to identify the configuration yielding the best uncertainty-aware performance. Importantly, both standard isotonic regression and the proposed method were applied to identical non-calibrated predictions from the same base neural network, ensuring fair comparison.

For a comprehensive evaluation, focal loss was included as a training-time calibration baseline. Since focal loss modifies the training objective rather than applying post-hoc calibration, a separate hyperparameter optimization was performed using focal loss. For focal loss experiments, only training/test splits were required, maintaining the same 30-run evaluation protocol.

Table \ref{tab:accuracy_result} summarizes the best performance achieved, reporting the highest F1-score obtained after fine-tuning.

\begin{sidewaystable}[!htbp]
\centering
\caption{Summary of performance of various calibration methods across various pre-trained models}
\label{tab:accuracy_result}
\begin{tabular}{lllllllll}
\hline
\multicolumn{1}{c}{Dataset} & \multicolumn{1}{c}{Calibration} & \multicolumn{1}{c}{Backbone} & \multicolumn{1}{c}{F1 Score} & \multicolumn{1}{c}{Accuracy} & \multicolumn{1}{c}{AUC} & \multicolumn{1}{c}{ECE} & \multicolumn{1}{c}{MCE} & \multicolumn{1}{c}{Brier Score} \\ \hline
CIFAR-10 & Non Cal & BiT & 91.236 ± 0.208 & 91.254 ± 0.202 & 95.141 ± 0.112 & 6.563 ± 0.237 & 21.626 ± 3.439 & 14.736 ± 0.187 \\
CIFAR-10 & Focal loss & BiT & 90.588 ± 0.374 & 90.608 ± 0.372 & 94.782 ± 0.207 & 0.611 ± 0.274 & 17.802 ± 14.687 & 13.855 ± 0.499 \\
CIFAR-10 & Iso Cal & BiT & 91.28 ± 0.166 & 91.295 ± 0.165 & 95.164 ± 0.092 & 0.532 ± 0.139 & 12.440 ± 8.357 & 12.911 ± 0.15 \\
CIFAR-10 & Dual Cal (0.65) & BiT & 91.624 ± 0.116 & 91.622 ± 0.118 & 95.345 ± 0.065 & 3.591 ± 0.689 & 27.316 ± 2.324 & 13.872 ± 0.302 \\
CIFAR-100-S & Non Cal & BiT & 78.728 ± 0.243 & 78.810 ± 0.238 & 88.847 ± 0.125 & 9.717 ± 0.797 & 25.382 ± 21.679 & 31.848 ± 0.188 \\
CIFAR-100-S & Focal loss & BiT & 75.651 ± 0.436 & 75.645 ± 0.402 & 87.182 ± 0.211 & 3.569 ± 0.689 & 11.879 ± 14.558 & 34.443 ± 0.528 \\
CIFAR-100-S & Iso Cal & BiT & 78.758 ± 0.231 & 78.843 ± 0.225 & 88.864 ± 0.118 & 1.963 ± 0.146 & 7.858 ± 3.024 & 30.000 ± 0.221 \\
CIFAR-100-S & Dual Cal (0.9) & BiT & 79.520 ± 0.236 & 79.619 ± 0.230 & 89.273 ± 0.121 & 6.458 ± 0.662 & 21.739 ± 18.436 & 29.845 ± 0.238 \\
CIFAR-100-F & Non Cal & BiT & 67.787 ± 0.366 & 68.109 ± 0.356 & 83.893 ± 0.180 & 15.952 ± 0.484 & 23.432 ± 1.522 & 46.124 ± 0.371 \\
CIFAR-100-F & Focal loss & BiT & 52.668 ± 0.614 & 53.357 ± 0.565 & 76.443 ± 0.285 & 6.606 ± 0.685 & 12.016 ± 1.378 & 60.801 ± 0.509 \\
CIFAR-100-F & Iso Cal & BiT & 68.130 ± 0.282 & 68.232 ± 0.264 & 83.955 ± 0.133 & 3.888 ± 0.394 & 10.028 ± 2.740 & 42.505 ± 0.248 \\
CIFAR-100-F & Dual Cal (1) & BiT & 68.705 ± 0.369 & 69.023 ± 0.361 & 84.355 ± 0.182 & 9.863 ± 0.497 & 17.602 ± 1.246 & 43.420 ± 0.329 \\
CIFAR-10 & Non Cal & CoAtNet & 93.129 ± 0.155 & 93.141 ± 0.154 & 96.189 ± 0.085 & 2.795 ± 0.145 & 25.738 ± 16.779 & 10.921 ± 0.098 \\
CIFAR-10 & Focal loss & CoAtNet & 92.370 ± 0.152 & 92.374 ± 0.152 & 95.763 ± 0.084 & 2.258 ± 0.141 & 17.058 ± 3.211 & 11.918 ± 0.138 \\
CIFAR-10 & Iso Cal & CoAtNet & 93.096 ± 0.164 & 93.103 ± 0.163 & 96.168 ± 0.090 & 0.174 ± 0.072 & 12.869 ± 8.214 & 10.260 ± 0.130 \\
CIFAR-10 & Dual Cal (0.6) & CoAtNet & 93.185 ± 0.147 & 93.194 ± 0.146 & 96.219 ± 0.081 & 0.363 ± 0.170 & 25.664 ± 20.497 & 10.313 ± 0.155 \\
CIFAR-100-S & Non Cal & CoAtNet & 80.174 ± 0.357 & 80.215 ± 0.344 & 89.586 ± 0.181 & 23.202 ± 0.593 & 34.669 ± 7.063 & 35.353 ± 0.540 \\
CIFAR-100-S & Focal loss & CoAtNet & 72.602 ± 0.569 & 72.668 ± 0.557 & 85.614 ± 0.293 & 8.754 ± 0.352 & 18.354 ± 13.359 & 40.102 ± 0.639 \\
CIFAR-100-S & Iso Cal & CoAtNet & 80.216 ± 0.355 & 80.250 ± 0.349 & 89.605 ± 0.183 & 2.176 ± 0.213 & 9.506 ± 3.752 & 28.405 ± 0.416 \\
CIFAR-100-S & Dual Cal (0.95) & CoAtNet & 80.377 ± 0.337 & 80.423 ± 0.331 & 89.696 ± 0.174 & 6.727 ± 1.718 & 25.074 ± 7.201 & 30.300 ± 0.748 \\
CIFAR-100-F & Non Cal & CoAtNet & 71.388 ± 0.314 & 71.57 ± 0.300 & 85.642 ± 0.152 & 13.824 ± 0.355 & 21.011 ± 0.981 & 41.229 ± 0.213 \\
CIFAR-100-F & Focal loss & CoAtNet & 64.737 ± 0.242 & 65.102 ± 0.239 & 82.374 ± 0.121 & 14.907 ± 0.306 & 21.572 ± 0.730 & 49.735 ± 0.266 \\
CIFAR-100-F & Iso Cal & CoAtNet & 71.447 ± 0.293 & 71.622 ± 0.289 & 85.668 ± 0.1463 & 2.773 ± 0.237 & 9.500 ± 3.044 & 38.428 ± 0.256 \\
CIFAR-100-F & Dual Cal (0.9) & CoAtNet & 70.680 ± 0.389 & 70.896 ± 0.408 & 85.301 ± 0.206 & 5.080 ± 2.099 & 12.530 ± 2.640 & 40.637 ± 0.410 \\ \hline
\end{tabular}%
\footnotetext{Non Cal: Non calibrated, Iso Cal: Standard Isotonic Regression Calibration, Dual Cal ($\beta$): Proposed Dual Isotonic Regression Calibration (under-confidence ratio)}
\end{sidewaystable}

As post-hoc calibration techniques operate exclusively on the predicted probability distributions, they are not expected to alter the underlying classification outcomes derived from the model. To verify this, performance metrics were reported after applying each calibration method. Across all datasets and backbone architectures, the results presented in Table \ref{tab:accuracy_result} confirm that classification performance remained stable following calibration. The observed variations in accuracy, F1 score, and AUC across methods were minimal within the range of expected stochastic fluctuations and did not reflect systematic gains or losses. The primary distinction between methods lies in their calibration quality, as measured by ECE, MCE, and Brier score. 

A direct comparison between the proposed dual calibration method and standard isotonic regression reveals expected trade-offs in calibration quality. As expected, standard isotonic regression achieved lower ECE, MCE, and Brier scores than the proposed approach, as it directly optimizes global calibration. In contrast, the proposed dual calibration method intentionally worsens calibration metrics by design, since putative incorrect predictions are proactively pushed toward uncertainty.

A comparison between the proposed dual calibration method and focal loss reveals a contrasting pattern in calibration performance across model architectures. On the BiT backbone, focal loss consistently achieved superior calibration metrics, yielding lower ECE and Brier scores than dual calibration across all datasets. However, this trend was reversed when evaluated on the CoAtNet backbone. In this setting, focal loss calibration performance degraded noticeably, while the dual calibration method maintained stable or improved calibration outcomes across all datasets.

These results highlight a key limitation of training-time calibration methods. By reweighting the cross-entropy loss based on confidence, focal loss alters the optimization dynamics, making its effectiveness highly dependent on model architecture and feature representation. The dramatic performance difference between BiT (ECE: $6.606  \pm  0.685$) and CoAtNet (ECE: $14.907  \pm  0.306$) on CIFAR-100-F exemplifies how architectural variations can fundamentally disrupt training-time calibration strategies. Most strikingly, focal loss achieved worse calibration than the non-calibrated baseline on CIFAR-100-F. In contrast, post-hoc calibration methods, including the proposed dual isotonic calibration, operate on the fixed output probability distributions after training completion. While different architectures produce different probability distributions that serve as inputs to post-hoc calibration, the calibration mechanism itself remains independent of the model's internal structure, gradient dynamics, or training peculiarities. 

Although the proposed dual calibration method exhibits ECE fluctuations across configurations, this variability primarily stems from the quality of stratification into putatively correct and incorrect samples and the deliberate miscalibration applied to the putatively incorrect group rather than architectural brittleness. The intentional underconfidence injection for this subset necessarily increases overall ECE relative to standard isotonic regression. Critically, despite this designed trade-off, the dual calibration method consistently achieved lower ECE than focal loss across CoAtNet architectures, suggesting that controlled, targeted miscalibration for safety purposes outperforms architecturally-sensitive training-time approaches that can unpredictably degrade global calibration quality. 

To assess whether differences in ECE between the proposed dual calibration method and focal loss were statistically significant, a Friedman test was conducted for each dataset–backbone pair. The Friedman test is a non-parametric alternative to repeated-measures ANOVA, suitable for comparing multiple methods over repeated runs. For each group, ECEs were ranked per run, and the test determined whether these rankings differed significantly. As shown in Table~\ref{tab:ece_stat_test}, the results confirmed significant differences among the methods across all datasets ($p<10^{-17}$), thereby justifying subsequent pairwise comparisons. Post-hoc pairwise comparisons were performed between the proposed method and other calibration methods as baselines (focal loss and standard isotonic regression) using the one-sided Wilcoxon signed-rank test, with Holm correction applied to control for multiple comparisons. In addition to p-values, effect sizes were reported using Cliff’s Delta, and median differences were computed to support interpretability. The results are presented in Table~\ref{tab:ece_Wilcoxon}. Of particular interest, within the CoAtNet architecture, proposed dual Calibration significantly outperformed focal loss across all three datasets (shown in bold values in Table \ref{tab:ece_Wilcoxon}, $p < 0.05$ after Holm correction).

\begin{table}[!htbp]
\centering
\caption{Friedman test results for ECE across all calibration methods and datasets.}
\label{tab:ece_stat_test}
\begin{tabular}{llllllll}
\hline
\multicolumn{1}{c}{} & \multicolumn{1}{c}{} & \multicolumn{2}{c}{Friedman} & \multicolumn{4}{c}{Average Rank} \\  \hline
\multicolumn{1}{c}{Dataset} & \multicolumn{1}{c}{Backbone} & \multicolumn{1}{c}{chi2} & \multicolumn{1}{c}{p-value} & \multicolumn{1}{c}{Non Cal} & \multicolumn{1}{c}{Focal loss} & \multicolumn{1}{c}{Iso Cal} & \multicolumn{1}{c}{Dual Cal} \\ 
CIFAR-10 & BiT & 81.36 & $1.57 \times 10^{-17}$ & 4 & 1.6 & 1.4 & 3 \\
CIFAR-100-F & BiT & 90 & $2.19 \times 10^{-19}$ & 4 & 2 & 1 & 3 \\
CIFAR-100-S & BiT & 90 & $2.19 \times 10^{-19}$ & 4 & 2 & 1 & 3 \\
CIFAR-10 & CoAtNet & 85.84 & $1.71 \times 10^{-18}$ & 4 & 3 & 1.14 & 1.87 \\
CIFAR-100-F & CoAtNet & 88.84 & $3.89 \times 10^{-19}$ & 3 & 4 & 1.04 & 1.97 \\
CIFAR-100-S & CoAtNet & 86.76 & $1.09 \times 10^{-18}$ & 4 & 2.9 & 1 & 2.1 \\ \hline
\end{tabular}
\end{table}

\begin{table}[!htbp]
\centering
\caption{Wilcoxon signed-rank test results comparing the proposed Dual Calibration method to baseline methods for ECE. Median differences are computed as (Proposed - Baseline). Negative values indicate lower calibration error for the proposed method. Holm-corrected p-values are reported to control for multiple comparisons, with significant results at $\alpha=0.05$.}
\label{tab:ece_Wilcoxon}
\begin{tabular}{llllllll}
\hline
\multicolumn{1}{c}{Dataset} & \multicolumn{1}{c}{Backbone} & \multicolumn{1}{c}{Baseline} & \multicolumn{1}{c}{\begin{tabular}[c]{@{}c@{}}Wilcoxon \\ Statistic\end{tabular}} & \multicolumn{1}{c}{Raw p-value} & \multicolumn{1}{c}{\begin{tabular}[c]{@{}c@{}}Median\\ Diff\end{tabular}} & \multicolumn{1}{c}{\begin{tabular}[c]{@{}c@{}}Cliff \\ Delta\end{tabular}} & \multicolumn{1}{c}{\begin{tabular}[c]{@{}c@{}}Holm-corrected\\ p-value\end{tabular}} \\ \hline
CIFAR10 & BiT & Focal loss & 465 & 1 & 0.02909 & 1 & 1 \\
CIFAR-100-S & BiT & Focal loss & 465 & 1 & 0.02667 & 1 & 1 \\
CIFAR100-F & BiT & Focal loss & 465 & 1 & 0.03448 & 1 & 1 \\
\textbf{CIFAR10} & \textbf{CoAtNet} & \textbf{Focal loss} & 0 & $9.31 \times 10^{-10}$ & \textbf{-0.0188} & \textbf{-1} & \textbf{$2.79 \times 10^{-9}$} \\
\textbf{CIFAR-100-S} & \textbf{CoAtNet} & \textbf{Focal loss} & 25 & $8.42 \times 10^{-7}$ & \textbf{-0.0235} & \textbf{-0.8111} & \textbf{$1.68 \times 10^{-6}$} \\
\textbf{CIFAR100-F} & \textbf{CoAtNet} & \textbf{Focal loss} & 0 & $9.31 \times 10^{-10}$ & \textbf{-0.1040} & \textbf{-1} & \textbf{$2.79 \times 10^{-9}$} \\
CIFAR10 & BiT & Iso Cal & 465 & 1 & 0.02954 & 1 & 1 \\
CIFAR-100-S & BiT & Iso Cal & 465 & 1 & 0.04514 & 1 & 1 \\
CIFAR100-F & BiT & Iso Cal & 465 & 1 & 0.05981 & 1 & 1 \\
CIFAR10 & CoAtNet & Iso Cal & 437 & 0.9999 & 0.00158 & 0.7044 & 0.9999 \\
CIFAR-100-S & CoAtNet & Iso Cal & 465 & 1 & 0.04406 & 1 & 1 \\
CIFAR100-F & CoAtNet & Iso Cal & 464 & 0.9999 & 0.01687 & 0.9088 & 0.9999 \\
CIFAR10 & BiT & Non Cal & 0 & $9.31 \times 10^{-10}$ & -0.0314 & -1 & $2.79 \times 10^{-9}$ \\
CIFAR-100-S & BiT & Non Cal & 0 & $9.31 \times 10^{-10}$ & -0.0295 & -1 & $2.79 \times 10^{-9}$ \\
CIFAR100-F & BiT & Non Cal & 0 & $9.31 \times 10^{-10}$ & -0.0601 & -1 & $2.79 \times 10^{-9}$ \\
CIFAR10 & CoAtNet & Non Cal & 0 & $9.31 \times 10^{-10}$ & -0.0243 & -1 & $2.79 \times 10^{-9}$ \\
CIFAR-100-S & CoAtNet & Non Cal & 0 & $9.31 \times 10^{-10}$ & -0.1665 & -1 & $2.79 \times 10^{-9}$ \\
CIFAR100-F & CoAtNet & Non Cal & 0 & $9.31 \times 10^{-10}$ & -0.0937 & -1 & $2.79 \times 10^{-9}$ \\ \hline
\end{tabular}
\end{table}

\begin{figure}[!htbp]
  \centering
  \captionsetup[subfloat]{font=tiny}
  \subfloat[Overall - Non Cal]{\includegraphics[width=0.32\textwidth]{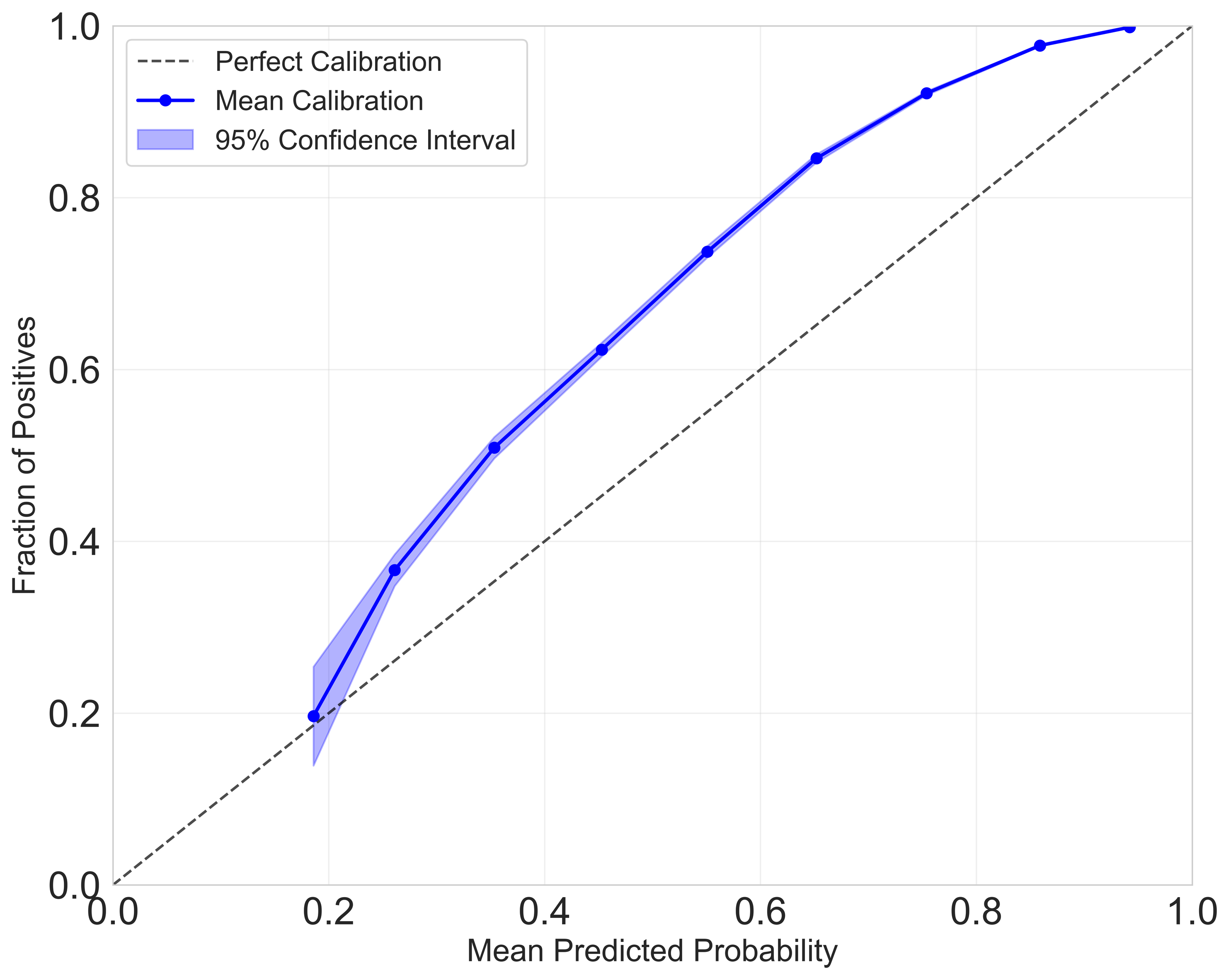}}
  \subfloat[Overall - Iso Cal]{\includegraphics[width=0.32\textwidth]{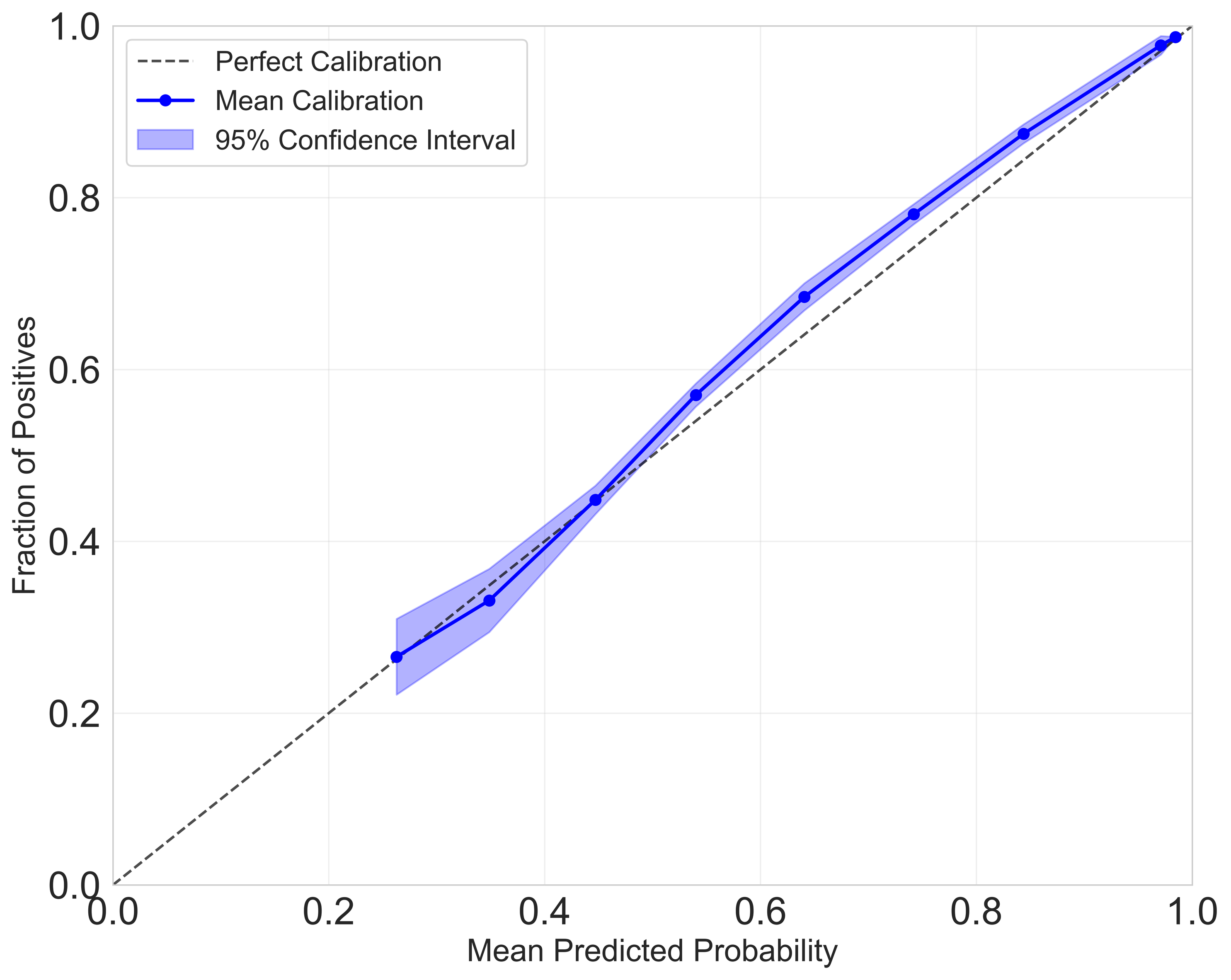}}
  \subfloat[Overall - Dual Cal]{\includegraphics[width=0.32\textwidth]{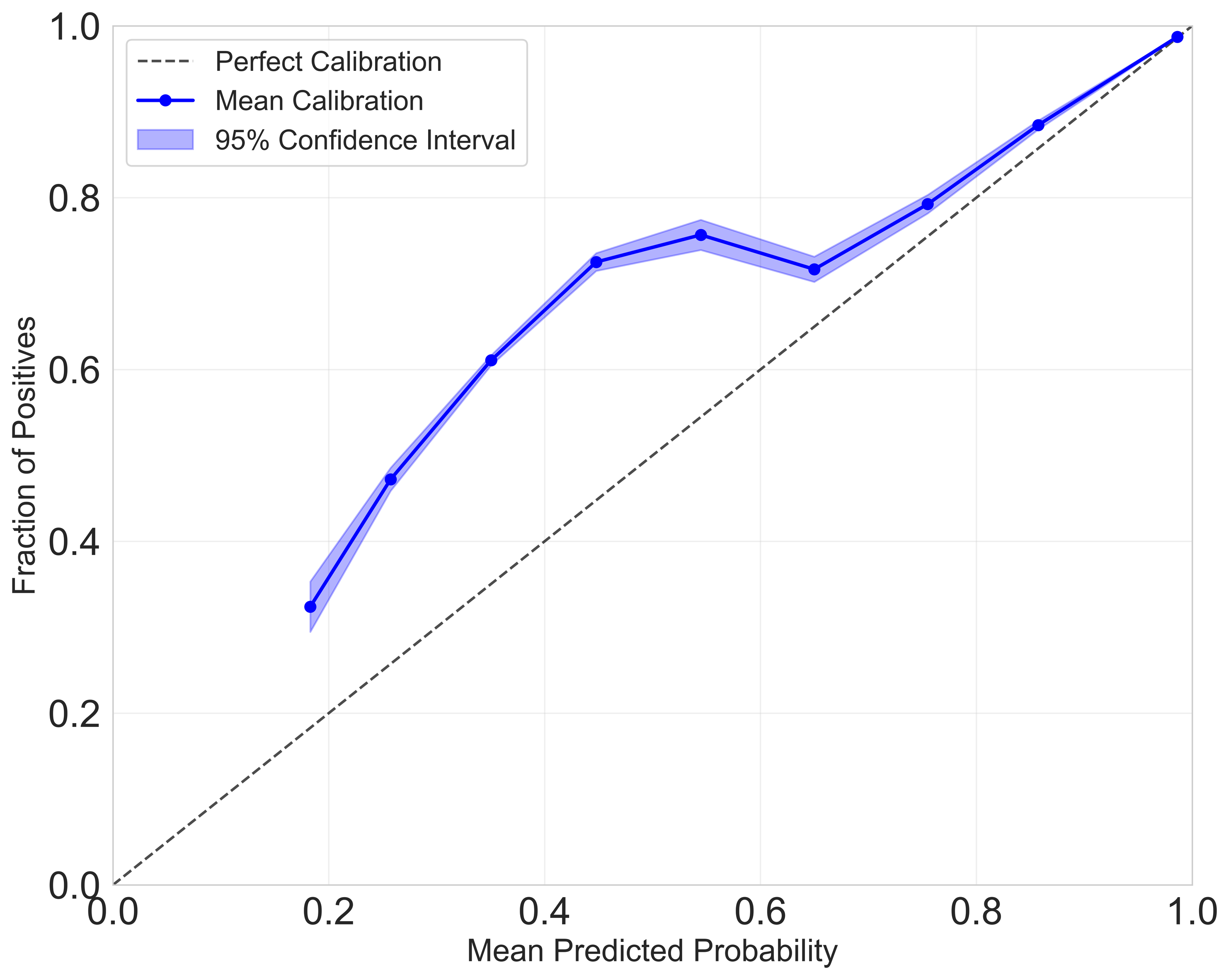}}\par
  \subfloat[Putatively Correct - Non Cal]{\includegraphics[width=0.32\textwidth]{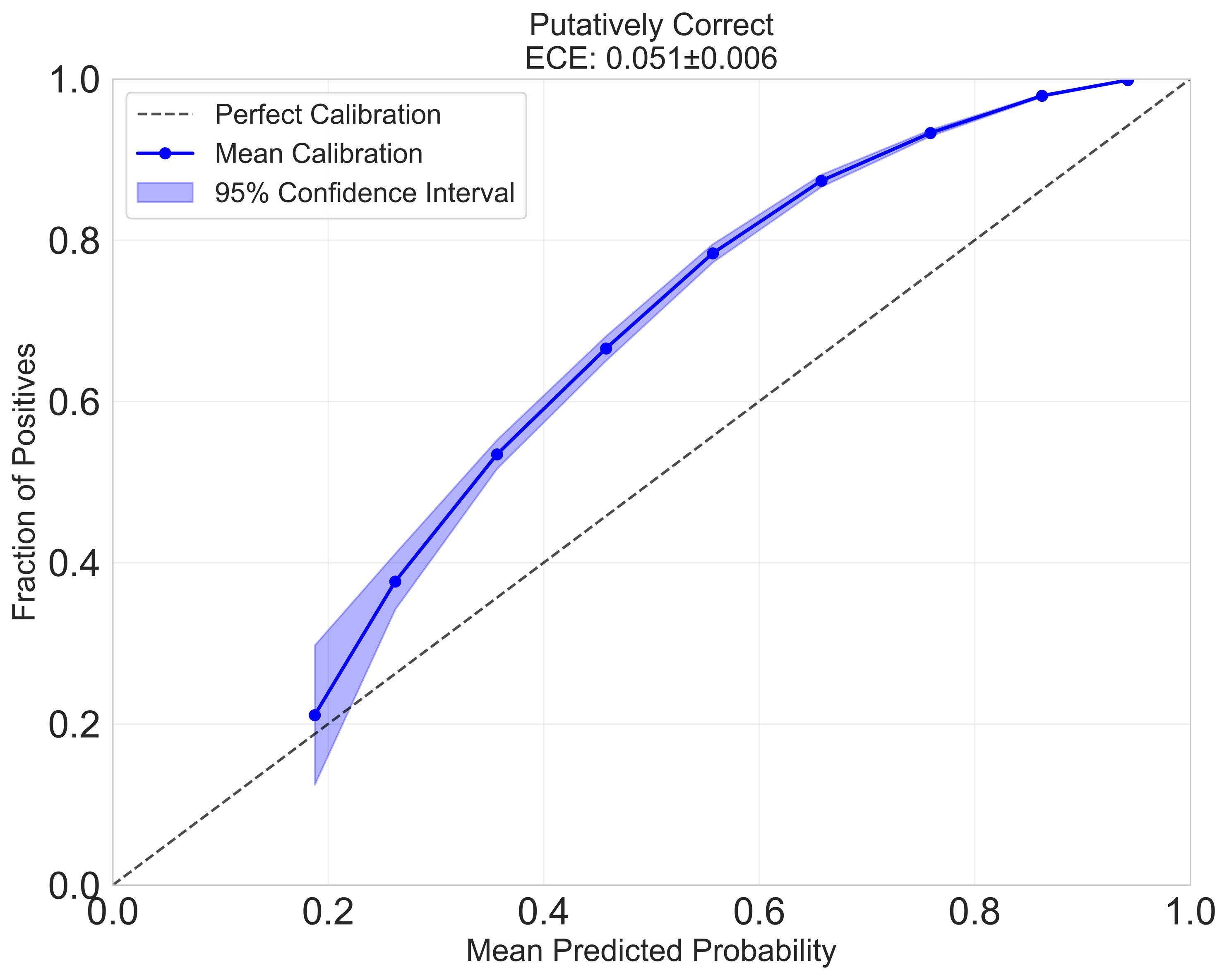}}
  \subfloat[Putatively Correct - Iso Cal]{\includegraphics[width=0.32\textwidth]{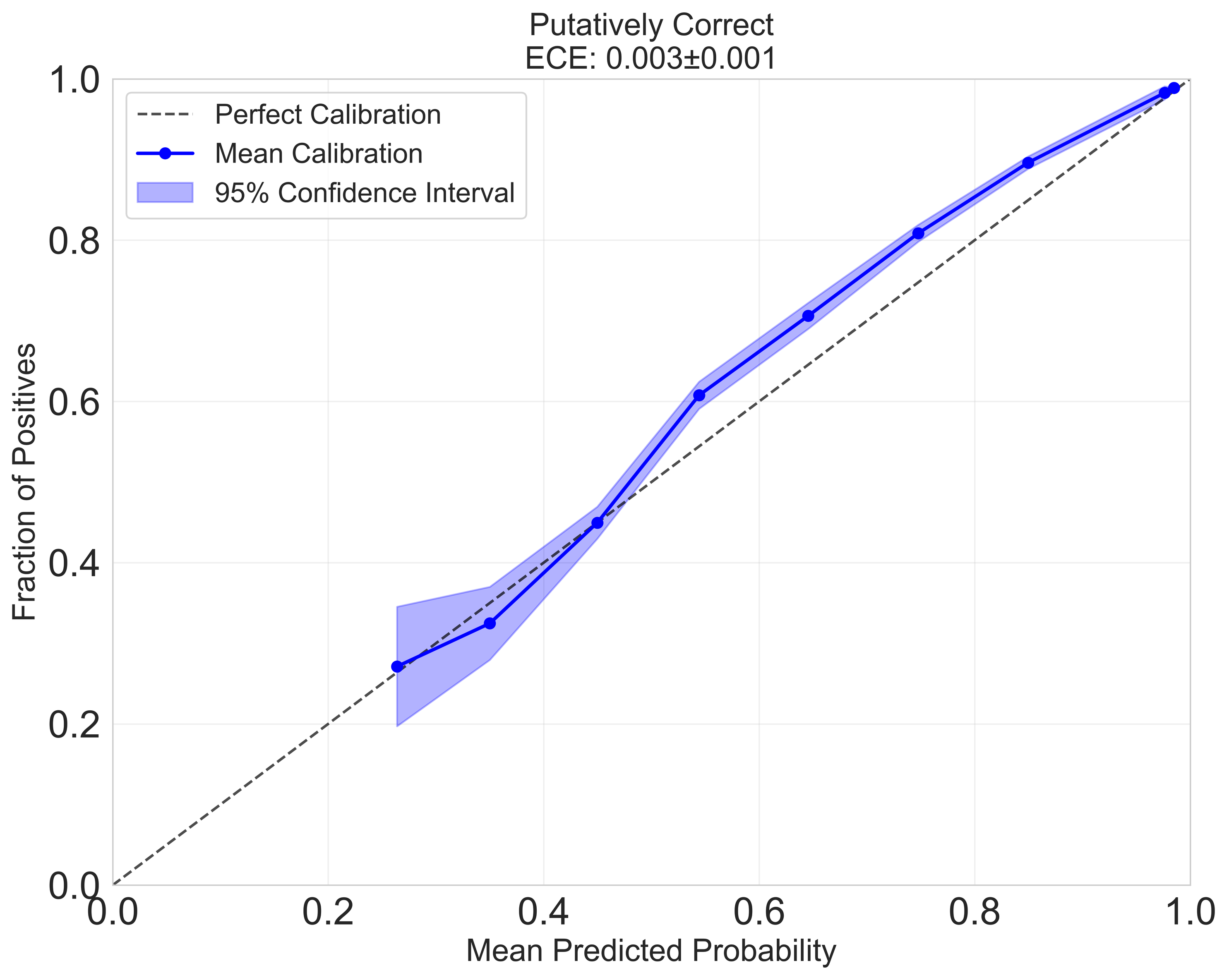}}
  \subfloat[Putatively Correct - Dual Cal]{\includegraphics[width=0.32\textwidth]{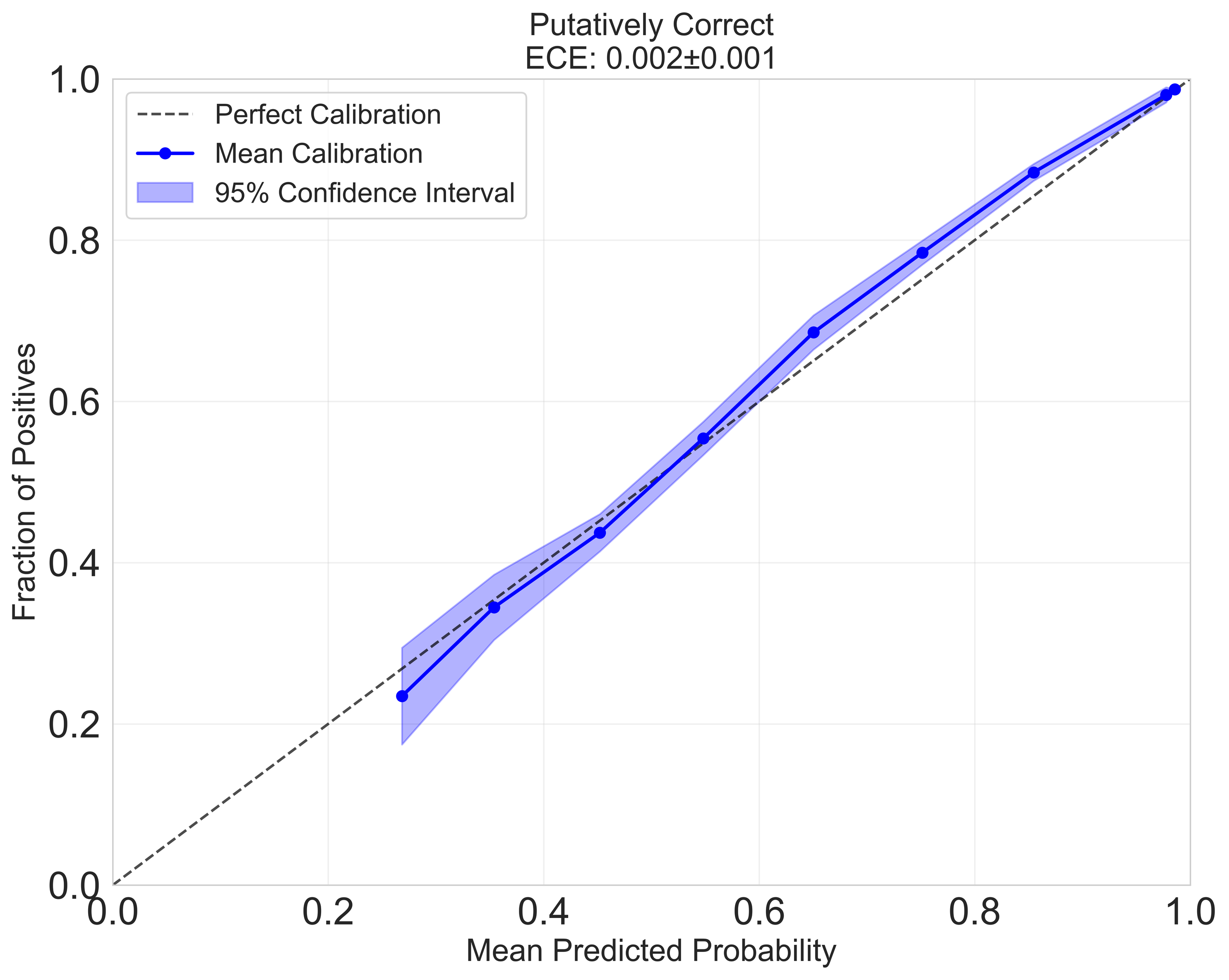}}\par
  \subfloat[Putatively Incorrect - Non Cal]{\includegraphics[width=0.32\textwidth]{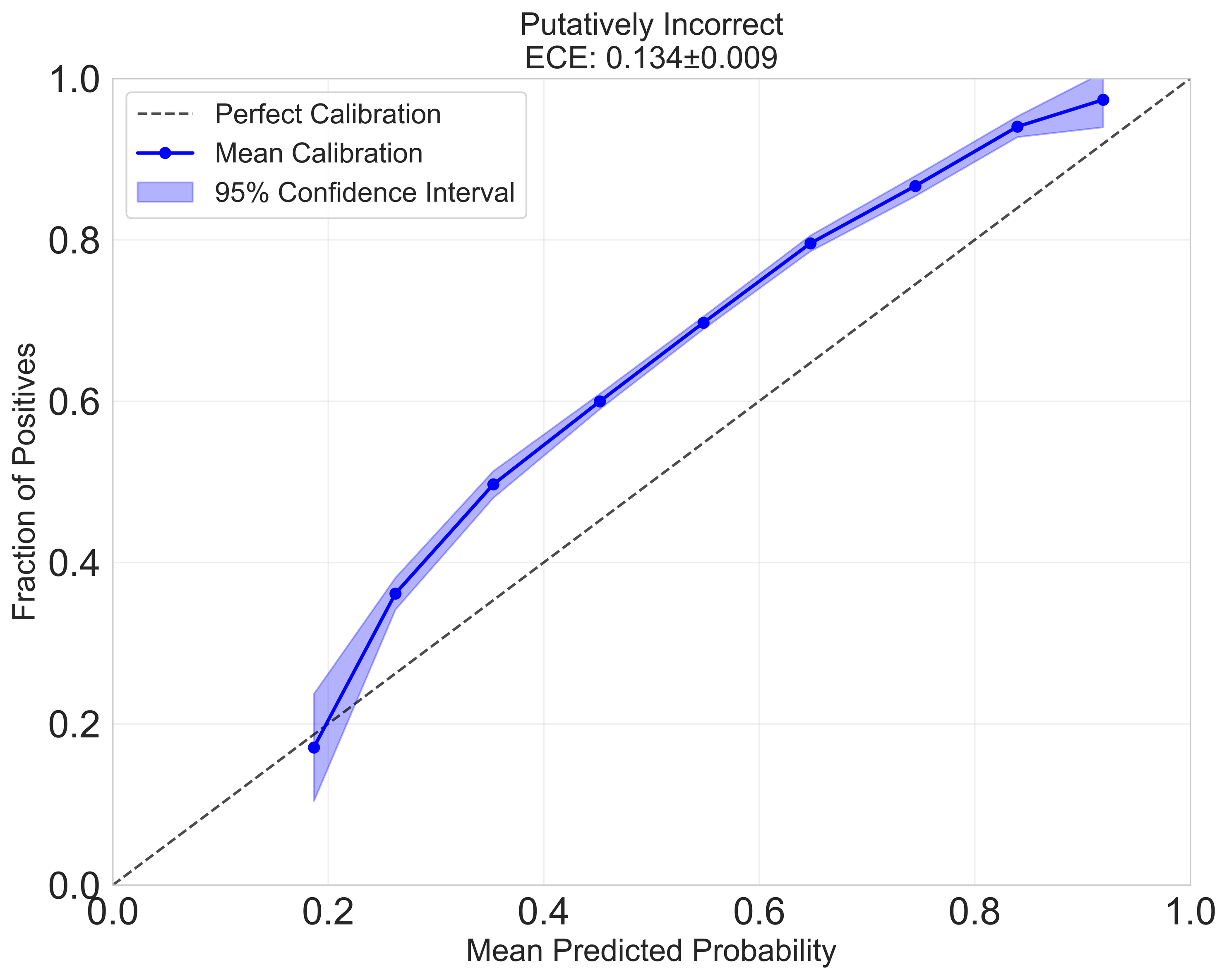}}
  \subfloat[Putatively Incorrect - Iso Cal]{\includegraphics[width=0.32\textwidth]{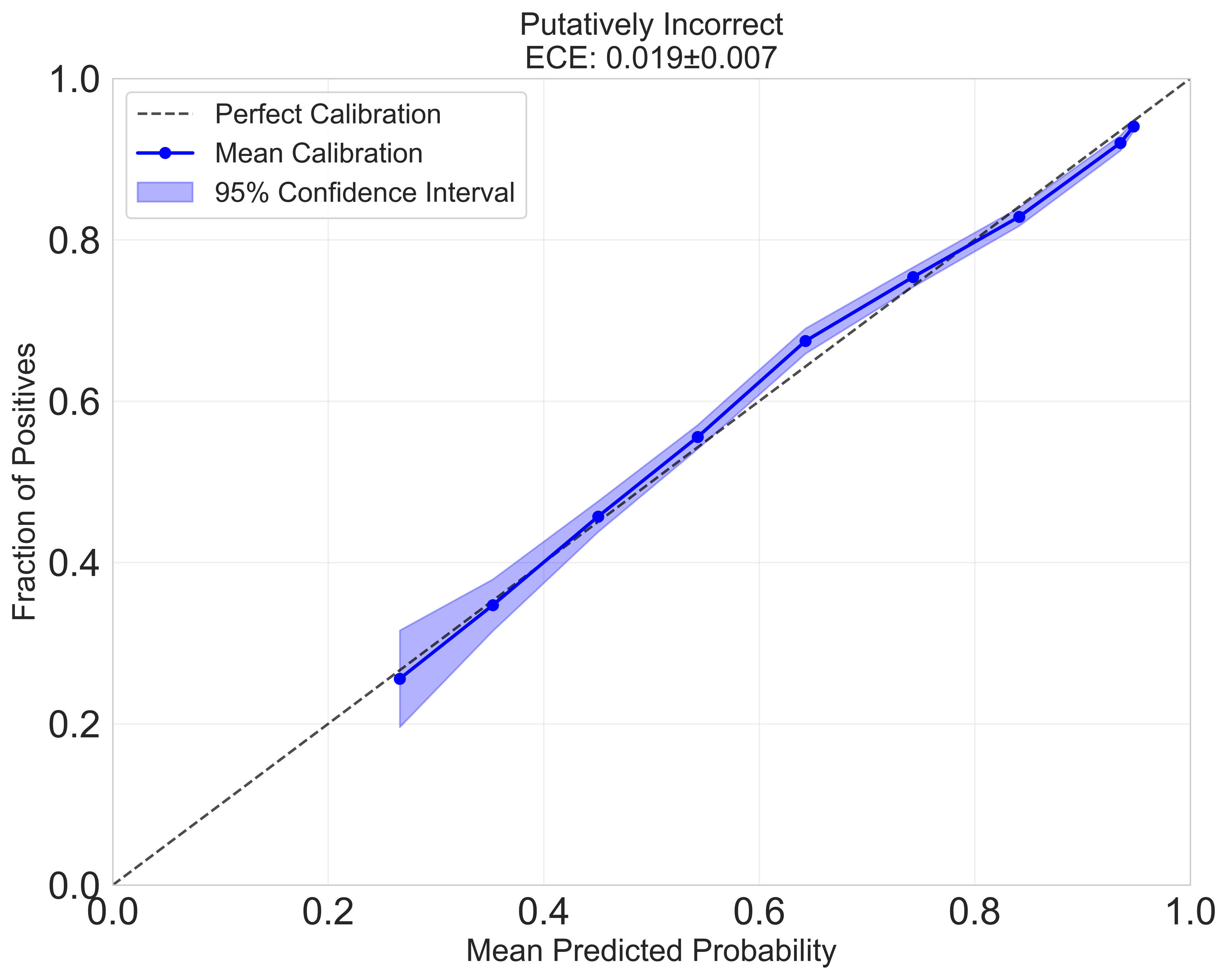}}
  \subfloat[Putatively Incorrect - Dual Cal]{\includegraphics[width=0.32\textwidth]{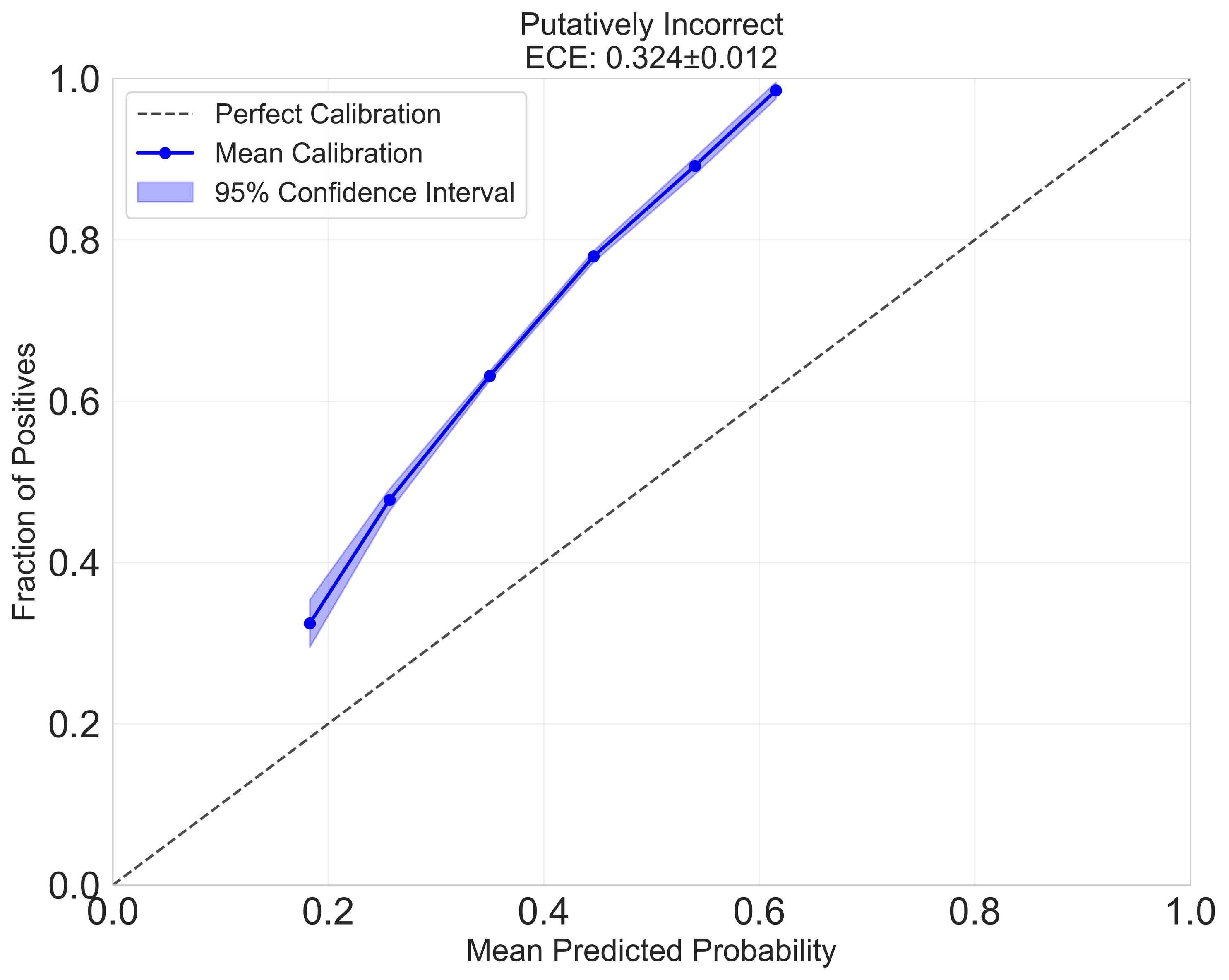}}
  \caption{Reliability diagrams CIFAR-10 with BiT backbone.}
  \label{fig:Reliability_diagram_BiT_Cifar10}
\end{figure}

\begin{figure}[!htbp]
  \centering
  \captionsetup[subfloat]{font=tiny}
  \subfloat[Overall - Non Cal]{\includegraphics[width=0.32\textwidth]{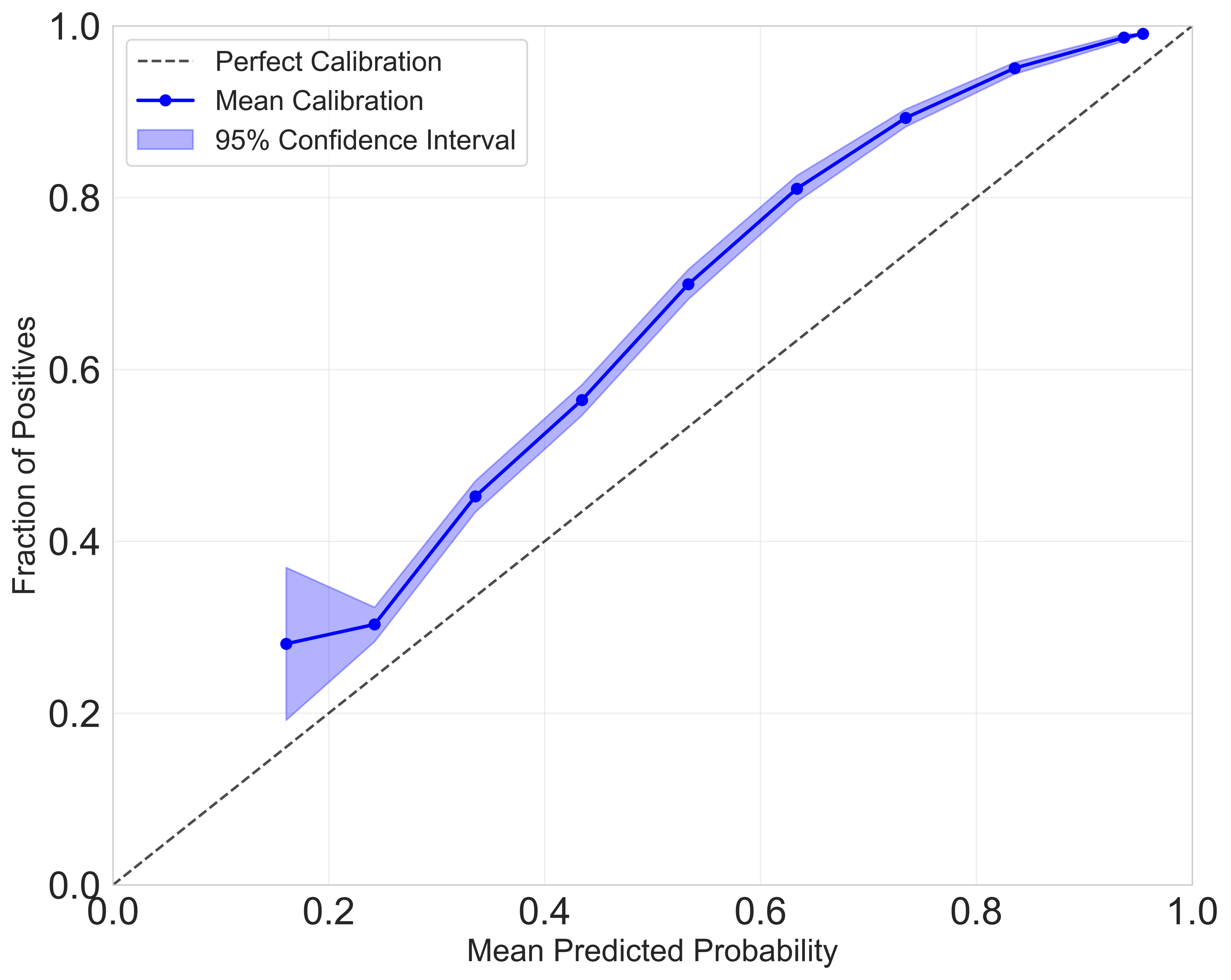}}
  \subfloat[Overall - Iso Cal]{\includegraphics[width=0.32\textwidth]{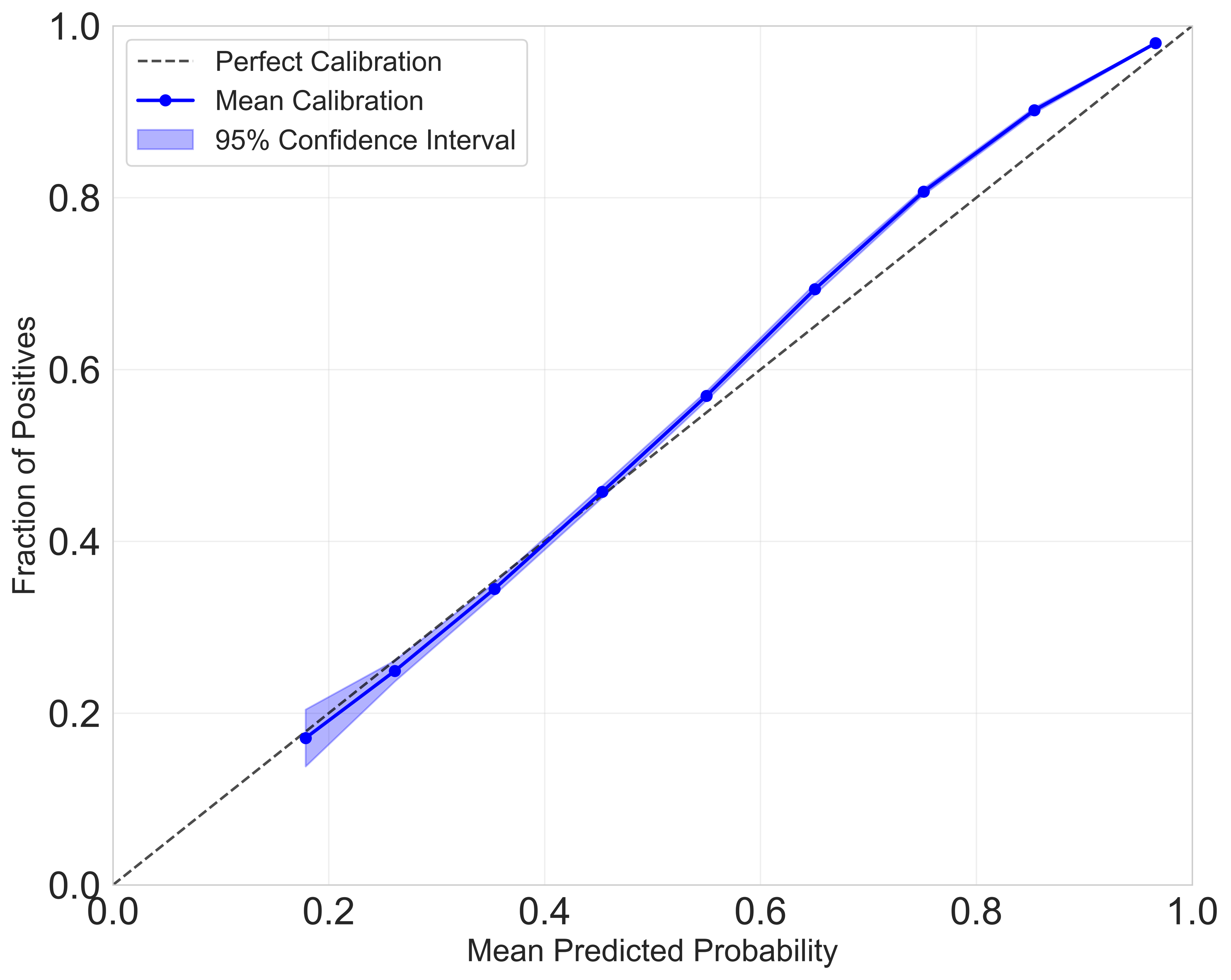}}
  \subfloat[Overall - Dual Cal]{\includegraphics[width=0.32\textwidth]{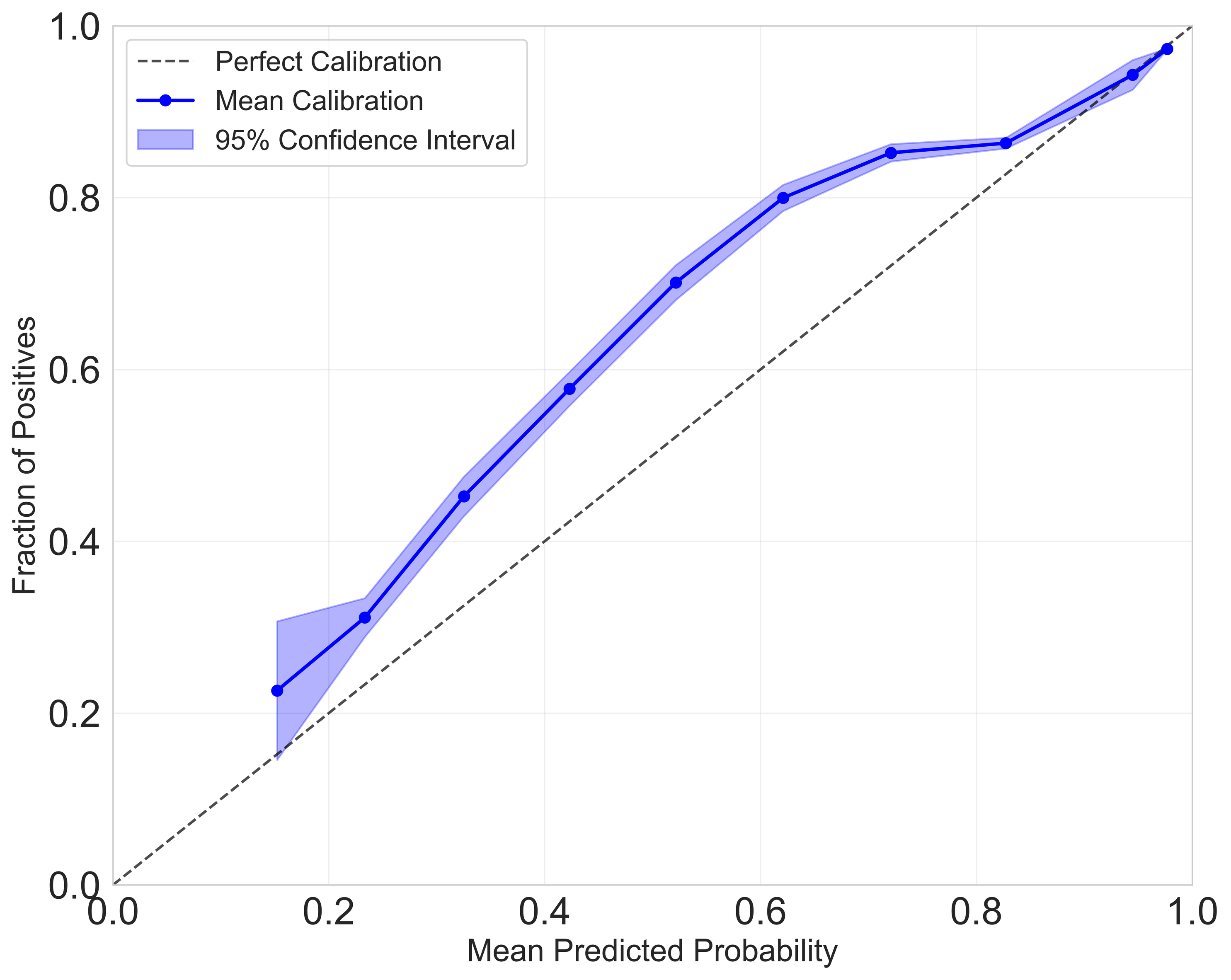}}\par
  \subfloat[Putatively Correct - Non Cal]{\includegraphics[width=0.32\textwidth]{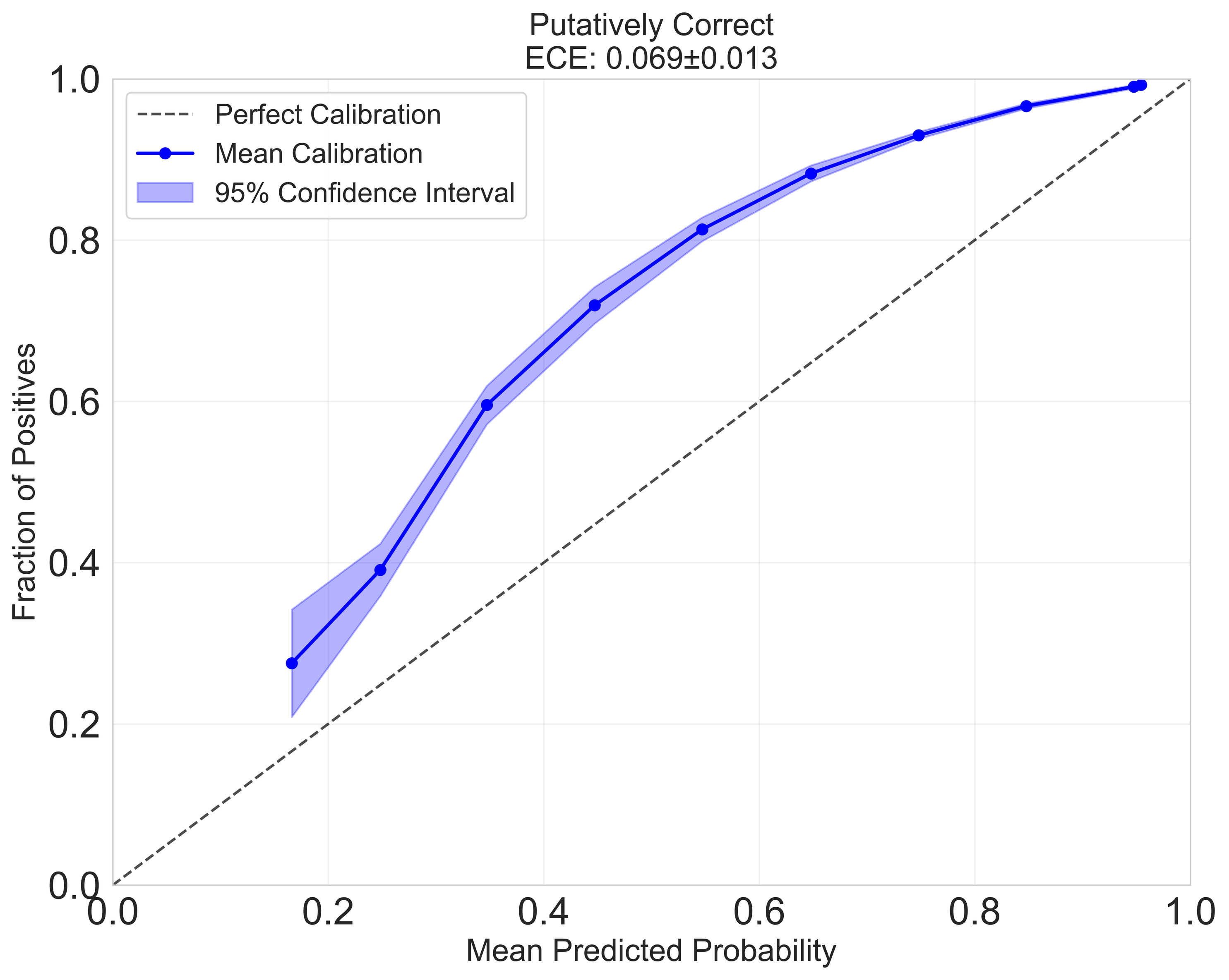}}
  \subfloat[Putatively Correct - Iso Cal]{\includegraphics[width=0.32\textwidth]{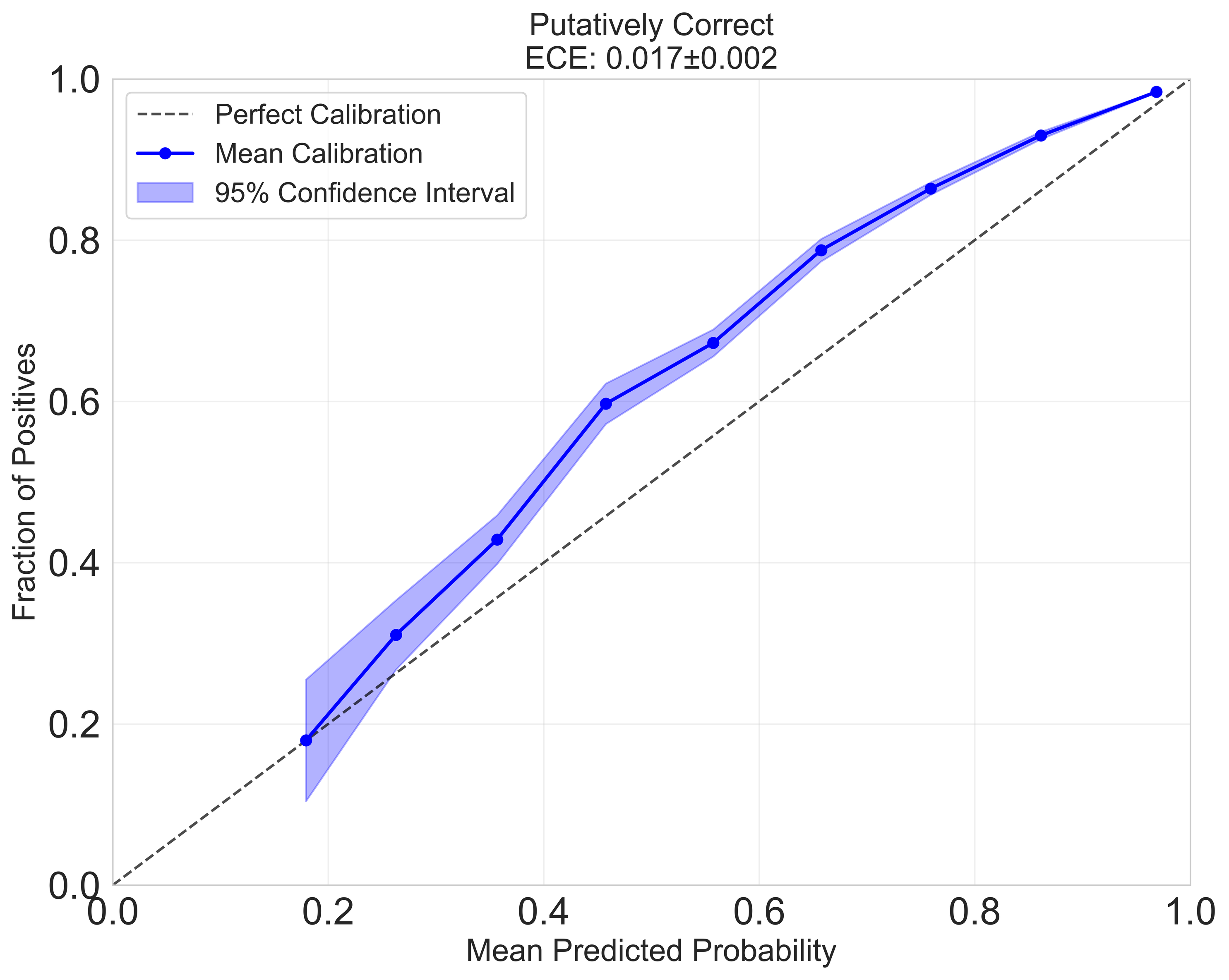}}
  \subfloat[Putatively Correct - Dual Cal]{\includegraphics[width=0.32\textwidth]{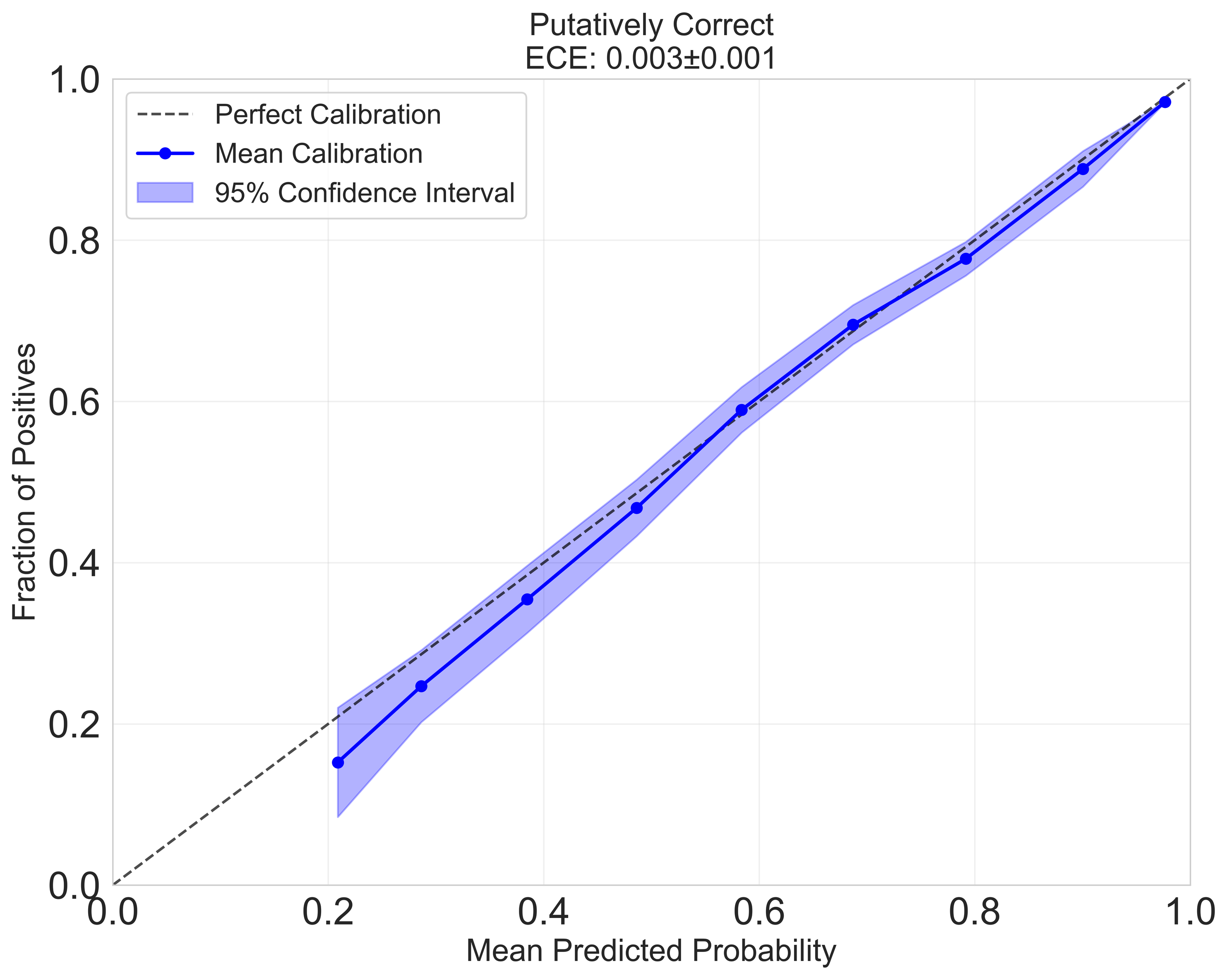}}\par
  \subfloat[Putatively Incorrect - Non Cal]{\includegraphics[width=0.32\textwidth]{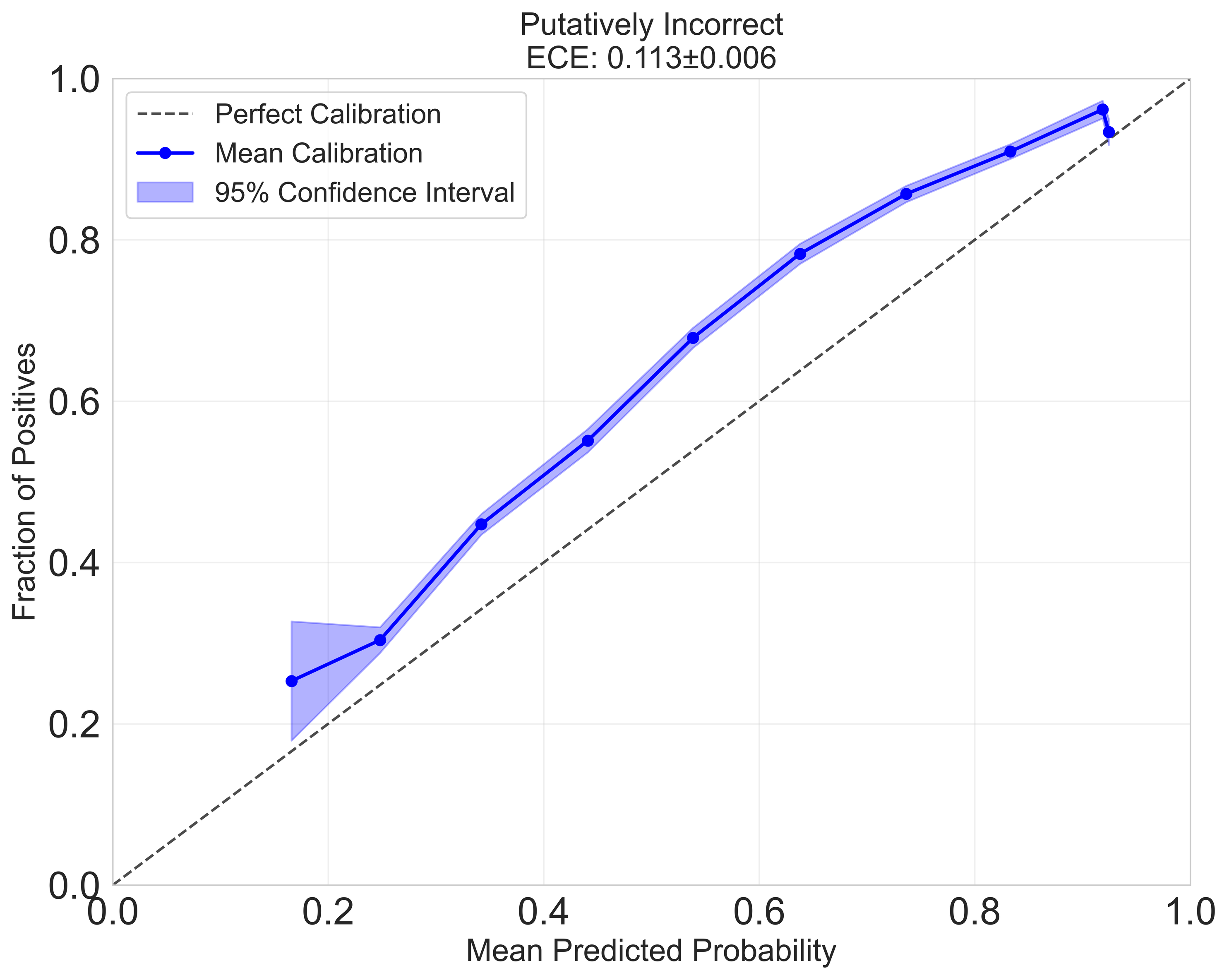}}
  \subfloat[Putatively Incorrect - Iso Cal]{\includegraphics[width=0.32\textwidth]{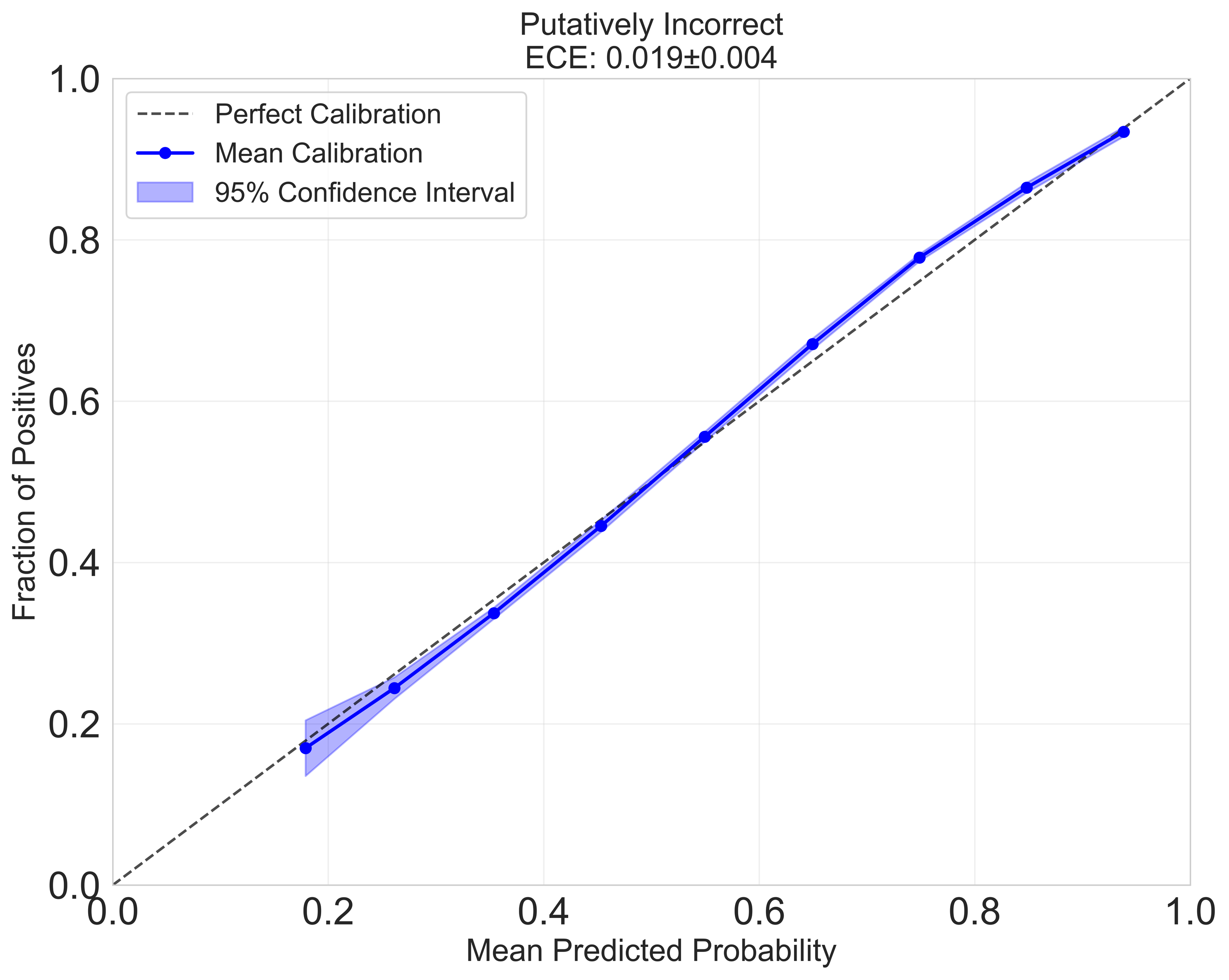}}
  \subfloat[Putatively Incorrect - Dual Cal]{\includegraphics[width=0.32\textwidth]{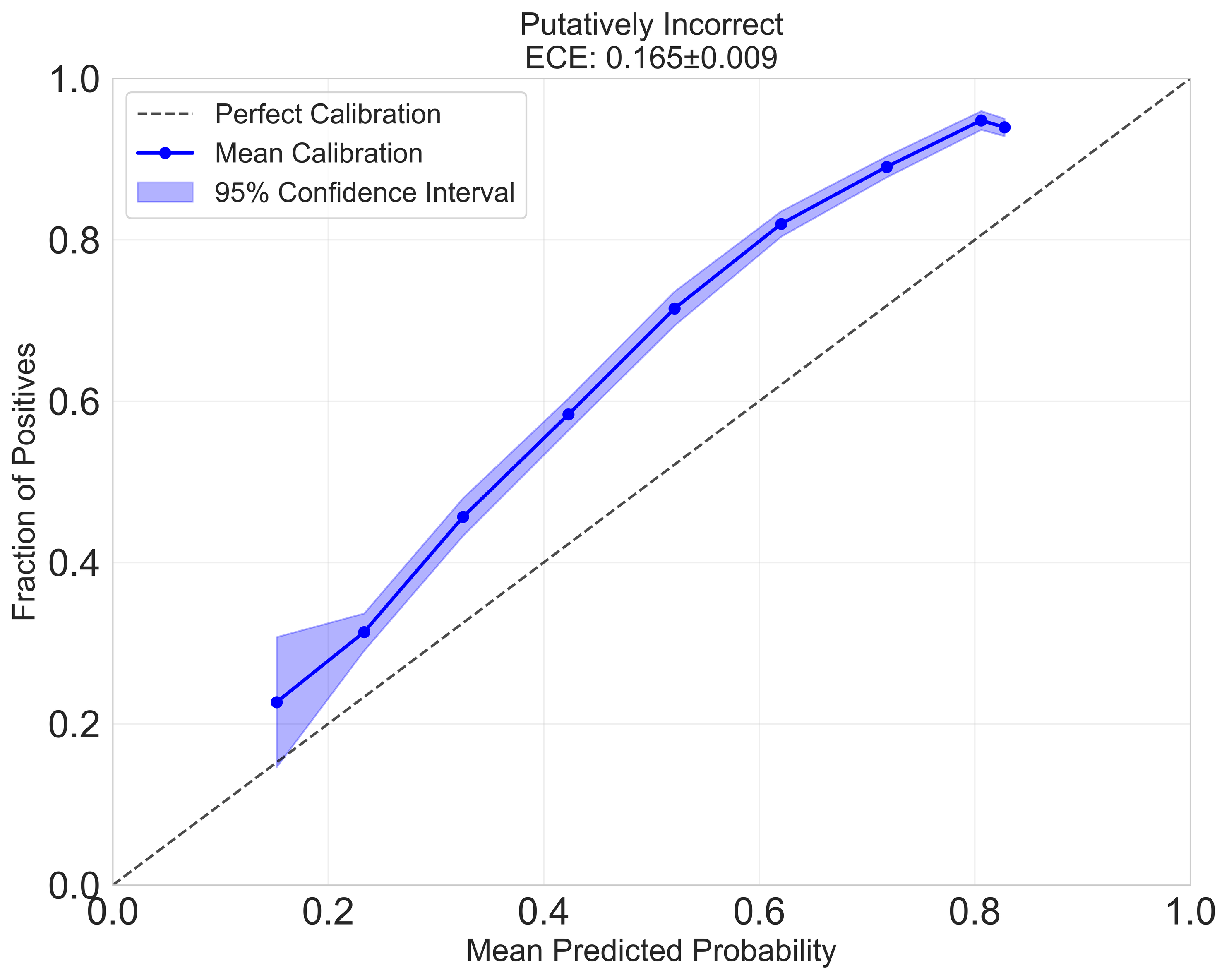}}
  \caption{Reliability diagrams CIFAR-100 superclasses with BiT backbone.}
  \label{fig:Reliability_diagram_BiT_Cifar100_coarse}
\end{figure}

\begin{figure}[!htbp]
  \centering
  \captionsetup[subfloat]{font=tiny}
  \subfloat[Overall - Non Cal]{\includegraphics[width=0.32\textwidth]{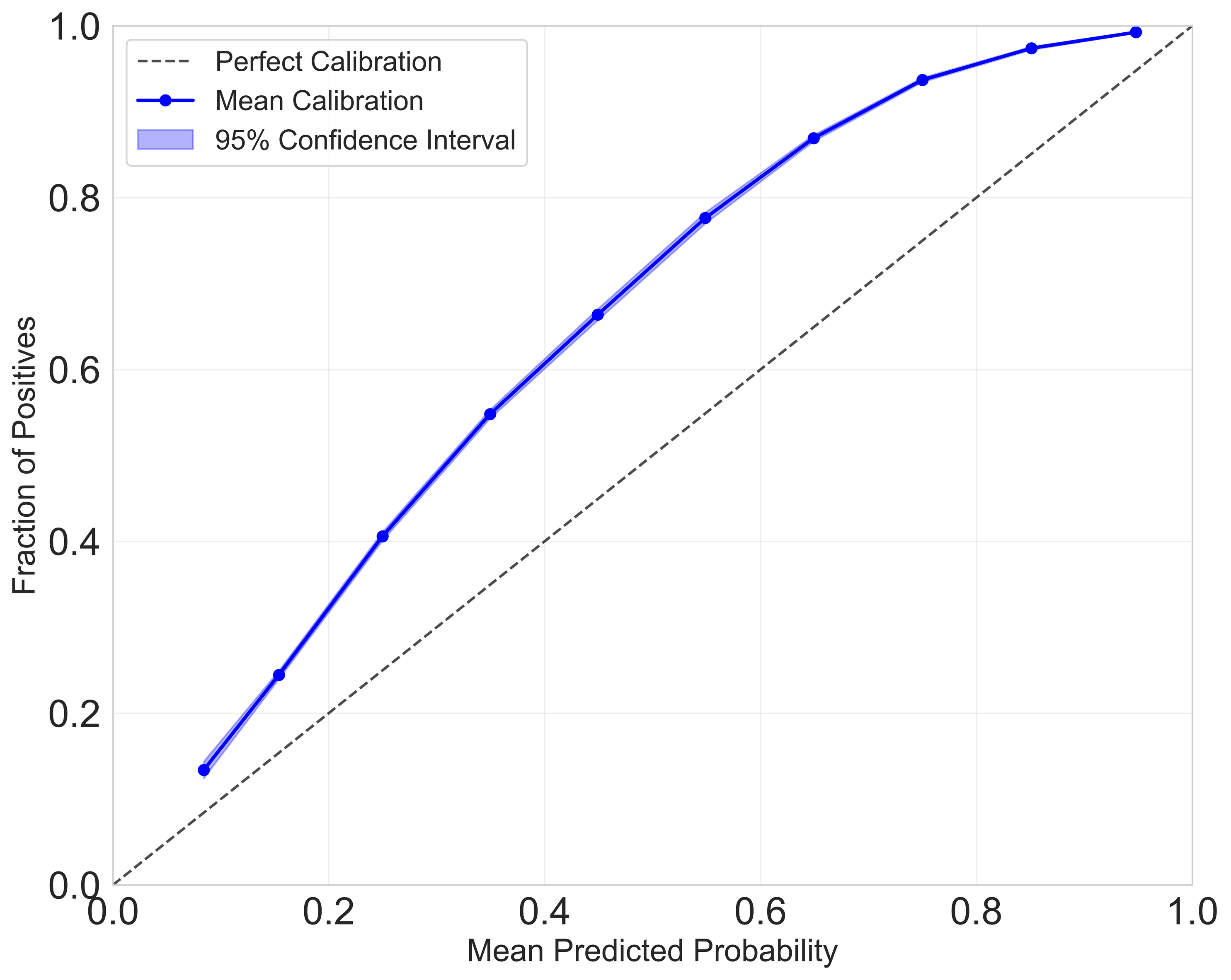}}
  \subfloat[Overall - Iso Cal]{\includegraphics[width=0.32\textwidth]{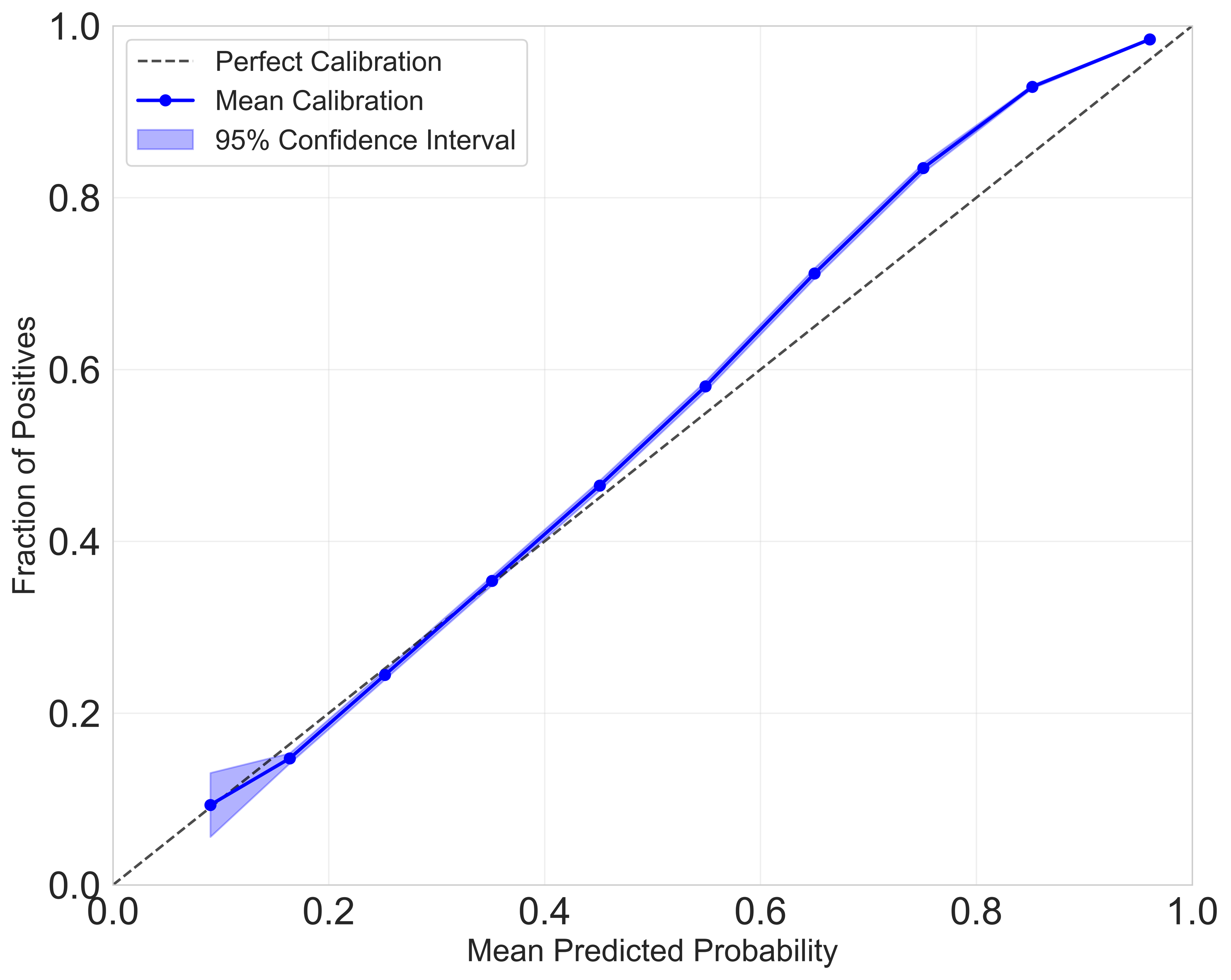}}
  \subfloat[Overall - Dual Cal]{\includegraphics[width=0.32\textwidth]{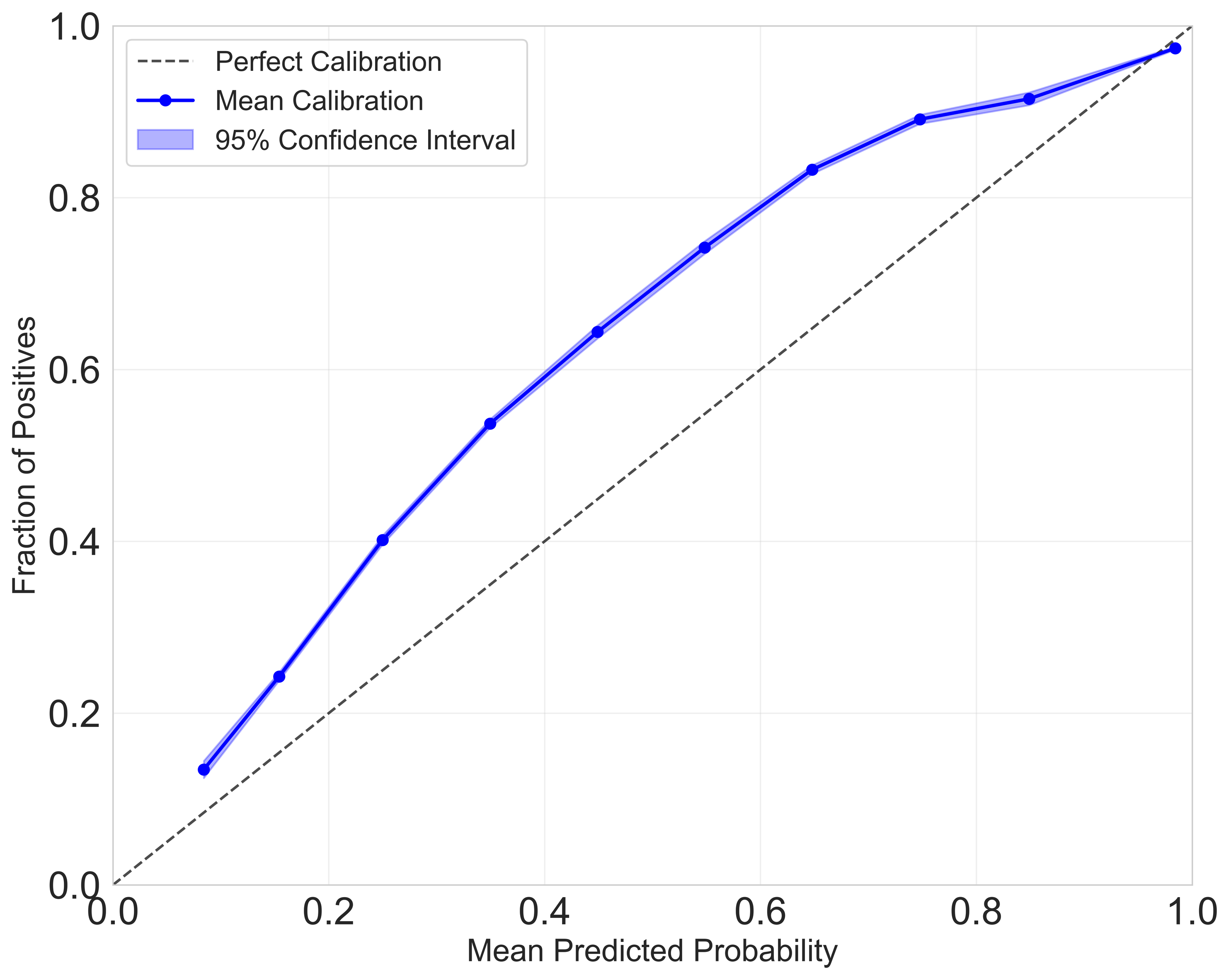}}\par
  \subfloat[Putatively Correct - Non Cal]{\includegraphics[width=0.32\textwidth]{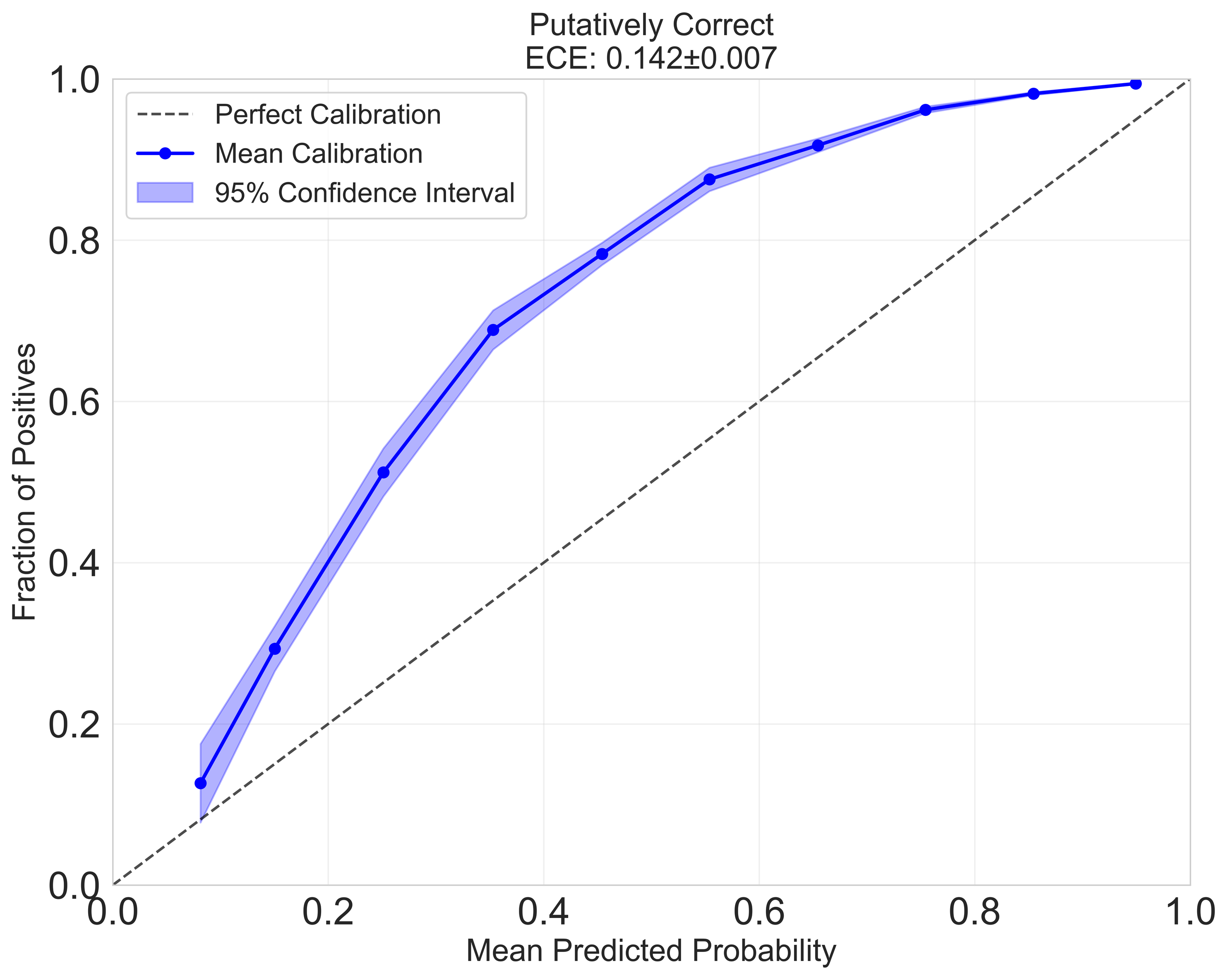}}
  \subfloat[Putatively Correct - Iso Cal]{\includegraphics[width=0.32\textwidth]{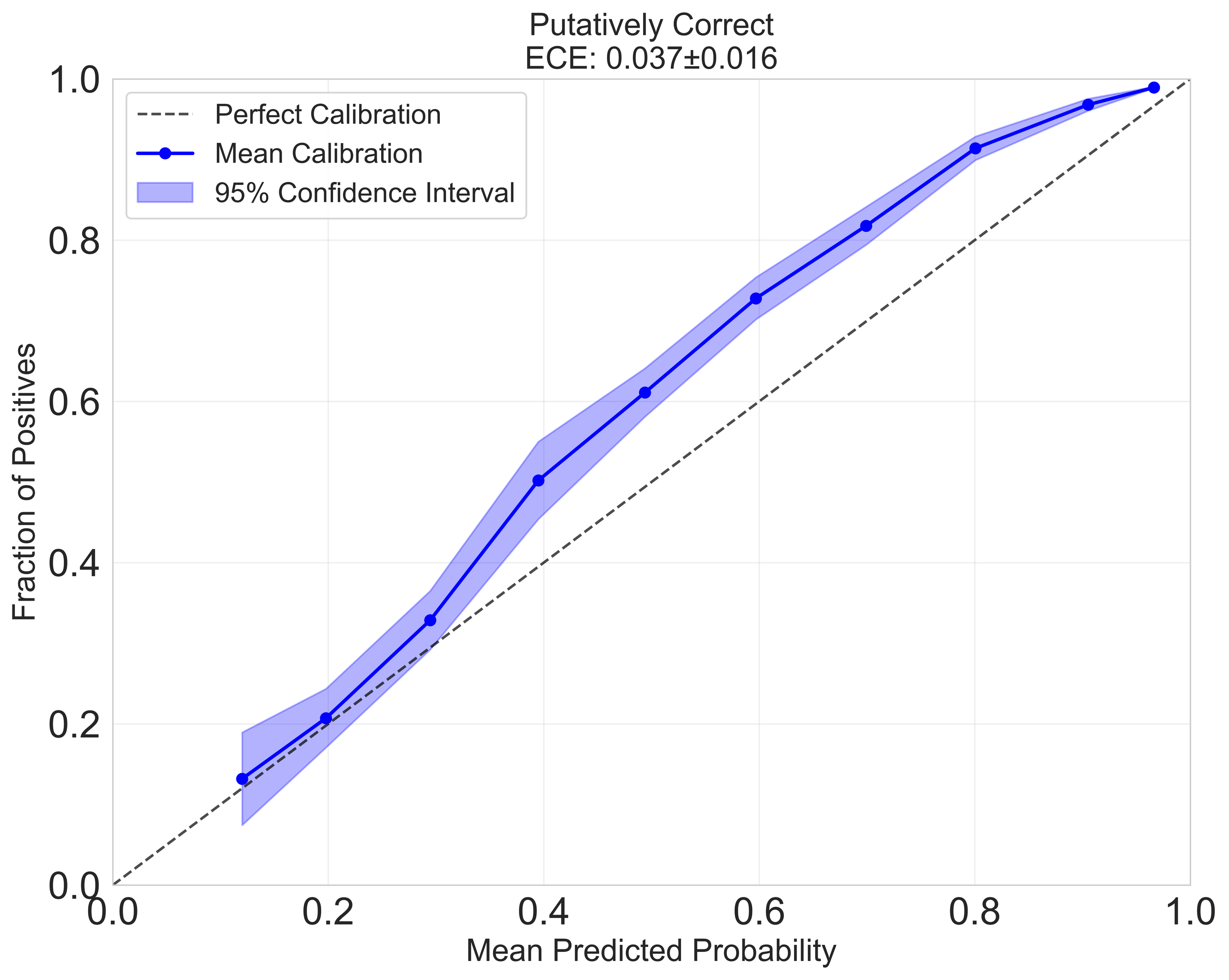}}
  \subfloat[Putatively Correct - Dual Cal]{\includegraphics[width=0.32\textwidth]{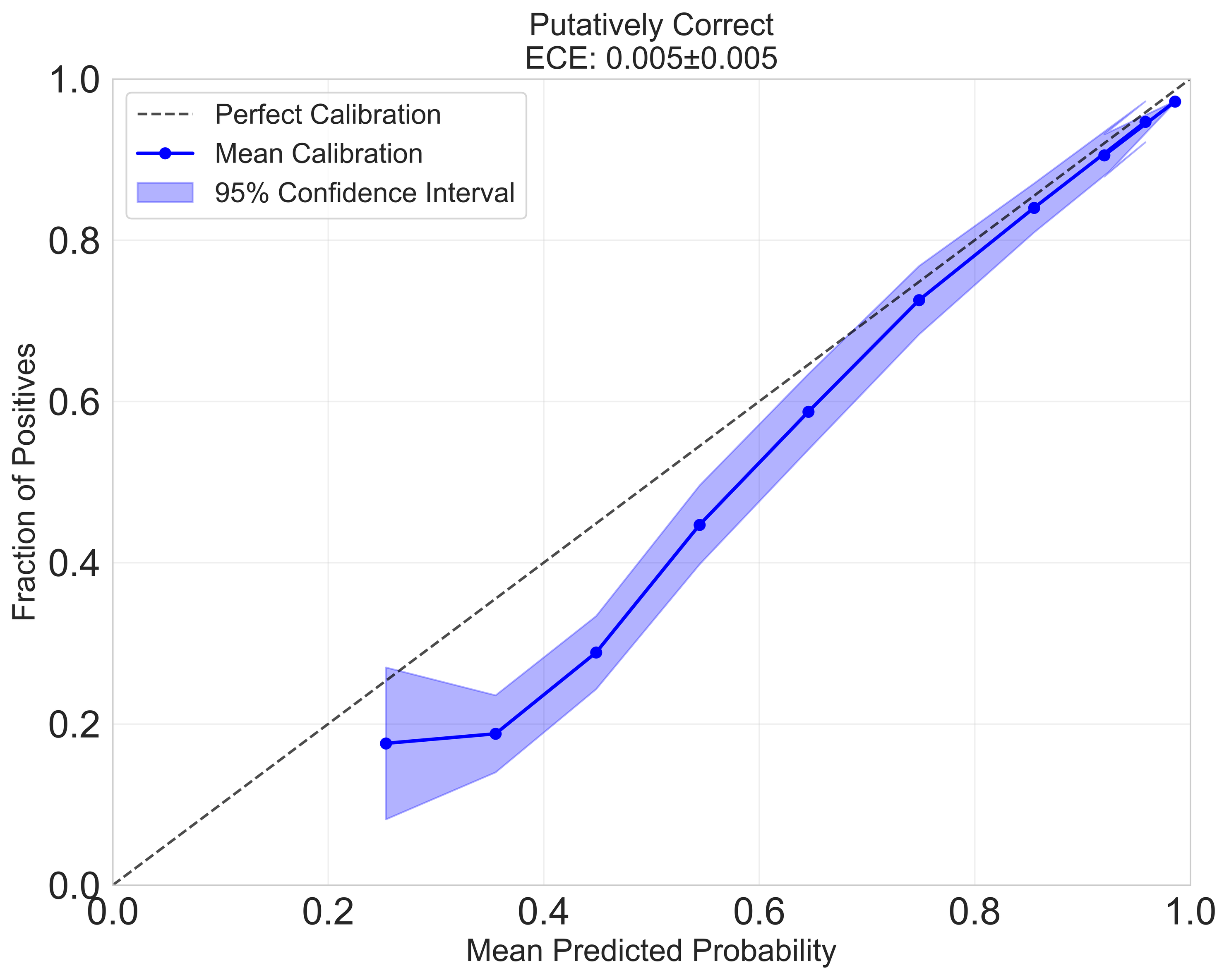}}\par
  \subfloat[Putatively Incorrect - Non Cal]{\includegraphics[width=0.32\textwidth]{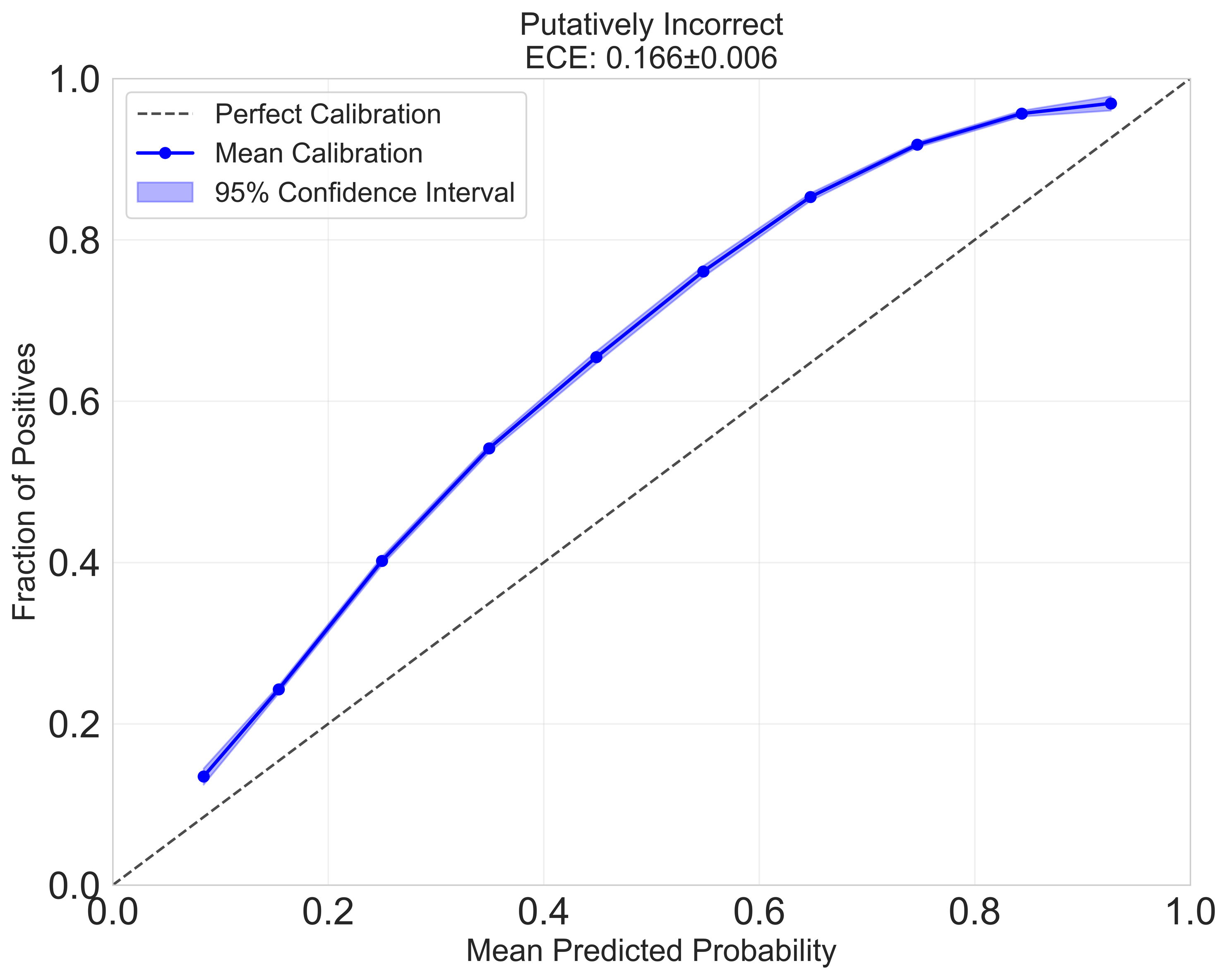}}
  \subfloat[Putatively Incorrect - Iso Cal]{\includegraphics[width=0.32\textwidth]{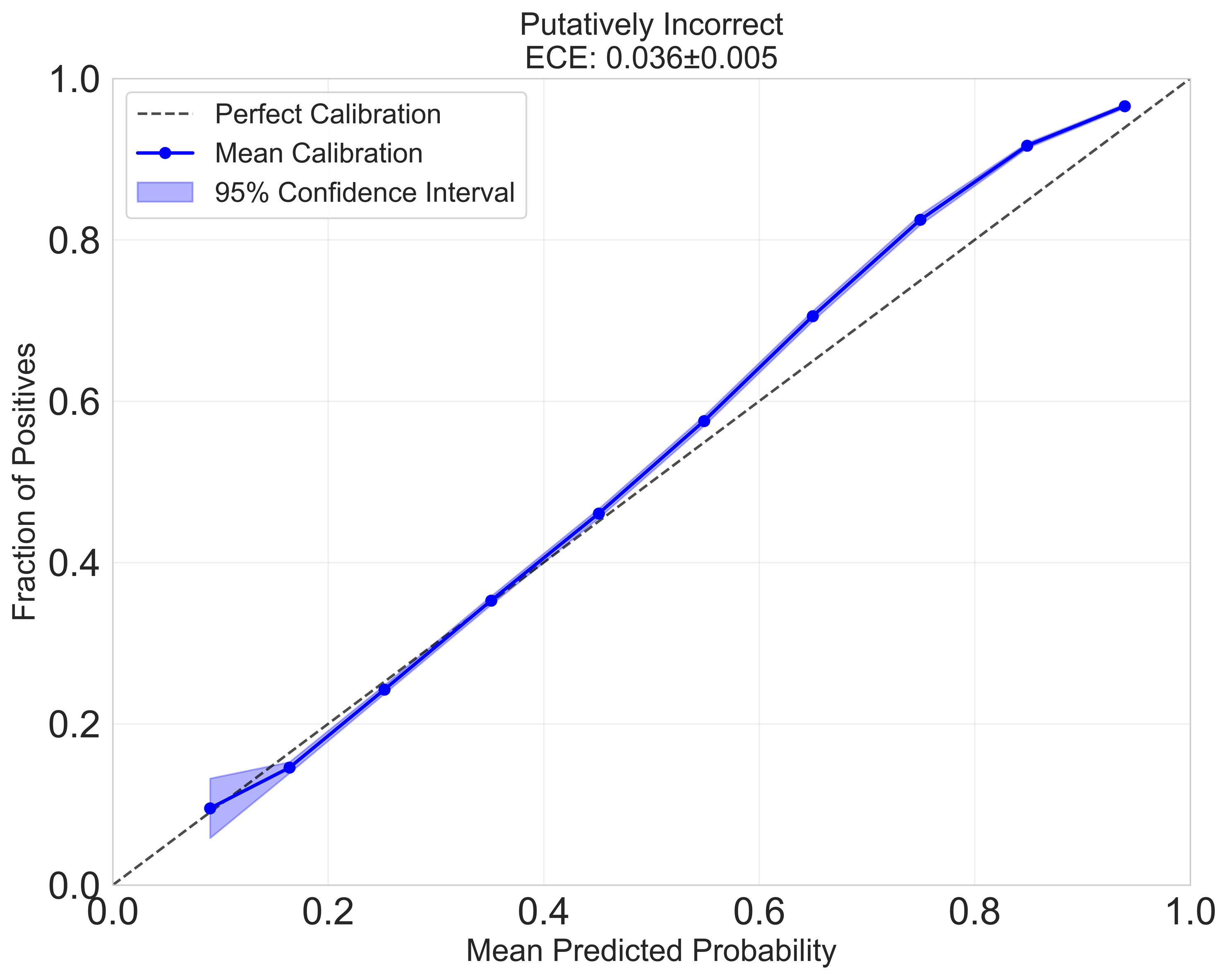}}
  \subfloat[Putatively Incorrect - Dual Cal]{\includegraphics[width=0.32\textwidth]{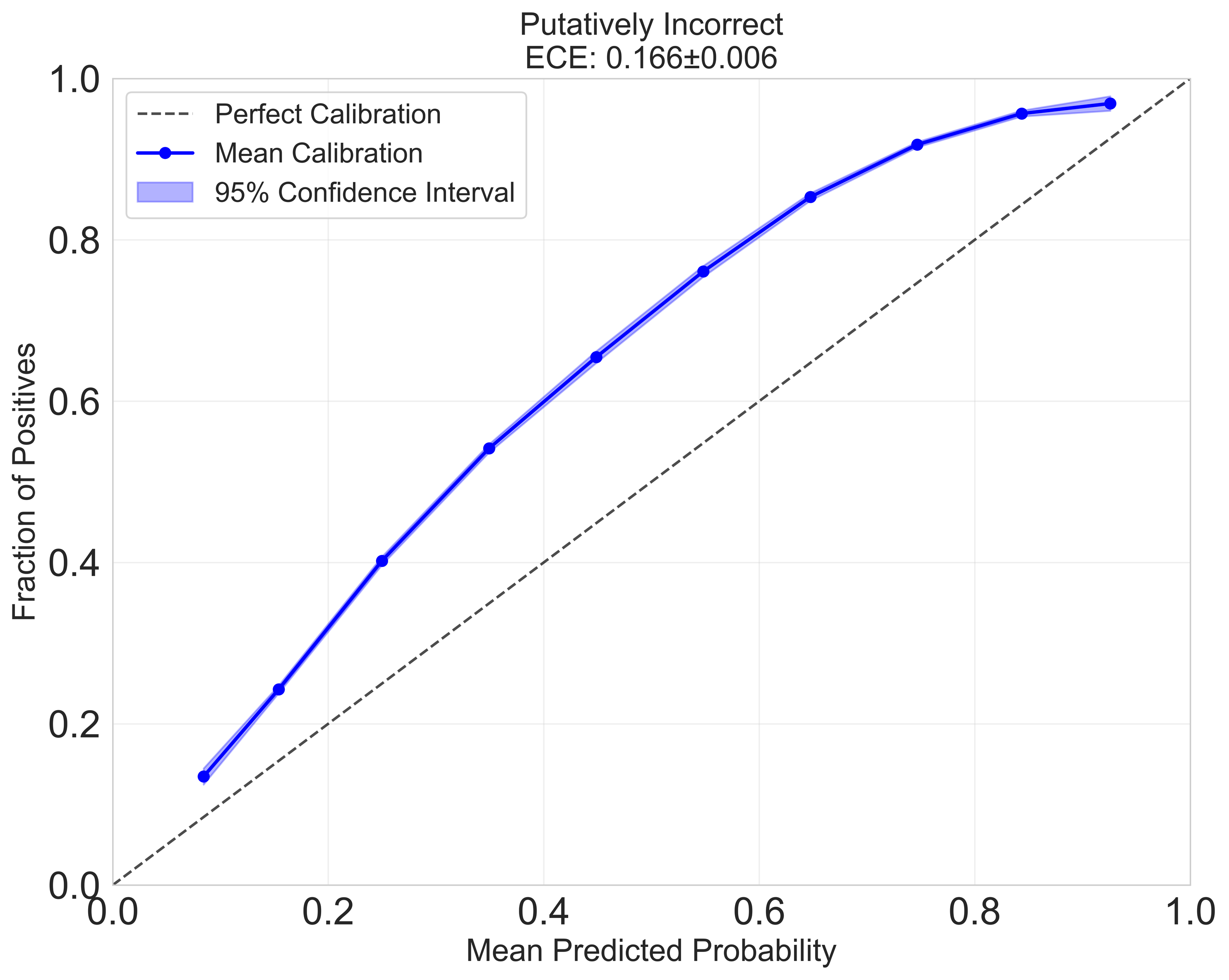}}
  \caption{Reliability diagrams CIFAR-100 fine-grained classes with BiT backbone.}
  \label{fig:Reliability_diagram_BiT_Cifar100}
\end{figure}

\begin{figure}[!htbp]
  \centering
  \captionsetup[subfloat]{font=tiny}
  \subfloat[Overall - Non Cal]{\includegraphics[width=0.32\textwidth]{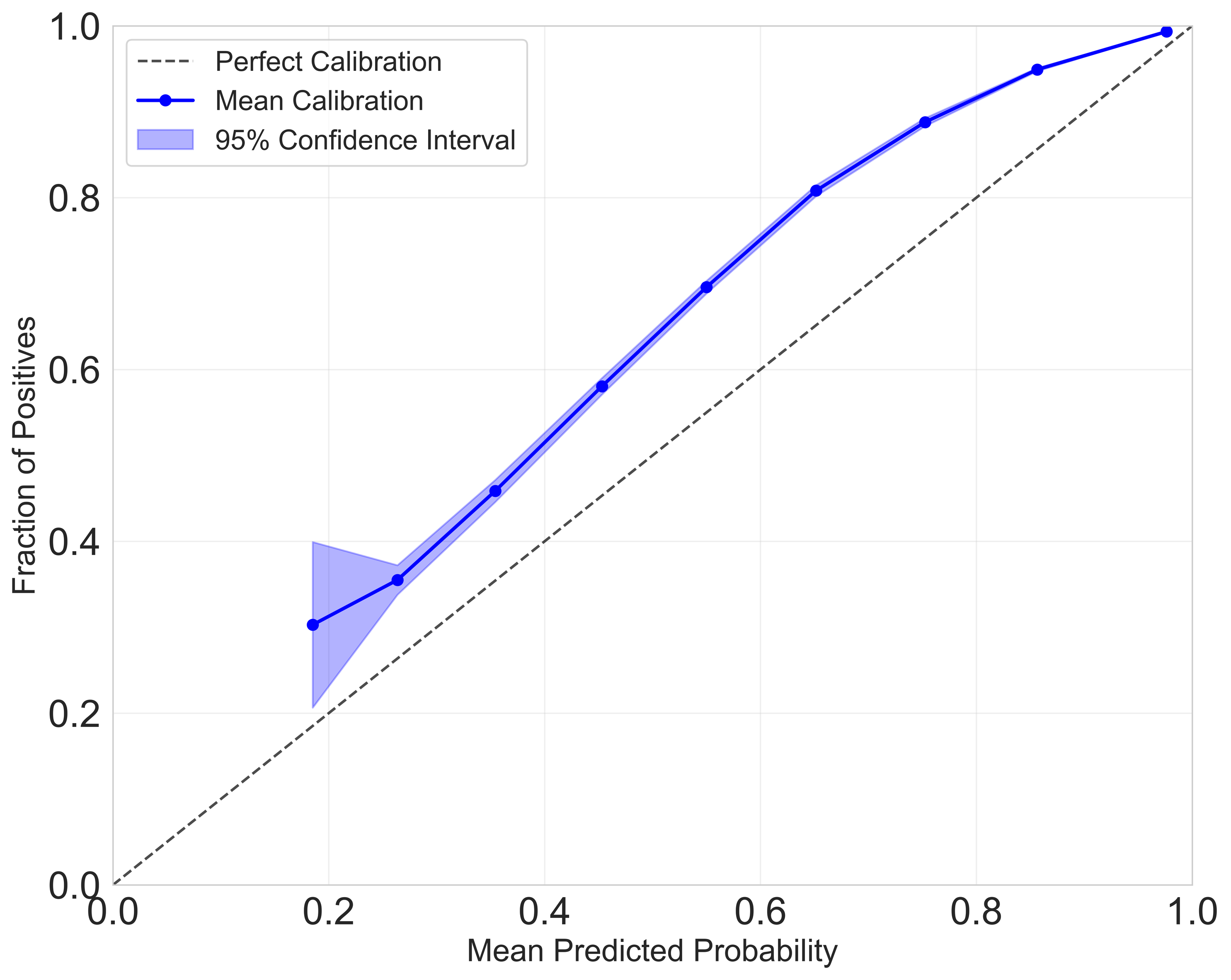}}
  \subfloat[Overall - Iso Cal]{\includegraphics[width=0.32\textwidth]{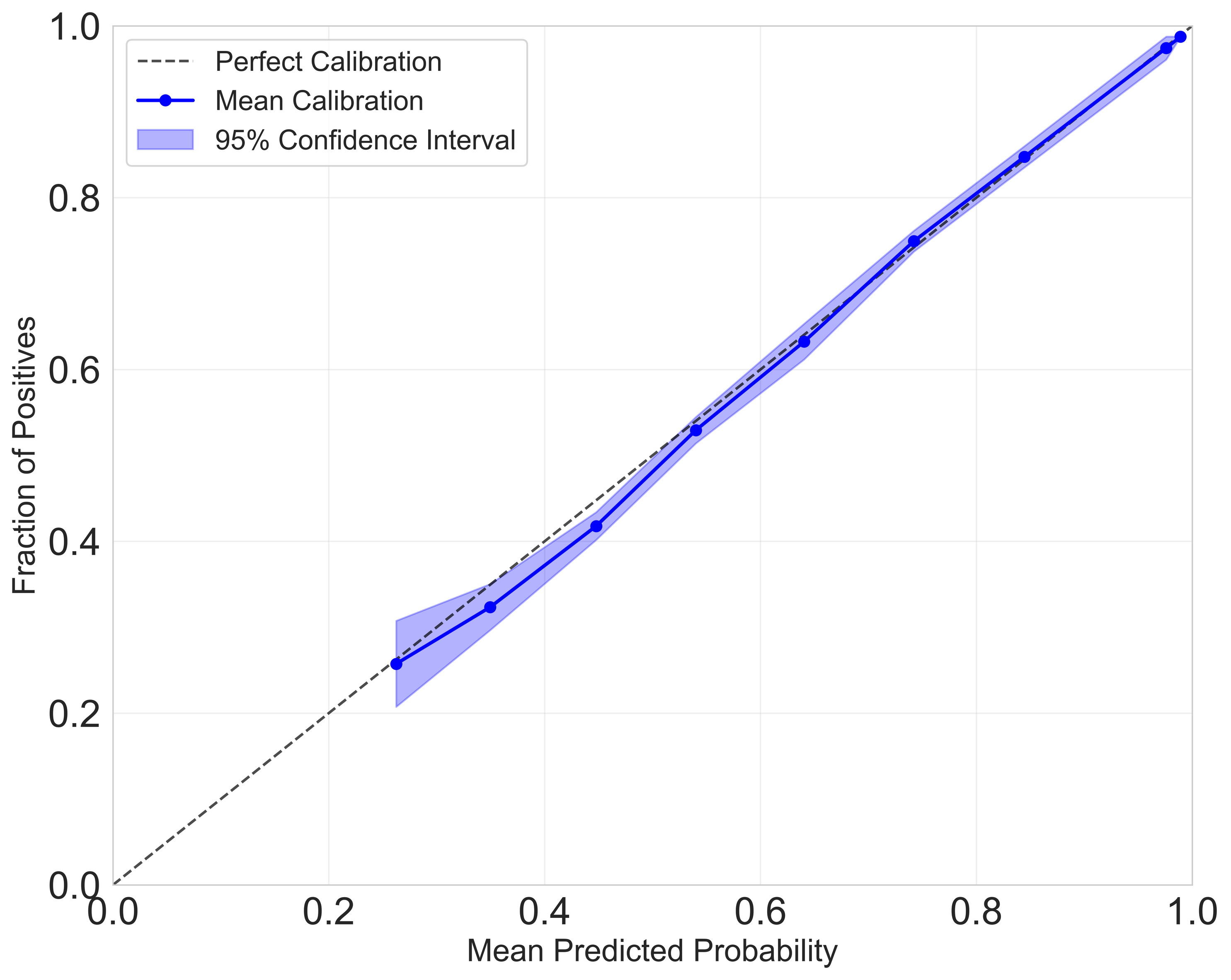}}
  \subfloat[Overall - Dual Cal]{\includegraphics[width=0.32\textwidth]{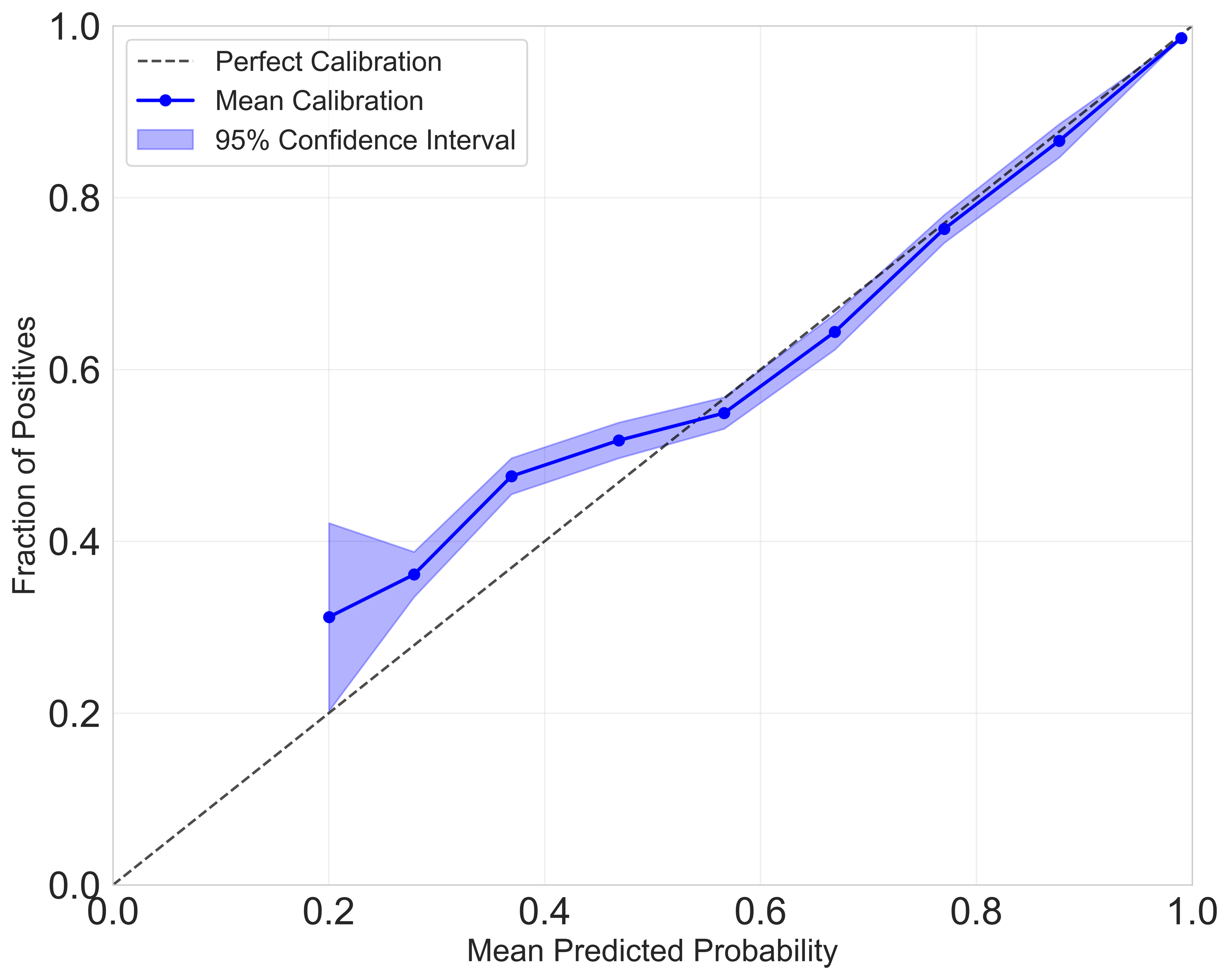}}\par
  \subfloat[Putatively Correct - Non Cal]{\includegraphics[width=0.32\textwidth]{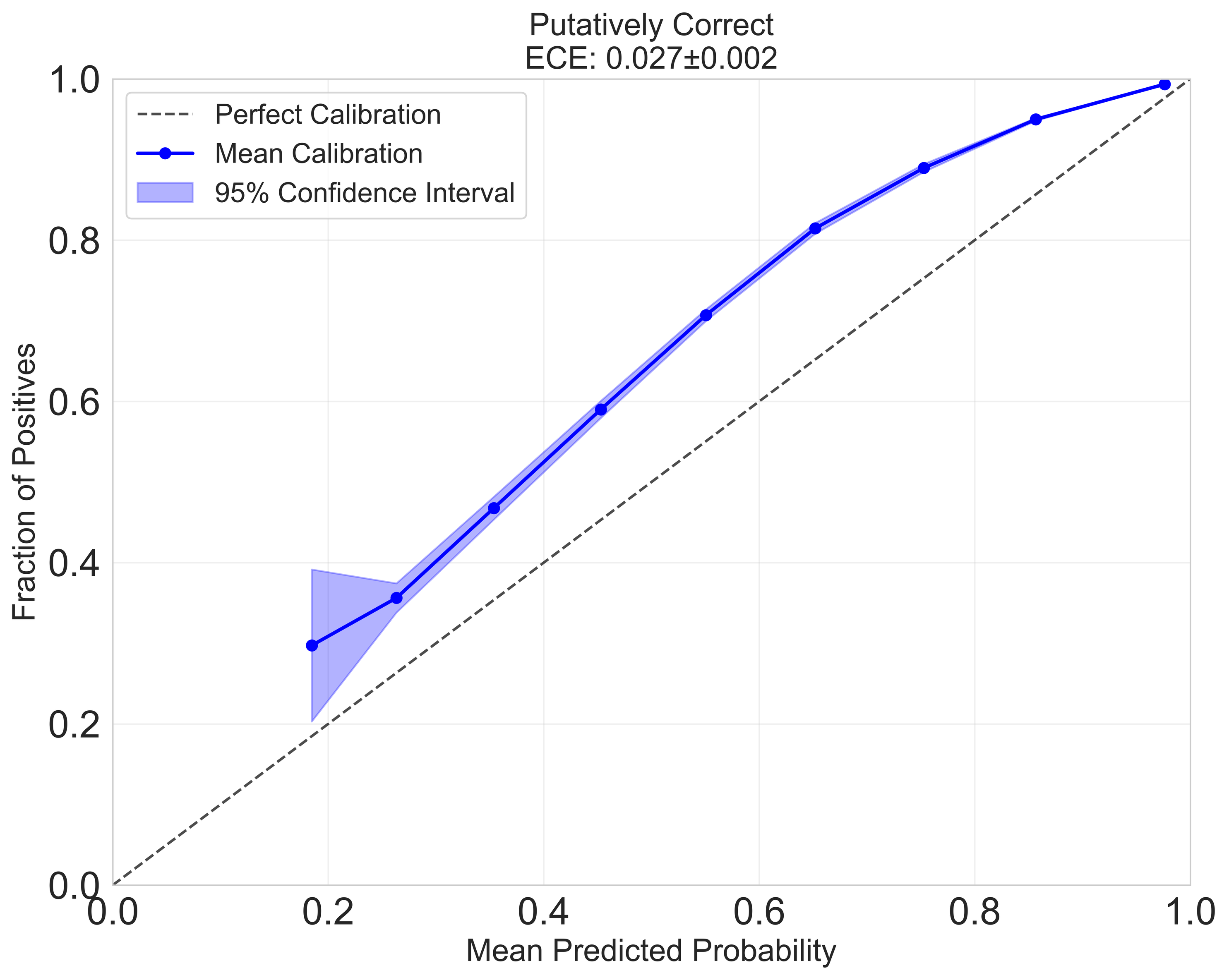}}
  \subfloat[Putatively Correct - Iso Cal]{\includegraphics[width=0.32\textwidth]{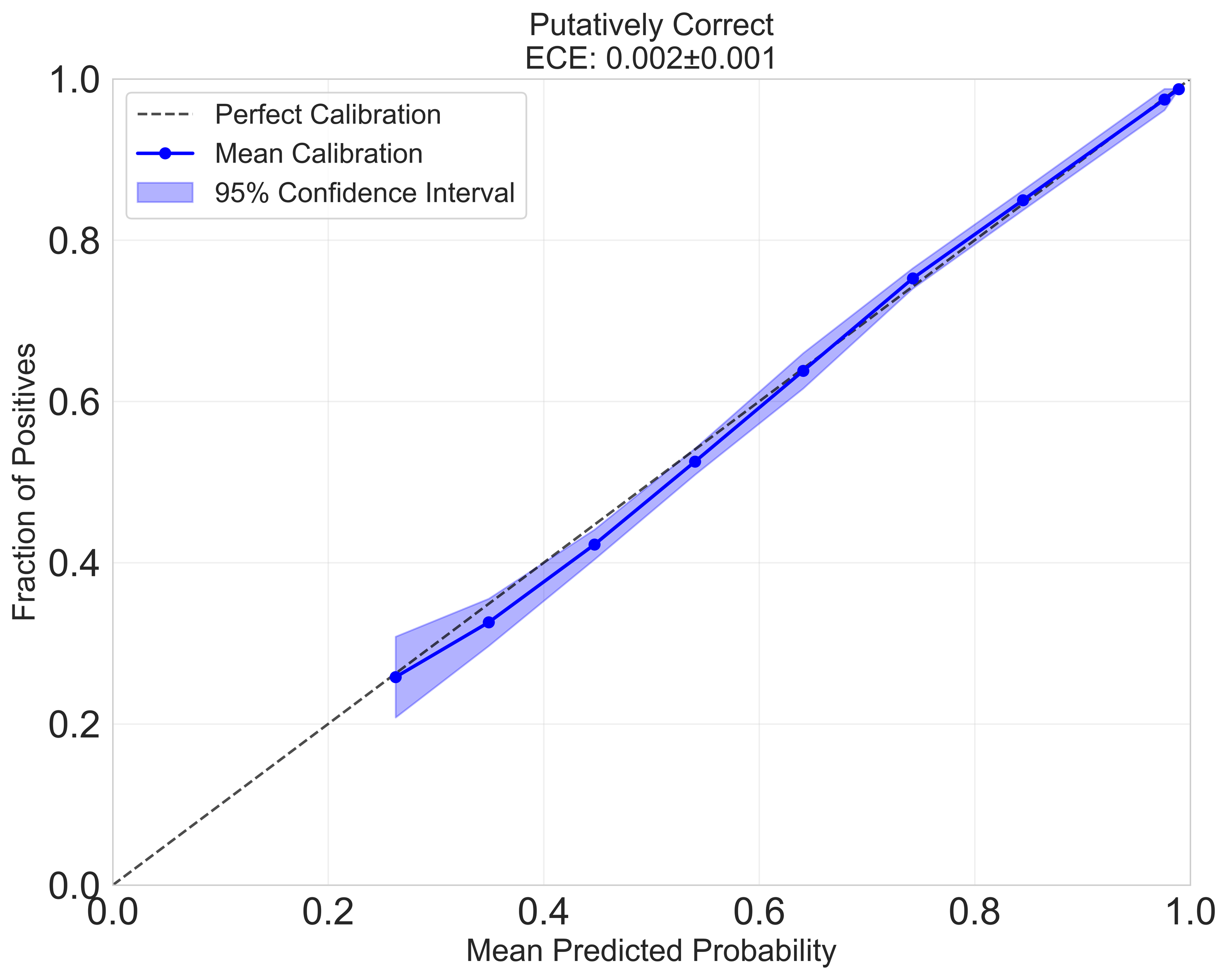}}
  \subfloat[Putatively Correct - Dual Cal]{\includegraphics[width=0.32\textwidth]{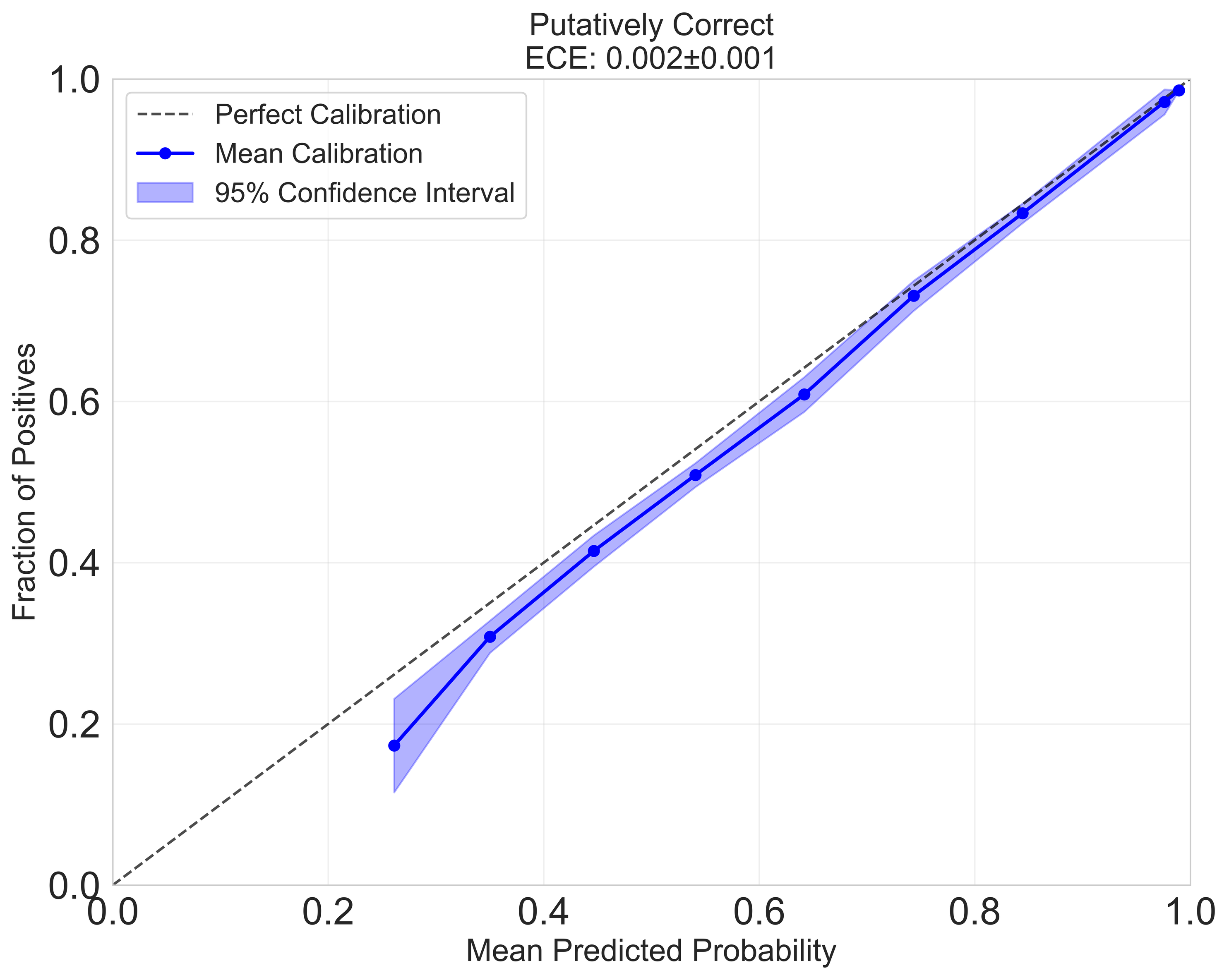}}\par
  \subfloat[Putatively Incorrect - Non Cal]{\includegraphics[width=0.32\textwidth]{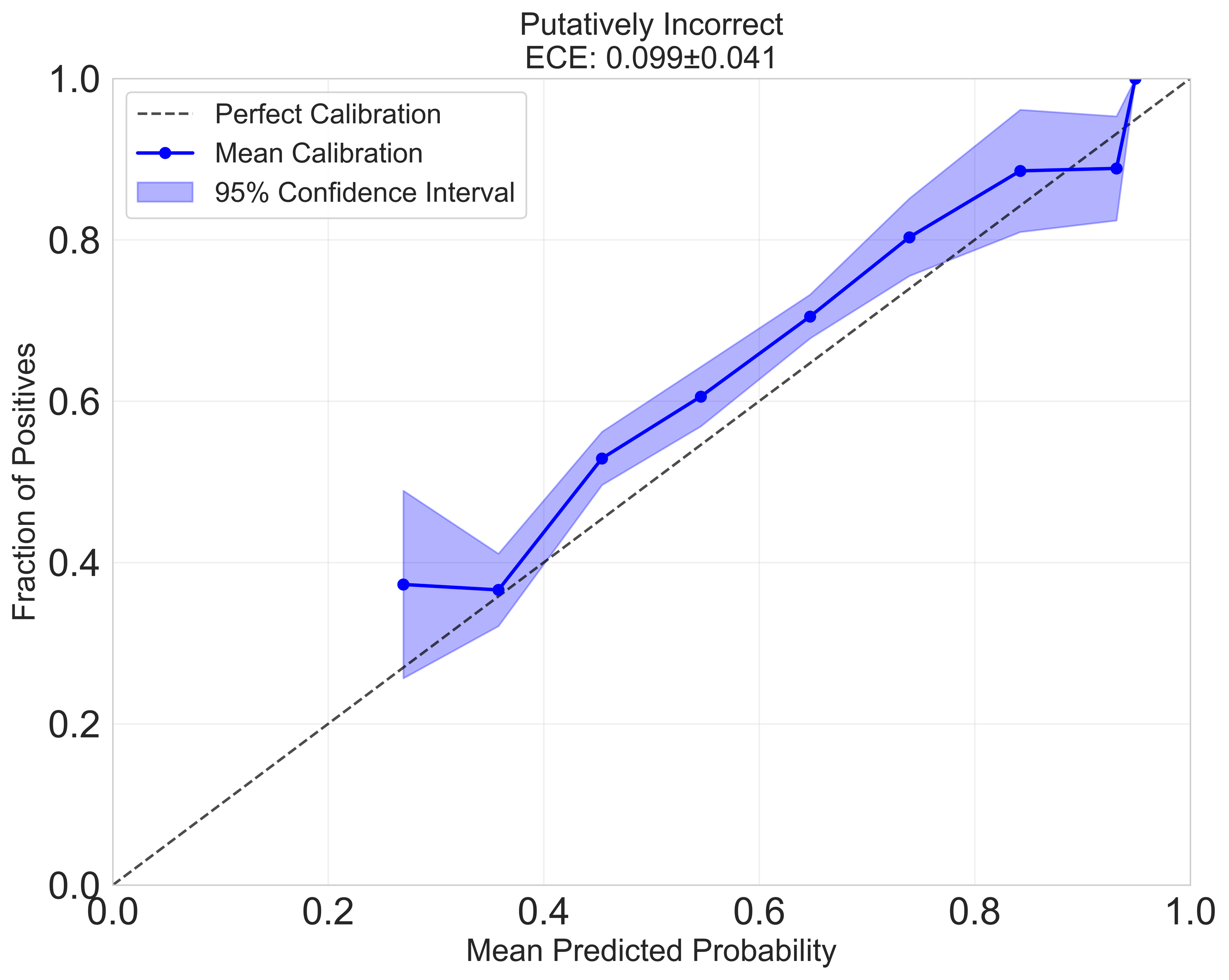}}
  \subfloat[Putatively Incorrect - Iso Cal]{\includegraphics[width=0.32\textwidth]{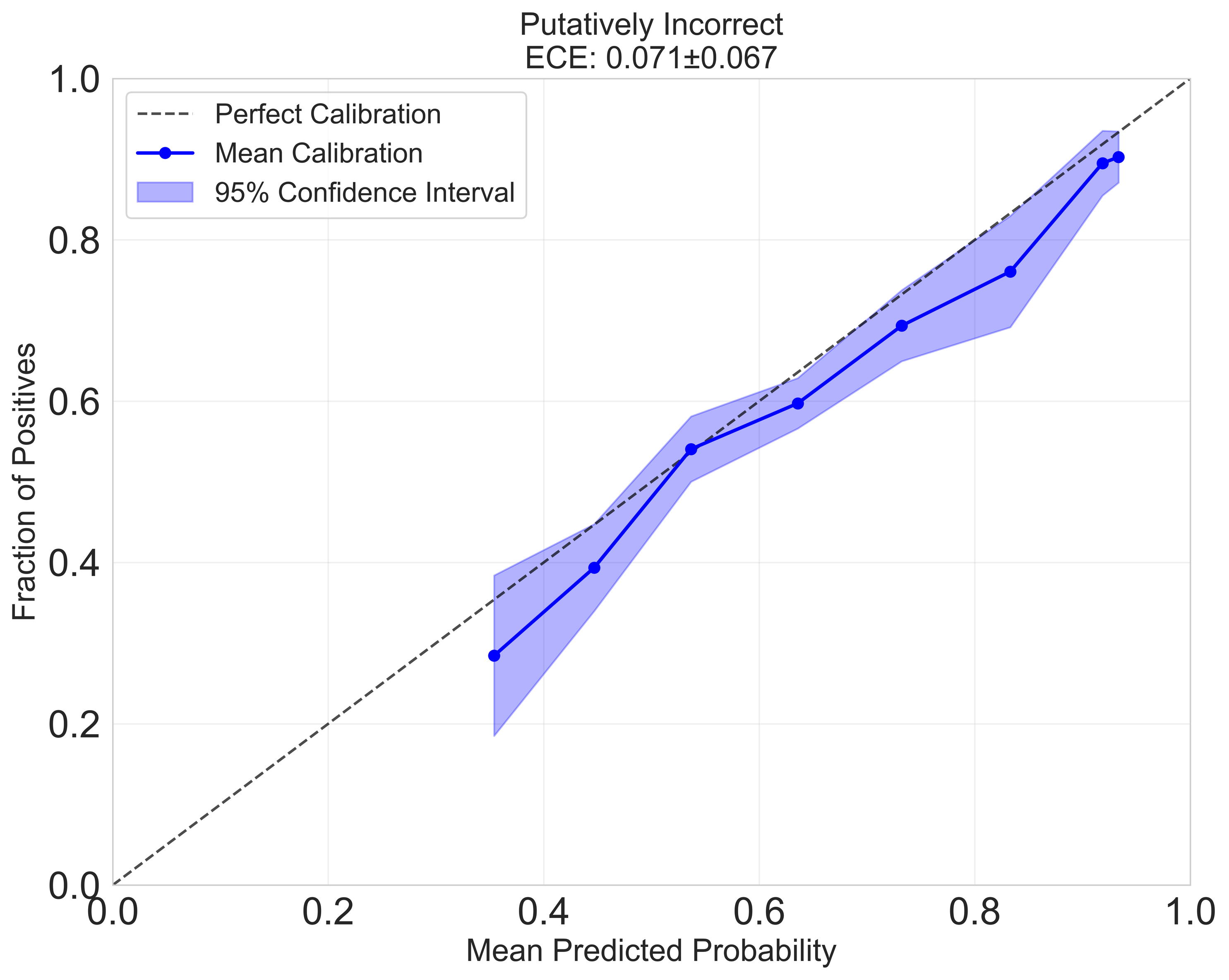}}
  \subfloat[Putatively Incorrect - Dual Cal]{\includegraphics[width=0.32\textwidth]{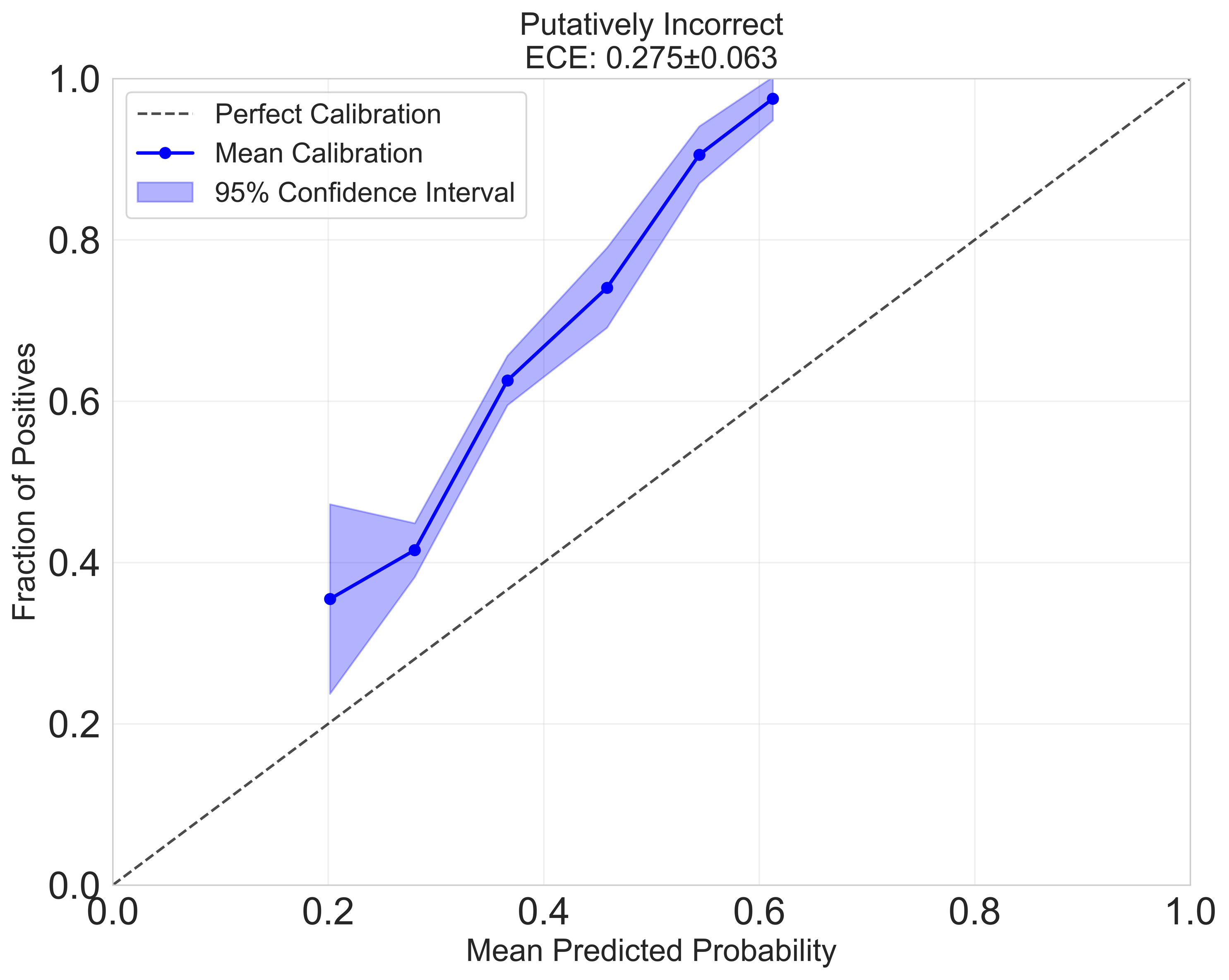}}
  \caption{Reliability diagrams CIFAR-10 with CoAtNet backbone.}
  \label{fig:Reliability_diagram_CoAtNet_Cifar10}
\end{figure}

\begin{figure}[!htbp]
  \centering
  \captionsetup[subfloat]{font=tiny}
  \subfloat[Overall - Non Cal]{\includegraphics[width=0.32\textwidth]{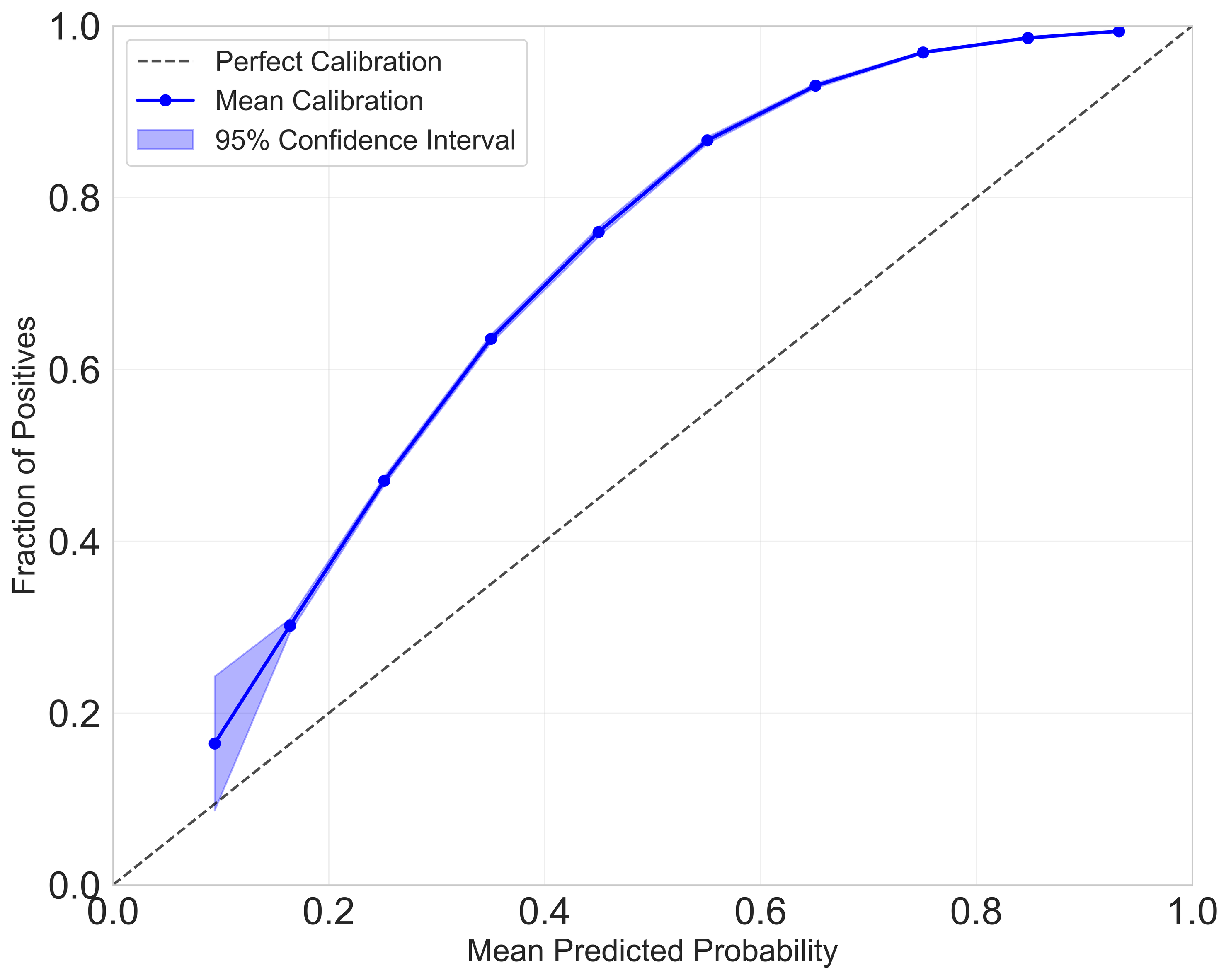}}
  \subfloat[Overall - Iso Cal]{\includegraphics[width=0.32\textwidth]{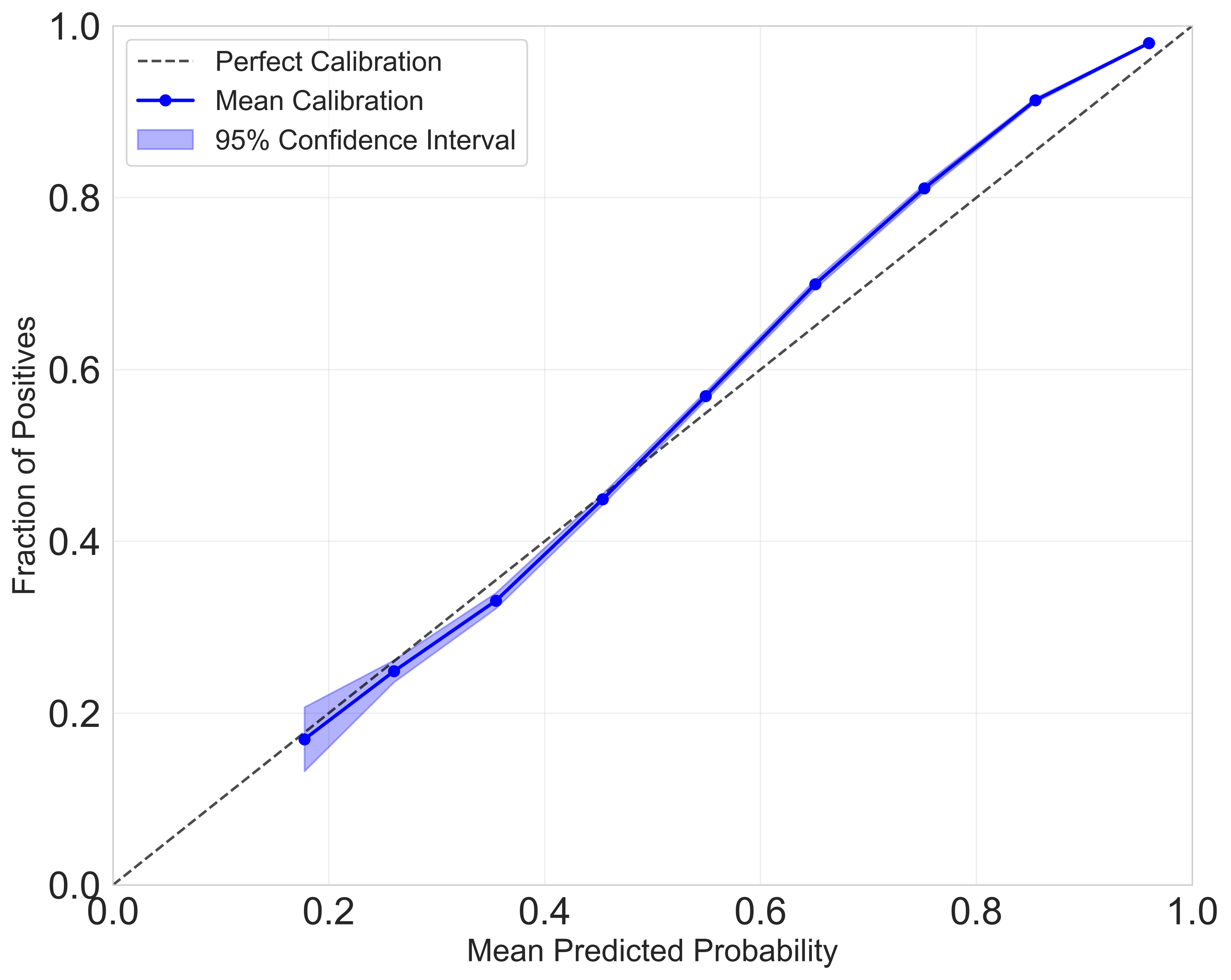}}
  \subfloat[Overall - Dual Cal]{\includegraphics[width=0.32\textwidth]{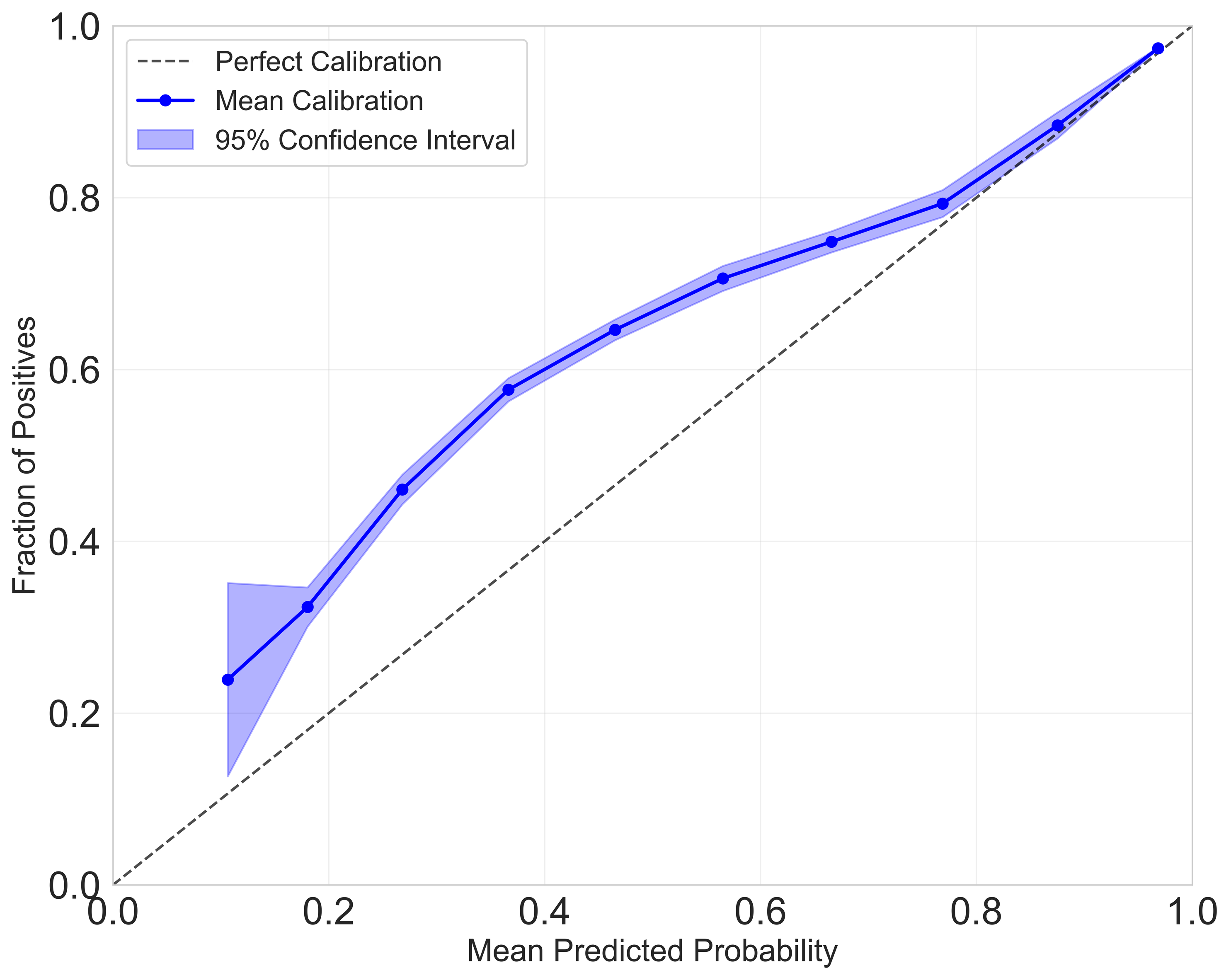}}\par
  \subfloat[Putatively Correct - Non Cal]{\includegraphics[width=0.32\textwidth]{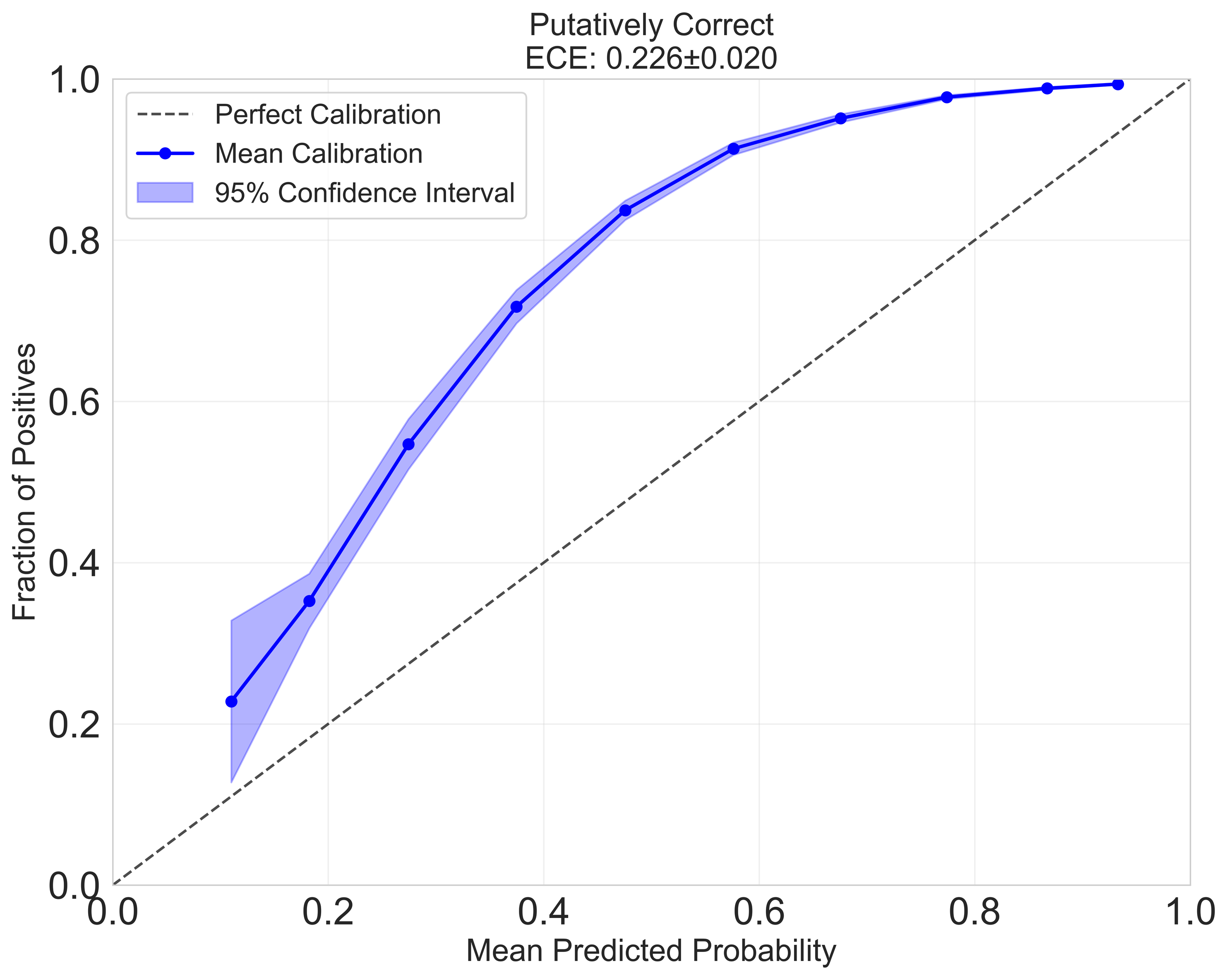}}
  \subfloat[Putatively Correct - Iso Cal]{\includegraphics[width=0.32\textwidth]{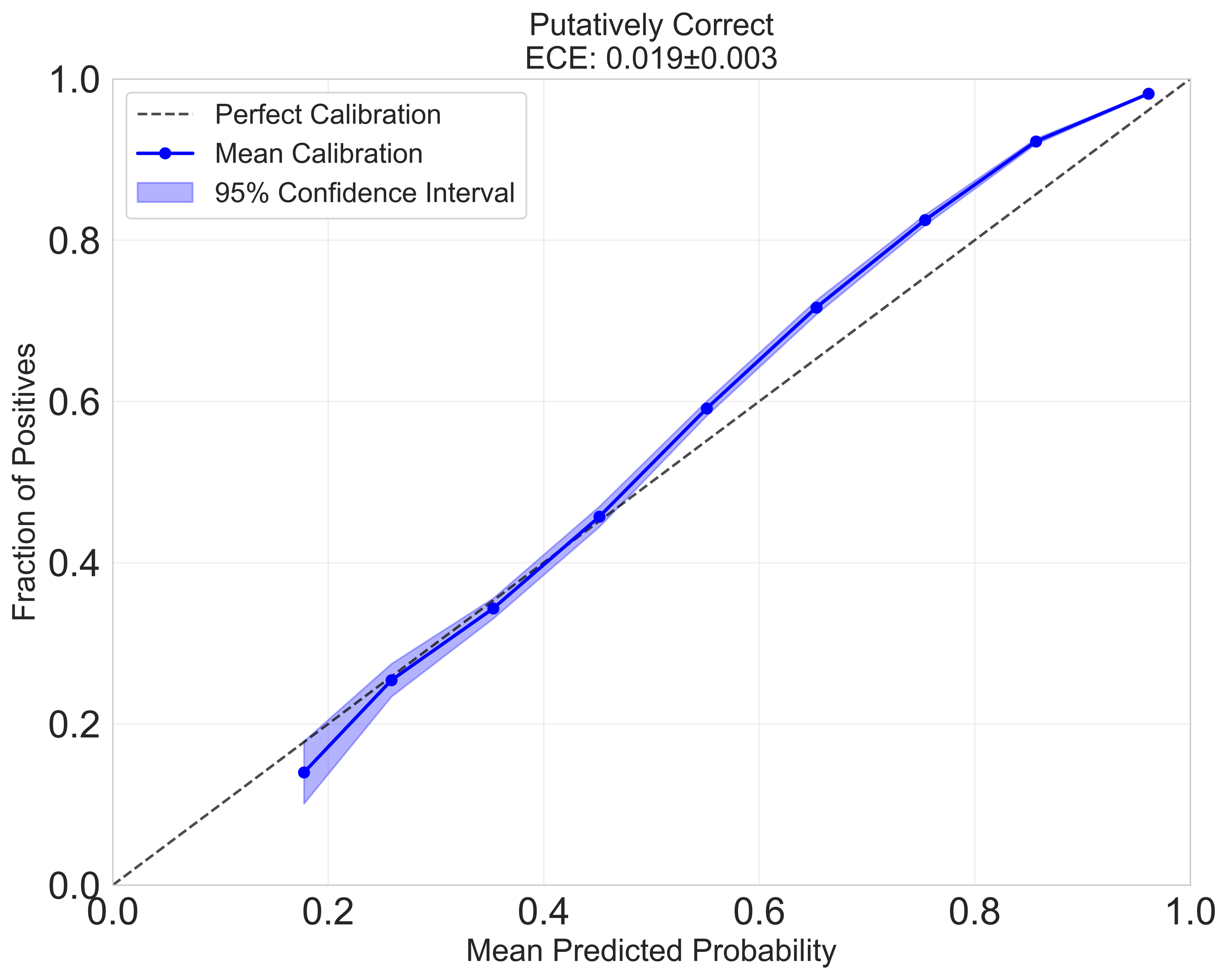}}
  \subfloat[Putatively Correct - Dual Cal]{\includegraphics[width=0.32\textwidth]{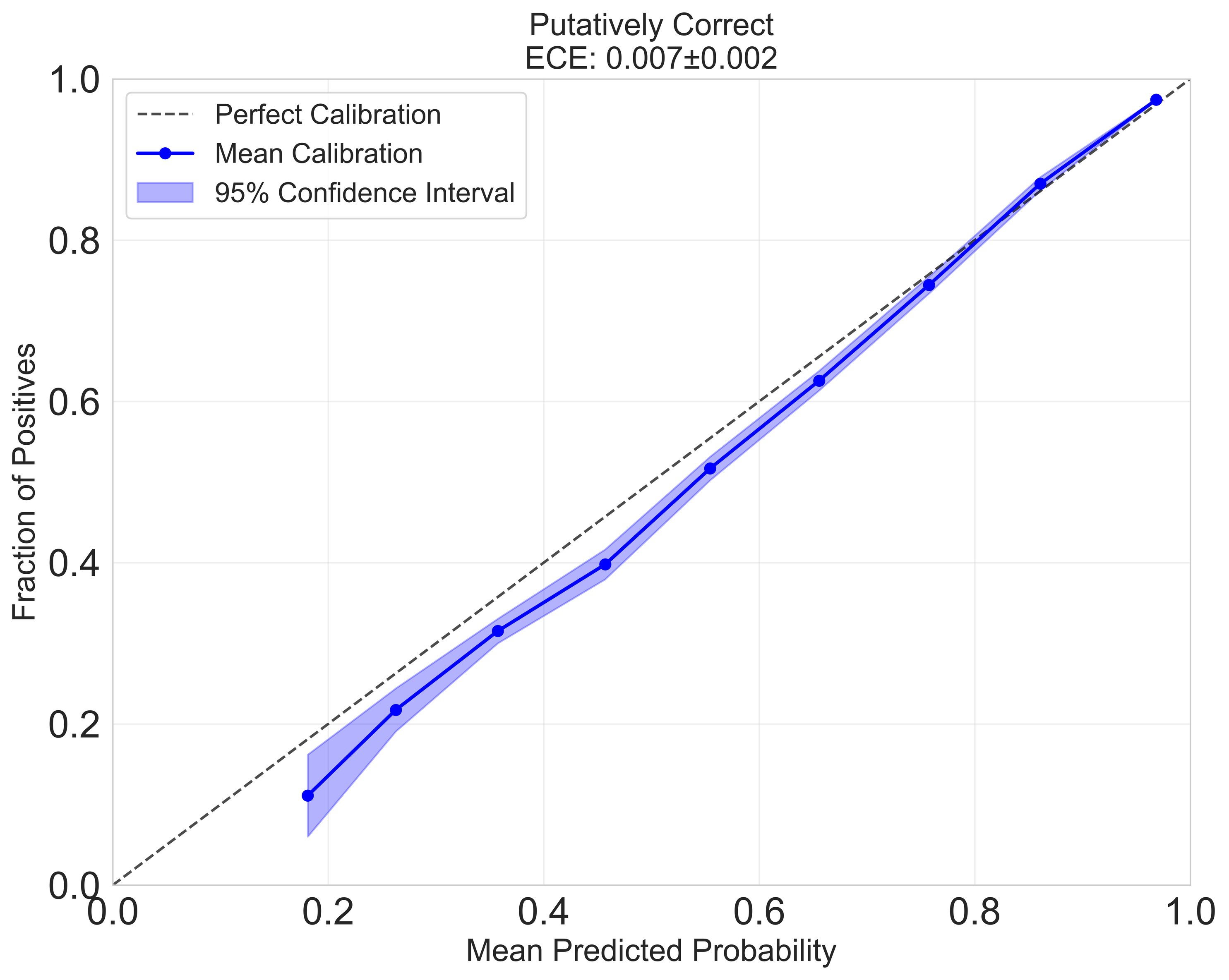}}\par
  \subfloat[Putatively Incorrect - Non Cal]{\includegraphics[width=0.32\textwidth]{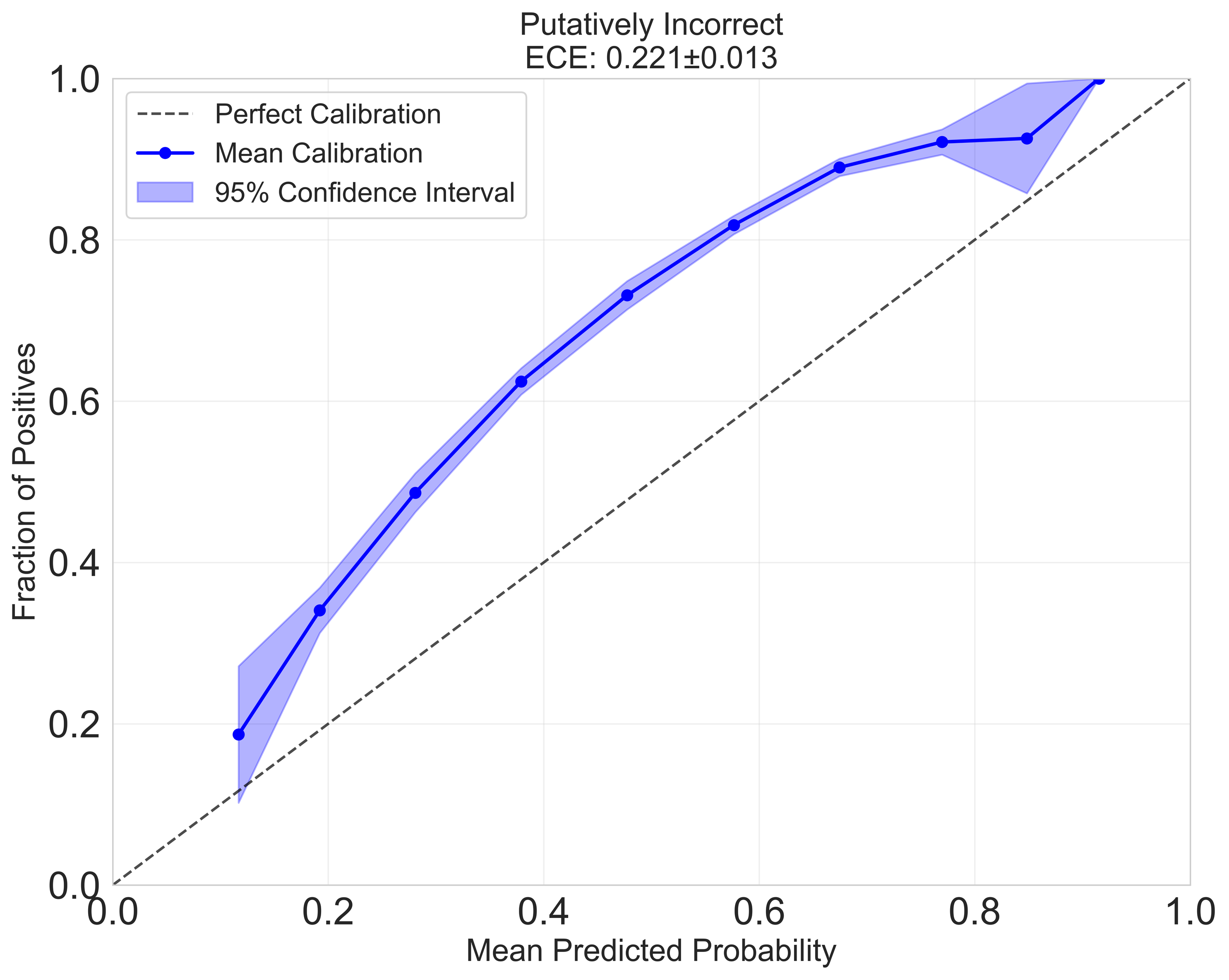}}
  \subfloat[Putatively Incorrect - Iso Cal]{\includegraphics[width=0.32\textwidth]{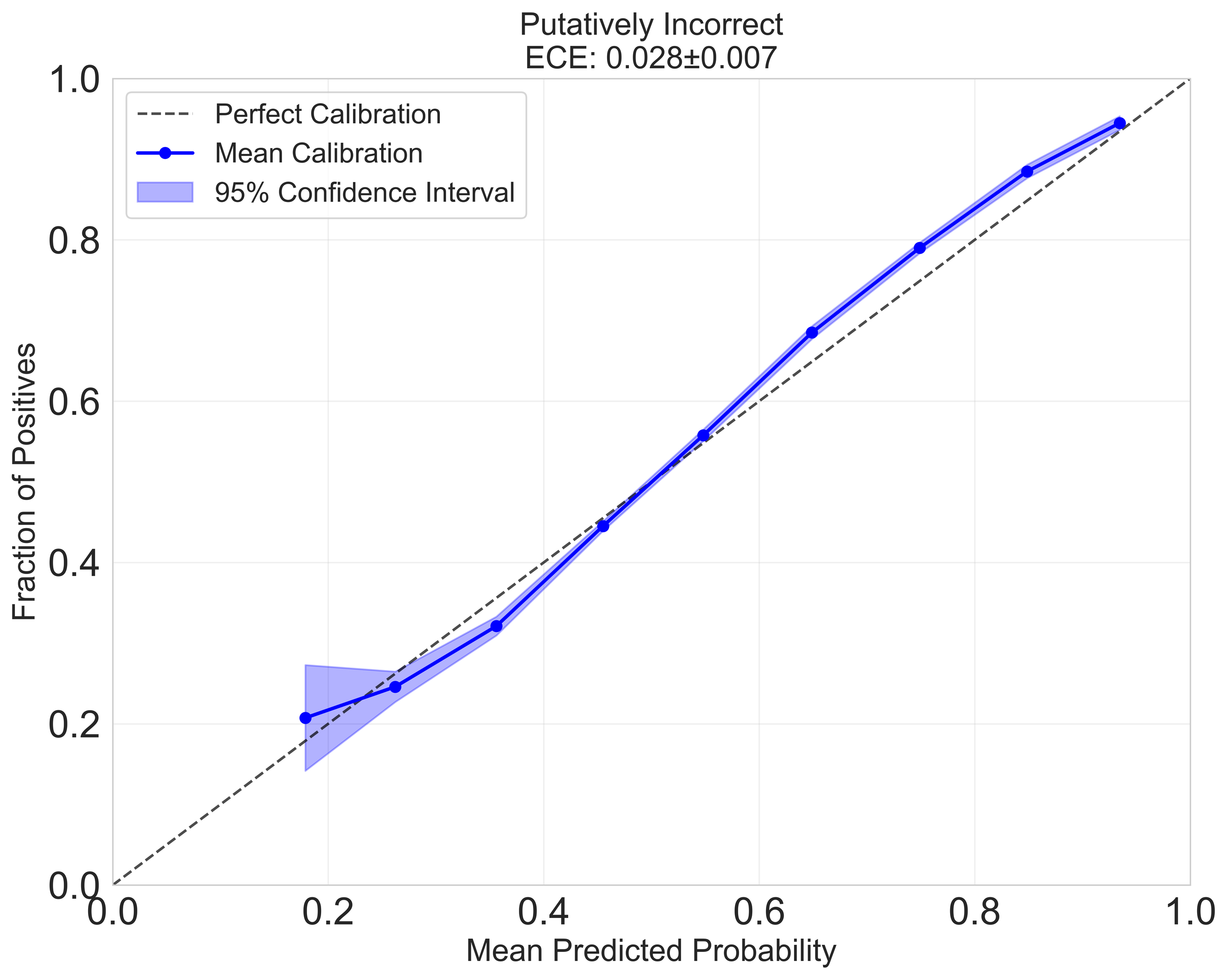}}
  \subfloat[Putatively Incorrect - Dual Cal]{\includegraphics[width=0.32\textwidth]{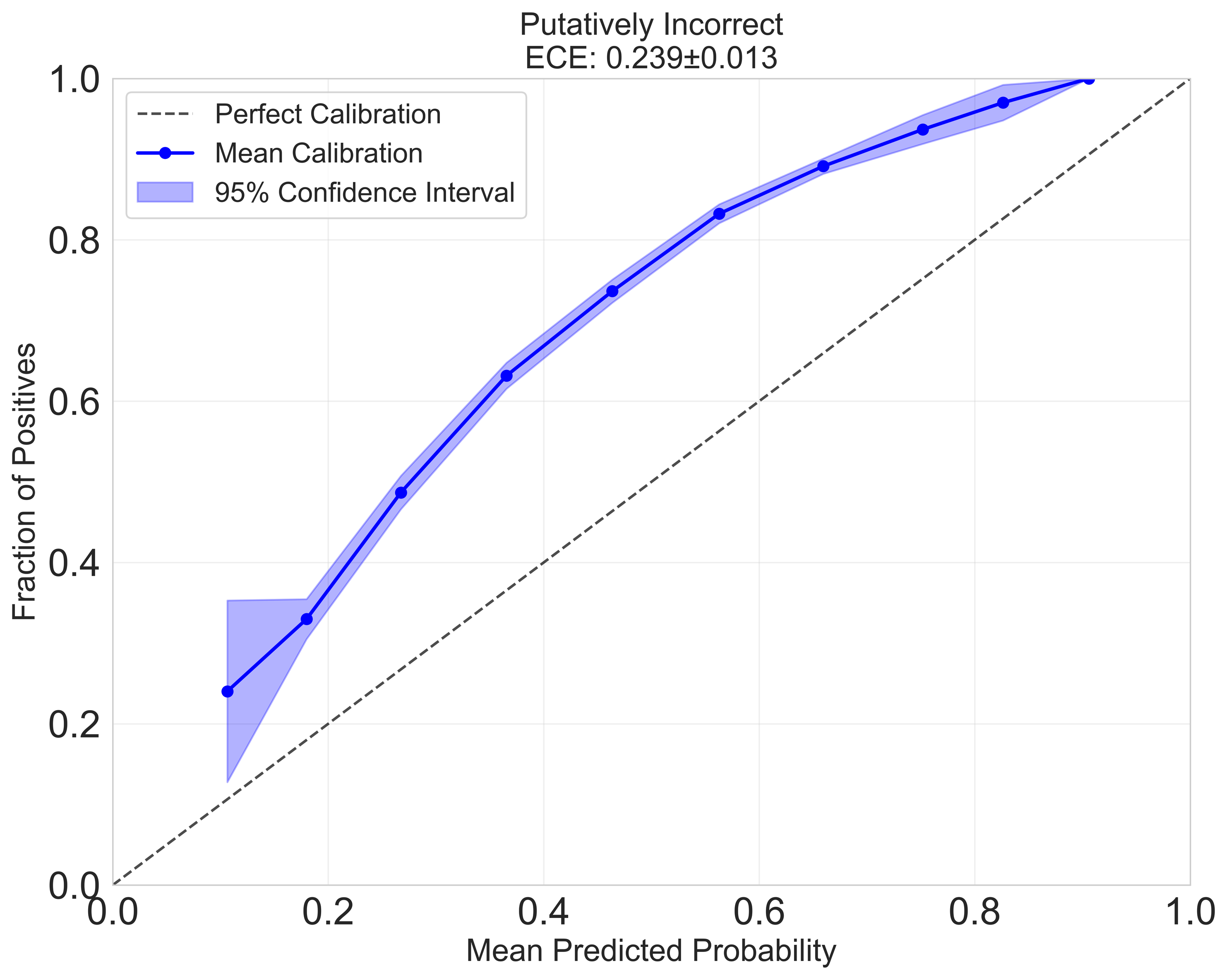}}
  \caption{Reliability diagrams CIFAR-100 superclasses with CoAtNet backbone.}
  \label{fig:Reliability_diagram_CoAtNet_Cifar100_coarse}
\end{figure}

\begin{figure}[!htbp]
  \centering
  \captionsetup[subfloat]{font=tiny}
  \subfloat[Overall - Non Cal]{\includegraphics[width=0.32\textwidth]{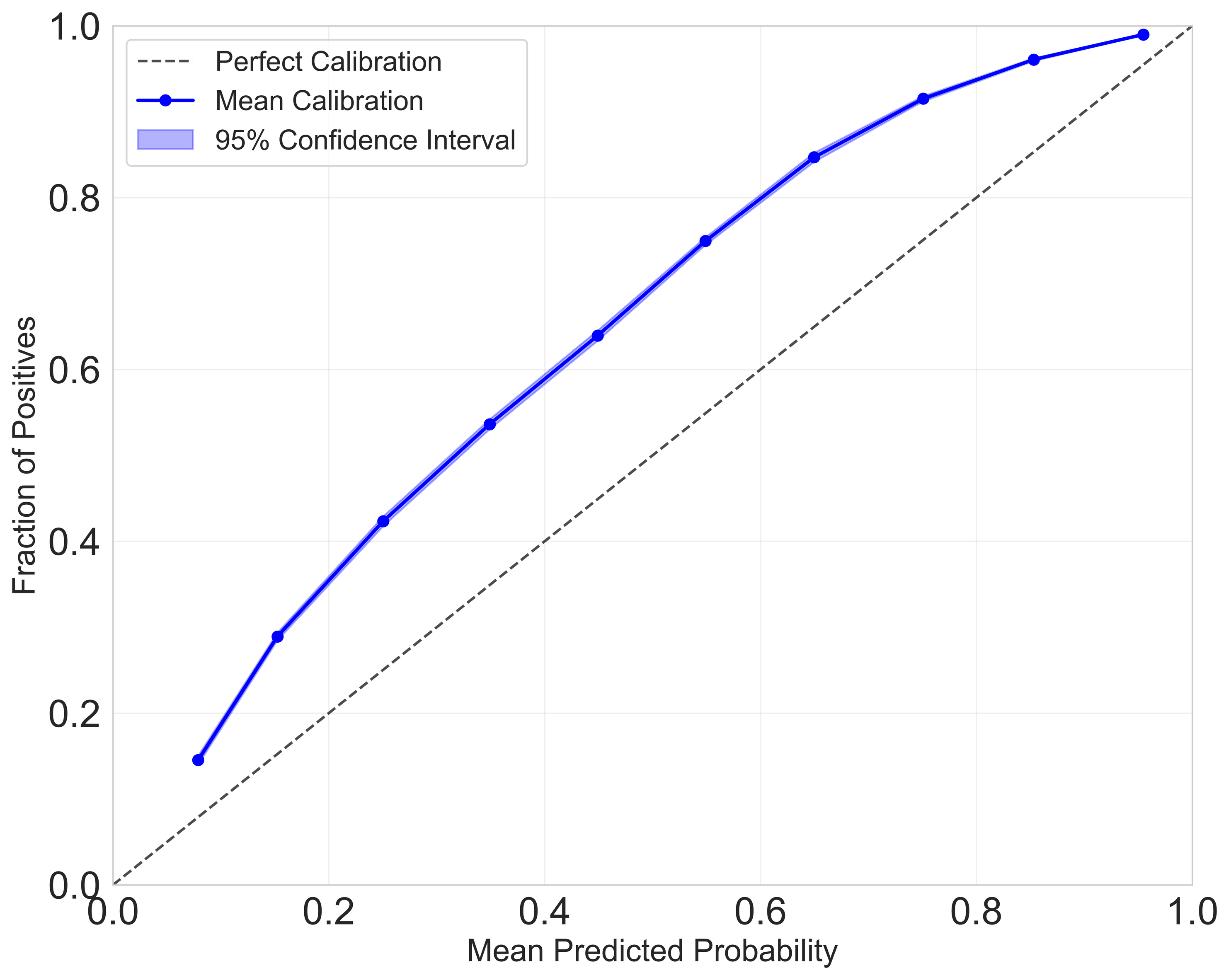}}
  \subfloat[Overall - Iso Cal]{\includegraphics[width=0.32\textwidth]{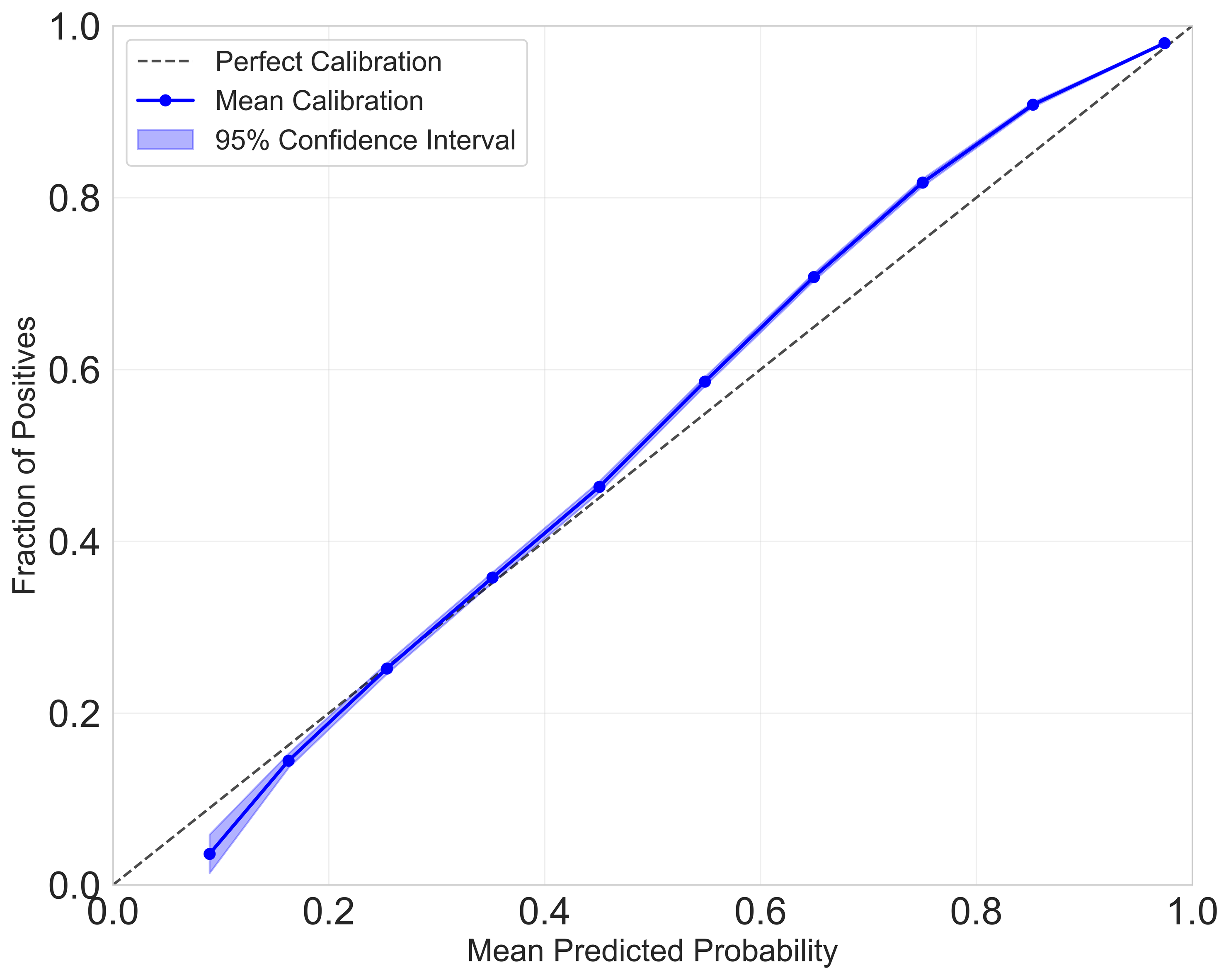}}
  \subfloat[Overall - Dual Cal]{\includegraphics[width=0.32\textwidth]{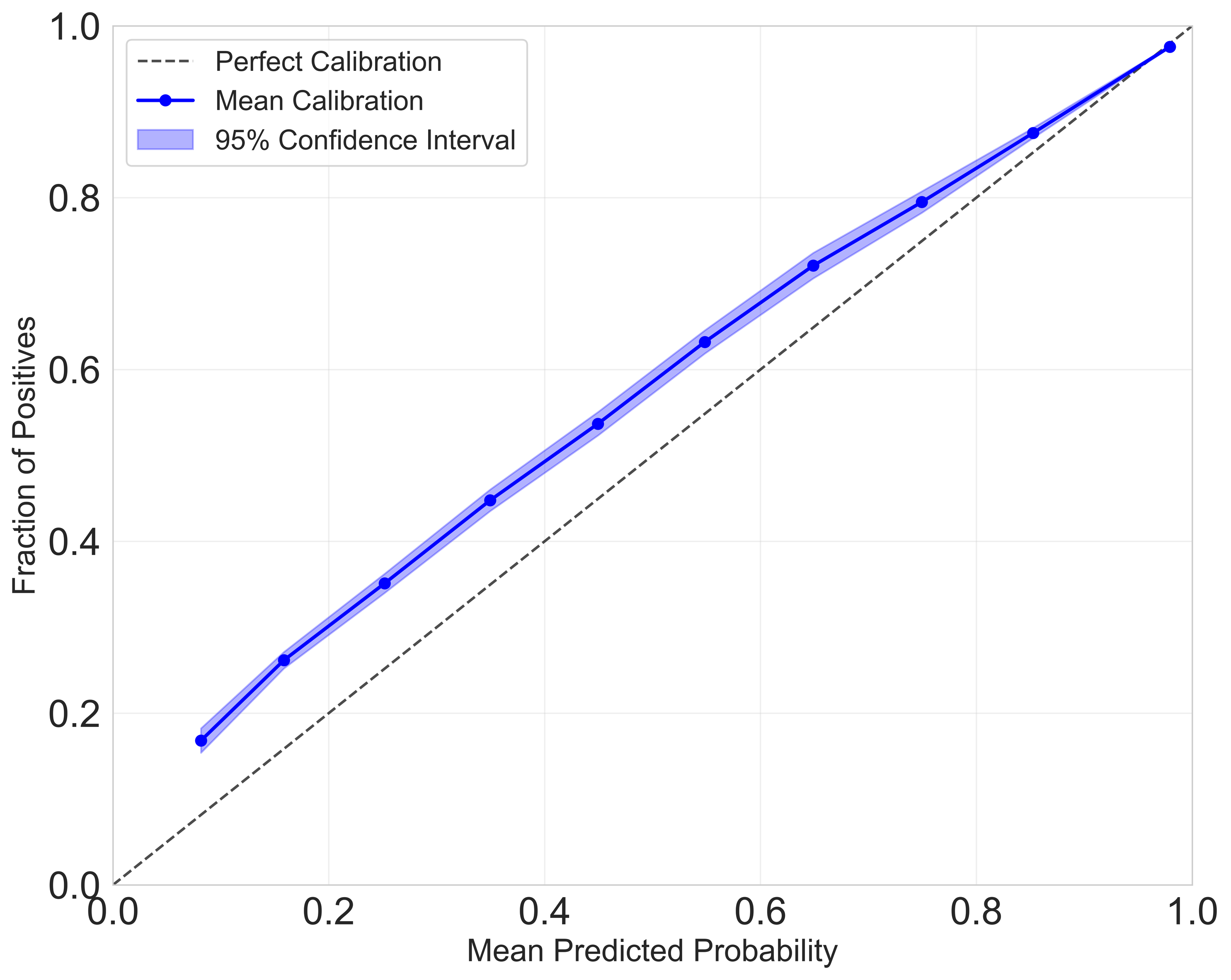}}\par
  \subfloat[Putatively Correct - Non Cal]{\includegraphics[width=0.32\textwidth]{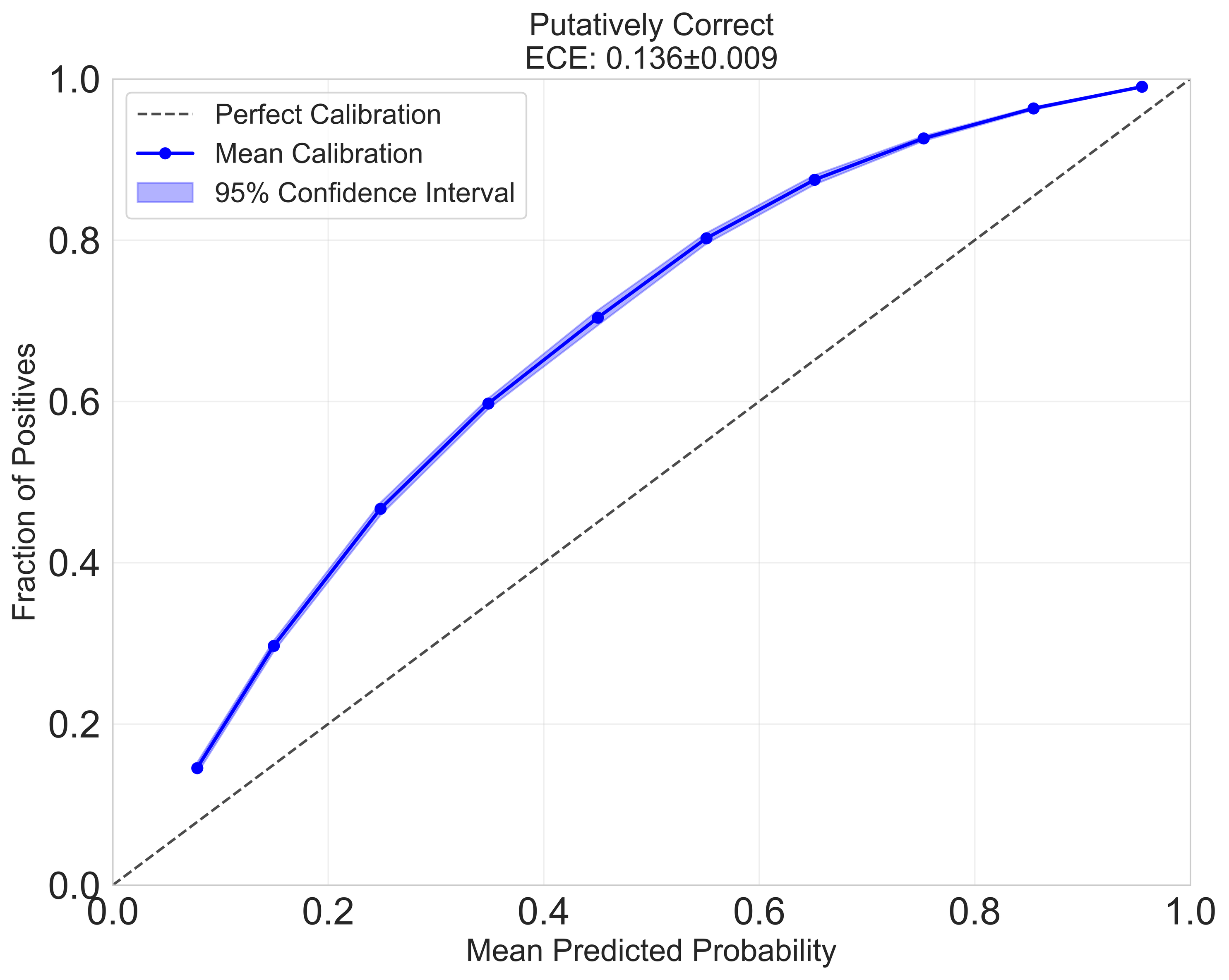}}
  \subfloat[Putatively Correct - Iso Cal]{\includegraphics[width=0.32\textwidth]{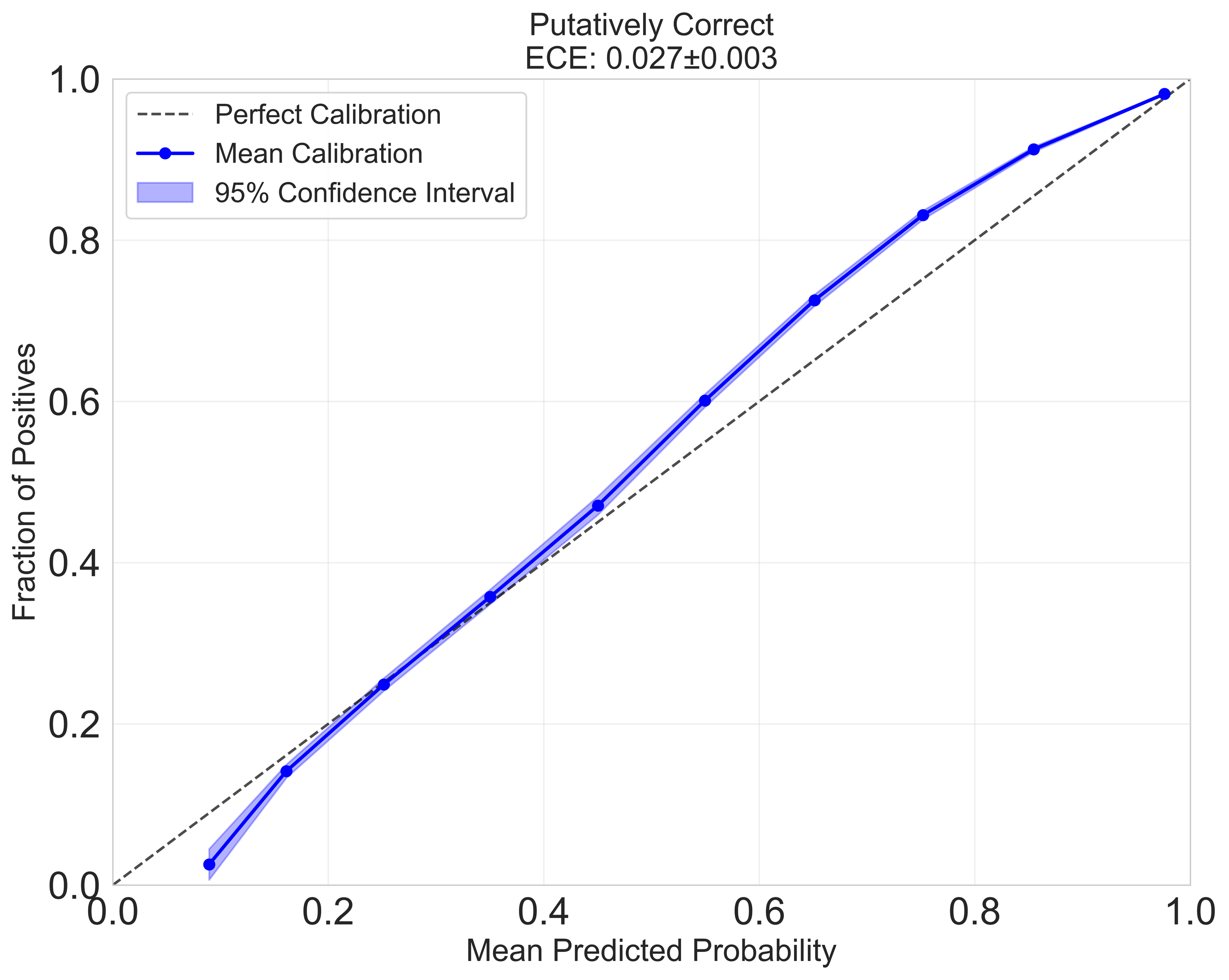}}
  \subfloat[Putatively Correct - Dual Cal]{\includegraphics[width=0.32\textwidth]{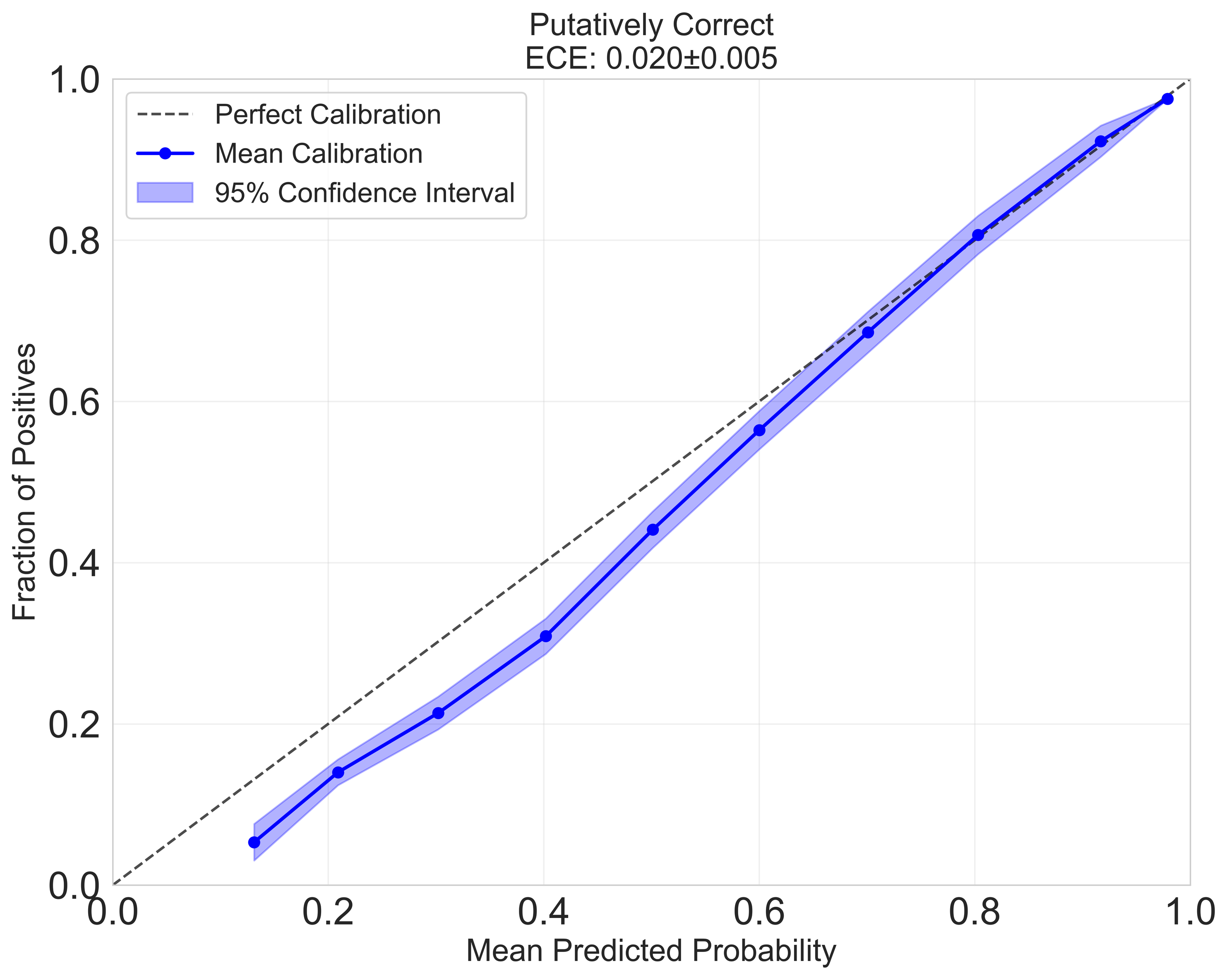}}\par
  \subfloat[Putatively Incorrect - Non Cal]{\includegraphics[width=0.32\textwidth]{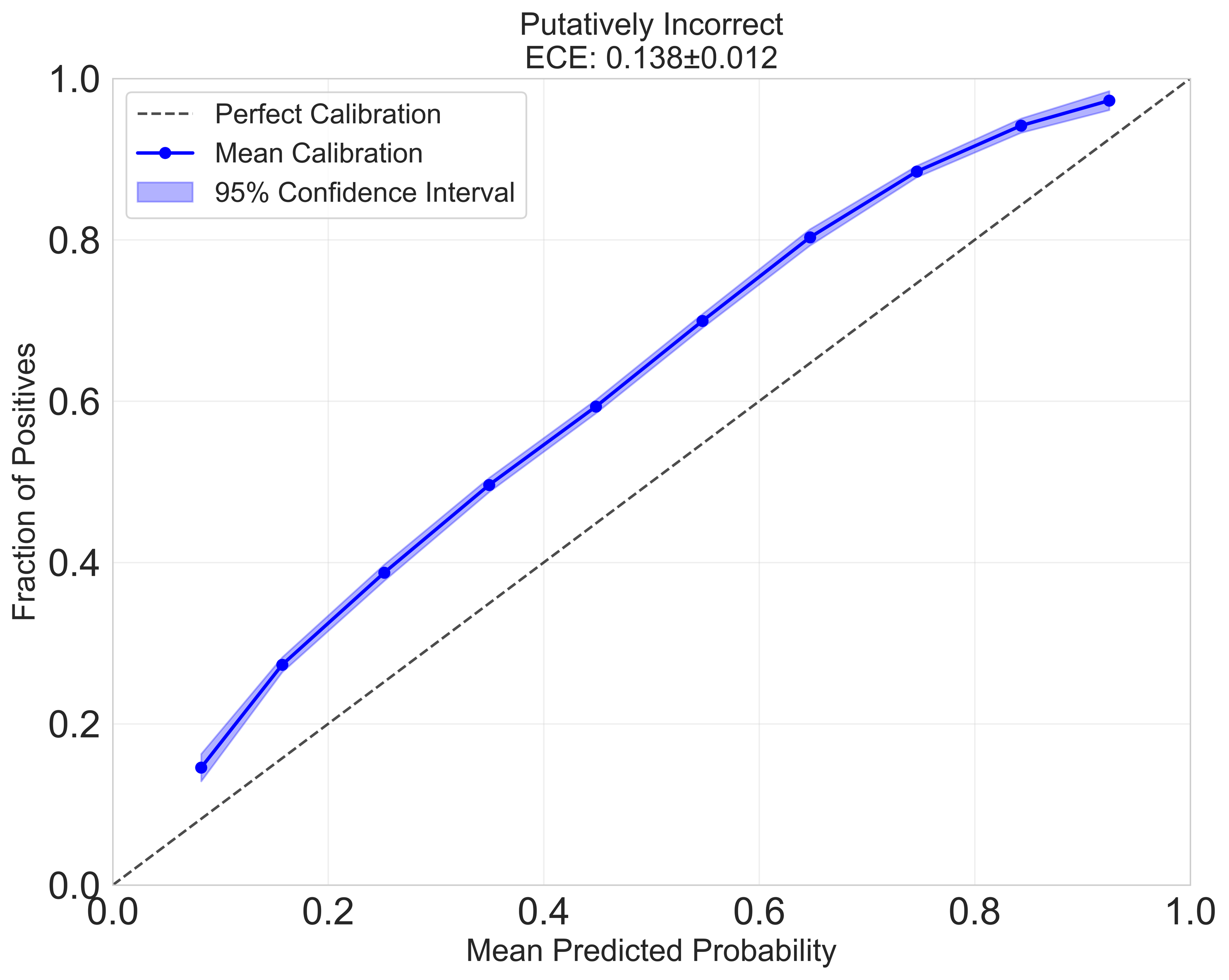}}
  \subfloat[Putatively Incorrect - Iso Cal]{\includegraphics[width=0.32\textwidth]{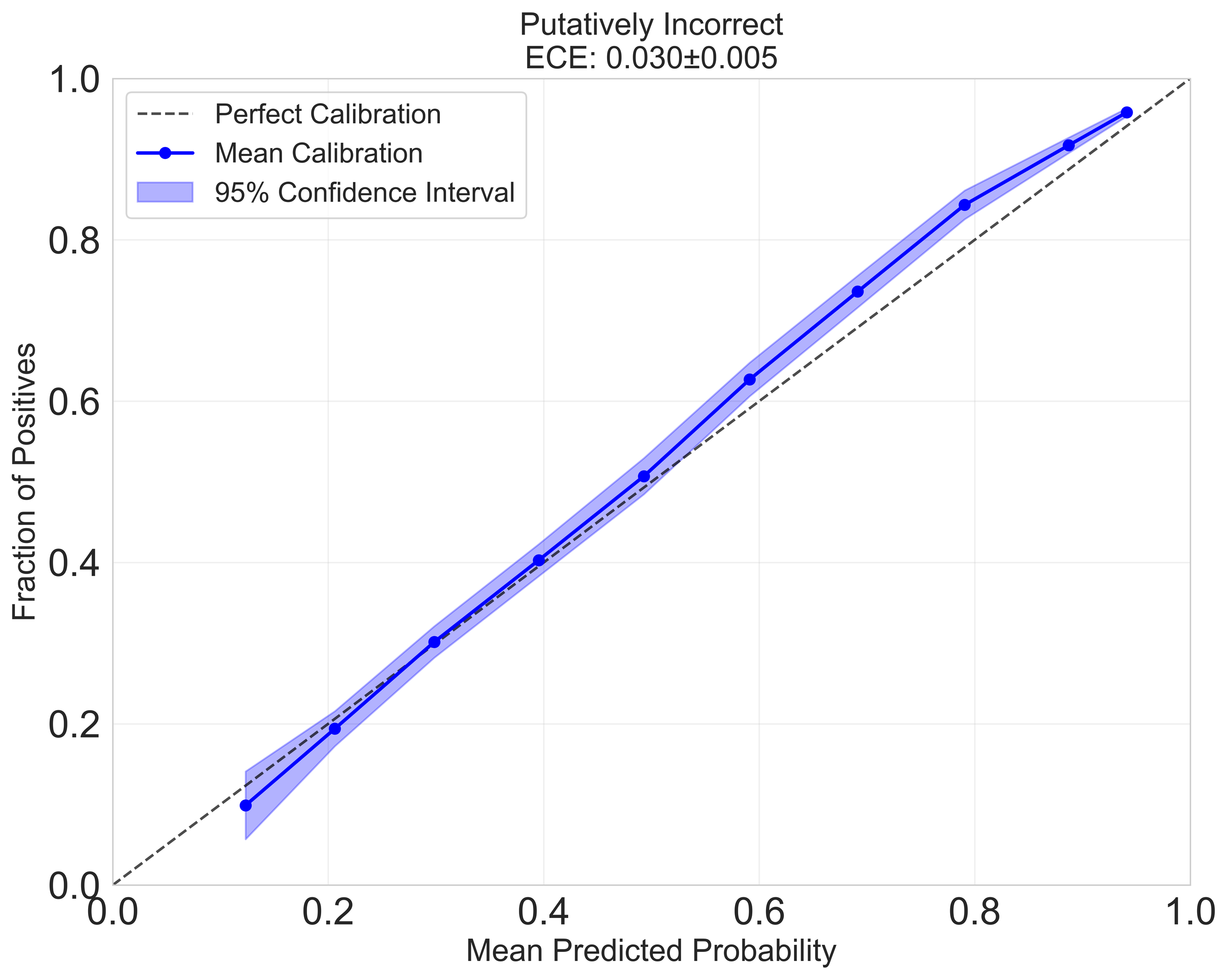}}
  \subfloat[Putatively Incorrect - Dual Cal]{\includegraphics[width=0.32\textwidth]{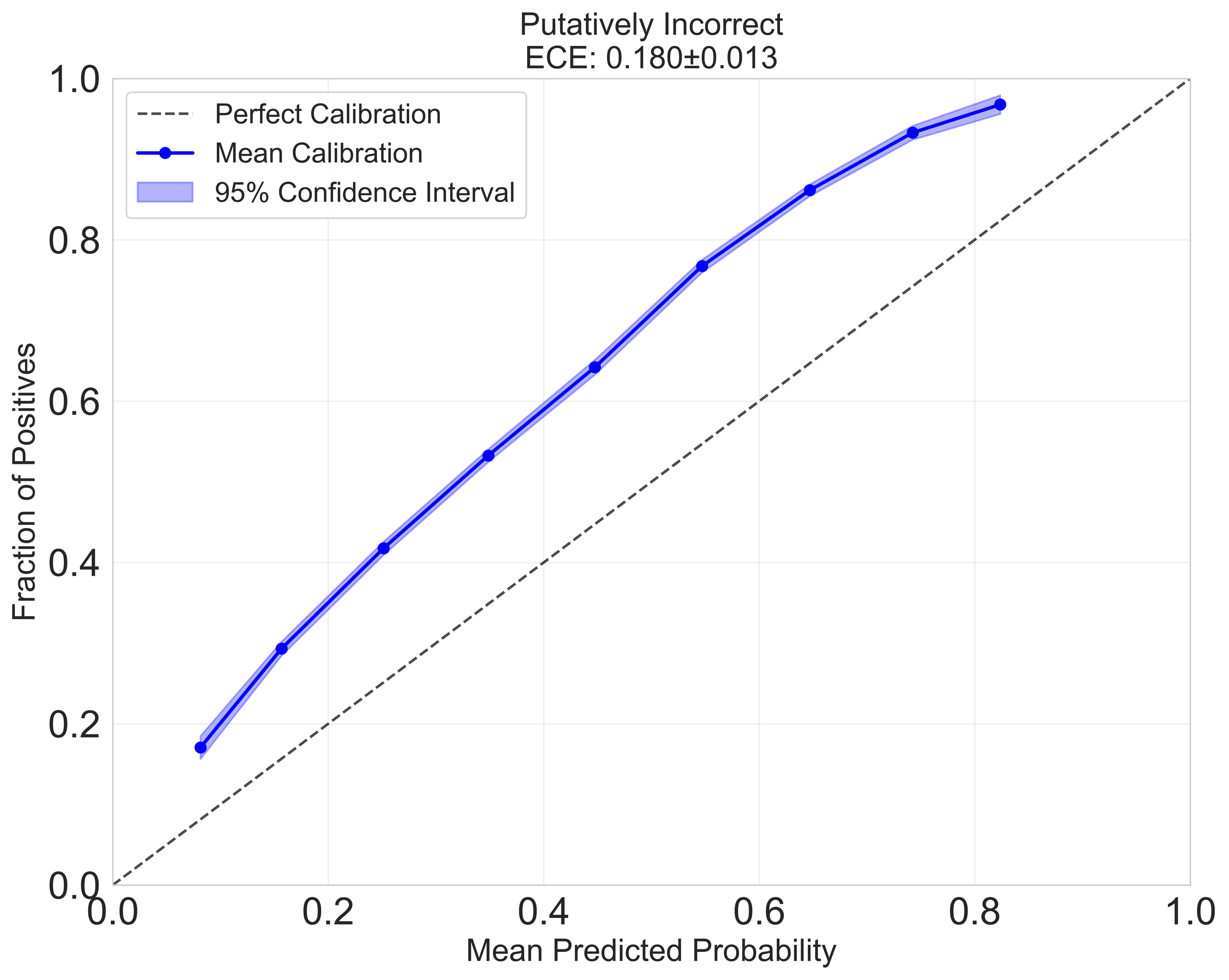}}
  \caption{Reliability diagrams CIFAR-100 fine-grained classes with CoAtNet backbone.}
  \label{fig:Reliability_diagram_CoAtNet_Cifar100F}
\end{figure}

According to the Friedman test (Table~\ref{tab:ece_stat_test}), isotonic calibration achieved the lowest overall ECE across all dataset–backbone combinations outperforming the proposed dual calibration in this metric. However, reliability diagrams illustrated by Figures~\ref{fig:Reliability_diagram_BiT_Cifar10}, \ref{fig:Reliability_diagram_BiT_Cifar100_coarse}, \ref{fig:Reliability_diagram_BiT_Cifar100}, \ref{fig:Reliability_diagram_CoAtNet_Cifar10}, \ref{fig:Reliability_diagram_CoAtNet_Cifar100_coarse} and \ref{fig:Reliability_diagram_CoAtNet_Cifar100F} offer deeper insight into the underlying mechanism of the proposed dual calibration method, explaining the observed increase in ECE.

These figures illustrate the calibration performance before any calibration, utilizing the standard isotonic calibration, and using the proposed dual calibration method. As part of the proposed method, test predictions were stratified into putatively correct and putatively incorrect groups, based on semantic conformal prediction. It should be noted that these groupings are not guaranteed to reflect the true correctness of each sample; rather, they rely on conformity scores, and the precision of this stratification is reported in Table~\ref{tab:conformal_result}.

It should be emphasized that, for the standard isotonic regression, displaying the reliability diagrams separately for the putatively correct and putatively incorrect groups does not imply that the calibrator was fit independently for each group. A single global calibrator was fitted across all samples; the same partitioning was then applied later solely for visualization and direct comparison with the proposed dual calibration method.

As shown in the reliability diagrams, the isotonic calibration method, applied uniformly across all samples, resulted in well-calibrated reliability diagrams with low ECE across both overall and group-specific views. In contrast, the dual calibration method applied isotonic regression separately within the two subsets. For the putatively correct group, calibration closely followed the standard isotonic regression behavior, yielding similar calibration curves and ECE values. However, for the putatively incorrect group, the method intentionally shifted predicted probabilities downward—often below 0.6—to reduce confidence predictions. This pattern of downward shift of mean predicted probability in reliability curves of the putatively incorrect group was consistently observed across other dataset–backbone pairs as well. Notably, in CIFAR-10 and CIFAR-100 superclasses with BiT ( Figures~\ref{fig:Reliability_diagram_BiT_Cifar10}, \ref{fig:Reliability_diagram_BiT_Cifar100_coarse}), CIFAR-10 and CIFAR-100 fine-grained with CoAtNet (Figures~\ref{fig:Reliability_diagram_CoAtNet_Cifar10}, \ref{fig:Reliability_diagram_CoAtNet_Cifar100F}). In this context, apparent overconfidence in the mid-range is inconsequential, since the goal is not to achieve precise probabilistic calibration but to intentionally reduce the confidence of samples expected to be incorrect. This downward adjustment increases predictive entropy, enabling these samples to be effectively recognized as uncertain during the uncertainty-aware decision process. Thus, the increased ECE should be interpreted not as a flaw, but as a designed trade-off that prioritizes trust calibration and detection of potentially erroneous predictions.


Although the effectiveness of the proposed dual calibration method is inherently dependent on the accuracy of the conformal prediction-based stratification step. Table~\ref{tab:conformal_result} reports the proportion of test samples flagged as putatively correct or incorrect, along with the accuracy of these groupings. For most dataset–backbone combinations, the putatively correct group constitutes the majority of the test set and exhibits high stratification accuracy, thereby enabling the proposed method to operate as intended. However, in the particularly challenging case of CIFAR-100 fine-grained classes with the BiT backbone, 72\% of the test samples were flagged as putatively incorrect, and only approximately 50\% of these were truly misclassified. This imbalance—characterized by a small, high-accuracy putatively correct group and a large, low-accuracy putatively incorrect group—limited the effectiveness of dual calibration and explains the near-overlap between the non-calibrated and dual calibration reliability diagrams in Figure.~\ref{fig:Reliability_diagram_BiT_Cifar100}. As expected, the performance of the proposed method is directly linked to the quality of the underlying conformal stratification.

The design choice of training two separate isotonic calibrators, rather than a single unified one, is also closely tied to this stratification structure. The putatively correct and putatively incorrect groups correspond to distinct reliability regimes with differing statistical properties. The putatively correct group contains predictions with high empirical accuracy, requiring only minor calibration adjustments to refine confidence alignment. In contrast, the putatively incorrect group consists of systematically unreliable predictions that necessitate deliberate confidence suppression to reduce false certainty and improve decision safety. As isotonic regression enforces a single monotonic mapping, the dominant distribution—typically the high-accuracy group—would bias the outcome, weakening both calibration quality and uncertainty discrimination. By maintaining separate calibrators, the proposed framework allows each model to specialize in its respective reliability domain, thereby achieving a principled balance between probabilistic calibration and uncertainty-aware safety.

\begin{table}[!htbp]
\centering
\caption{Test set stratification by conformal prediction: percentage of samples flagged as putatively correct or incorrect, along with corresponding conformal stratification performance.}
\label{tab:conformal_result}
\begin{tabular}{llllll}
\hline
\multicolumn{1}{c}{Dataset} & \multicolumn{1}{c}{Backbone} & \multicolumn{1}{c}{\begin{tabular}[c]{@{}c@{}}Putative correct\\ group size\end{tabular}} & \multicolumn{1}{c}{\begin{tabular}[c]{@{}c@{}}Putative correct\\ group accuracy\end{tabular}} & \multicolumn{1}{c}{\begin{tabular}[c]{@{}c@{}}Putative incorrect\\ group size\end{tabular}} & \multicolumn{1}{c}{\begin{tabular}[c]{@{}c@{}}Putative incorrect\\ group accuracy\end{tabular}} \\ \hline
CIFAR-10 & BiT & 85.495 ± 2.065 & 95.378 ± 0.486 & 14.504 ± 2.065 & 66.804 ± 2.029 \\
CIFAR-10 & CoAtNet & 98.149 ± 0.906 & 93.744 ± 0.272 & 1.850 ± 0.906 & 60.001 ± 4.777 \\
CIFAR-100-S & BiT & 52.866 ± 1.959 & 93.834 ± 0.634 & 47.133 ± 1.959 & 61.946 ± 0.998 \\
CIFAR-100-S & CoAtNet & 69.829 ± 5.148 & 87.727 ± 0.976 & 30.170 ± 5.148 & 62.724 ± 1.817 \\
CIFAR-100-F & BiT & 27.172 ± 2.095 & 93.863 ± 0.881 & 72.827 ± 2.095 & 58.497 ± 0.925 \\
CIFAR-100-F & CoAtNet & 65.62 ± 7.953 & 79.470 ± 2.220 & 34.38 ± 7.953 & 56.674 ± 1.799 \\ \hline
\end{tabular}%
\end{table}

The ultimate goal of the proposed method is framed in terms of trustworthiness improvement through uncertainty-aware predictions. Accordingly, its effectiveness in uncertainty-aware decision making is assessed in the following section. 

To effectively evaluate the proposed dual calibration method's performance in reducing confidently incorrect predictions, the analysis primarily focused on two key metrics: FC\% and UG-Mean. The FC\% directly quantifies the proportion of incorrect predictions that are inappropriately flagged as certain, representing the most critical failure mode for safety-critical applications. However, evaluating FC\% in isolation could be misleading, as methods might trivially achieve low FC by indiscriminately flagging most predictions as uncertain, thereby sacrificing operational utility. The UG-Mean metric addresses this concern by providing a balanced measure that combines the UTPR, which captures the model's ability to confidently classify correct predictions, with the complement of the (1-UFPR), which represents the capacity to appropriately flag incorrect predictions as uncertain. This dual-metric approach ensures that FC reduction is achieved through intelligent uncertainty-awareness rather than excessive conservatism. 

Furthermore, the non-calibrated baseline's occasionally superior FC performance must be contextualized within its poor calibration quality (high ECE), as uncalibrated probability distributions lack the reliability necessary for consistent uncertainty-aware decision-making in practical deployments. The analysis therefore, examines whether the proposed method achieves meaningful FC reduction while maintaining competitive UG-Mean scores, thereby demonstrating both enhanced safety through reduced confident errors and preserved operational efficiency through appropriate confidence in correct predictions. Since uncertainty-aware metrics (FC\%, UG-Mean) are threshold-dependent, requiring an entropy cutoff to categorize predictions as certain or uncertain, the evaluation was performed at multiple threshold values ($\tau$ = 0.2, 0.3, 0.4, 0.5, 0.6) to comprehensively assess calibration method performance across different decision stringency levels.

Table~\ref{tab:uncertainty_performance_02} reports the uncertainty-aware performance at an entropy threshold of 0.2. At this highly conservative setting, FC values remained below 1\% for most methods, while TC rates were also low—indicating that such a stringent threshold is unsuitable for evaluating practical uncertainty-aware decision-making.

The non-calibrated baseline yielded the lowest FC\% values in several configurations, achieving 0.045\% for CIFAR-10 with BiT and 0.055\% for CIFAR-100-S with CoAtNet. However, these apparently strong safety scores coincided with very poor TC performance (40.072\% and 6.607\%, respectively), revealing that the low FC rates stemmed from excessive overall uncertainty rather than selective identification of unreliable predictions.

In contrast, the proposed dual calibration method maintained a more balanced behavior. For instance, on CIFAR-10 with BiT, it achieved an FC of 0.422\%—intermediate between the non-calibrated (0.045\%) and focal loss (1.158\%) baselines—while attaining substantially higher TC (63.601\%) than the non-calibrated model. The UG-Mean metric further confirmed this balance, with the dual calibration method scoring 81.426 for CIFAR-10 BiT compared to 66.082 for non-calibrated and 80.928 for standard isotonic calibration, demonstrating competitive performance even without minimizing FC.

The focal loss method exhibited the highest variability, with FC values ranging from 0.313\% (CIFAR-100-F BiT) to 1.158\% (CIFAR-10 BiT). Moreover, its TC rates were consistently lower than those of post-hoc calibration methods—most notably in CIFAR-100-F BiT, where TC dropped to 9.347\% despite maintaining a low FC of 0.313\%. This suggests that focal loss tends to over-regularize confidence, producing models that are excessively uncertain and consequently less useful in uncertainty-aware decision contexts.

\begin{sidewaystable}[!htbp]
\centering
\caption{Summary of uncertainty-aware performance metrics for different calibration methods across various backbones at entropy threshold $\tau = 0.2$.}
\label{tab:uncertainty_performance_02}
\tiny
\begin{tabular}{lllllllllll}
\hline
\multicolumn{1}{c}{Dataset} & \multicolumn{1}{c}{Calibration} & \multicolumn{1}{c}{Backbone} & \multicolumn{1}{c}{UAcc} & \multicolumn{1}{c}{TC \%} & \multicolumn{1}{c}{TU \%} & \multicolumn{1}{c}{FC \%} & \multicolumn{1}{c}{FU \%} & \multicolumn{1}{c}{UTPR} & \multicolumn{1}{c}{UFPR} & \multicolumn{1}{c}{UG-Mean} \\ \hline
CIFAR-10 & Non Cal & BiT & 48.763 ± 1.297 & 40.072 ± 1.311 & 8.691 ± 0.206 & 0.045 ± 0.021 & 51.191 ± 1.301 & 43.908 ± 1.427 & 0.518 ± 0.239 & 66.082 ± 1.065 \\
CIFAR-10 & Focal loss & BiT & 81.825 ± 2.877 & 73.968 ± 3.463 & 7.857 ± 0.629 & 1.158 ± 0.497 & 17.016 ± 3.329 & 81.292 ± 3.684 & 12.899 ± 5.607 & 84.019 ± 1.445 \\
CIFAR-10 & Iso Cal & BiT & 70.976 ± 0.695 & 62.691 ± 0.714 & 8.284 ± 0.186 & 0.399 ± 0.048 & 28.626 ± 0.706 & 68.653 ± 0.771 & 4.598 ± 0.579 & 80.928 ± 0.454 \\
CIFAR-10 & Dual Cal & BiT & 71.902 ± 0.769 & 63.601 ± 0.81 & 8.301 ± 0.211 & 0.422 ± 0.058 & 27.676 ± 0.799 & 69.679 ± 0.873 & 4.838 ± 0.673 & 81.426 ± 0.431 \\
CIFAR-100-S & Non Cal & BiT & 49.013 ± 0.746 & 28.237 ± 0.808 & 20.776 ± 0.26 & 0.414 ± 0.057 & 50.573 ± 0.773 & 35.829 ± 0.999 & 1.955 ± 0.274 & 59.263 ± 0.783 \\
CIFAR-100-S & Focal loss & BiT & 57.796 ± 1.24 & 35.838 ± 1.31 & 21.959 ± 0.38 & 0.87 ± 0.141 & 41.334 ± 1.296 & 46.439 ± 1.666 & 3.805 ± 0.568 & 66.825 ± 1.14 \\
CIFAR-100-S & Iso Cal & BiT & 58.366 ± 0.493 & 38.231 ± 0.566 & 20.135 ± 0.266 & 0.907 ± 0.083 & 40.727 ± 0.522 & 48.419 ± 0.673 & 4.312 ± 0.403 & 68.065 ± 0.427 \\
CIFAR-100-S & Dual Cal & BiT & 59.278 ± 0.868 & 38.9 ± 0.881 & 20.378 ± 0.266 & 0.854 ± 0.099 & 39.869 ± 0.908 & 49.386 ± 1.124 & 4.022 ± 0.471 & 68.842 ± 0.724 \\
CIFAR-100-F & Non Cal & BiT & 50.798 ± 0.782 & 19.459 ± 0.755 & 31.339 ± 0.375 & 0.58 ± 0.105 & 48.622 ± 0.832 & 28.582 ± 1.112 & 1.816 ± 0.326 & 52.964 ± 0.99 \\
CIFAR-100-F & Focal loss & BiT & 50.952 ± 0.793 & 9.347 ± 0.661 & 41.606 ± 0.609 & 0.313 ± 0.08 & 48.734 ± 0.818 & 16.092 ± 1.118 & 0.747 ± 0.191 & 39.942 ± 1.347 \\
CIFAR-100-F & Iso Cal & BiT & 58.91 ± 0.738 & 28.764 ± 0.79 & 30.146 ± 0.3 & 1.201 ± 0.097 & 39.889 ± 0.77 & 41.897 ± 1.121 & 3.831 ± 0.314 & 63.469 ± 0.811 \\
CIFAR-100-F & Dual Cal & BiT & 56.975 ± 1.118 & 25.777 ± 1.064 & 31.198 ± 0.384 & 0.836 ± 0.136 & 42.189 ± 1.203 & 37.929 ± 1.62 & 2.609 ± 0.415 & 60.762 ± 1.222 \\
CIFAR-10 & Non Cal & CoAtNet & 75.307 ± 0.469 & 68.957 ± 0.469 & 6.35 ± 0.162 & 0.511 ± 0.04 & 24.182 ± 0.479 & 74.037 ± 0.507 & 7.461 ± 0.641 & 82.771 ± 0.346 \\
CIFAR-10 & Focal loss & CoAtNet & 74.699 ± 0.465 & 68.275 ± 0.478 & 6.424 ± 0.129 & 0.488 ± 0.031 & 24.813 ± 0.475 & 73.344 ± 0.508 & 7.06 ± 0.467 & 82.562 ± 0.292 \\
CIFAR-10 & Iso Cal & CoAtNet & 83.205 ± 0.322 & 77.203 ± 0.288 & 6.002 ± 0.171 & 0.841 ± 0.05 & 15.953 ± 0.326 & 82.875 ± 0.337 & 12.304 ± 0.816 & 85.25 ± 0.431 \\
CIFAR-10 & Dual Cal & CoAtNet & 83.531 ± 0.331 & 77.535 ± 0.265 & 5.996 ± 0.171 & 0.86 ± 0.056 & 15.61 ± 0.335 & 83.242 ± 0.342 & 12.547 ± 0.868 & 85.32 ± 0.465 \\
CIFAR-100-S & Non Cal & CoAtNet & 26.323 ± 0.766 & 6.607 ± 0.813 & 19.715 ± 0.391 & 0.055 ± 0.018 & 73.622 ± 0.774 & 8.234 ± 1.002 & 0.279 ± 0.088 & 28.598 ± 1.798 \\
CIFAR-100-S & Focal loss & CoAtNet & 35.678 ± 0.644 & 13.594 ± 0.758 & 22.084 ± 0.418 & 0.163 ± 0.028 & 64.159 ± 0.648 & 17.481 ± 0.929 & 0.733 ± 0.128 & 41.643 ± 1.096 \\
CIFAR-100-S & Iso Cal & CoAtNet & 57.171 ± 0.871 & 38.285 ± 1.035 & 18.886 ± 0.393 & 0.744 ± 0.072 & 42.085 ± 0.884 & 47.633 ± 1.182 & 3.794 ± 0.393 & 67.689 ± 0.812 \\
CIFAR-100-S & Dual Cal & CoAtNet & 61.378 ± 1.473 & 42.596 ± 1.66 & 18.783 ± 0.427 & 0.941 ± 0.093 & 37.68 ± 1.511 & 53.058 ± 1.963 & 4.779 ± 0.514 & 71.064 ± 1.262 \\
CIFAR-100-F & Non Cal & CoAtNet & 55.726 ± 0.49 & 27.957 ± 0.399 & 27.769 ± 0.333 & 0.672 ± 0.063 & 43.602 ± 0.495 & 39.069 ± 0.571 & 2.364 ± 0.222 & 61.76 ± 0.446 \\
CIFAR-100-F & Focal loss & CoAtNet & 47.448 ± 0.547 & 14.794 ± 0.575 & 32.654 ± 0.286 & 0.232 ± 0.034 & 52.319 ± 0.565 & 22.043 ± 0.834 & 0.706 ± 0.104 & 46.775 ± 0.872 \\
CIFAR-100-F & Iso Cal & CoAtNet & 66.115 ± 0.485 & 39.956 ± 0.408 & 26.159 ± 0.329 & 1.845 ± 0.088 & 32.04 ± 0.461 & 55.498 ± 0.571 & 6.59 ± 0.332 & 71.999 ± 0.414 \\
CIFAR-100-F & Dual Cal & CoAtNet & 64.203 ± 2.015 & 36.829 ± 2.156 & 27.374 ± 0.364 & 1.286 ± 0.195 & 34.511 ± 2.184 & 51.626 ± 3.031 & 4.486 ± 0.685 & 70.181 ± 1.922 \\ \hline
\end{tabular}
\end{sidewaystable}

\begin{sidewaystable}[!htbp]
\centering
\caption{Summary of uncertainty-aware performance metrics for different calibration methods across various backbones at entropy threshold $\tau = 0.3$.}
\label{tab:uncertainty_performance_03}
\tiny
\begin{tabular}{lllllllllll}
\hline
\multicolumn{1}{c}{Dataset} & \multicolumn{1}{c}{Calibration} & \multicolumn{1}{c}{Backbone} & \multicolumn{1}{c}{UAcc} & \multicolumn{1}{c}{TC \%} & \multicolumn{1}{c}{TU \%} & \multicolumn{1}{c}{FC \%} & \multicolumn{1}{c}{FU \%} & \multicolumn{1}{c}{UTPR} & \multicolumn{1}{c}{UFPR} & \multicolumn{1}{c}{UG-Mean} \\ \hline
CIFAR-10 & Non Cal & BiT & 67.373 ± 0.903 & 58.958 ± 0.936 & 8.415 ± 0.208 & 0.322 ± 0.047 & 32.305 ± 0.91 & 64.602 ± 1.003 & 3.686 ± 0.543 & 78.877 ± 0.612 \\
CIFAR-10 & Focal loss & BiT & 86.765 ± 1.809 & 79.961 ± 2.518 & 6.804 ± 0.748 & 2.211 ± 0.611 & 11.024 ± 2.372 & 87.88 ± 2.624 & 24.615 ± 7.055 & 81.229 ± 2.999 \\
CIFAR-10 & Iso Cal & BiT & 80.535 ± 0.454 & 72.989 ± 0.579 & 7.546 ± 0.203 & 1.137 ± 0.103 & 18.328 ± 0.53 & 79.929 ± 0.587 & 13.105 ± 1.221 & 83.335 ± 0.389 \\
CIFAR-10 & Dual Cal & BiT & 79.653 ± 0.798 & 71.93 ± 0.827 & 7.723 ± 0.226 & 1.0 ± 0.094 & 19.347 ± 0.839 & 78.804 ± 0.911 & 11.47 ± 1.115 & 83.521 ± 0.537 \\
CIFAR-100-S & Non Cal & BiT & 61.54 ± 0.578 & 42.079 ± 0.662 & 19.461 ± 0.281 & 1.729 ± 0.138 & 36.731 ± 0.664 & 53.393 ± 0.827 & 8.16 ± 0.664 & 70.022 ± 0.406 \\
CIFAR-100-S & Focal loss & BiT & 68.629 ± 0.968 & 48.389 ± 1.114 & 20.24 ± 0.401 & 2.589 ± 0.273 & 28.782 ± 1.047 & 62.703 ± 1.363 & 11.334 ± 1.101 & 74.555 ± 0.766 \\
CIFAR-100-S & Iso Cal & BiT & 70.111 ± 0.459 & 52.3 ± 0.513 & 17.812 ± 0.299 & 3.23 ± 0.163 & 26.658 ± 0.478 & 66.237 ± 0.603 & 15.354 ± 0.805 & 74.876 ± 0.413 \\
CIFAR-100-S & Dual Cal & BiT & 64.73 ± 0.997 & 45.018 ± 1.105 & 19.712 ± 0.289 & 1.52 ± 0.165 & 33.75 ± 1.101 & 57.152 ± 1.392 & 7.158 ± 0.779 & 72.835 ± 0.721 \\
CIFAR-100-F & Non Cal & BiT & 60.954 ± 0.804 & 31.467 ± 0.854 & 29.488 ± 0.347 & 2.431 ± 0.212 & 36.614 ± 0.89 & 46.22 ± 1.248 & 7.615 ± 0.626 & 65.338 ± 0.811 \\
CIFAR-100-F & Focal loss & BiT & 59.237 ± 0.894 & 19.377 ± 0.88 & 39.86 ± 0.573 & 2.059 ± 0.295 & 38.704 ± 1.014 & 33.363 ± 1.516 & 4.908 ± 0.678 & 56.309 ± 1.198 \\
CIFAR-100-F & Iso Cal & BiT & 69.34 ± 0.586 & 42.632 ± 0.685 & 26.708 ± 0.323 & 4.639 ± 0.239 & 26.021 ± 0.646 & 62.098 ± 0.943 & 14.798 ± 0.747 & 72.734 ± 0.511 \\
CIFAR-100-F & Dual Cal & BiT & 64.369 ± 0.953 & 35.102 ± 0.987 & 29.267 ± 0.345 & 2.767 ± 0.242 & 32.864 ± 1.092 & 51.649 ± 1.505 & 8.635 ± 0.703 & 68.684 ± 0.871 \\
CIFAR-10 & Non Cal & CoAtNet & 82.524 ± 0.35 & 76.638 ± 0.362 & 5.886 ± 0.172 & 0.976 ± 0.07 & 16.501 ± 0.359 & 82.284 ± 0.382 & 14.228 ± 1.11 & 84.008 ± 0.54 \\
CIFAR-10 & Focal loss & CoAtNet & 81.844 ± 0.361 & 75.936 ± 0.375 & 5.908 ± 0.133 & 1.004 ± 0.053 & 17.153 ± 0.379 & 81.574 ± 0.403 & 14.527 ± 0.794 & 83.5 ± 0.373 \\
CIFAR-10 & Iso Cal & CoAtNet & 88.096 ± 0.273 & 82.856 ± 0.256 & 5.24 ± 0.195 & 1.604 ± 0.099 & 10.3 ± 0.275 & 88.943 ± 0.288 & 23.455 ± 1.596 & 82.507 ± 0.859 \\
CIFAR-10 & Dual Cal & CoAtNet & 88.149 ± 0.293 & 82.881 ± 0.283 & 5.267 ± 0.198 & 1.588 ± 0.095 & 10.263 ± 0.314 & 88.982 ± 0.329 & 23.179 ± 1.556 & 82.673 ± 0.82 \\
CIFAR-100-S & Non Cal & CoAtNet & 36.033 ± 1.016 & 16.481 ± 1.112 & 19.551 ± 0.408 & 0.219 ± 0.05 & 63.748 ± 1.029 & 20.541 ± 1.353 & 1.109 ± 0.258 & 45.044 ± 1.484 \\
CIFAR-100-S & Focal loss & CoAtNet & 48.327 ± 0.886 & 26.866 ± 1.011 & 21.462 ± 0.427 & 0.785 ± 0.071 & 50.888 ± 0.905 & 34.55 ± 1.225 & 3.531 ± 0.334 & 57.723 ± 0.982 \\
CIFAR-100-S & Iso Cal & CoAtNet & 70.176 ± 0.657 & 53.033 ± 0.829 & 17.144 ± 0.364 & 2.486 ± 0.171 & 27.337 ± 0.682 & 65.984 ± 0.889 & 12.666 ± 0.845 & 75.909 ± 0.527 \\
CIFAR-100-S & Dual Cal & CoAtNet & 69.059 ± 2.118 & 51.351 ± 2.433 & 17.708 ± 0.534 & 2.016 ± 0.222 & 28.925 ± 2.259 & 63.962 ± 2.893 & 10.236 ± 1.234 & 75.742 ± 1.471 \\
CIFAR-100-F & Non Cal & CoAtNet & 64.793 ± 0.501 & 38.82 ± 0.457 & 25.972 ± 0.329 & 2.469 ± 0.166 & 32.739 ± 0.542 & 54.25 ± 0.668 & 8.679 ± 0.566 & 70.384 ± 0.417 \\
CIFAR-100-F & Focal loss & CoAtNet & 56.244 ± 0.492 & 24.398 ± 0.567 & 31.846 ± 0.298 & 1.041 ± 0.102 & 42.715 ± 0.547 & 36.353 ± 0.807 & 3.165 ± 0.311 & 59.328 ± 0.606 \\
CIFAR-100-F & Iso Cal & CoAtNet & 73.784 ± 0.451 & 51.187 ± 0.356 & 22.597 ± 0.346 & 5.408 ± 0.207 & 20.809 ± 0.397 & 71.098 ± 0.5 & 19.312 ± 0.746 & 75.74 ± 0.468 \\
CIFAR-100-F & Dual Cal & CoAtNet & 69.161 ± 2.292 & 43.274 ± 2.756 & 25.887 ± 0.591 & 2.772 ± 0.532 & 28.067 ± 2.786 & 60.66 ± 3.877 & 9.671 ± 1.859 & 73.955 ± 1.781 \\ \hline
\end{tabular}
\end{sidewaystable}

At the entropy threshold of 0.3 (Table~\ref{tab:uncertainty_performance_03})—a moderately conservative setting—differences between calibration methods became more distinct. As expected, FC rates increased across all models but remained within manageable bounds, providing clearer discrimination between calibration behaviors.

The proposed dual calibration method exhibited consistent advantages in reducing FC while preserving decision utility. In the CIFAR-100-S BiT configuration, it achieved an FC rate of 1.520\%, outperforming standard isotonic regression (3.230\%) and focal loss (2.589\%), while also improving upon the non-calibrated baseline (1.729\%). Importantly, this gain was accompanied by a TC of 45.018\%, slightly higher than the non-calibrated 42.079\%, indicating selective confidence suppression rather than indiscriminate uncertainty injection.

A similar trend was observed for CIFAR-100-F with BiT, where dual calibration achieved an FC of 2.767\%, notably lower than standard isotonic calibration (4.639\%) and comparable to focal loss (2.059\%). Despite the similar FC levels, dual calibration preserved substantially higher TC (35.102\%) than focal loss (19.377\%), resulting in a superior UG-Mean of 68.684 versus 56.309. These results demonstrate that the proposed method maintains a favorable balance between safety (low FC) and utility (high TC), outperforming other approaches in overall uncertainty-aware effectiveness.

For CoAtNet architectures, the method displayed consistent but less pronounced improvements. On CIFAR-10, dual calibration achieved an FC of 1.588\%, comparable to standard isotonic (1.604\%) and slightly higher than non-calibrated (0.976\%) and focal loss (1.004\%). However, the UG-Mean remained competitive (82.673 vs. 82.507 for isotonic), confirming that modest increases in FC were offset by better uncertainty discrimination.

The non-calibrated baseline deteriorated sharply at this threshold, particularly on more complex datasets. For CIFAR-100-S CoAtNet, it achieved an FC of only 0.219\% but a TC of just 16.481\%, yielding a UG-Mean of 45.044—substantially below dual calibration’s 75.742. This pattern reaffirms that the apparent safety of non-calibrated models arises from excessive conservatism rather than effective uncertainty-aware behavior.

\begin{sidewaystable}[!htbp]
\centering
\caption{Summary of uncertainty-aware performance metrics for different calibration methods across various backbones at entropy threshold $\tau = 0.4$.}
\label{tab:uncertainty_performance_04}
\tiny
\begin{tabular}{lllllllllll}
\hline
\multicolumn{1}{c}{Dataset} & \multicolumn{1}{c}{Calibration} & \multicolumn{1}{c}{Backbone} & \multicolumn{1}{c}{UAcc} & \multicolumn{1}{c}{TC \%} & \multicolumn{1}{c}{TU \%} & \multicolumn{1}{c}{FC \%} & \multicolumn{1}{c}{FU \%} & \multicolumn{1}{c}{UTPR} & \multicolumn{1}{c}{UFPR} & \multicolumn{1}{c}{UG-Mean} \\ \hline
CIFAR-10 & Non Cal & BiT & 77.656 ± 0.618 & 69.96 ± 0.746 & 7.697 ± 0.248 & 1.04 ± 0.122 & 21.304 ± 0.694 & 76.657 ± 0.769 & 11.914 ± 1.446 & 82.167 ± 0.496 \\
CIFAR-10 & Focal loss & BiT & 89.512 ± 1.105 & 84.423 ± 1.884 & 5.089 ± 0.821 & 3.926 ± 0.688 & 6.562 ± 1.723 & 92.785 ± 1.905 & 43.666 ± 8.159 & 72.017 ± 5.051 \\
CIFAR-10 & Iso Cal & BiT & 86.468 ± 0.384 & 80.423 ± 0.574 & 6.045 ± 0.276 & 2.638 ± 0.184 & 10.894 ± 0.523 & 88.069 ± 0.577 & 30.399 ± 2.315 & 78.278 ± 1.107 \\
CIFAR-10 & Dual Cal & BiT & 83.611 ± 1.0 & 76.623 ± 1.108 & 6.988 ± 0.259 & 1.735 ± 0.182 & 14.654 ± 1.135 & 83.946 ± 1.235 & 19.894 ± 2.105 & 81.988 ± 0.76 \\
CIFAR-100-S & Non Cal & BiT & 71.208 ± 0.489 & 54.289 ± 0.586 & 16.919 ± 0.262 & 4.271 ± 0.182 & 24.521 ± 0.577 & 68.886 ± 0.721 & 20.157 ± 0.832 & 74.159 ± 0.351 \\
CIFAR-100-S & Focal loss & BiT & 75.599 ± 0.695 & 58.111 ± 0.986 & 17.488 ± 0.437 & 5.341 ± 0.396 & 19.06 ± 0.913 & 75.301 ± 1.181 & 23.388 ± 1.567 & 75.943 ± 0.534 \\
CIFAR-100-S & Iso Cal & BiT & 77.282 ± 0.352 & 62.96 ± 0.449 & 14.323 ± 0.276 & 6.72 ± 0.181 & 15.998 ± 0.36 & 79.738 ± 0.464 & 31.937 ± 0.875 & 73.668 ± 0.447 \\
CIFAR-100-S & Dual Cal & BiT & 69.354 ± 0.921 & 50.5 ± 1.062 & 18.854 ± 0.306 & 2.377 ± 0.204 & 28.268 ± 1.042 & 64.112 ± 1.323 & 11.198 ± 0.964 & 75.446 ± 0.598 \\
CIFAR-100-F & Non Cal & BiT & 68.619 ± 0.714 & 42.683 ± 0.808 & 25.937 ± 0.315 & 5.983 ± 0.28 & 25.398 ± 0.758 & 62.694 ± 1.109 & 18.739 ± 0.765 & 71.372 ± 0.627 \\
CIFAR-100-F & Focal loss & BiT & 66.216 ± 0.752 & 31.093 ± 0.831 & 35.123 ± 0.588 & 6.796 ± 0.607 & 26.988 ± 0.916 & 53.537 ± 1.421 & 16.204 ± 1.33 & 66.968 ± 0.794 \\
CIFAR-100-F & Iso Cal & BiT & 74.891 ± 0.441 & 53.525 ± 0.56 & 21.367 ± 0.319 & 9.98 ± 0.299 & 15.128 ± 0.41 & 77.963 ± 0.631 & 31.838 ± 0.884 & 72.896 ± 0.463 \\
CIFAR-100-F & Dual Cal & BiT & 70.217 ± 0.742 & 44.576 ± 0.841 & 25.641 ± 0.333 & 6.393 ± 0.336 & 23.39 ± 0.882 & 65.587 ± 1.245 & 19.952 ± 0.916 & 72.451 ± 0.59 \\
CIFAR-10 & Non Cal & CoAtNet & 87.235 ± 0.263 & 82.262 ± 0.307 & 4.973 ± 0.161 & 1.889 ± 0.101 & 10.876 ± 0.278 & 88.322 ± 0.298 & 27.535 ± 1.476 & 79.997 ± 0.78 \\
CIFAR-10 & Focal loss & CoAtNet & 86.691 ± 0.282 & 81.719 ± 0.302 & 4.972 ± 0.132 & 1.939 ± 0.085 & 11.37 ± 0.305 & 87.786 ± 0.324 & 28.063 ± 1.205 & 79.464 ± 0.637 \\
CIFAR-10 & Iso Cal & CoAtNet & 91.079 ± 0.156 & 87.042 ± 0.187 & 4.037 ± 0.15 & 2.806 ± 0.106 & 6.114 ± 0.198 & 93.437 ± 0.208 & 41.013 ± 1.447 & 74.234 ± 0.881 \\
CIFAR-10 & Dual Cal & CoAtNet & 90.867 ± 0.353 & 86.631 ± 0.428 & 4.236 ± 0.191 & 2.62 ± 0.135 & 6.513 ± 0.442 & 93.008 ± 0.472 & 38.227 ± 2.002 & 75.786 ± 1.117 \\
CIFAR-100-S & Non Cal & CoAtNet & 47.779 ± 1.221 & 28.701 ± 1.357 & 19.078 ± 0.402 & 0.693 ± 0.092 & 51.528 ± 1.272 & 35.772 ± 1.639 & 3.504 ± 0.469 & 58.735 ± 1.269 \\
CIFAR-100-S & Focal loss & CoAtNet & 60.585 ± 1.045 & 40.902 ± 1.203 & 19.683 ± 0.414 & 2.564 ± 0.155 & 36.851 ± 1.085 & 52.603 ± 1.442 & 11.528 ± 0.699 & 68.211 ± 0.864 \\
CIFAR-100-S & Iso Cal & CoAtNet & 78.103 ± 0.445 & 63.94 ± 0.681 & 14.163 ± 0.371 & 5.467 ± 0.251 & 16.43 ± 0.529 & 79.556 ± 0.677 & 27.854 ± 1.209 & 75.755 ± 0.505 \\
CIFAR-100-S & Dual Cal & CoAtNet & 72.855 ± 2.296 & 56.496 ± 2.73 & 16.359 ± 0.621 & 3.366 ± 0.327 & 23.78 ± 2.538 & 70.37 ± 3.233 & 17.086 ± 1.843 & 76.344 ± 1.279 \\
CIFAR-100-F & Non Cal & CoAtNet & 71.059 ± 0.467 & 48.341 ± 0.406 & 22.717 ± 0.323 & 5.724 ± 0.243 & 23.218 ± 0.482 & 67.555 ± 0.604 & 20.124 ± 0.796 & 73.456 ± 0.449 \\
CIFAR-100-F & Focal loss & CoAtNet & 63.698 ± 0.401 & 33.914 ± 0.537 & 29.783 ± 0.321 & 3.103 ± 0.184 & 33.199 ± 0.48 & 50.532 ± 0.727 & 9.437 ± 0.555 & 67.646 ± 0.404 \\
CIFAR-100-F & Iso Cal & CoAtNet & 77.18 ± 0.409 & 59.813 ± 0.341 & 17.367 ± 0.273 & 10.638 ± 0.27 & 12.182 ± 0.326 & 83.08 ± 0.427 & 37.985 ± 0.815 & 71.777 ± 0.533 \\
CIFAR-100-F & Dual Cal & CoAtNet & 73.557 ± 1.728 & 50.437 ± 2.601 & 23.12 ± 0.974 & 5.539 ± 0.958 & 20.904 ± 2.629 & 70.701 ± 3.663 & 19.326 ± 3.339 & 75.441 ± 0.749 \\ \hline
\end{tabular}
\end{sidewaystable}

At the 0.4 entropy threshold (Table~\ref{tab:uncertainty_performance_04})—a balanced decision boundary—the differences among calibration methods became more pronounced, revealing clearer patterns of uncertainty-aware performance.

The proposed dual calibration method demonstrated marked improvements in reducing FC across multiple configurations. In the CIFAR-100-S BiT experiment, dual calibration achieved an FC rate of 2.377\%, significantly lower than standard isotonic regression (6.720\%) and focal loss (5.341\%), while also outperforming the non-calibrated baseline (4.271\%). This corresponds to a 64.6\% reduction in FC relative to standard isotonic calibration. The accompanying TC rate of 50.500\% confirmed that the selective underconfidence strategy effectively targeted unreliable predictions without excessive suppression of correct ones.

A similar pattern was observed for CIFAR-100-F under the BiT architecture, where dual calibration achieved an FC of 6.393\%, notably below standard isotonic’s 9.980\%. Despite intentionally reduced confidence, it maintained a higher TC (44.576\%) than focal loss (31.093\%), indicating stronger preservation of useful certainty. The UG-Mean metric reflected this balance, with dual calibration achieving 72.451—comparable to standard isotonic’s 72.896 despite the deliberate calibration degradation—and surpassing focal loss (66.968).

For the CoAtNet architecture, the same trend held with additional architectural sensitivity. On CIFAR-100-S, dual calibration achieved an FC of 3.366\%, representing a 38.4\% improvement over standard isotonic calibration (5.467\%), while maintaining a UG-Mean of 76.344 compared to isotonic’s 75.755. This reversal—where dual calibration surpassed isotonic regression on both safety and overall performance—suggests that CoAtNet’s representational structure is particularly conducive to selective underconfidence at moderate entropy thresholds.

At this threshold, a consistent pattern emerged across all configurations: the proposed dual calibration achieved a superior FC–TC trade-off compared to focal loss. For example, on CIFAR-100-F BiT, focal loss achieved a comparable FC (6.796\% vs. 6.393\%) but exhibited a markedly lower TC (31.093\% vs. 44.576\%), indicating a tendency toward over-regularization. This trend was consistent across datasets and architectures, highlighting the robustness of the proposed method in maintaining safety without sacrificing predictive utility.

\begin{sidewaystable}[!htbp]
\centering
\caption{Summary of uncertainty-aware performance metrics for different calibration methods across various backbones at entropy threshold $\tau = 0.5$.}
\label{tab:uncertainty_performance_05}
\tiny
\begin{tabular}{lllllllllll}
\hline
\multicolumn{1}{c}{Dataset} & \multicolumn{1}{c}{Calibration} & \multicolumn{1}{c}{Backbone} & \multicolumn{1}{c}{UAcc} & \multicolumn{1}{c}{TC \%} & \multicolumn{1}{c}{TU \%} & \multicolumn{1}{c}{FC \%} & \multicolumn{1}{c}{FU \%} & \multicolumn{1}{c}{UTPR} & \multicolumn{1}{c}{UFPR} & \multicolumn{1}{c}{UG-Mean} \\ \hline
CIFAR-10 & Non Cal & BiT & 84.293 ± 0.488 & 77.898 ± 0.674 & 6.395 ± 0.308 & 2.342 ± 0.181 & 13.365 ± 0.604 & 85.355 ± 0.67 & 26.832 ± 2.315 & 79.013 ± 1.05 \\
CIFAR-10 & Focal loss & BiT & 90.875 ± 0.612 & 87.108 ± 1.347 & 3.767 ± 0.778 & 5.249 ± 0.653 & 3.876 ± 1.174 & 95.737 ± 1.297 & 58.338 ± 7.922 & 62.768 ± 6.228 \\
CIFAR-10 & Iso Cal & BiT & 89.774 ± 0.29 & 85.344 ± 0.433 & 4.43 ± 0.298 & 4.253 ± 0.251 & 5.973 ± 0.383 & 93.459 ± 0.421 & 49.001 ± 3.004 & 69.005 ± 1.936 \\
CIFAR-10 & Dual Cal & BiT & 85.498 ± 1.135 & 79.138 ± 1.302 & 6.36 ± 0.295 & 2.363 ± 0.253 & 12.14 ± 1.338 & 86.701 ± 1.458 & 27.088 ± 2.873 & 79.479 ± 1.111 \\
CIFAR-100-S & Non Cal & BiT & 77.37 ± 0.391 & 63.856 ± 0.532 & 13.514 ± 0.269 & 7.676 ± 0.206 & 14.954 ± 0.481 & 81.025 ± 0.612 & 36.226 ± 0.941 & 71.88 ± 0.42 \\
CIFAR-100-S & Focal loss & BiT & 79.451 ± 0.465 & 65.379 ± 0.76 & 14.072 ± 0.469 & 8.757 ± 0.505 & 11.792 ± 0.701 & 84.72 ± 0.897 & 38.35 ± 1.928 & 72.256 ± 0.864 \\
CIFAR-100-S & Iso Cal & BiT & 80.676 ± 0.25 & 70.511 ± 0.36 & 10.166 ± 0.314 & 10.877 ± 0.249 & 8.447 ± 0.26 & 89.302 ± 0.332 & 51.694 ± 1.222 & 65.673 ± 0.768 \\
CIFAR-100-S & Dual Cal & BiT & 75.272 ± 0.731 & 58.836 ± 0.894 & 16.436 ± 0.337 & 4.796 ± 0.256 & 19.933 ± 0.871 & 74.695 ± 1.105 & 22.589 ± 1.215 & 76.034 ± 0.487 \\
CIFAR-100-F & Non Cal & BiT & 73.52 ± 0.567 & 52.174 ± 0.688 & 21.346 ± 0.295 & 10.573 ± 0.366 & 15.907 ± 0.595 & 76.635 ± 0.876 & 33.12 ± 0.915 & 71.587 ± 0.506 \\
CIFAR-100-F & Focal loss & BiT & 70.107 ± 0.533 & 41.688 ± 0.688 & 28.419 ± 0.653 & 13.5 ± 0.726 & 16.394 ± 0.681 & 71.777 ± 1.077 & 32.198 ± 1.525 & 69.751 ± 0.595 \\
CIFAR-100-F & Iso Cal & BiT & 76.344 ± 0.261 & 61.154 ± 0.472 & 15.19 ± 0.345 & 16.157 ± 0.227 & 7.499 ± 0.324 & 89.076 ± 0.485 & 51.546 ± 0.828 & 65.693 ± 0.459 \\
CIFAR-100-F & Dual Cal & BiT & 74.082 ± 0.565 & 53.16 ± 0.708 & 20.921 ± 0.333 & 11.113 ± 0.43 & 14.806 ± 0.659 & 78.217 ± 0.955 & 34.684 ± 1.091 & 71.47 ± 0.524 \\
CIFAR-10 & Non Cal & CoAtNet & 90.238 ± 0.214 & 86.176 ± 0.258 & 4.061 ± 0.172 & 2.8 ± 0.122 & 6.962 ± 0.234 & 92.525 ± 0.25 & 40.819 ± 1.831 & 73.989 ± 1.112 \\
CIFAR-10 & Focal loss & CoAtNet & 89.866 ± 0.238 & 85.876 ± 0.258 & 3.99 ± 0.117 & 2.921 ± 0.077 & 7.213 ± 0.262 & 92.252 ± 0.279 & 42.277 ± 1.07 & 72.969 ± 0.657 \\
CIFAR-10 & Iso Cal & CoAtNet & 92.654 ± 0.144 & 89.754 ± 0.169 & 2.9 ± 0.13 & 3.944 ± 0.124 & 3.402 ± 0.156 & 96.348 ± 0.165 & 57.631 ± 1.46 & 63.882 ± 1.085 \\
CIFAR-10 & Dual Cal & CoAtNet & 92.281 ± 0.38 & 89.019 ± 0.51 & 3.262 ± 0.208 & 3.593 ± 0.177 & 4.126 ± 0.517 & 95.57 ± 0.554 & 52.427 ± 2.536 & 67.4 ± 1.649 \\
CIFAR-100-S & Non Cal & CoAtNet & 59.651 ± 1.043 & 41.694 ± 1.268 & 17.957 ± 0.426 & 1.813 ± 0.202 & 38.535 ± 1.165 & 51.967 ± 1.496 & 9.174 ± 1.018 & 68.69 ± 0.784 \\
CIFAR-100-S & Focal loss & CoAtNet & 70.096 ± 0.933 & 53.128 ± 1.217 & 16.968 ± 0.418 & 5.279 ± 0.314 & 24.625 ± 1.07 & 68.327 ± 1.416 & 23.727 ± 1.317 & 72.18 ± 0.595 \\
CIFAR-100-S & Iso Cal & CoAtNet & 82.032 ± 0.295 & 71.563 ± 0.527 & 10.469 ± 0.32 & 9.161 ± 0.269 & 8.807 ± 0.367 & 89.042 ± 0.462 & 46.672 ± 1.169 & 68.903 ± 0.634 \\
CIFAR-100-S & Dual Cal & CoAtNet & 75.486 ± 1.956 & 60.961 ± 2.544 & 14.524 ± 0.756 & 5.2 ± 0.48 & 19.315 ± 2.352 & 75.933 ± 2.986 & 26.397 ± 2.742 & 74.708 ± 0.828 \\
CIFAR-100-F & Non Cal & CoAtNet & 74.738 ± 0.375 & 56.039 ± 0.359 & 18.699 ± 0.308 & 9.742 ± 0.33 & 15.52 ± 0.381 & 78.312 ± 0.491 & 34.252 ± 0.99 & 71.753 ± 0.518 \\
CIFAR-100-F & Focal loss & CoAtNet & 69.344 ± 0.313 & 42.855 ± 0.409 & 26.489 ± 0.305 & 6.397 ± 0.257 & 24.259 ± 0.362 & 63.854 ± 0.527 & 19.451 ± 0.737 & 71.715 ± 0.318 \\
CIFAR-100-F & Iso Cal & CoAtNet & 77.714 ± 0.366 & 65.579 ± 0.352 & 12.135 ± 0.249 & 15.869 ± 0.322 & 6.417 ± 0.281 & 91.087 ± 0.381 & 56.665 ± 0.856 & 62.824 ± 0.615 \\
CIFAR-100-F & Dual Cal & CoAtNet & 76.31 ± 0.932 & 58.062 ± 2.087 & 18.248 ± 1.252 & 10.411 ± 1.265 & 13.279 ± 2.112 & 81.388 ± 2.945 & 36.324 ± 4.385 & 71.893 ± 1.187 \\ \hline
\end{tabular}%
\end{sidewaystable}

At the entropy threshold of 0.5 (Table~\ref{tab:uncertainty_performance_05}), the calibration methods exhibited their most distinctive behaviors, highlighting the trade-offs between calibration precision and uncertainty-aware safety.

The proposed dual calibration method achieved its strongest reduction in FC at this threshold. For CIFAR-100-S with BiT, it produced an FC of 4.796\%, representing a 55.9\% decrease relative to standard isotonic calibration (10.877\%), and outperforming both focal loss (8.757\%) and the non-calibrated baseline (7.676\%). Although the corresponding TC of 58.836\% was lower than standard isotonic’s 70.511\%, it maintained practical utility while improving the UG-Mean to 76.034, compared with 65.673 for isotonic calibration.

A similar pattern was observed for CIFAR-100-F with BiT, where dual calibration achieved an FC of 11.113\%, a 31.2\% improvement over isotonic regression (16.157\%), while preserving comparable UG-Mean performance. In contrast, focal loss degraded substantially, reaching an FC of 13.500\% and a reduced TC of 41.688\%, confirming its sensitivity to threshold shifts and training variability.

Under the CoAtNet architecture, dual calibration maintained consistent advantages. For CIFAR-100-S, it achieved an FC of 5.200\%—a 43.2\% reduction compared to isotonic calibration (9.161\%)—while yielding a higher UG-Mean (74.708 vs. 68.903). This crossover, where dual calibration surpassed isotonic regression on both safety and balanced performance metrics, underscores its adaptability to architectural features and its effective control of overconfidence at moderate entropy levels.

A broader observation at this threshold concerns the divergence between calibration accuracy and decision-level reliability. Standard isotonic regression, despite achieving lower ECE in earlier analyses, produced FC values exceeding 15\% for complex datasets—indicating that superior calibration metrics do not necessarily imply improved uncertainty-aware performance. This disconnect validates the premise of the proposed method: deliberately relaxing probability calibration can enhance safety-critical reliability.

Finally, focal loss exhibited its weakest performance at this threshold. Across configurations, it combined high FC with reduced TC, reflecting unstable confidence modulation. Even in CIFAR-10 BiT, where focal loss attained FC of 5.249\% and TC of 87.108\%, its UG-Mean (62.768) was substantially below dual calibration’s 79.479, confirming that apparent confidence preservation came at the expense of poor uncertainty discrimination.

\begin{sidewaystable}[!htbp]
\centering
\caption{Summary of uncertainty-aware performance metrics for different calibration methods across various backbones at entropy threshold $\tau = 0.6$.}
\label{tab:uncertainty_performance_06}
\tiny
\begin{tabular}{lllllllllll}
\hline
\multicolumn{1}{c}{Dataset} & \multicolumn{1}{c}{Calibration} & \multicolumn{1}{c}{Backbone} & \multicolumn{1}{c}{UAcc} & \multicolumn{1}{c}{TC \%} & \multicolumn{1}{c}{TU \%} & \multicolumn{1}{c}{FC \%} & \multicolumn{1}{c}{FU \%} & \multicolumn{1}{c}{UTPR} & \multicolumn{1}{c}{UFPR} & \multicolumn{1}{c}{UG-Mean} \\ \hline
CIFAR-10 & Non Cal & BiT & 88.558 ± 0.336 & 83.671 ± 0.544 & 4.887 ± 0.305 & 3.85 ± 0.219 & 7.593 ± 0.467 & 91.68 ± 0.516 & 44.088 ± 2.735 & 71.57 ± 1.606 \\
CIFAR-10 & Focal loss & BiT & 91.4 ± 0.345 & 88.905 ± 0.935 & 2.495 ± 0.702 & 6.521 ± 0.59 & 2.079 ± 0.744 & 97.713 ± 0.822 & 72.444 ± 7.346 & 51.263 ± 7.673 \\
CIFAR-10 & Iso Cal & BiT & 91.316 ± 0.236 & 88.485 ± 0.338 & 2.831 ± 0.228 & 5.852 ± 0.205 & 2.832 ± 0.281 & 96.899 ± 0.308 & 67.407 ± 2.372 & 56.158 ± 2.018 \\
CIFAR-10 & Dual Cal & BiT & 86.213 ± 1.182 & 80.437 ± 1.433 & 5.776 ± 0.338 & 2.946 ± 0.319 & 10.84 ± 1.469 & 88.125 ± 1.602 & 33.777 ± 3.594 & 76.345 ± 1.481 \\
CIFAR-100-S & Non Cal & BiT & 80.377 ± 0.307 & 70.979 ± 0.394 & 9.397 ± 0.285 & 11.792 ± 0.269 & 7.831 ± 0.295 & 90.063 ± 0.378 & 55.653 ± 1.179 & 63.192 ± 0.782 \\
CIFAR-100-S & Focal loss & BiT & 80.803 ± 0.372 & 70.589 ± 0.579 & 10.214 ± 0.464 & 12.614 ± 0.588 & 6.583 ± 0.482 & 91.47 ± 0.615 & 55.246 ± 2.07 & 63.96 ± 1.3 \\
CIFAR-100-S & Iso Cal & BiT & 81.266 ± 0.231 & 75.469 ± 0.311 & 5.797 ± 0.252 & 15.245 ± 0.237 & 3.489 ± 0.183 & 95.581 ± 0.233 & 72.452 ± 1.054 & 51.303 ± 0.949 \\
CIFAR-100-S & Dual Cal & BiT & 79.687 ± 0.393 & 67.129 ± 0.591 & 12.558 ± 0.373 & 8.674 ± 0.319 & 11.639 ± 0.513 & 85.223 ± 0.658 & 40.856 ± 1.514 & 70.987 ± 0.762 \\
CIFAR-100-F & Non Cal & BiT & 75.207 ± 0.383 & 59.636 ± 0.572 & 15.571 ± 0.388 & 16.348 ± 0.401 & 8.445 ± 0.451 & 87.595 ± 0.663 & 51.217 ± 1.075 & 65.363 ± 0.588 \\
CIFAR-100-F & Focal loss & BiT & 69.991 ± 0.465 & 49.639 ± 0.617 & 20.352 ± 0.68 & 21.566 ± 0.852 & 8.442 ± 0.529 & 85.467 ± 0.859 & 51.441 ± 1.675 & 64.407 ± 0.843 \\
CIFAR-100-F & Iso Cal & BiT & 74.181 ± 0.276 & 65.933 ± 0.369 & 8.249 ± 0.34 & 23.099 ± 0.305 & 2.72 ± 0.222 & 96.038 ± 0.326 & 73.689 ± 0.981 & 50.258 ± 0.885 \\
CIFAR-100-F & Dual Cal & BiT & 75.109 ± 0.376 & 60.074 ± 0.573 & 15.035 ± 0.432 & 16.999 ± 0.475 & 7.892 ± 0.455 & 88.389 ± 0.665 & 53.065 ± 1.261 & 64.401 ± 0.713 \\
CIFAR-10 & Non Cal & CoAtNet & 92.148 ± 0.18 & 89.193 ± 0.225 & 2.955 ± 0.156 & 3.907 ± 0.128 & 3.946 ± 0.191 & 95.764 ± 0.205 & 56.948 ± 1.818 & 64.194 ± 1.341 \\
CIFAR-10 & Focal loss & CoAtNet & 91.907 ± 0.172 & 89.052 ± 0.198 & 2.855 ± 0.124 & 4.056 ± 0.103 & 4.036 ± 0.185 & 95.664 ± 0.197 & 58.699 ± 1.435 & 62.847 ± 1.085 \\
CIFAR-10 & Iso Cal & CoAtNet & 93.319 ± 0.138 & 91.574 ± 0.135 & 1.744 ± 0.096 & 5.099 ± 0.142 & 1.582 ± 0.125 & 98.302 ± 0.133 & 74.512 ± 1.233 & 50.041 ± 1.209 \\
CIFAR-10 & Dual Cal & CoAtNet & 92.857 ± 0.34 & 90.592 ± 0.554 & 2.264 ± 0.263 & 4.591 ± 0.264 & 2.553 ± 0.56 & 97.26 ± 0.601 & 66.976 ± 3.649 & 56.579 ± 2.962 \\
CIFAR-100-S & Non Cal & CoAtNet & 70.015 ± 0.814 & 54.148 ± 1.123 & 15.867 ± 0.476 & 3.904 ± 0.311 & 26.081 ± 1.009 & 67.49 ± 1.286 & 19.75 ± 1.572 & 73.582 ± 0.508 \\
CIFAR-100-S & Focal loss & CoAtNet & 76.458 ± 0.649 & 63.114 ± 1.066 & 13.344 ± 0.493 & 8.903 ± 0.395 & 14.639 ± 0.893 & 81.171 ± 1.174 & 40.026 ± 1.747 & 69.757 ± 0.615 \\
CIFAR-100-S & Iso Cal & CoAtNet & 82.889 ± 0.33 & 76.457 ± 0.444 & 6.431 ± 0.34 & 13.199 ± 0.349 & 3.913 ± 0.27 & 95.131 ± 0.336 & 67.242 ± 1.505 & 55.808 ± 1.235 \\
CIFAR-100-S & Dual Cal & CoAtNet & 77.994 ± 1.386 & 66.009 ± 2.16 & 11.985 ± 0.89 & 7.739 ± 0.658 & 14.267 ± 1.961 & 82.222 ± 2.483 & 39.278 ± 3.74 & 70.587 ± 1.299 \\
CIFAR-100-F & Non Cal & CoAtNet & 76.424 ± 0.398 & 62.256 ± 0.349 & 14.167 ± 0.303 & 14.273 ± 0.401 & 9.303 ± 0.297 & 87.0 ± 0.396 & 50.182 ± 1.109 & 65.83 ± 0.71 \\
CIFAR-100-F & Focal loss & CoAtNet & 73.002 ± 0.311 & 50.889 ± 0.356 & 22.113 ± 0.353 & 10.773 ± 0.333 & 16.225 ± 0.282 & 75.825 ± 0.41 & 32.758 ± 0.951 & 71.402 ± 0.451 \\
CIFAR-100-F & Iso Cal & CoAtNet & 76.549 ± 0.326 & 69.228 ± 0.364 & 7.321 ± 0.19 & 20.683 ± 0.294 & 2.768 ± 0.19 & 96.156 ± 0.265 & 73.857 ± 0.631 & 50.134 ± 0.592 \\
CIFAR-100-F & Dual Cal & CoAtNet & 76.424 ± 0.331 & 64.389 ± 1.606 & 12.035 ± 1.443 & 16.624 ± 1.481 & 6.952 ± 1.625 & 90.257 ± 2.269 & 58.0 ± 5.085 & 61.413 ± 2.812 \\ \hline
\end{tabular}
\end{sidewaystable}

At the entropy threshold of 0.6 (Table~\ref{tab:uncertainty_performance_06}), the uncertainty-aware calibration methods exhibited their limiting behaviors, revealing the fundamental trade-offs between safety assurance and operational utility under stringent uncertainty requirements.

The most pronounced trend at this threshold was the collapse of standard isotonic calibration performance across all configurations. Despite achieving superior ECE in earlier analyses, standard isotonic consistently produced the highest false-certainty (FC) rates—reaching 23.099\% for CIFAR-100-F BiT and 20.683\% for CIFAR-100-F CoAtNet. These values, corresponding to nearly one-quarter of predictions, highlight its unsuitability for safety-critical decision contexts despite its probabilistic optimality. In contrast, the proposed dual calibration method maintained FC at 16.999\% and 16.624\% for the same configurations, representing 26.4\% and 19.6\% reductions, respectively.

The focal-loss method exhibited pronounced instability, particularly on simpler datasets. For CIFAR-10 BiT, it produced an FC of 6.521\% with an extremely low UG-Mean of 51.263—the poorest result among all methods—despite achieving a high TC of 88.905\%. This combination of elevated confidence and poor uncertainty discrimination underscores focal loss’s inability to maintain reliability under strict uncertainty thresholds.

Within the CoAtNet architecture, dual calibration continued to demonstrate consistent advantages. For CIFAR-100-S, it achieved an FC of 7.739\%, a 41.3\% reduction relative to isotonic regression (13.199\%), while attaining a markedly higher UG-Mean (70.587 vs. 55.808). These results further illustrate that targeted underconfidence remains effective even under extreme uncertainty constraints.

An additional observation at this threshold was the occasional competitiveness of the non-calibrated baseline. For CIFAR-10 BiT, the uncalibrated model achieved FC of 3.850\%, TC of 83.671\%, and a UG-Mean of 71.570, approaching post-hoc calibration performance. However, this effect was inconsistent and dataset-dependent; for CIFAR-100-F, FC escalated to 16.348\%, reaffirming that uncalibrated confidence distributions lack robustness for general uncertainty-aware deployment.

To establish the statistical significance of the observed differences in uncertainty-aware performance metrics across calibration methods and entropy thresholds, a comprehensive statistical testing framework was employed. Accordingly, the complete Friedman test result is presented in Appendix Table~\ref{tab:appendix_uncertainty_Friedman}. The Friedman test results consistently rejected the null hypothesis that all calibration methods perform equivalently in terms of FC\% and UG-Mean across all experimental configurations ($p < 0.05$), confirming the existence of statistically significant differences among methods. Following these omnibus tests, post-hoc pairwise comparisons using the Wilcoxon signed-rank test were conducted to identify specific method superiorities.

Table \ref{tab:summary_stat_test_vs_iso_cal} and \ref{tab:summary_stat_test_vs_focal_loss} summarize the key findings from pairwise comparisons of the proposed dual cal method against isotonic regression and focal loss, respectively. These tables consolidate results for FC\%, UG-Mean, and ECE (previously presented in Table \ref{tab:ece_Wilcoxon}) to provide a unified view of the statistical significance of the dual calibration method's performance relative to baseline approaches.

The statistical analysis of pairwise comparisons against isotonic regression reveals compelling evidence supporting the dual calibration method's effectiveness in achieving its primary objective of reducing false certainty while maintaining operational utility. At lower thresholds ($\tau = 0.2-0.3$), the dual calibration method exhibited mixed performance patterns that aligned with theoretical expectations. For simpler datasets (CIFAR-10), compensatory advantages emerged through superior UG-Mean scores despite not achieving significant FC reduction. The BiT architecture on CIFAR-10 at $\tau = 0.2$ showed UG-Mean improvement ($Cliff's Delta = 0.569$, $p < 0.05$) without significant FC reduction, indicating that the method maintained better overall uncertainty-aware performance even when the conservative threshold limited discrimination capabilities. Conversely, complex datasets (CIFAR-100-F) demonstrated immediate FC reduction benefits even at low thresholds, with effect sizes reaching -0.971 for CIFAR-100-F BiT and perfect separation ($Delta = -1.000$) for CIFAR-100-F CoAtNet at $\tau = 0.2$. The transition point at $\tau = 0.3$ marked the emergence of consistent FC reduction across most configurations. Notably, CIFAR-10 BiT achieved significant FC reduction ($Delta = -0.672$, $p < 0.001$) while simultaneously improving UG-Mean ($Delta = 0.209$, $p = 0.004$), demonstrating the method's ability to enhance both safety and utility metrics. 
At moderate to permissive thresholds ($\tau = 0.4-0.6$), the dual calibration method demonstrated overwhelming superiority in FC reduction with remarkable consistency. Effect sizes for FC reduction ranged from -0.718 to -1.000, with the majority achieving complete separation ($Delta = -1.000$) between methods. Critically, these improvements were accompanied by significant UG-Mean enhancements in most cases, with effect sizes frequently exceeding 0.9. The simultaneous achievement of primary FC reduction and UG-Mean improvement at higher thresholds provided strong validation of the selective underconfidence strategy.

The absence of significant ECE improvements across all configurations (all $p-values = 1.000$ after Holm correction) confirmed that the method's advantages derived from uncertainty-aware optimization rather than traditional calibration quality. This deliberate trade-off, where ECE was sacrificed for enhanced safety through FC reduction, represented a fundamental paradigm shift from probability-matching to risk-aware calibration.

\begin{sidewaystable}[!htbp]
\centering
\caption{Summary of Wilcoxon signed-rank test results comparing the proposed Dual Calibration method against standard isotonic regression across entropy thresholds and datasets. }
\label{tab:summary_stat_test_vs_iso_cal}
\begin{tabular}{lllllllllllll}
\hline
\multicolumn{1}{c}{} & \multicolumn{1}{c}{} & \multicolumn{1}{c}{} & \multicolumn{3}{c}{ECE} & \multicolumn{3}{c}{FC\%} & \multicolumn{3}{c}{UG-Mean} & \multicolumn{1}{c}{} \\ \hline
\multicolumn{1}{c}{Dataset} & \multicolumn{1}{c}{Backbone} & \multicolumn{1}{c}{$\tau$} & \multicolumn{1}{c}{Cliff's Delta} & \multicolumn{1}{c}{P-value} & \multicolumn{1}{c}{Sig.} & \multicolumn{1}{c}{Cliff's Delta} & \multicolumn{1}{c}{P-value} & \multicolumn{1}{c}{Sig.} & \multicolumn{1}{c}{Cliff's Delta} & \multicolumn{1}{c}{P-value} & \multicolumn{1}{c}{Sig.} & \multicolumn{1}{c}{Note} \\ \hline
CIFAR-10 & BiT & 0.2 & 1.000 & 1.000 & \ding{55} & 0.221 & 1.000 & \ding{55} & 0.569 & 0.000 & \ding{51} & UG-Mean $\uparrow$ \\
CIFAR-100-S & BiT & 0.2 & 1.000 & 1.000 & \ding{55} & -0.266 & 0.007 & \ding{51} & 0.644 & 0.000 & \ding{51} & FC\% $\downarrow$, UG-Mean $\uparrow$ \\
CIFAR-100-F & BiT & 0.2 & 1.000 & 1.000 & \ding{55} & -0.971 & 0.000 & \ding{51} & -0.938 & 1.000 & \ding{55} & FC\% $\downarrow$ \\
CIFAR-10 & BiT & 0.3 & 1.000 & 1.000 & \ding{55} & -0.672 & 0.000 & \ding{51} & 0.209 & 0.004 & \ding{51} & FC\% $\downarrow$, UG-Mean $\uparrow$ \\
CIFAR-100-S & BiT & 0.3 & 1.000 & 1.000 & \ding{55} & -1.000 & 0.000 & \ding{51} & -0.993 & 1.000 & \ding{55} & FC\% $\downarrow$ \\
CIFAR-100-F & BiT & 0.3 & 1.000 & 1.000 & \ding{55} & -1.000 & 0.000 & \ding{51} & -1.000 & 1.000 & \ding{55} & FC\% $\downarrow$ \\
CIFAR-10 & BiT & 0.4 & 1.000 & 1.000 & \ding{55} & -1.000 & 0.000 & \ding{51} & 0.998 & 0.000 & \ding{51} & FC\% $\downarrow$, UG-Mean $\uparrow$ \\
CIFAR-100-S & BiT & 0.4 & 1.000 & 1.000 & \ding{55} & -1.000 & 0.000 & \ding{51} & 0.991 & 0.000 & \ding{51} & FC\% $\downarrow$, UG-Mean $\uparrow$ \\
CIFAR-100-F & BiT & 0.4 & 1.000 & 1.000 & \ding{55} & -1.000 & 0.000 & \ding{51} & -0.418 & 1.000 & \ding{55} & FC\% $\downarrow$ \\
CIFAR-10 & BiT & 0.5 & 1.000 & 1.000 & \ding{55} & -1.000 & 0.000 & \ding{51} & 1.000 & 0.000 & \ding{51} & FC\% $\downarrow$, UG-Mean $\uparrow$ \\
CIFAR-100-S & BiT & 0.5 & 1.000 & 1.000 & \ding{55} & -1.000 & 0.000 & \ding{51} & 1.000 & 0.000 & \ding{51} & FC\% $\downarrow$, UG-Mean $\uparrow$ \\
CIFAR-100-F & BiT & 0.5 & 1.000 & 1.000 & \ding{55} & -1.000 & 0.000 & \ding{51} & 1.000 & 0.000 & \ding{51} & FC\% $\downarrow$, UG-Mean $\uparrow$ \\
CIFAR-10 & BiT & 0.6 & 1.000 & 1.000 & \ding{55} & -1.000 & 0.000 & \ding{51} & 1.000 & 0.000 & \ding{51} & FC\% $\downarrow$, UG-Mean $\uparrow$ \\
CIFAR-100-S & BiT & 0.6 & 1.000 & 1.000 & \ding{55} & -1.000 & 0.000 & \ding{51} & 1.000 & 0.000 & \ding{51} & FC\% $\downarrow$, UG-Mean $\uparrow$ \\
CIFAR-100-F & BiT & 0.6 & 1.000 & 1.000 & \ding{55} & -1.000 & 0.000 & \ding{51} & 1.000 & 0.000 & \ding{51} & FC\% $\downarrow$, UG-Mean $\uparrow$ \\
CIFAR-10 & CoAtNet & 0.2 & 0.704 & 1.000 & \ding{55} & 0.166 & 1.000 & \ding{55} & 0.102 & 0.013 & \ding{51} & UG-Mean $\uparrow$ (negilible effect) \\
CIFAR-100-S & CoAtNet & 0.2 & 1.000 & 1.000 & \ding{55} & 0.899 & 1.000 & \ding{55} & 0.936 & 0.000 & \ding{51} & UG-Mean $\uparrow$ \\
CIFAR-100-F & CoAtNet & 0.2 & 0.909 & 1.000 & \ding{55} & -1.000 & 0.000 & \ding{51} & -0.780 & 1.000 & \ding{55} & FC\% $\downarrow$ \\
CIFAR-10 & CoAtNet & 0.3 & 0.704 & 1.000 & \ding{55} & -0.119 & 0.244 & \ding{55} & 0.138 & 0.025 & \ding{51} & UG-Mean $\uparrow$ (negilible effect) \\
CIFAR-100-S & CoAtNet & 0.3 & 1.000 & 1.000 & \ding{55} & -0.940 & 0.000 & \ding{51} & 0.071 & 0.484 & \ding{55} & FC\% $\downarrow$, UG-Mean $\uparrow$ \\
CIFAR-100-F & CoAtNet & 0.3 & 0.909 & 1.000 & \ding{55} & -1.000 & 0.000 & \ding{51} & -0.827 & 1.000 & \ding{55} & FC\% $\downarrow$ \\
CIFAR-10 & CoAtNet & 0.4 & 0.704 & 1.000 & \ding{55} & -0.718 & 0.000 & \ding{51} & 0.700 & 0.000 & \ding{51} & FC\% $\downarrow$, UG-Mean $\uparrow$ \\
CIFAR-100-S & CoAtNet & 0.4 & 1.000 & 1.000 & \ding{55} & -1.000 & 0.000 & \ding{51} & 0.484 & 0.004 & \ding{51} & FC\% $\downarrow$, UG-Mean $\uparrow$ \\
CIFAR-100-F & CoAtNet & 0.4 & 0.909 & 1.000 & \ding{55} & -1.000 & 0.000 & \ding{51} & 0.996 & 0.000 & \ding{51} & FC\% $\downarrow$, UG-Mean $\uparrow$ \\
CIFAR-10 & CoAtNet & 0.5 & 0.704 & 1.000 & \ding{55} & -0.886 & 0.000 & \ding{51} & 0.929 & 0.000 & \ding{51} & FC\% $\downarrow$, UG-Mean $\uparrow$ \\
CIFAR-100-S & CoAtNet & 0.5 & 1.000 & 1.000 & \ding{55} & -1.000 & 0.000 & \ding{51} & 1.000 & 0.000 & \ding{51} & FC\% $\downarrow$, UG-Mean $\uparrow$ \\
CIFAR-100-F & CoAtNet & 0.5 & 0.909 & 1.000 & \ding{55} & -1.000 & 0.000 & \ding{51} & 1.000 & 0.000 & \ding{51} & FC\% $\downarrow$, UG-Mean $\uparrow$ \\
CIFAR-10 & CoAtNet & 0.6 & 0.704 & 1.000 & \ding{55} & -0.906 & 0.000 & \ding{51} & 0.984 & 0.000 & \ding{51} & FC\% $\downarrow$, UG-Mean $\uparrow$ \\
CIFAR-100-S & CoAtNet & 0.6 & 1.000 & 1.000 & \ding{55} & -1.000 & 0.000 & \ding{51} & 1.000 & 0.000 & \ding{51} & FC\% $\downarrow$, UG-Mean $\uparrow$ \\
CIFAR-100-F & CoAtNet & 0.6 & 0.909 & 1.000 & \ding{55} & -1.000 & 0.000 & \ding{51} & 1.000 & 0.000 & \ding{51} & FC\% $\downarrow$, UG-Mean $\uparrow$ \\ \hline
\end{tabular}
\end{sidewaystable}

The pairwise comparison against focal loss reveals a fundamentally different performance profile compared to the isotonic calibration baseline, highlighting the architectural dependency inherent in training-time calibration methods.
The results demonstrate a striking dichotomy based on model architecture. For BiT models, the dual calibration method achieved consistent and significant FC reduction across nearly all thresholds, with effect sizes predominantly reaching -1.000 (complete separation). This pattern held from $\tau = 0.2$ through $\tau = 0.6$, with 11 out of 15 BiT comparisons achieving primary FC reduction. The accompanying UG-Mean improvements in many cases (e.g., CIFAR-10 BiT at $\tau \geq  0.4$ showing both FC reduction and UG-Mean superiority) indicated that the method successfully addressed focal loss's tendency toward overconfident incorrect predictions in this architecture. The threshold progression revealed focal loss's increasing instability at permissive thresholds. While some configurations showed competitive performance at $\tau = 0.2-0.3$, the method's effectiveness deteriorated at higher thresholds.

Conversely, CoAtNet architectures exhibited an entirely different pattern, where focal loss generally maintained lower FC rates, forcing the dual calibration to rely on compensatory advantages. Remarkably, dual calibration achieved significant ECE improvements over focal loss in CoAtNet models (Cliff's Delta = -1.000, $p < 0.001$) across multiple configurations, contradicting expectations given that focal loss operates during training while dual calibration intentionally degrades ECE. This paradoxical result suggested that focal loss's training-time modifications disrupted CoAtNet's inherent calibration properties more severely than the intentional underconfidence injection of dual calibration.

The contrasting results between isotonic and focal loss comparisons validated the theoretical framework underlying dual calibration. Against isotonic calibration, the method needed to demonstrate that selective underconfidence provided safety benefits worth the calibration degradation. Against focal loss, the method often achieved multiple simultaneous improvements, suggesting that post-hoc calibration's architectural independence provided more reliable uncertainty quantification than training-time modifications.

The superiority of the proposed dual calibration method becomes increasingly evident as the entropy threshold increases from the conservative value of 0.2 to the more lenient value of 0.6. This raises a practical question: which threshold should be selected? Although no standardized method for threshold selection has been established in the literature, one approach is to maximize UAcc, which measures overall correctness in uncertainty-aware classification. As shown in the results, UAcc increases with higher thresholds, reaching its maximum at $\tau = 0.6$. However, at this threshold, UG-Mean declines substantially. At $\tau = 0.5$, both UAcc and UG-Mean approach their optimal values simultaneously, suggesting this threshold provides the best trade-off. Notably, the dual calibration method demonstrates its most pronounced advantages at this threshold. It is important to note that UAcc was not used as the primary comparison metric throughout this evaluation because it exhibits bias toward the majority class—in this case, TC, which dominates the sample distribution. Consequently, high UAcc values can coexist with elevated FC rates, which represent the most critical safety failures in uncertainty-aware systems. The UG-Mean metric addresses this limitation by providing a balanced assessment that simultaneously considers the model's ability to maintain confidence in correct predictions (captured by UTPR) and its capacity to appropriately flag incorrect predictions as uncertain (captured by 1-UFPR). Therefore, UG-Mean serves as the more appropriate metric for evaluating the safety-utility trade-off in uncertainty-aware calibration.

\begin{sidewaystable}[!htbp]
\centering
\caption{Summary of Wilcoxon signed-rank test results comparing the proposed Dual Calibration method against focal loss across entropy thresholds and datasets.}
\label{tab:summary_stat_test_vs_focal_loss}
\begin{tabular}{lllllllllllll}
\hline
\multicolumn{1}{c}{} & \multicolumn{1}{c}{} & \multicolumn{1}{c}{} & \multicolumn{3}{c}{ECE} & \multicolumn{3}{c}{FC\%} & \multicolumn{3}{c}{UG-Mean} & \multicolumn{1}{c}{} \\ \hline
\multicolumn{1}{c}{Dataset} & \multicolumn{1}{c}{Backbone} & \multicolumn{1}{c}{$\tau$} & \multicolumn{1}{c}{Cliff's Delta} & \multicolumn{1}{c}{P-value} & \multicolumn{1}{c}{Sig.} & \multicolumn{1}{c}{Cliff's Delta} & \multicolumn{1}{c}{P-value} & \multicolumn{1}{c}{Sig.} & \multicolumn{1}{c}{Cliff's Delta} & \multicolumn{1}{c}{P-value} & \multicolumn{1}{c}{Sig.} & \multicolumn{1}{c}{Note} \\ \hline
CIFAR-10 & BiT & 0.2 & 1.000 & 1.000 & \ding{55} & -1.000 & 0.000 & \ding{51} & -0.933 & 1.000 & \ding{55} & FC\% $\downarrow$ \\
CIFAR-100-S & BiT & 0.2 & 1.000 & 1.000 & \ding{55} & -0.006 & 0.897 & \ding{55} & 0.851 & 0.000 & \ding{51} & UG-Mean $\uparrow$ \\
CIFAR-100-F & BiT & 0.2 & 1.000 & 1.000 & \ding{55} & 1.000 & 1.000 & \ding{55} & 1.000 & 0.000 & \ding{51} & UG-Mean $\uparrow$ \\
CIFAR-10 & BiT & 0.3 & 1.000 & 1.000 & \ding{55} & -1.000 & 0.000 & \ding{51} & 0.753 & 0.000 & \ding{51} & FC\% $\downarrow$, UG-Mean $\uparrow$ \\
CIFAR-100-S & BiT & 0.3 & 1.000 & 1.000 & \ding{55} & -1.000 & 0.000 & \ding{51} & -0.918 & 1.000 & \ding{55} & FC\% $\downarrow$ \\
CIFAR-100-F & BiT & 0.3 & 1.000 & 1.000 & \ding{55} & 0.932 & 1.000 & \ding{55} & 1.000 & 0.000 & \ding{51} & UG-Mean $\uparrow$ \\
CIFAR-10 & BiT & 0.4 & 1.000 & 1.000 & \ding{55} & -1.000 & 0.000 & \ding{51} & 1.000 & 0.000 & \ding{51} & FC\% $\downarrow$, UG-Mean $\uparrow$ \\
CIFAR-100-S & BiT & 0.4 & 1.000 & 1.000 & \ding{55} & -1.000 & 0.000 & \ding{51} & -0.480 & 0.998 & \ding{55} & FC\% $\downarrow$ \\
CIFAR-100-F & BiT & 0.4 & 1.000 & 1.000 & \ding{55} & -0.386 & 0.006 & \ding{51} & 1.000 & 0.000 & \ding{51} & FC\% $\downarrow$, UG-Mean $\uparrow$ \\
CIFAR-10 & BiT & 0.5 & 1.000 & 1.000 & \ding{55} & -1.000 & 0.000 & \ding{51} & 1.000 & 0.000 & \ding{51} & FC\% $\downarrow$, UG-Mean $\uparrow$ \\
CIFAR-100-S & BiT & 0.5 & 1.000 & 1.000 & \ding{55} & -1.000 & 0.000 & \ding{51} & 1.000 & 0.000 & \ding{51} & FC\% $\downarrow$, UG-Mean $\uparrow$ \\
CIFAR-100-F & BiT & 0.5 & 1.000 & 1.000 & \ding{55} & -1.000 & 0.000 & \ding{51} & 0.971 & 0.000 & \ding{51} & FC\% $\downarrow$, UG-Mean $\uparrow$ \\
CIFAR-10 & BiT & 0.6 & 1.000 & 1.000 & \ding{55} & -1.000 & 0.000 & \ding{51} & 1.000 & 0.000 & \ding{51} & FC\% $\downarrow$, UG-Mean $\uparrow$ \\
CIFAR-100-S & BiT & 0.6 & 1.000 & 1.000 & \ding{55} & -1.000 & 0.000 & \ding{51} & 1.000 & 0.000 & \ding{51} & FC\% $\downarrow$, UG-Mean $\uparrow$ \\
CIFAR-100-F & BiT & 0.6 & 1.000 & 1.000 & \ding{55} & -1.000 & 0.000 & \ding{51} & -0.007 & 0.887 & \ding{55} & FC\% $\downarrow$ \\
CIFAR-10 & CoAtNet & 0.2 & -1.000 & 0.000 & \ding{51} & 1.000 & 1.000 & \ding{55} & 1.000 & 0.000 & \ding{51} & UG-Mean $\uparrow$, ECE $\downarrow$ \\
CIFAR-100-S & CoAtNet & 0.2 & -0.811 & 0.000 & \ding{51} & 1.000 & 1.000 & \ding{55} & 1.000 & 0.000 & \ding{51} & UG-Mean $\uparrow$, ECE $\downarrow$ \\
CIFAR-100-F & CoAtNet & 0.2 & -1.000 & 0.000 & \ding{51} & 1.000 & 1.000 & \ding{55} & 1.000 & 0.000 & \ding{51} & UG-Mean $\uparrow$, ECE $\downarrow$ \\
CIFAR-10 & CoAtNet & 0.3 & -1.000 & 0.000 & \ding{51} & 1.000 & 1.000 & \ding{55} & -0.578 & 1.000 & \ding{55} & ECE $\downarrow$ \\
CIFAR-100-S & CoAtNet & 0.3 & -0.811 & 0.000 & \ding{51} & 1.000 & 1.000 & \ding{55} & 1.000 & 0.000 & \ding{51} & UG-Mean $\uparrow$, ECE $\downarrow$ \\
CIFAR-100-F & CoAtNet & 0.3 & -1.000 & 0.000 & \ding{51} & 0.998 & 1.000 & \ding{55} & 1.000 & 0.000 & \ding{51} & UG-Mean $\uparrow$, ECE $\downarrow$ \\
CIFAR-10 & CoAtNet & 0.4 & -1.000 & 0.000 & \ding{51} & 1.000 & 1.000 & \ding{55} & -0.989 & 1.000 & \ding{55} & ECE $\downarrow$ \\
CIFAR-100-S & CoAtNet & 0.4 & -0.811 & 0.000 & \ding{51} & 0.946 & 1.000 & \ding{55} & 1.000 & 0.000 & \ding{51} & UG-Mean $\uparrow$, ECE $\downarrow$ \\
CIFAR-100-F & CoAtNet & 0.4 & -1.000 & 0.000 & \ding{51} & 0.952 & 1.000 & \ding{55} & 1.000 & 0.000 & \ding{51} & UG-Mean $\uparrow$, ECE $\downarrow$ \\
CIFAR-10 & CoAtNet & 0.5 & -1.000 & 0.000 & \ding{51} & 1.000 & 1.000 & \ding{55} & -1.000 & 1.000 & \ding{55} & ECE $\downarrow$ \\
CIFAR-100-S & CoAtNet & 0.5 & -0.811 & 0.000 & \ding{51} & -0.146 & 0.229 & \ding{55} & 0.982 & 0.000 & \ding{51} & UG-Mean $\uparrow$, ECE $\downarrow$ \\
CIFAR-100-F & CoAtNet & 0.5 & -1.000 & 0.000 & \ding{51} & 1.000 & 1.000 & \ding{55} & 0.047 & 0.543 & \ding{55} & ECE $\downarrow$ \\
CIFAR-10 & CoAtNet & 0.6 & -1.000 & 0.000 & \ding{51} & 0.942 & 1.000 & \ding{55} & -0.931 & 1.000 & \ding{55} & ECE $\downarrow$ \\
CIFAR-100-S & CoAtNet & 0.6 & -0.811 & 0.000 & \ding{51} & -0.883 & 0.000 & \ding{51} & 0.487 & 0.001 & \ding{51} & FC\% $\downarrow$, UG-Mean $\uparrow$, ECE $\downarrow$ \\
CIFAR-100-F & CoAtNet & 0.6 & -1.000 & 0.000 & \ding{51} & 1.000 & 1.000 & \ding{55} & -1.000 & 1.000 & \ding{55} & ECE $\downarrow$ \\ \hline
\end{tabular}
\end{sidewaystable}

\begin{figure}[!htbp]
  \centering
  \captionsetup[subfloat]{font=tiny}
  \subfloat[CIFAR-10, BiT]{\includegraphics[width=0.9\textwidth]{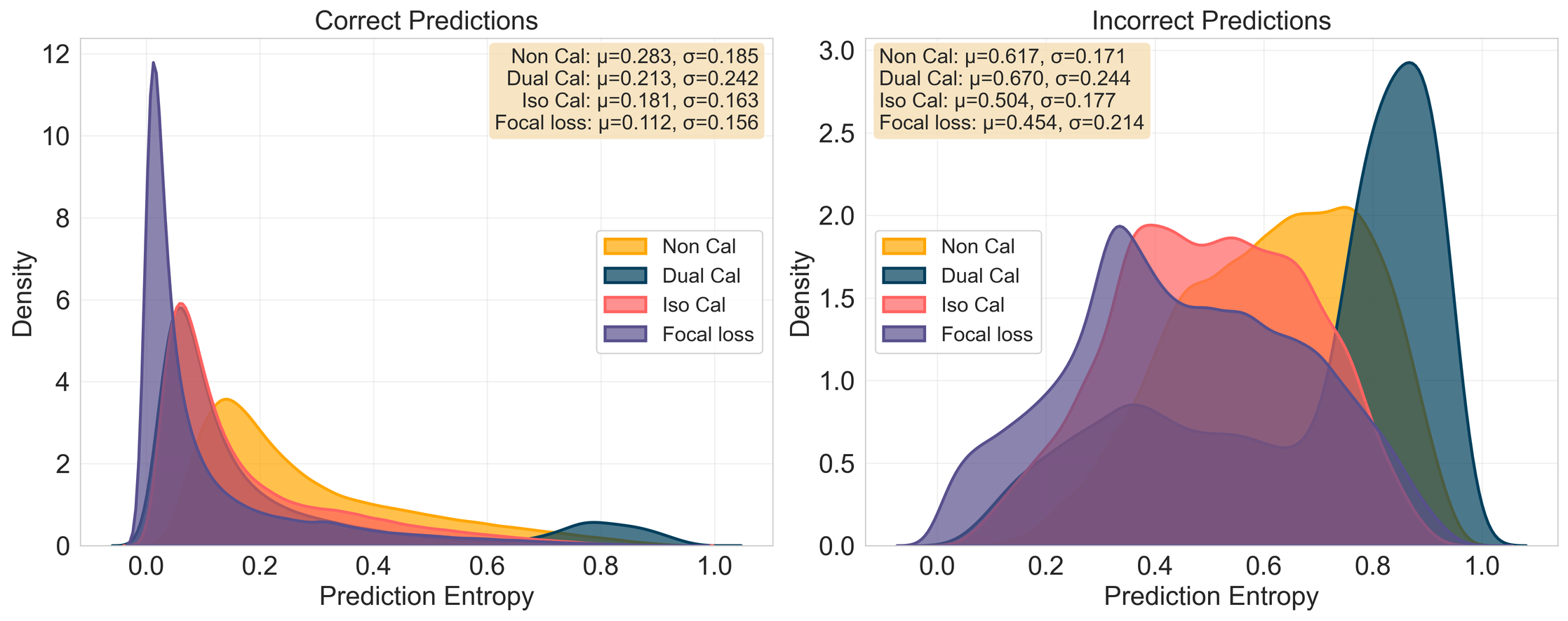}}\par
  \subfloat[CIFAR-100-S, BiT]{\includegraphics[width=0.9\textwidth]{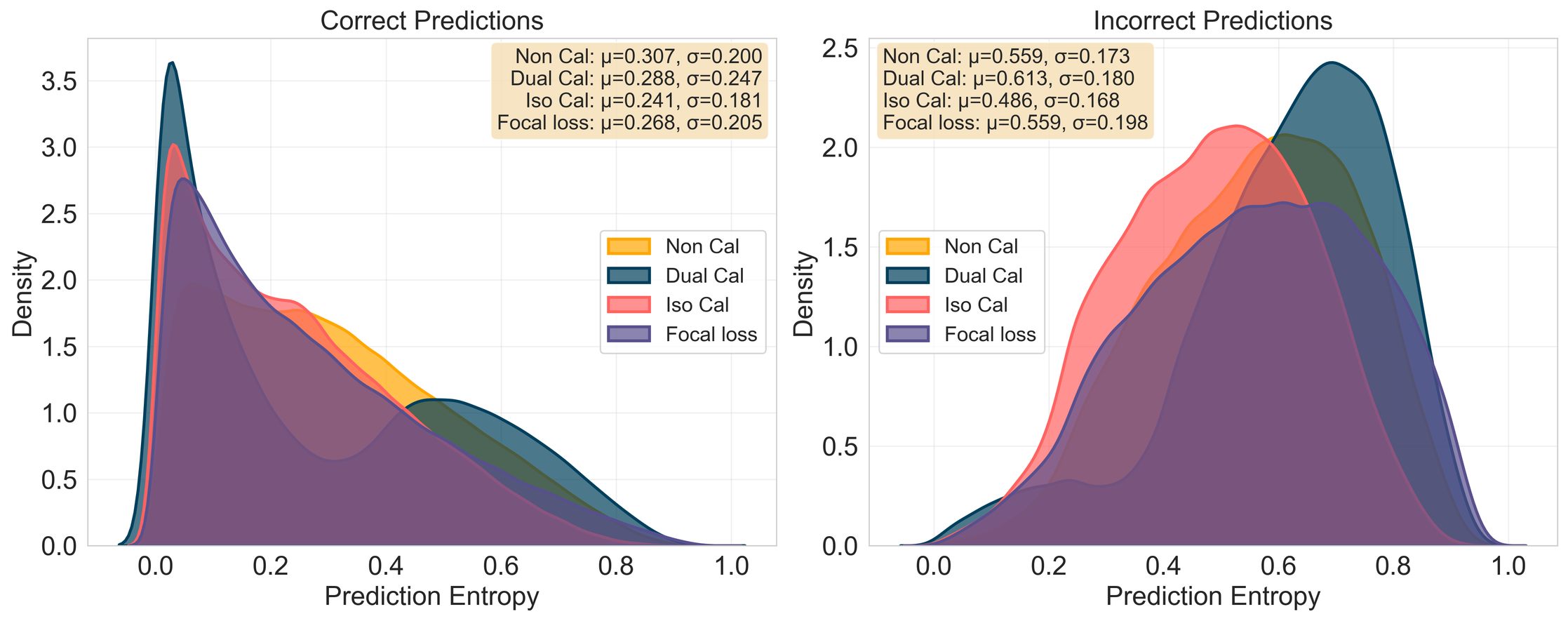}}\par
  \subfloat[CIFAR100-F, BiT]{\includegraphics[width=0.9\textwidth]{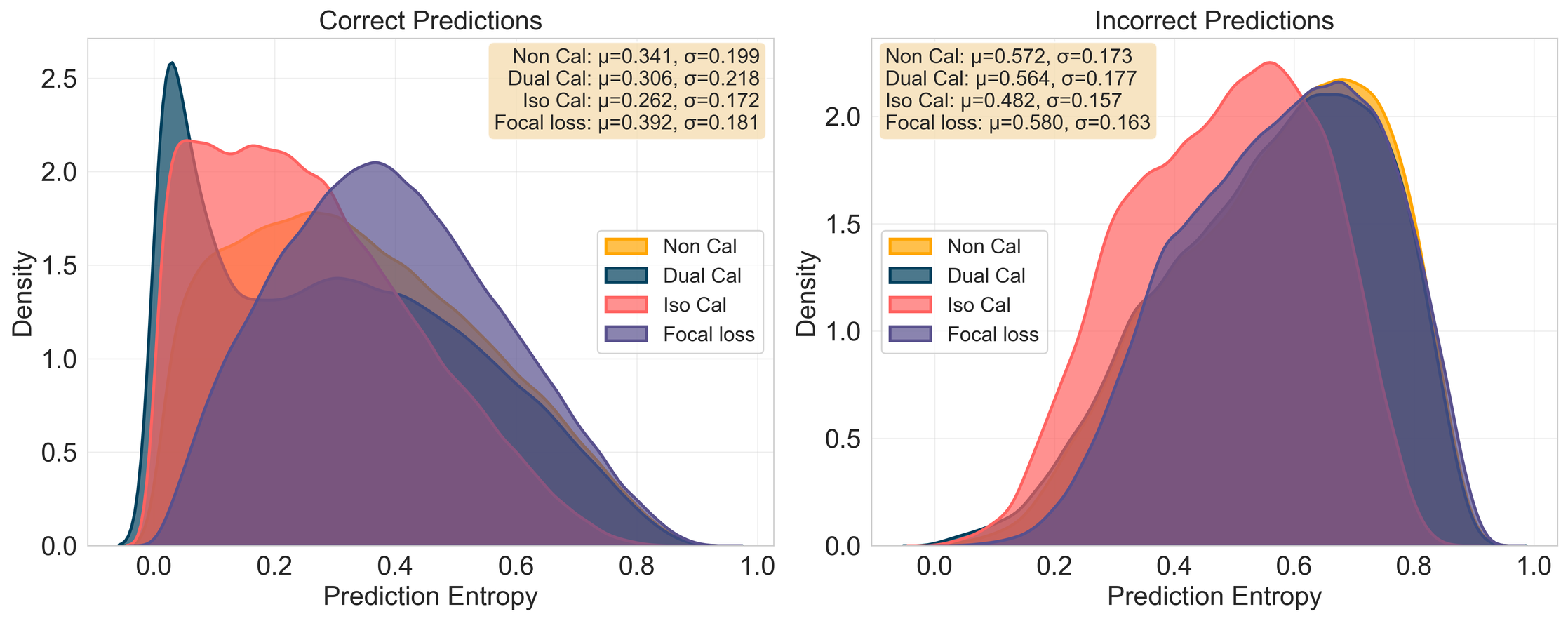}}\par
  \caption{Distribution of prediction entropy for correct (left) and incorrect (right) predictions across different datasets with BiT backbone across different calibration methods.}
  \label{fig:PredictionEntropy_bit}
\end{figure}

\begin{figure}[!htbp]
  \centering
  \captionsetup[subfloat]{font=tiny}
  \subfloat[CIFAR-10, CoAtNet]{\includegraphics[width=0.9\textwidth]{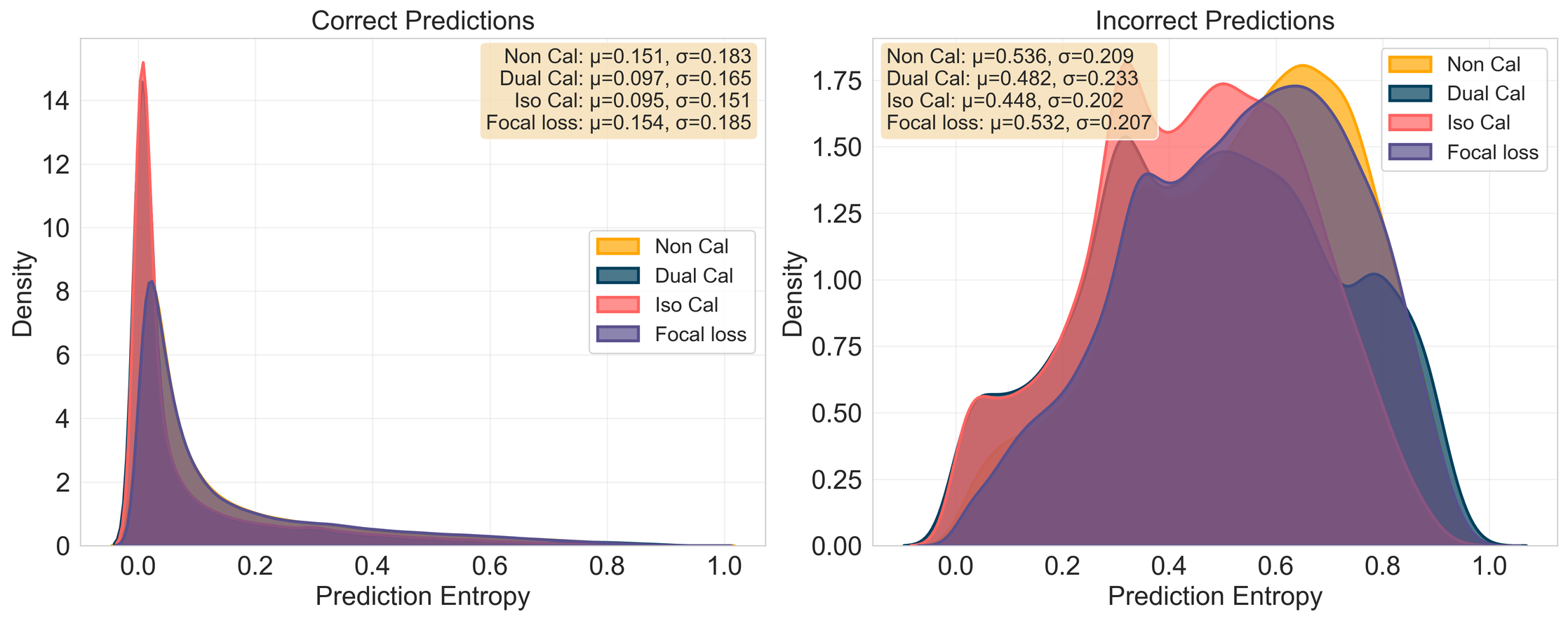}}\par
  \subfloat[CIFAR-100-S, CoAtNet]{\includegraphics[width=0.9\textwidth]{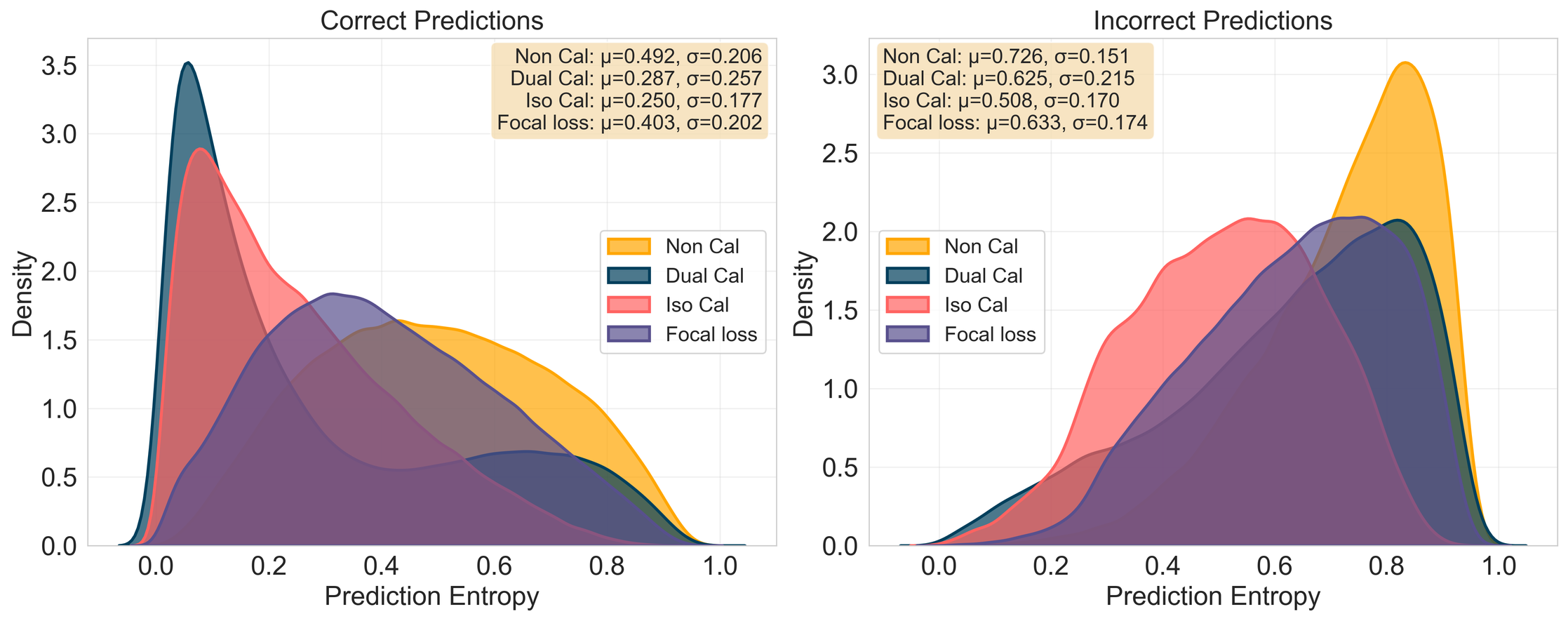}}\par
  \subfloat[CIFAR-100-F, CoAtNet]{\includegraphics[width=0.9\textwidth]{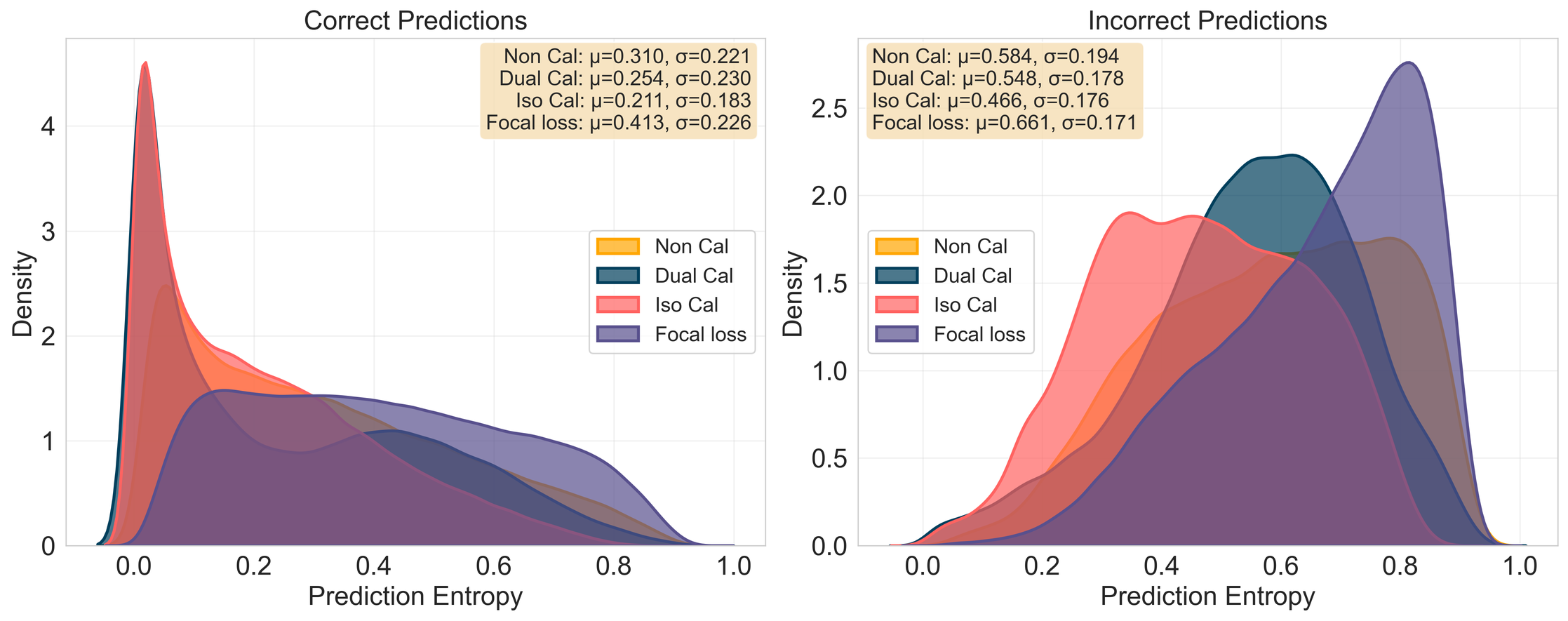}}\par
  \caption{Distribution of prediction entropy for correct (left) and incorrect (right) predictions across different datasets with CoAtNet backbone across different calibration methods.}
  \label{fig:PredictionEntropy_coatnet}
\end{figure}

Having established the comparative performance trends across entropy thresholds, the next analysis examines the internal behavior driving these outcomes through the distribution of prediction entropy for correct and incorrect samples. Figures~\ref{fig:PredictionEntropy_bit} and \ref{fig:PredictionEntropy_coatnet} illustrate the distribution of prediction entropy for correct and incorrect samples across different calibration methods. Notably, the proposed dual calibration method demonstrates a distinct shift in the entropy of incorrect predictions toward the higher end of the uncertainty spectrum (i.e., entropy $\approx$ 1.0). This shift reflects the mechanism by which dual calibration effectively reduces false certainty: by promoting uncertainty in samples likely to be incorrect.

It is important to emphasize that this analysis is based on true correctness labels (i.e., actual correct vs. incorrect predictions), in contrast to earlier stratifications based on putative correctness derived from conformal prediction. Therefore, this visualization directly captures the true impact of calibration on prediction trustworthiness.

The success of dual calibration in reducing FC is visually explained by its ability to increase entropy — and thus model caution — where it matters most: in the truly incorrect predictions.

As evidenced in the prediction entropy plots for correctly classified samples, the entropy distributions are not entirely concentrated near zero (i.e., in the region of high certainty). Instead, a secondary peak is observed in the higher-entropy (uncertain) region. This phenomenon is particularly noticeable for the CIFAR-100-Coarse dataset with CoAtNet and BiT backbones. The underlying cause of this shift can be attributed to the conformal prediction step misclassifying certain correctly predicted instances as putatively incorrect. Consequently, these samples are subjected to the incorrect-specific calibration, which inadvertently increases their entropy and displaces them from the certainty region.

This unintended entropy inflation within the correct prediction group diminishes the separability between correct and incorrect predictions in the entropy space. Enhancing the accuracy of the conformal prediction stage—particularly in its ability to reliably distinguish between correct and incorrect predictions—has the potential to sharpen this separation. Specifically, it would encourage entropy distributions for correct predictions to remain distinctly peaked near zero (reflecting high certainty), while those for incorrect predictions would remain concentrated at higher entropy levels, thereby improving trust-aware decision-making.

\subsection{Ablation Study}
An ablation study is conducted to investigate the influence of the two critical hyperparameters on the performance of the proposed dual calibration framework. Specifically, the analysis examines $K$, representing the number of nearest neighbors in the semantic conformal prediction component, and $\beta$, denoting the underconfidence ratio applied during calibration. 

To isolate the individual contribution of each parameter, the analysis is partitioned into two sequential investigations. Section \ref{sec:ablation_k} evaluates the impact of the $K$ through systematic variation while holding $\beta = 0.9$ constant, thus characterizing the sensitivity of the flagging mechanism to the proximity threshold. Section \ref{sec:ablation_beta} examines the influence of the $\beta$  by varying its value while maintaining $K=20$ fixed, thereby isolating the effect of the calibration strategy on model performance. To avoid exhaustive and repetitive analysis, the ablation discussion is therefore restricted to three key metrics: ECE, FC\%, and TC\% at the entropy threshold $\tau=0.5$ to characterize the primary behavioral dynamics of the proposed framework. The selection of FC and TC is motivated by their foundational role in uncertainty-aware evaluation, which together provide sufficient insight into how the proposed dual calibration framework balances calibration and uncertainty-aware decisions.

\subsubsection{Sensitivity of Conformal Prediction to Neighborhood Size $K$} \label{sec:ablation_k}
The neighborhood size parameter K in the proximity-based conformal prediction mechanism directly influences the composition and quality of sample stratification, which subsequently affects the effectiveness of the dual calibration approach. This parameter affects the accuracy of sample stratification into putatively correct and putatively incorrect groups; second, these stratification outcomes directly impact the effectiveness of the proposed dual calibration approach, as the quality of targeted calibration depends on the precision of sample separation.

\begin{figure}[!htbp]
  \centering
  \captionsetup[subfloat]{font=tiny}
  \subfloat[Size of putatively correct samples]{\includegraphics[width=0.5\textwidth]{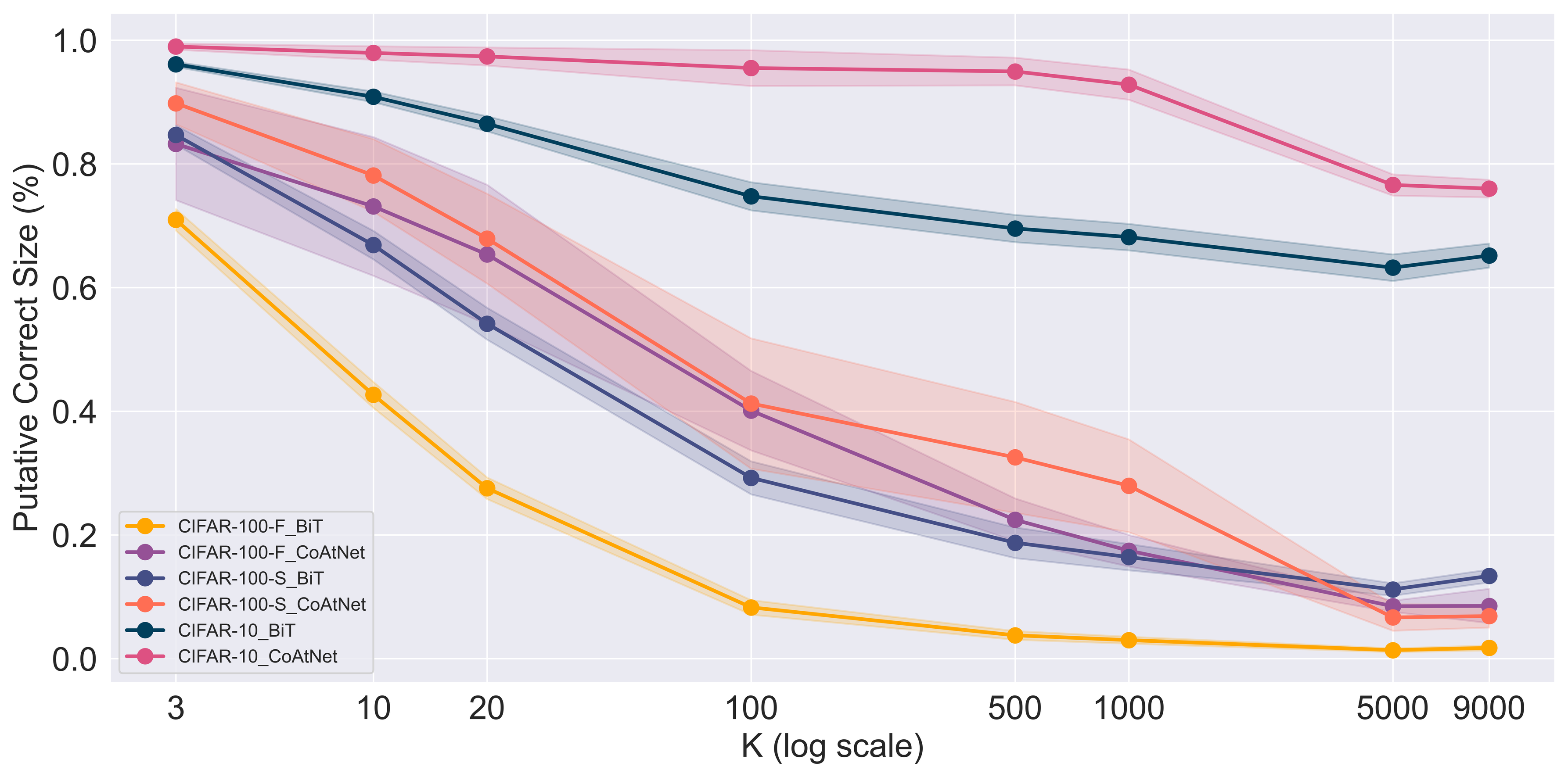}}
  \subfloat[Accuracy of putatively correct samples]{\includegraphics[width=0.5\textwidth]{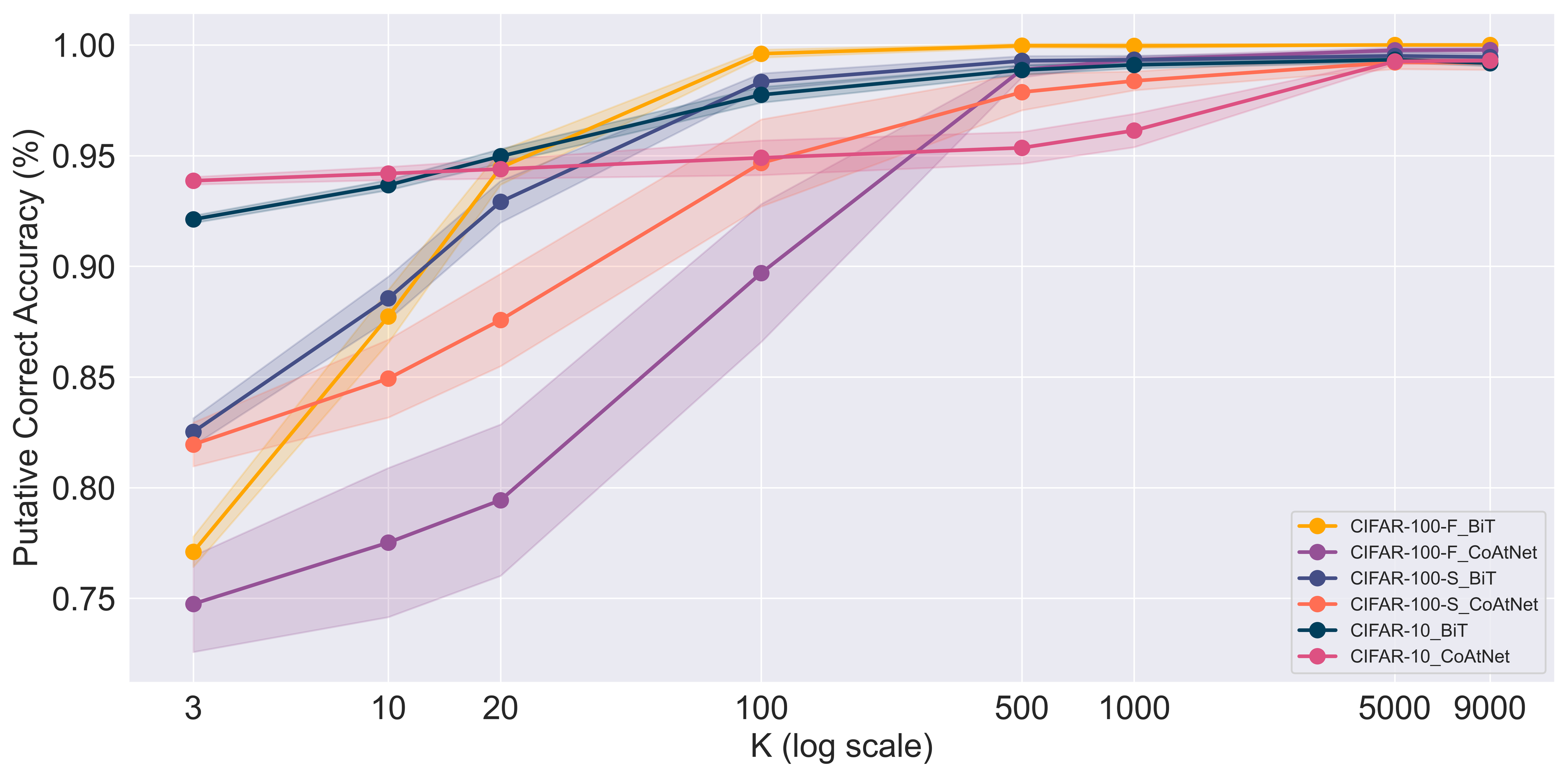}}\par
  \subfloat[Size of putatively incorrect samples]{\includegraphics[width=0.5\textwidth]{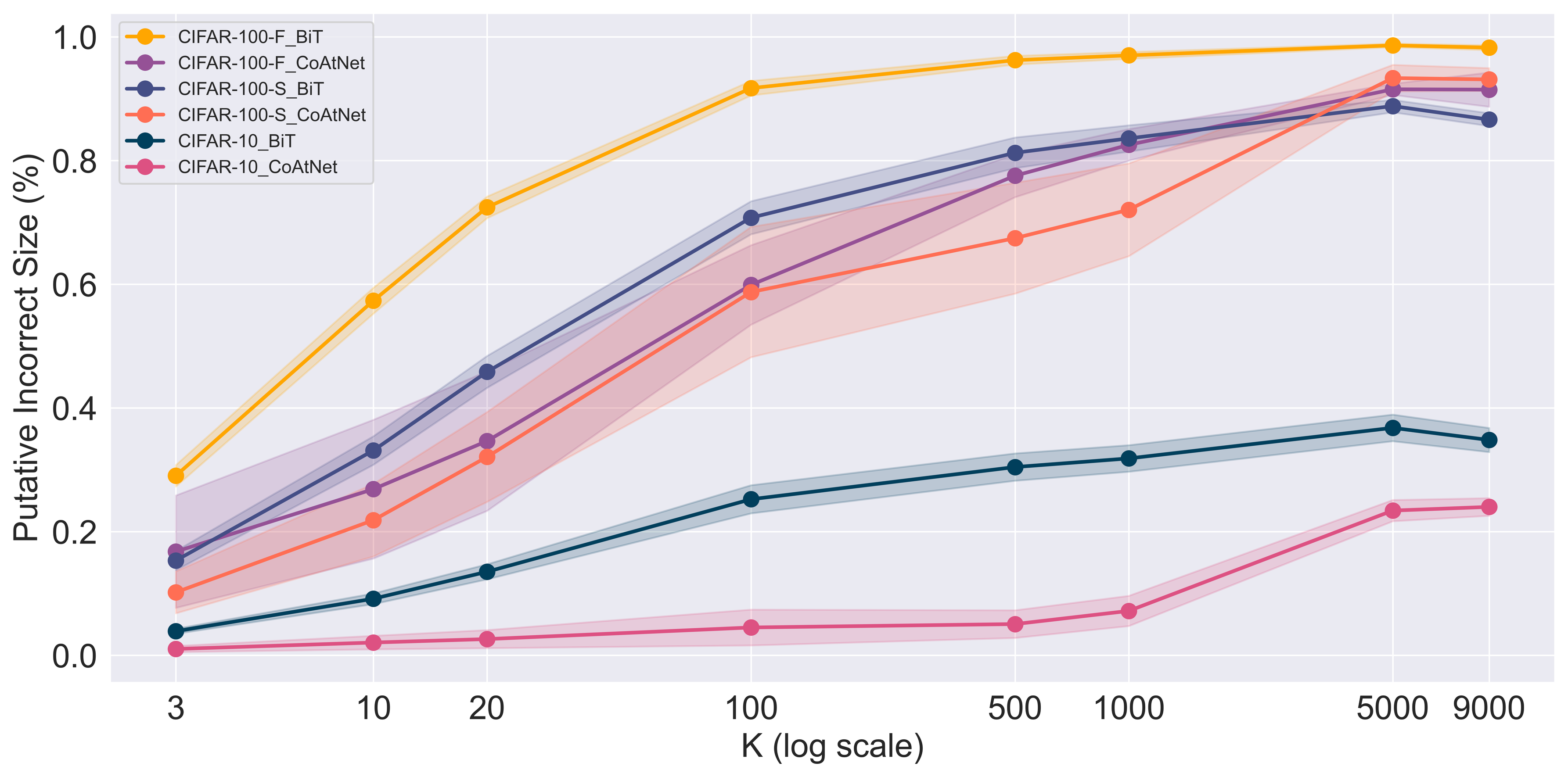}}
  \subfloat[Accuracy of putatively incorrect samples]{\includegraphics[width=0.5\textwidth]{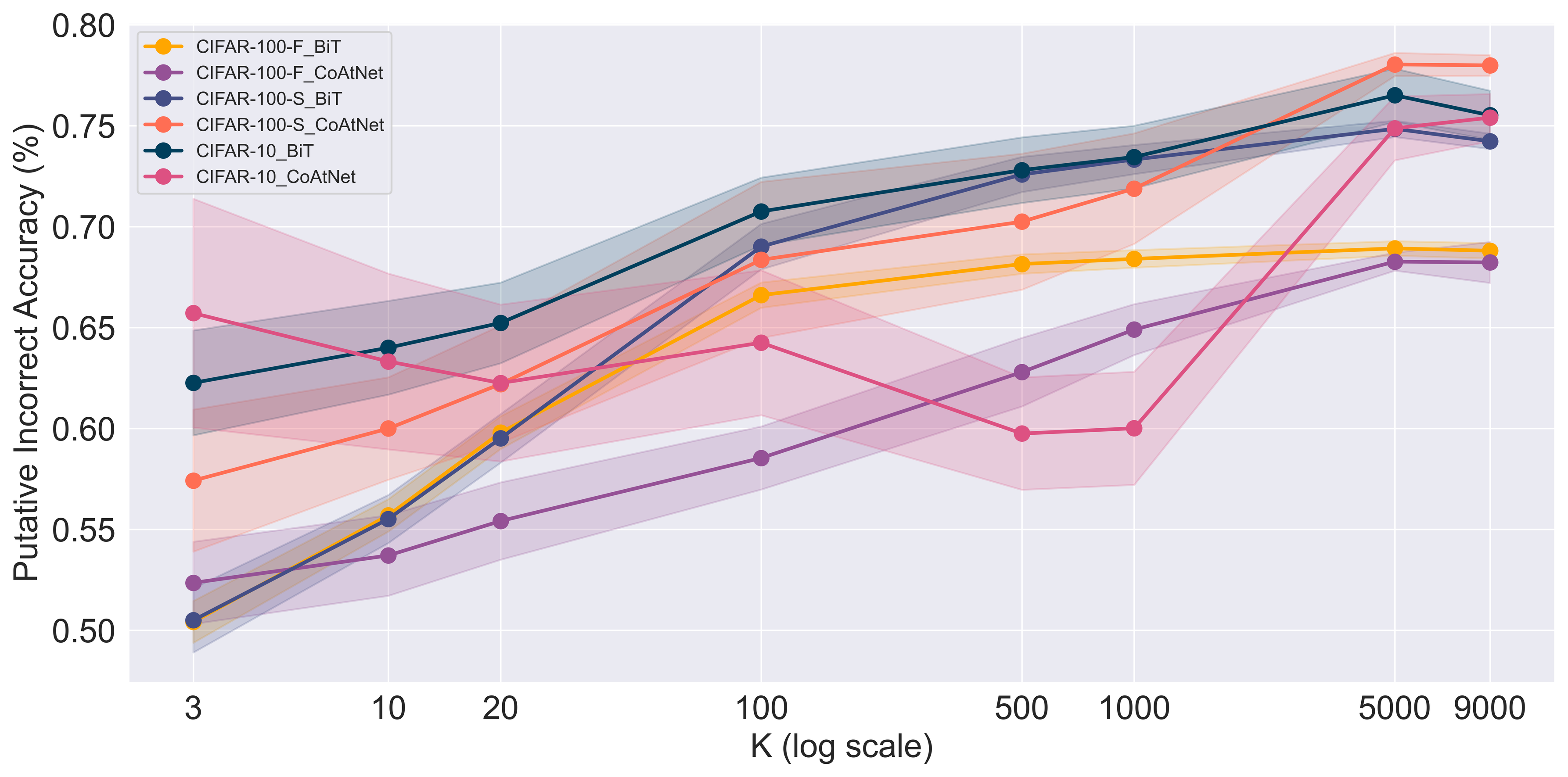}}\par
  \caption{Effect of neighborhood size K on sample stratification into putatively correct and putatively incorrect groups ($\beta = 0.9$ fixed).}
  \label{fig:ablation_impact_k_conformal_result}
\end{figure}

The ablation results are presented in Figures \ref{fig:ablation_impact_k_conformal_result}, where the effect of varying the neighborhood size $K$ (log-scaled for clarity) on the stratification performance is illustrated. The stratification ideally aims to assign all correctly classified samples into the putatively correct group and all misclassified samples into the putatively incorrect group. For instance, as shown in Table \ref{tab:conformal_result}, the baseline accuracy of the BiT model on CIFAR-10 is 91.326\%, implying that an ideal stratification would place 91.326\% of the samples into the putatively correct group and 8.674\% into the putatively incorrect group, with both groups achieving 100\% accuracy internally.

However, the empirical results reveal a clear departure from this ideal behavior as $K$ increases. For CIFAR-100-S and CIFAR-100-F across both backbones, the size of the putatively incorrect group grows substantially beyond the true proportion of misclassified samples once $K>100$. This indicates that a considerable fraction of correctly predicted samples are being mis-flagged as putatively incorrect. Naturally, this expansion of the putatively incorrect group coincides with a reduction in the size of the putatively correct group. Despite this, the accuracy of the putatively correct group increases with $K$, approaching near-perfect separation for larger $K$.

Concurrently, the accuracy of the putatively incorrect group also appears to increase with 
$K$. This increase was a natural consequence of the stratification design: once the putatively correct group became highly purified (i.e., containing almost no misclassified samples), all misclassified samples were concentrated in the putatively incorrect group, which inherently raised its accuracy.

\begin{figure}[!htbp]
  \centering
  \captionsetup[subfloat]{font=tiny}
  \subfloat[Sensitivity of ECE to $K$]{\includegraphics[width=0.5\textwidth]{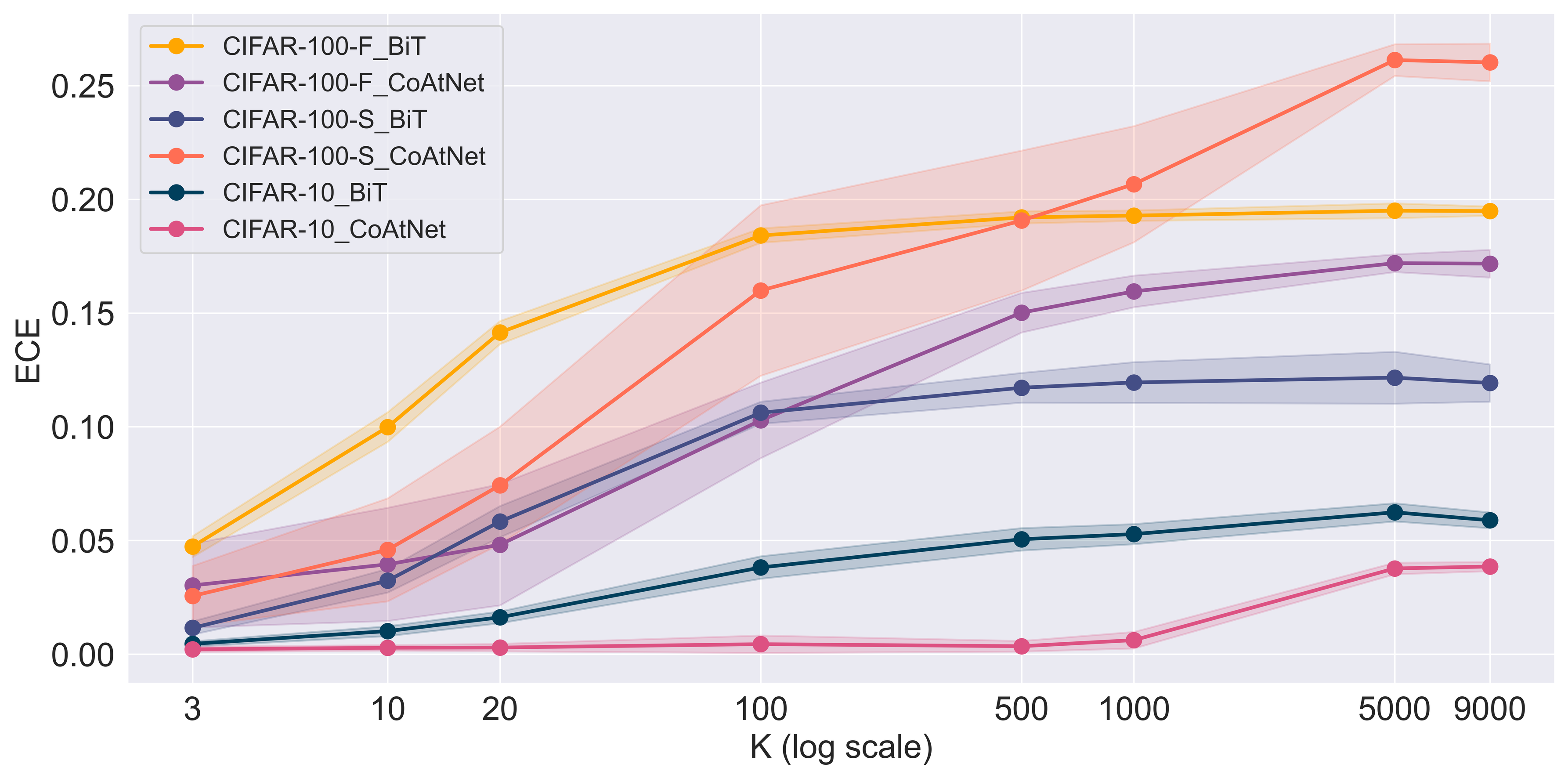}}\par
  \subfloat[Sensitivity of FC\% to $K$]{\includegraphics[width=0.5\textwidth]{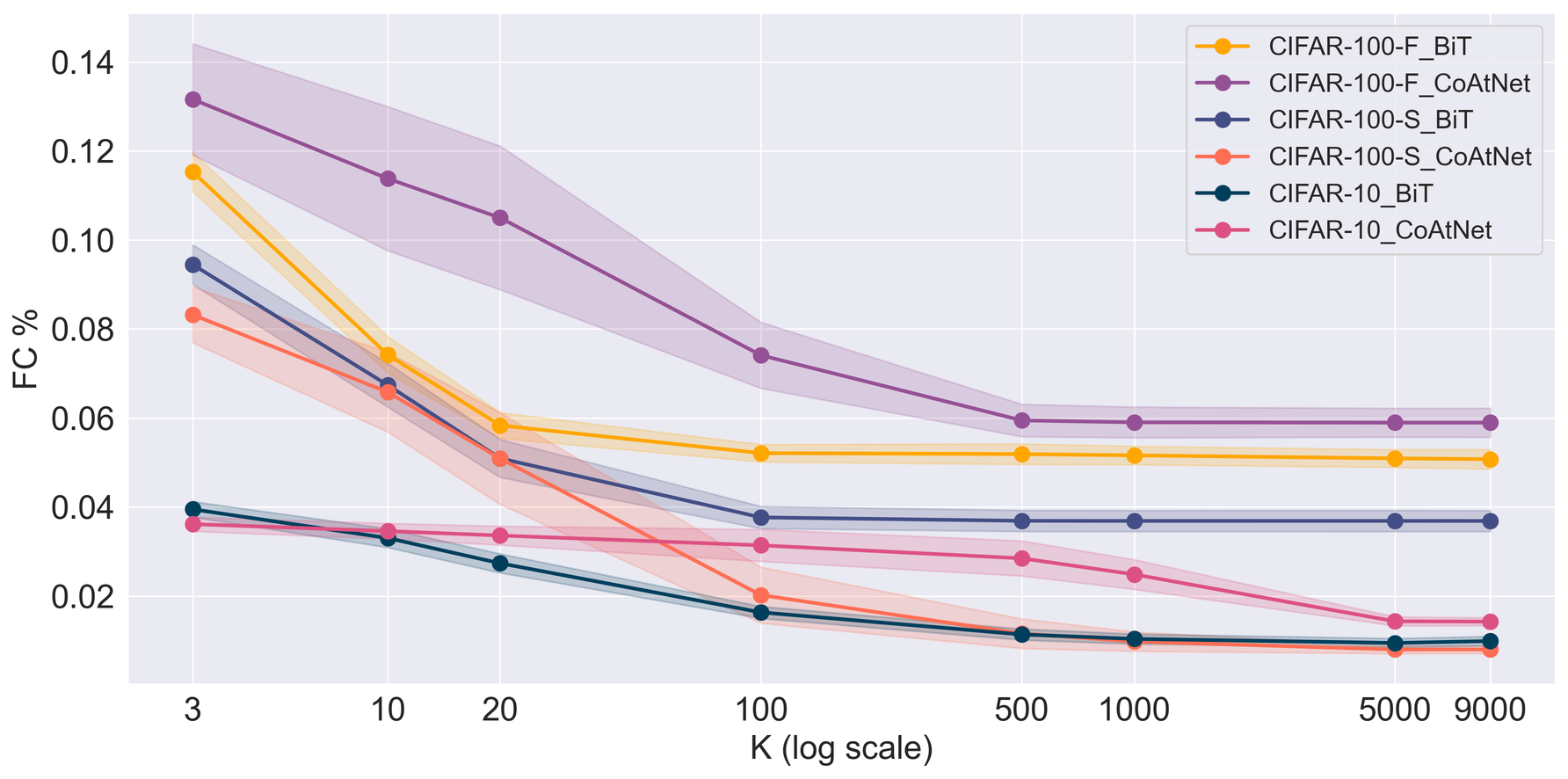}}
  \subfloat[Sensitivity of TC\% to $K$]{\includegraphics[width=0.5\textwidth]{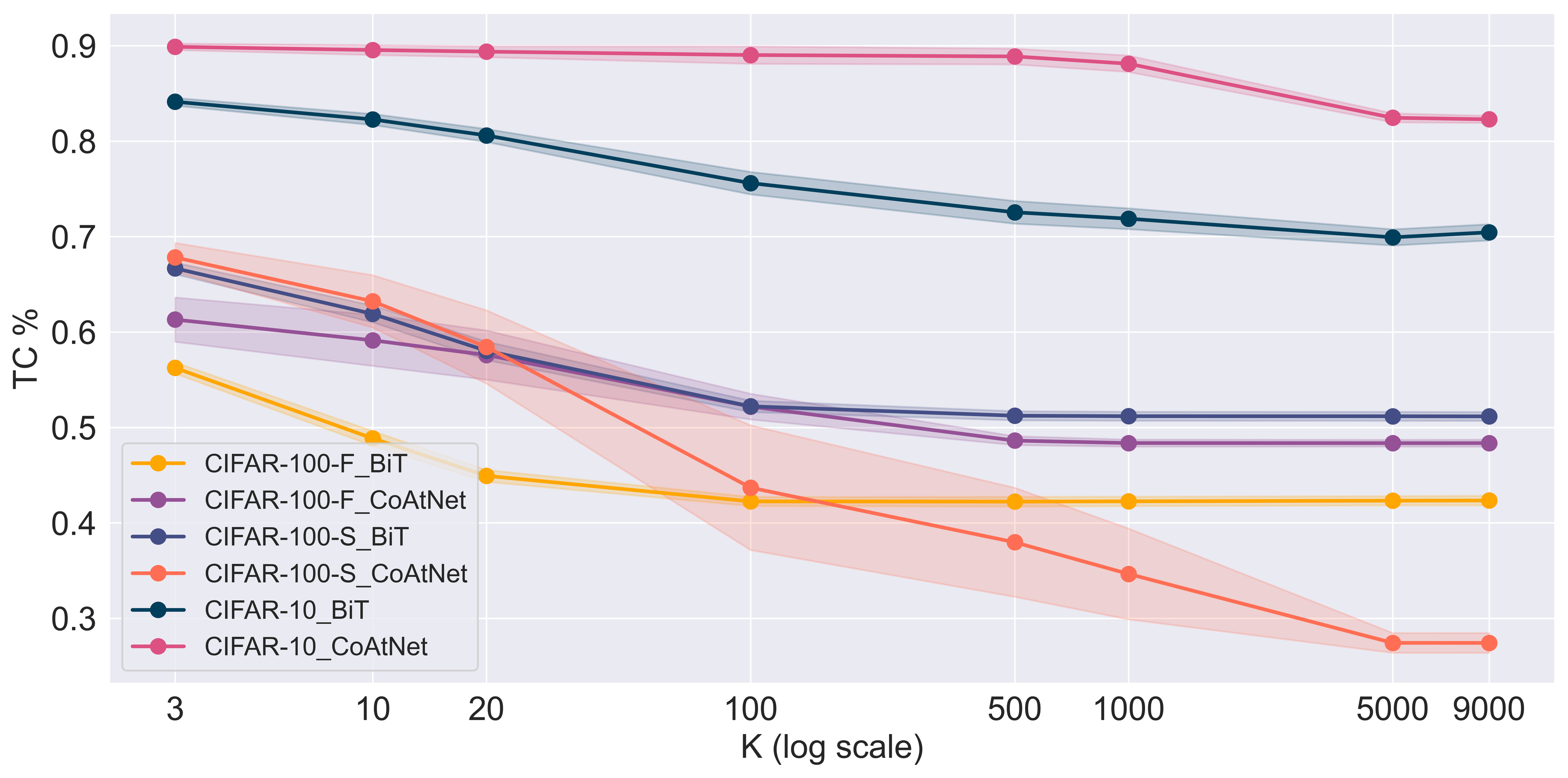}}
  \caption{Sensitivity of ECE, FC\%, and TC\% to conformal prediction neighborhood size K ($\beta = 0.9$ fixed).}
  \label{fig:ablation_impact_k_performance_result}
\end{figure}

These dynamics, however, were associated with notable calibration consequences. The proposed dual calibration method pushes prediction probabilities of putative incorrect samples toward uniform distributions, thereby inflating the calibration error. As more samples were allocated to this group at higher values of k, the overall ECE rose steadily, as shown in Figure \ref{fig:ablation_impact_k_performance_result}, particularly for CIFAR-100 models. This pattern illustrates the trade-off: while stratification purified the putatively correct group, the price was paid in elevated miscalibration among samples assigned as putatively incorrect.

Finally, the reduction in FC\% with increasing $K$ (Figure \ref{fig:ablation_impact_k_performance_result}) is explained by the same mechanism. As $K$ grows, a larger fraction of samples are flagged into the putatively incorrect group and their probabilities are shifted toward uncertainty. This reduces the number of overconfident but incorrect predictions, thereby decreasing FC\%.

However, the expansion of the putatively incorrect group comes at the cost of a decrease in the TC rate. As shown in Figures \ref{fig:ablation_impact_k_performance_result}, the TC percentage consistently decreases with larger values of $k$, confirming that a portion of the correct predictions are treated as uncertain.

Overall, these results demonstrate that increasing k enhances the purification of the putatively correct set and improves FC detection, but at the expense of higher calibration error and lower TC. The balance of these effects highlights that moderate values of k offer the most favorable trade-offs, while excessively large neighborhoods degrade reliability by over-allocating correct predictions to the uncertain region.

\subsubsection{Sensitivity to Underconfidence Calibration Ratio $\beta$} \label{sec:ablation_beta}
The behavior of ECE across different values of the underconfidence factor $\beta$ reflects the trade-off between fidelity to the model’s original probabilities and the degree of regularization toward maximum entropy. When $\beta = 0$, the transformation collapses all predictions to the uniform distribution $1/C$. In this extreme case, confidence is completely removed, and calibration error is minimal since predicted probabilities trivially match the uniform expectation, but discriminative information is lost.

As $\beta$ increases from 0 to approximately 0.2, a sharp rise in ECE is observed across all datasets and backbones. This increase arises because predictions are partially influenced by the original model outputs, which may be systematically overconfident, while still being strongly pulled toward uniform. This “hybrid” distribution introduces mismatch with the true empirical frequencies, thereby inflating calibration error.

Beyond $\beta \approx 0.2$, ECE decreases monotonically as $\beta \to 1$. In this regime, the transformation preserves more of the original predicted probabilities, while still applying sufficient regularization to temper extreme overconfidence. As a result, calibrated probabilities align more closely with empirical correctness, leading to improved calibration.

The sensitivity analysis with respect to the underconfidence factor $\beta$ reveals distinct behavioral trends across the uncertainty metrics. At low values of $\beta$ (close to zero), the calibrated probabilities are strongly mixed with the uniform distribution, resulting in highly attenuated confidence scores. Under this regime, the proportion of false confident predictions (FC\%) remains minimal, since incorrect predictions are rarely assigned high confidence. 

However, the variation of TC with the $\beta$ is strongly tied to the stratification presented in Table \ref{tab:conformal_result}. At $k=20$, except for CIFAR-100-F with BiT, the majority of samples are allocated to the putatively correct group, which also achieves a high classification accuracy (e.g., 93.8\% for CIFAR-100-S with BiT). In contrast, the putatively incorrect group exhibits markedly lower accuracy, typically around 60\%.

When $\beta$ is small the proportion of samples identified as true confident stems almost exclusively from the putatively correct group. Correctly predicted samples that have been mistakenly placed into the putatively incorrect group are heavily regularized toward maximum uncertainty, thereby preventing their detection as true confident. Consequently, the overall TC percentage remains essentially fixed at the level of low $\beta$.

\begin{figure}[!htbp]
  \centering
  \captionsetup[subfloat]{font=tiny}
  \subfloat[Sensitivity of ECE to $\beta$]{\includegraphics[width=0.5\textwidth]{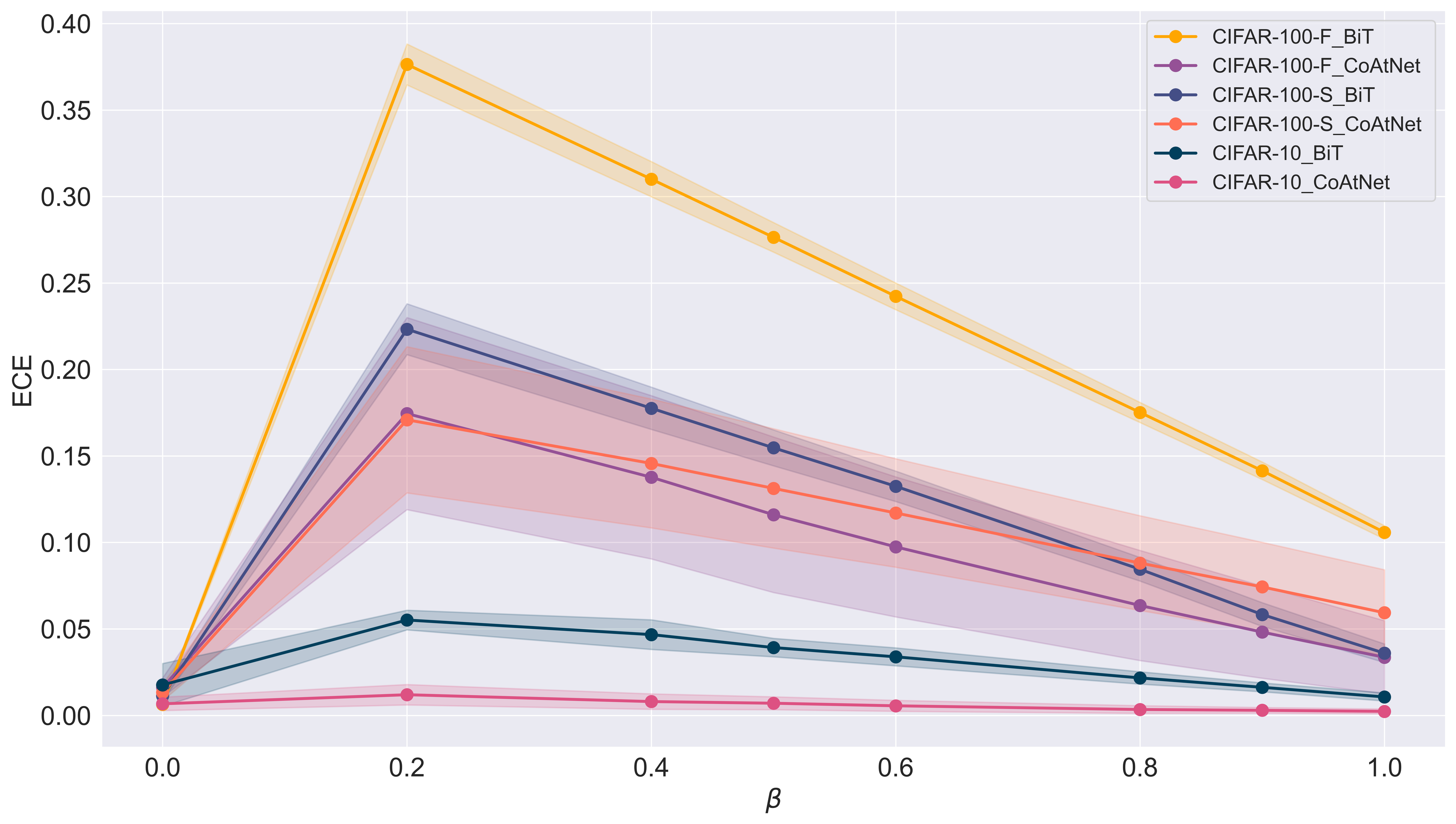}}\par
  \subfloat[Sensitivity of FC\% to $\beta$]{\includegraphics[width=0.5\textwidth]{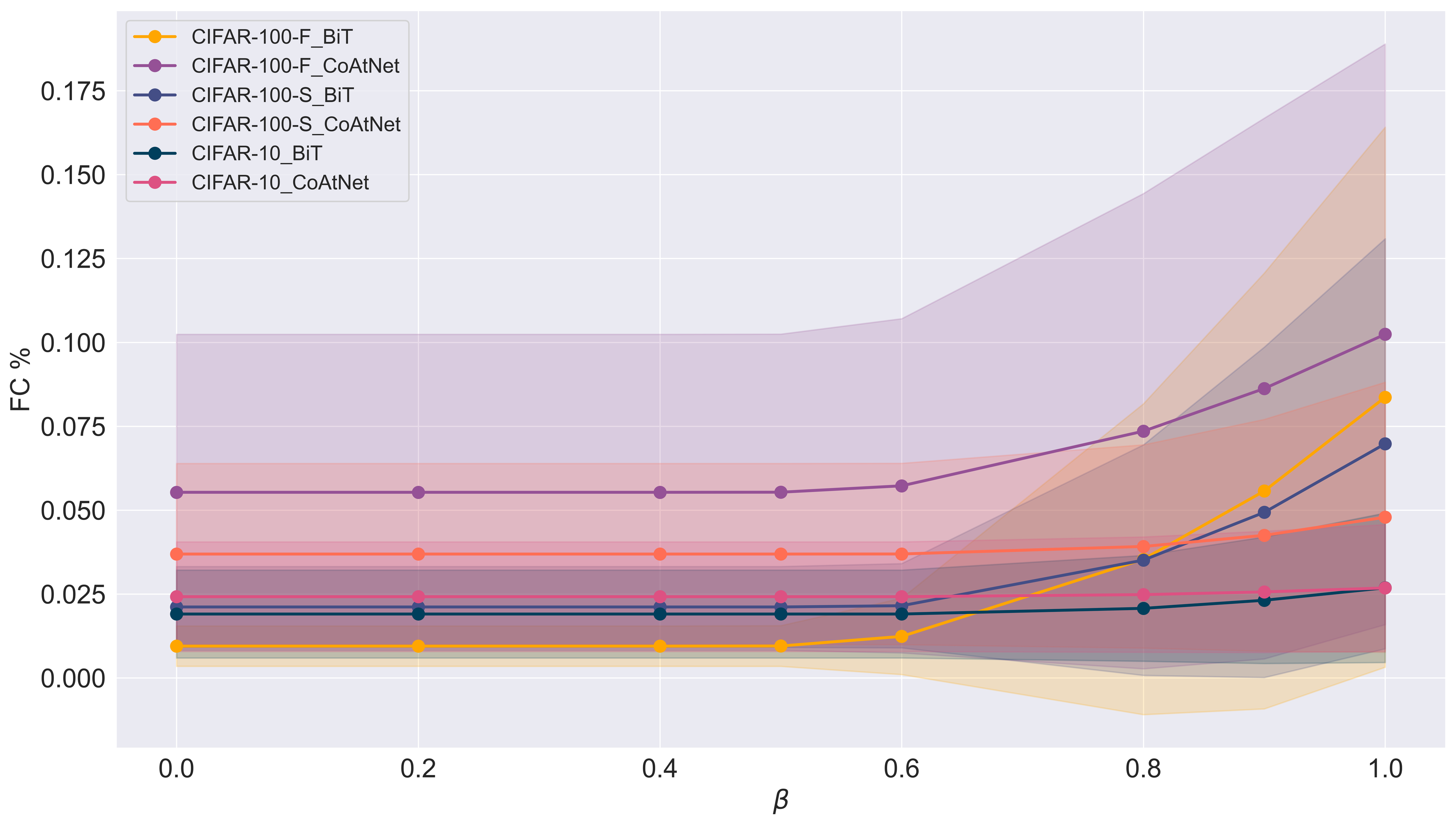}}
  \subfloat[Sensitivity of TC\% to $\beta$]{\includegraphics[width=0.5\textwidth]{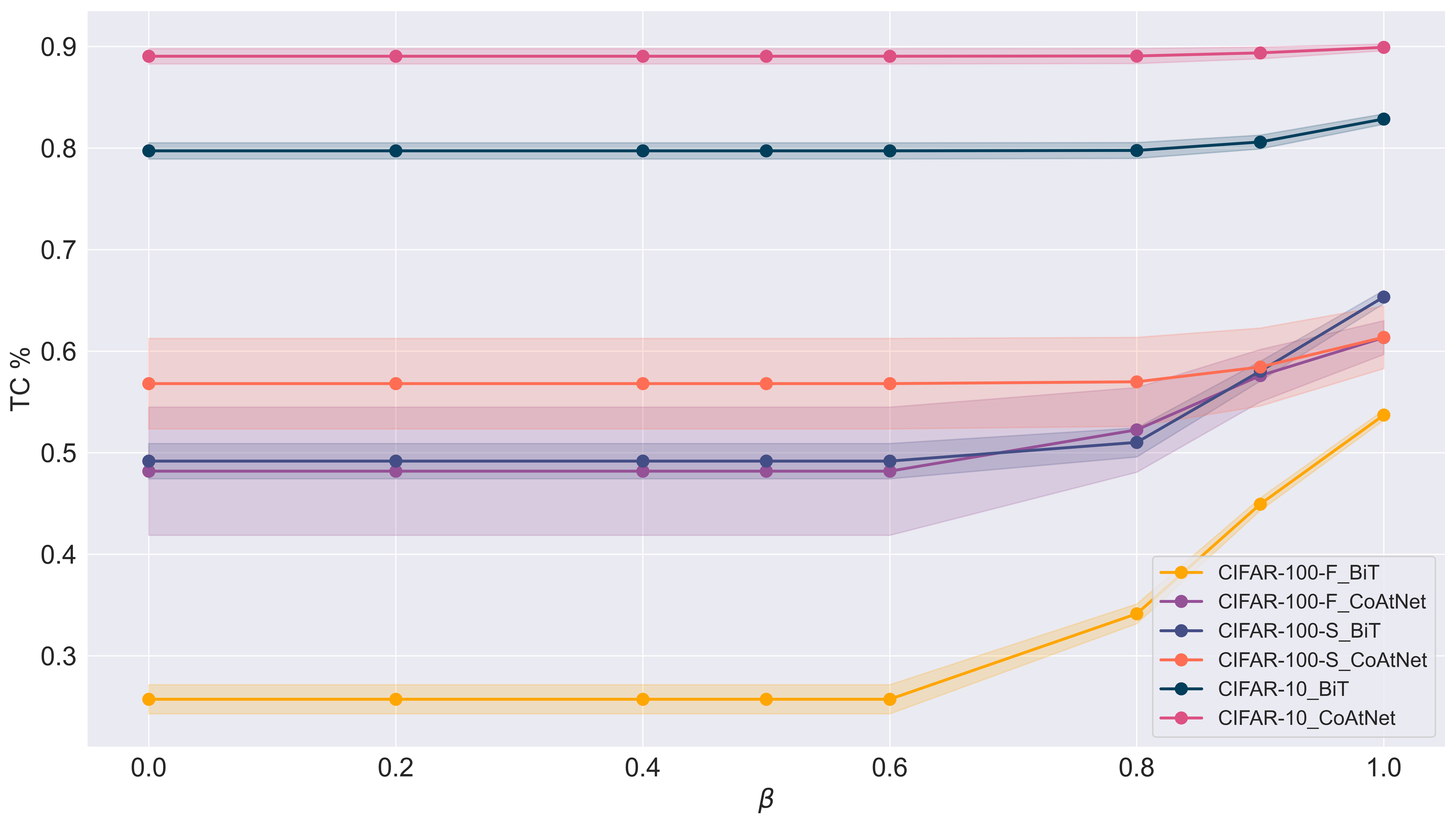}}
  \caption{Impact of underconfidence parameter $\beta$ on ECE, FC\%, TC\% (neighborhood size $K = 20$ fixed).}
  \label{fig:ablation_impact_beta_performance_result}
\end{figure}

As $\beta$ increases, the calibration targets preserve more of the original predicted probabilities. This allows the correctly predicted samples that were initially misplaced in the putatively incorrect group to regain proximity to their true class probability and be reassigned as true confident. This explains the observed increase in TC with larger $\beta$ values, most prominently for CIFAR-100-F, BiT, where the fraction of putatively incorrect samples is substantial. Thus, the trend in Figure \ref{fig:ablation_impact_beta_performance_result} can be interpreted as a recovery effect: increasing $\beta$ reduces the excessive penalization of correct samples in the putatively incorrect group, enabling their detection as TC. This effect is weaker in datasets such as CIFAR-10, where the putatively correct group already dominates the stratification and TC remains consistently high across all $\beta$ values.

The sensitivity experiments highlight two central design rules for the dual calibration framework. First, the neighborhood size $K$ governs the quality of conformal stratification: moderate $K$ values enhance the separation of correct and incorrect predictions, but overly large $K$ mis-flags correct samples as putatively incorrect, inflating ECE and reducing the TC. Second, the underconfidence factor $\beta$ regulates how strongly predictions are pulled toward maximum entropy: very small $\beta$ values overly suppress confidence and prevent the recovery of correct samples in the putatively incorrect group, while larger $\beta$ values allow calibrated probabilities to realign with true correctness, reducing ECE and increasing TC.

Together, these results establish that optimal performance emerges from balancing $K$ and $\beta$, where $K$ is large enough to purify the putatively correct set without over-penalizing correct samples, and $\beta$ is high enough to preserve predictive signal while still tempering overconfidence. This trade-off ensures low false confidence, controlled calibration error, and robust detection of true confident samples across datasets and backbones.

\section{Conclusion} \label{sec:Conclusion}
This work addresses a critical gap in neural network calibration by introducing an uncertainty-aware post-hoc framework that bridges calibration improvement with effective decision-making under uncertainty. Unlike conventional calibration methods that apply uniform transformations across all predictions, the proposed approach leverages proximity-based conformal prediction to stratify samples based on predicted correctness, enabling targeted calibration strategies tailored to prediction reliability.
The framework implements a dual-pathway calibration mechanism: standard isotonic regression calibrates confidence in putatively correct predictions, while underconfidence-regularized calibration reduces confidence toward uniform distributions for putatively incorrect predictions. This instance-level adaptivity addresses the fundamental limitation of global calibration methods, which overlook the safety-critical distinction between refining reliable predictions versus detecting unreliable ones for further review.

The comprehensive empirical evaluation established three fundamental contributions of the proposed dual calibration framework. First, the results validated that optimal traditional calibration (minimal ECE) and effective uncertainty-aware decision making represent distinct—and sometimes competing—objectives. Standard isotonic regression consistently achieved superior ECE yet demonstrated the high FC rates at permissive thresholds.

The dual calibration method's intentional ECE degradation yielded substantial FC reductions at moderate thresholds ($\tau$=0.5-0.6), confirming that targeted underconfidence-regularization provides measurable safety benefits that justify departing from probability-matching optimality. Second, the architectural dependency analysis revealed that training-time calibration methods suffer from fundamental brittleness—focal loss exhibited dramatic performance divergence across architectures. The dual calibration method maintained consistent performance patterns regardless of architecture, demonstrating the inherent advantage of post-hoc approaches that operate independently of gradient dynamics and feature learning processes.

Third, ablation studies established clear design principles: moderate neighborhood values optimize conformal stratification by balancing group purity against correct-sample misclassification, while underconfidence ratios ($\beta$=0.6-0.9) preserve discriminative information while tempering overconfidence. Critically, framework effectiveness emerged progressively with entropy thresholds increased, achieving maximum benefits.

The work demonstrates a successful paradigm shift from calibration as probability alignment to risk-aware confidence adjustment. Statistical validation confirmed this transition yields simultaneous FC reduction and maintained operational utility, representing an intelligent reallocation of calibration resources toward safety-critical objectives. This advancement holds particular significance for high-stakes domains such as medical diagnosis, where confidently incorrect predictions can result in severe consequences.

However, several limitations present opportunities for future research. First, the framework's effectiveness is fundamentally constrained by conformal prediction quality. Degraded stratification accuracy substantially diminishes performance gains. Future work could explore adaptive conformal prediction mechanisms or ensemble-based stratification to enhance separation reliability. Second, the method introduces two hyperparameters requiring dataset-specific tuning, potentially limiting deployment ease. Developing automated hyperparameter selection strategies, such as meta-learning approaches that dynamically adjust parameters based on stratification quality metrics, would significantly improve practical deployability. As the present study focused on image classification, exploring the framework’s applicability to other modalities, including natural language offers an important avenue for future investigation.


\section*{Declarations}
The authors have no relevant financial or non-financial interests to disclose.

\begin{appendices}

\section{Appendix A: Friedman Test Results across Calibration Methods}\label{secA1}

\begin{table}[!h]
\centering
\caption{Friedman test results across calibration methods for uncertainty-aware performance metrics: FC\% and UG-Mean. The test evaluates whether there are statistically significant differences in the ranks of methods across datasets and backbones for each metric.}
\label{tab:appendix_uncertainty_Friedman}
\begin{tabular}{llllclllll}
\hline
\multicolumn{1}{c}{} & \multicolumn{1}{c}{} & \multicolumn{1}{c}{} & \multicolumn{1}{c}{} & \multicolumn{2}{c}{Friedman} & \multicolumn{4}{c}{AvgRank} \\ \hline
\multicolumn{1}{c}{Dataset} & \multicolumn{1}{c}{Backbone} & \multicolumn{1}{c}{Metric} & \multicolumn{1}{c}{$\tau$} & $\chi^2$ & \multicolumn{1}{c}{P-value} & \multicolumn{1}{c}{Non Cal} & \multicolumn{1}{c}{Focal loss} & \multicolumn{1}{c}{Iso Cal} & \multicolumn{1}{c}{Dual Cal} \\ \hline
CIFAR-10 & BiT & FC \% & 0.2 & 84.092 & $4.07 \times10^{-18}$ & 1.00 & 4.00 & 2.28 & 2.72 \\
CIFAR-10 & BiT & FC \% & 0.3 & 88.840 & $3.89 \times10^{-19}$ & 1.00 & 4.00 & 2.97 & 2.03 \\
CIFAR-10 & BiT & FC \% & 0.4 & 90.000 & $2.19 \times10^{-19}$ & 1.00 & 4.00 & 3.00 & 2.00 \\
CIFAR-10 & BiT & FC \% & 0.5 & 81.160 & $1.73 \times10^{-17}$ & 1.43 & 4.00 & 3.00 & 1.57 \\
CIFAR-10 & BiT & FC \% & 0.6 & 87.760 & $6.63 \times10^{-19}$ & 2.00 & 3.93 & 3.07 & 1.00 \\
CIFAR-100-S & BiT & FC \% & 0.2 & 56.525 & $3.25 \times10^{-12}$ & 1.00 & 2.90 & 3.27 & 2.83 \\
CIFAR-100-S & BiT & FC \% & 0.3 & 84.532 & $3.27 \times10^{-18}$ & 1.85 & 3.03 & 3.97 & 1.15 \\
CIFAR-100-S & BiT & FC \% & 0.4 & 90.000 & $2.19 \times10^{-19}$ & 2.00 & 3.00 & 4.00 & 1.00 \\
CIFAR-100-S & BiT & FC \% & 0.5 & 88.840 & $3.89 \times10^{-19}$ & 2.03 & 2.97 & 4.00 & 1.00 \\
CIFAR-100-S & BiT & FC \% & 0.6 & 87.760 & $6.63 \times10^{-19}$ & 2.07 & 2.93 & 4.00 & 1.00 \\
CIFAR-100-F & BiT & FC \% & 0.2 & 90.000 & $2.19 \times10^{-19}$ & 2.00 & 1.00 & 4.00 & 3.00 \\
CIFAR-100-F & BiT & FC \% & 0.3 & 85.840 & $1.71 \times10^{-18}$ & 1.87 & 1.13 & 4.00 & 3.00 \\
CIFAR-100-F & BiT & FC \% & 0.4 & 73.240 & $8.64 \times10^{-16}$ & 1.17 & 2.53 & 4.00 & 2.30 \\
CIFAR-100-F & BiT & FC \% & 0.5 & 90.000 & $2.19 \times10^{-19}$ & 1.00 & 3.00 & 4.00 & 2.00 \\
CIFAR-100-F & BiT & FC \% & 0.6 & 88.840 & $3.89 \times10^{-19}$ & 1.00 & 3.03 & 3.97 & 2.00 \\
CIFAR-10 & CoAtNet & FC \% & 0.2 & 77.653 & $9.78 \times10^{-17}$ & 1.68 & 1.32 & 3.22 & 3.78 \\
CIFAR-10 & CoAtNet & FC \% & 0.3 & 73.384 & $8.04 \times10^{-16}$ & 1.38 & 1.62 & 3.57 & 3.43 \\
CIFAR-10 & CoAtNet & FC \% & 0.4 & 81.040 & $1.84 \times10^{-17}$ & 1.40 & 1.60 & 3.98 & 3.02 \\
CIFAR-10 & CoAtNet & FC \% & 0.5 & 85.840 & $1.71 \times10^{-18}$ & 1.13 & 1.87 & 4.00 & 3.00 \\
CIFAR-10 & CoAtNet & FC \% & 0.6 & 83.640 & $5.08 \times10^{-18}$ & 1.17 & 1.87 & 4.00 & 2.97 \\
CIFAR-100-S & CoAtNet & FC \% & 0.2 & 90.000 & $2.19 \times10^{-19}$ & 1.00 & 2.00 & 3.00 & 4.00 \\
CIFAR-100-S & CoAtNet & FC \% & 0.3 & 90.000 & $2.19 \times10^{-19}$ & 1.00 & 2.00 & 4.00 & 3.00 \\
CIFAR-100-S & CoAtNet & FC \% & 0.4 & 88.840 & $3.89 \times10^{-19}$ & 1.00 & 2.03 & 4.00 & 2.97 \\
CIFAR-100-S & CoAtNet & FC \% & 0.5 & 81.640 & $1.37 \times10^{-17}$ & 1.00 & 2.63 & 4.00 & 2.37 \\
CIFAR-100-S & CoAtNet & FC \% & 0.6 & 88.840 & $3.89 \times10^{-19}$ & 1.00 & 2.97 & 4.00 & 2.03 \\
CIFAR-100-F & CoAtNet & FC \% & 0.2 & 90.000 & $2.19 \times10^{-19}$ & 2.00 & 1.00 & 4.00 & 3.00 \\
CIFAR-100-F & CoAtNet & FC \% & 0.3 & 83.560 & $5.29 \times10^{-18}$ & 2.23 & 1.00 & 4.00 & 2.77 \\
CIFAR-100-F & CoAtNet & FC \% & 0.4 & 79.480 & $3.97 \times10^{-17}$ & 2.57 & 1.03 & 4.00 & 2.40 \\
CIFAR-100-F & CoAtNet & FC \% & 0.5 & 84.240 & $3.78 \times10^{-18}$ & 2.20 & 1.00 & 4.00 & 2.80 \\
CIFAR-100-F & CoAtNet & FC \% & 0.6 & 86.760 & $1.09 \times10^{-18}$ & 2.10 & 1.00 & 4.00 & 2.90 \\
CIFAR-10 & BiT & UG-Mean & 0.2 & 85.360 & $2.17 \times10^{-18}$ & 4.00 & 1.07 & 2.93 & 2.00 \\
CIFAR-10 & BiT & UG-Mean & 0.3 & 68.720 & $8.02 \times10^{-15}$ & 3.87 & 2.93 & 1.87 & 1.33 \\
CIFAR-10 & BiT & UG-Mean & 0.4 & 81.040 & $1.84 \times10^{-17}$ & 1.47 & 4.00 & 3.00 & 1.53 \\
CIFAR-10 & BiT & UG-Mean & 0.5 & 77.480 & $1.07 \times10^{-16}$ & 1.63 & 3.87 & 3.13 & 1.37 \\
CIFAR-10 & BiT & UG-Mean & 0.6 & 82.000 & $1.14 \times10^{-17}$ & 2.00 & 3.67 & 3.33 & 1.00 \\
CIFAR-100-S & BiT & UG-Mean & 0.2 & 79.120 & $4.74 \times10^{-17}$ & 4.00 & 2.87 & 1.93 & 1.20 \\
CIFAR-100-S & BiT & UG-Mean & 0.3 & 81.160 & $1.73 \times10^{-17}$ & 4.00 & 1.57 & 1.43 & 3.00 \\
CIFAR-100-S & BiT & UG-Mean & 0.4 & 76.280 & $1.93 \times10^{-16}$ & 3.13 & 1.27 & 3.83 & 1.77 \\
CIFAR-100-S & BiT & UG-Mean & 0.5 & 82.000 & $1.14 \times10^{-17}$ & 2.67 & 2.33 & 4.00 & 1.00 \\
CIFAR-100-S & BiT & UG-Mean & 0.6 & 81.040 & $1.84 \times10^{-17}$ & 2.53 & 2.47 & 4.00 & 1.00 \\
CIFAR-100-F & BiT & UG-Mean & 0.2 & 90.000 & $2.19 \times10^{-19}$ & 3.00 & 4.00 & 1.00 & 2.00 \\
CIFAR-100-F & BiT & UG-Mean & 0.3 & 90.000 & $2.19 \times10^{-19}$ & 3.00 & 4.00 & 1.00 & 2.00 \\
CIFAR-100-F & BiT & UG-Mean & 0.4 & 83.560 & $5.29 \times10^{-18}$ & 3.00 & 4.00 & 1.23 & 1.77 \\
CIFAR-100-F & BiT & UG-Mean & 0.5 & 84.240 & $3.78 \times10^{-18}$ & 1.20 & 3.00 & 4.00 & 1.80 \\
CIFAR-100-F & BiT & UG-Mean & 0.6 & 69.960 & $4.35 \times10^{-15}$ & 1.23 & 2.33 & 4.00 & 2.43 \\
CIFAR-10 & CoAtNet & UG-Mean & 0.2 & 74.080 & $5.71 \times10^{-16}$ & 3.37 & 3.63 & 1.70 & 1.30 \\
CIFAR-10 & CoAtNet & UG-Mean & 0.3 & 51.240 & $4.35 \times10^{-11}$ & 1.30 & 2.13 & 3.43 & 3.13 \\
CIFAR-10 & CoAtNet & UG-Mean & 0.4 & 82.000 & $1.14 \times10^{-17}$ & 1.33 & 1.67 & 4.00 & 3.00 \\
CIFAR-10 & CoAtNet & UG-Mean & 0.5 & 83.560 & $5.29 \times10^{-18}$ & 1.23 & 1.77 & 4.00 & 3.00 \\
CIFAR-10 & CoAtNet & UG-Mean & 0.6 & 84.520 & $3.29 \times10^{-18}$ & 1.13 & 1.90 & 4.00 & 2.97 \\
CIFAR-100-S & CoAtNet & UG-Mean & 0.2 & 88.840 & $3.89 \times10^{-19}$ & 4.00 & 3.00 & 1.97 & 1.03 \\
CIFAR-100-S & CoAtNet & UG-Mean & 0.3 & 81.000 & $1.87 \times10^{-17}$ & 4.00 & 3.00 & 1.50 & 1.50 \\
CIFAR-100-S & CoAtNet & UG-Mean & 0.4 & 82.960 & $7.11 \times10^{-18}$ & 4.00 & 3.00 & 1.73 & 1.27 \\
CIFAR-100-S & CoAtNet & UG-Mean & 0.5 & 81.000 & $1.87 \times10^{-17}$ & 3.50 & 2.00 & 3.50 & 1.00 \\
CIFAR-100-S & CoAtNet & UG-Mean & 0.6 & 84.240 & $3.78 \times10^{-18}$ & 1.00 & 2.80 & 4.00 & 2.20 \\
CIFAR-100-F & CoAtNet & UG-Mean & 0.2 & 87.760 & $6.63 \times10^{-19}$ & 3.00 & 4.00 & 1.07 & 1.93 \\
CIFAR-100-F & CoAtNet & UG-Mean & 0.3 & 88.840 & $3.89 \times10^{-19}$ & 2.97 & 4.00 & 1.00 & 2.03 \\
CIFAR-100-F & CoAtNet & UG-Mean & 0.4 & 88.840 & $3.89 \times10^{-19}$ & 1.97 & 4.00 & 3.00 & 1.03 \\
CIFAR-100-F & CoAtNet & UG-Mean & 0.5 & 54.280 & $9.78 \times10^{-12}$ & 1.97 & 2.10 & 4.00 & 1.93 \\
CIFAR-100-F & CoAtNet & UG-Mean & 0.6 & 86.760 & $1.09 \times10^{-18}$ & 2.10 & 1.00 & 4.00 & 2.90 \\ \hline
\end{tabular}
\end{table}

\end{appendices}

\bibliography{Refs}

\end{document}